%% file: ms.tex
\pdfoutput=1
\documentclass{article}

\usepackage[preprint]{neurips_2022}

\usepackage[T1]{fontenc}    
\usepackage{nicefrac}       
\usepackage{microtype}      

\usepackage[utf8]{inputenc}
\usepackage{graphicx}
\usepackage{geometry}
\usepackage{latexsym}
\usepackage{amsfonts}
\usepackage{mathrsfs}
\usepackage{amssymb,amscd}
\usepackage[all]{xy}
\usepackage{amsmath}
\usepackage{fancyhdr}
\usepackage{fancybox}
\usepackage[dvipsnames]{xcolor}
\usepackage{array}
\usepackage{makecell}
\usepackage{multirow}
\usepackage{booktabs}
\usepackage{arydshln}
\usepackage{algorithm}
\usepackage{algorithmic}
\usepackage{url}
\usepackage{natbib}
\usepackage[colorlinks=true,allcolors=blue]{hyperref}
\usepackage{caption}
\usepackage{subcaption}
\usepackage{overpic}

\usepackage{tikz}
\usetikzlibrary{matrix}

\input{header}

\title{Approximate Data Deletion in Generative Models}
\author{%
  Zhifeng Kong \\
  University of California San Diego\\
  La Jolla, CA 92093 \\
  \texttt{z4kong@eng.ucsd.edu} \\
  \And
  Scott Alfeld \\
  Amherst College\\
  Amherst, MA, 01002\\
  \texttt{salfeld@amherst.edu}
}

\begin{document}

\maketitle

\begin{abstract}
Users have the right to have their data deleted by third-party learned systems, as codified by recent legislation such as the General Data Protection Regulation (GDPR) and the California Consumer Privacy Act (CCPA). Such data deletion can be accomplished by full re-training, but this incurs a high computational cost for modern machine learning models. To avoid this cost, many approximate data deletion methods have been developed for supervised learning. Unsupervised learning, in contrast, remains largely an open problem when it comes to (approximate or exact) efficient data deletion. In this paper, we propose a density-ratio-based framework for generative models. Using this framework, we introduce a fast method for approximate data deletion and a statistical test for estimating whether or not training points have been deleted. We provide theoretical guarantees under various learner assumptions and empirically demonstrate our methods across a variety of generative methods. 
\end{abstract}

\section{Introduction}\label{sec: intro}

Machine learning has proved to be an increasingly powerful tool.
With this power comes responsibility and there are growing concerns in academia, government, and the private sector about user privacy and responsible data management.
Recent regulations (e.g., GDPR and CCPA) have introduced a \textit{right to erasure} whereby a user may request that their data is deleted from a database.
While deleting user data from a simple database is straightforward, a savvy attacker might still be able to reverse-engineer the data by examining a machine learning model trained on it \citep{balle2021reconstructing}.
Re-training a model from scratch (after deleting the requested data) is computationally expensive, especially for modern deep learning methods (sometimes taking days or weeks to train \citep{karras2020analyzing}).
This has motivated \textit{machine unlearning} \citep{cao2015towards} where learned models are altered (in a computationally cheap way) to emulate the re-training process.
In this paper we focus on machine unlearning for generative modeling, a class of unsupervised learning methods that learn the probability distribution from data.

Prior work in supervised learning proposed \textit{approximate data deletion} to approximate the re-trained model without actually performing the re-training \citep{guo2019certified,neel2021descent,sekhari2021remember,izzo2021approximate}.
While these methods have achieved great success, approximate data deletion for \textit{un}supervised learning largely remains an open question.
In this paper we present a framework for generative models wherein we model the updated training set (the original with a subset removed) as a collection of i.i.d. samples from a perturbed distribution.
Using this framework we present two novel contributions:
\begin{enumerate}
\vspace{-0.5em}
    \item We propose a fast method for approximate data deletion for generative models.
    \item We provide a statistical test for estimating whether or not training data have been deleted from a generative model given only sample access to the model.
\vspace{-0.5em}
\end{enumerate}
For both contributions, we provide theoretical guarantees under a variety of learner assumptions and perform empirical investigations.

The supervised and unsupervised settings have two major differences in the context of data deletion.
The first is the definition -- what does it mean to effectively delete training data?
In the supervised setting, it means the function that maps data to targets (i.e., the classifier) approximates the re-trained function, while in generative models the goal is to approximate the re-trained generative distribution.
The other difference is the user's capability when evaluating data deletion.
In the supervised setting, one can construct an input sample and query its predicted target.
In contrast, we consider the setting where a user can only draw samples from a generative model and then investigate the empirical distribution to evaluate the effectiveness of approximate data deletion.

In Section~\ref{sec: framework} we present our density-ratio-based framework.
We present our primary contributions in Sections~\ref{sec: DRE} and~\ref{sec: stat test}.
We then perform empirical investigations (Section~\ref{sec: experiments}) on real and synthetic data for both our fast deletion method and statistical test.
We discuss related work in Section~\ref{sec: related work} and conclude with a discussion of future work in Section~\ref{sec: conclusion}.

\section{A Density Ratio Based Framework for Data Deletion}\label{sec: framework}


Let \(p_*\) be a distribution over $\mathbb{R}^d$ and $X$ be $N$ i.i.d. samples from $p_*$.
We consider a generative learning algorithm \(\mA\) which aims to model \(p_*\). 
We denote the distribution \(\mA\) learns from \(X\) as $p_{\mA(X)}$, and we refer to $\hat{p}=p_{\mA(X)}$ as the pre-trained model.
Let $X'\subset X$ be $N'$ samples we would like to delete from $\hat{p}$, and $\hat{p}'=p_{\mA(X\setminus X')}$ be the ground-truth re-trained model. A notation table is provided in Appendix \ref{appendix: notation}. In this paper, we present solutions to two problems: 
\begin{enumerate}
\vspace{-0.5em}
    \item Fast deletion: given $\hat{p}$, approximate $\hat{p}'$ more efficiently than full re-training.
    \item Deletion test: assuming $q\in\{\hat{p},\hat{p}'\}$, test whether $q=\hat{p}$ or $q=\hat{p}'$ by drawing samples.
\vspace{-0.5em}
\end{enumerate}

\subsection{Framework}

\begin{figure}[!t]
\vspace{-1.2em}
\begin{minipage}[t!]{.7\textwidth}
    \centering
    \begin{tikzpicture}
        \matrix (m) [matrix of math nodes,row sep=1em,column sep=2em,minimum width=0.5em,nodes={anchor=center,color=blue}]
        {p_* & X & \hat{p} & \mD(\hat{p},X,X') \\
         & & \hat{\rho}_{\mE} & \\
         p_*' & X\setminus X' & & \hat{p}' \\};
        \path[-stealth]
        (m-1-1) edge node [above] {i.i.d.} (m-1-2)
        (m-3-1) edge node [above] {i.i.d.} (m-3-2)
        (m-1-1) edge node [left] {$\times\rho_*$} (m-3-1)
        (m-1-2) edge node [left] {$-X'$} (m-3-2)
        (m-1-2) edge node [above] {$\mA$} (m-1-3)
        (m-3-2) edge node [above] {$\mA$} (m-3-4)
        (m-1-2) edge node [below=2mm] {DRE} (m-2-3)
        (m-3-2) edge node [below] {} (m-2-3)
        (m-1-3) edge node [above] {$\times\hat{\rho}_{\mE}$} (m-1-4)
        (m-2-3) edge node [below] {} (m-1-4.west)
        (m-1-4) edge node [left] {goal: $\approx$} (m-3-4)
        ;
    \end{tikzpicture}
    \captionsetup{width=0.93\textwidth}
    \vspace{-0.8em}
    \captionof{figure}{Our density-ratio-based framework. We train a DRE $\hat{\rho}_{\mE}$ between $X$ and $X\setminus X'$. We then multiply it to the pre-trained model $\hat{p}$ to obtain the approximated model $\mD(\hat{p},X,X')$ that approximates the re-trained model $\hat{p}'$. We model $X\setminus X'$ to be i.i.d. samples from $p_*'$ which enables theoretical guarantees.}
    \label{fig: framework}
\end{minipage}%
\begin{minipage}[t!]{.3\textwidth}
\begin{algorithm}[H]
    \centering
    \caption{Sampling from the approximated model}
    \label{alg: rej sampling}
    \begin{algorithmic}[1]
        \WHILE{True}
        \STATE Sample $y\sim\hat{p}$ and \\
        $u\sim\texttt{Uniform}([0,1])$.
        \IF{$\hat{\rho}_{\mE}(y) > B\cdot u$}
            \STATE \textbf{return} $y$
        \ENDIF 
        \ENDWHILE
    \end{algorithmic}
\end{algorithm}
\end{minipage}
\vspace{-1.5em}
\end{figure}

In this paper, we propose a density-ratio-based framework to perform fast (approximate) deletion and our deletion test. The density ratio between two distributions $\mu_1$ and $\mu_2$ on $\mathbb{R}^d$ is defined as $\rho(\mu_1,\mu_2): \mathbb{R}^d\rightarrow\mathbb{R}^+, x\mapsto \mu_2(x) / \mu_1(x)$, where we choose this order to make the theory cleaner. Let $\hat{\rho} = \rho(\hat{p},\hat{p}')$ be the density ratio between pre-trained and re-trained models. In our proposed framework, we learn a density ratio estimator (DRE) $\hat{\rho}_{\mE}=\hat{\rho}_{\mE}(X, X\setminus X')$ between $X$ and $X\setminus X'$ to approximate $\hat{\rho}$. 
Then, to perform fast deletion we define the approximated model $\mD(\hat{p},X,X'): \mathbb{R}^d\rightarrow\mathbb{R}^+, x\mapsto \hat{\rho}_{\mE}(x) \cdot \hat{p}(x)$, which we abbreviate as $\hat{\rho}_{\mE} \cdot \hat{p}$ for conciseness.

Core to both our method of fast deletion and our deletion test is our DRE based framework (summarized in Fig. \ref{fig: framework}).
We model $X\setminus X'$ to be a set of i.i.d. samples from some distribution we denote as $p_*'$, and define $\rho_*=\rho(p_*,p_*')$.
We assume $\|\rho_*\|_{\infty} \leq \infty$.
Intuitively,  deleting some samples from $p_*$ will only increase likelihood of regions far from these samples by at most a constant factor, and reduce likelihood of regions around these samples. 
We also assume $N'\ll N$ -- only a small fraction of training samples are to be deleted.
Intuitively, this means that the pre-trained and re-trained models are likely similar.
This allows us to provide approximation bounds for consistent learning algorithms $\mA$.
In Section \ref{sec: approx theory} we derive such bounds for various forms of consistency.

In the supervised setting, approximate deletion can be done by altering the pre-trained model to be closer to the (never computed) re-trained model.
In contrast, we alter the process of sampling from the unsupervised pre-trained model to simulate sampling from the re-trained model.
Drawing samples from the approximated model is done in two steps: first draw samples from $\hat{p}$, and then perform rejection sampling according to $\hat{\rho}_{\mE}$. 
Note that this procedure requires there exists a known constant $B\geq \|\hat{\rho}_{\mE}\|_{\infty}$, which we discuss further in  Section \ref{sec: practical theory}.
We present this procedure in Alg. \ref{alg: rej sampling}.

\subsection{Approximation under Consistency}\label{sec: approx theory}

A learning algorithm \(\mA\) is said to be \textit{consistent} if $p_{\mA(X)}$ converges to $p_*$ as $N\rightarrow\infty$ \citep{wied2012consistency}, where a specific type of convergence leads to a specific definition of consistency. If $\mA$ is consistent, then we have $\hat{p}\approx p_*$ and $\hat{p}'\approx p_*'$ for large $N$.
In this section, we derive DREs for two kinds of consistency to achieve approximated deletion: $\hat{\rho}_{\mE}$ such that the approximated model $\mD(\hat{p},X,X'):=\hat{\rho}_{\mE}\cdot\hat{p}\approx\hat{p}'$.

In \textbf{Def.} \ref{def: RC}, we introduce ratio consistency, which bounds the density ratio between true and learned distributions. We show in \textbf{Thm.} \ref{thm: approx under RC} that approximation in $L_1$ distance can be achieved in this case. We then look at a more practical total variation consistency in \textbf{Def.} \ref{def: TVC}, which bounds the total variation distance (half of $L_1$ distance) between true and learned distributions. We show in \textbf{Thm.} \ref{thm: approx under TVC} that approximation in expectation is achieved in this case.

\begin{definition}[Ratio Consistent (RC)]\label{def: RC}
    We say $\mA$ is $(c_N,\delta_N)$-RC if for any distribution $\mu$, with probability at least $1-\delta_N$, it holds that $\|\log\rho(p_{\mA(Z)}, \mu)\|_{\infty} \leq \log c_N$, where $Z$ contains $N$ i.i.d. samples from $\mu$, and $c_N\rightarrow1$, $\delta_N\rightarrow0$ as $N\rightarrow\infty$.
\end{definition}

\begin{theorem}[Approximation under RC]\label{thm: approx under RC}
    If $\mA$ is $(c_N,\delta_N)$-RC, then there exists a DRE $\hat{\rho}_{\mE}$ such that with probability at least $1 - 2(\delta_N+\delta_{N-N'})$, it holds that $\|\hat{\rho}_{\mE}\cdot\hat{p}-\hat{p}'\|_1\leq4(c_N+c_{N-N'}-2)$.
\end{theorem}

\begin{definition}[Total Variation Consistent (TVC)]\label{def: TVC}
    We say $\mA$ is $(\epsilon_N,\delta_N)$-TVC if for any distribution $\mu$, with probability at least $1-\delta_N$, it holds that $\|p_{\mA(Z)} - \mu\|_1 \leq \epsilon_N$, where $Z$ contains $N$ i.i.d. samples from $\mu$, and $\epsilon_N\rightarrow0$, $\delta_N\rightarrow0$ as $N\rightarrow\infty$.
\end{definition}

\begin{theorem}[Approximation under TVC]\label{thm: approx under TVC}
    Define $\|h\|_{1,\mu}=\int_x \mu(x)|h(x)|dx$. 
    If $\mA$ is $(\epsilon_N,\delta_N)$-TVC, then there exists a DRE $\hat{\rho}_{\mE}$ such that with probability at least $1 - 2(\delta_N+\delta_{N-N'})$, it holds that $\|\hat{\rho}_{\mE}\cdot\hat{p}-\hat{p}'\|_{1,\hat{p}}\leq2(\epsilon_{N-N'}+\|\rho_*\|_{\infty}\epsilon_N)$.
\end{theorem}

Full proofs are provided in Appendix \ref{appendix: approx theory}.
For each, the high level idea is to choose a fixed consistent algorithm $\mA_0$, and define $\hat{\rho}_{\mE}(Z_1,Z_2) = \rho(p_{\mA_0(Z_1)}, p_{\mA_0(Z_2)})$. This yields $\hat{\rho}_{\mE}(X, X\setminus X')\approx\rho_*\approx\hat{\rho}$ and therefore $\mD(\hat{p},X,X')=\hat{\rho}_{\mE}\cdot\hat{p}\approx\hat{p}'$.

\subsection{Practicability under Stability}\label{sec: practical theory}

Running Alg. \ref{alg: rej sampling} in practice requires $\|\hat{\rho}_{\mE}\|_{\infty}$ to be finite (see Line 3 of Alg~\ref{alg: rej sampling}). To have $\hat{\rho}_{\mE}\approx \hat{\rho}$, we need $\|\hat{\rho}\|_{\infty}$ to be finite. In this section, we study several stability conditions of the learning algorithm $\mA$ that guarantee this practicability. 

We organize these stability conditions in the order from strong to weak. 
In \textbf{Def.} \ref{def: DP} -- \ref{def: LBLI}, we discuss several strong, classic stability conditions that guarantee $\|\hat{\rho}\|_{\infty}$ to be small (see \textbf{Thm.} \ref{thm: prac under LBLI}). We then introduce ratio stability, a concept crafted for our framework, in \textbf{Def.} \ref{def: RS}.
Ratio stability bounds the difference between two $\log$ density ratios of true and learned distributions, which intuitively indicates the learning algorithm has a stable bias. We discuss its connection with ratio consistency in \textbf{Thm.} \ref{thm: RC RS}, and bound the difference between $\|\hat{\rho}\|_{\infty}$ and $\|\rho_*\|_{\infty}$ in \textbf{Thm.} \ref{thm: prac under RS}. Finally, we discuss a special type of error stability \citep{bousquet2002stability} in \textbf{Def.} \ref{def: ES}, and show a concentration bound on $\hat{\rho}$ in \textbf{Thm.} \ref{thm: prac under ES}.

\begin{definition}[Differentially Private (DP) \citep{dwork2006calibrating}]\label{def: DP}
    We say $\mA$ is $\epsilon$-DP if for any adjacent sets $Z_0$ and $Z_1$, and any test set $\hat{Z}$, it holds that $e^{-\epsilon} p_{\mA(Z_1)}(\hat{Z}) \leq p_{\mA(Z_0)}(\hat{Z}) \leq e^{\epsilon} p_{\mA(Z_1)}(\hat{Z})$.
\end{definition}

\begin{definition}[Uniformly Stable (US) \citep{bousquet2002stability}]\label{def: US}
    We say $\mA$ is $\epsilon$-US if for any set $Z$, $z\in Z$, and test sample $\hat{z}$, it holds that $|\log p_{\mA(Z\setminus\{z\})}(\hat{z}) - \log p_{\mA(Z)}(\hat{z})| \leq \epsilon$.
\end{definition}

\begin{definition}[Lower Bounded in Likelihood Influence (LBLI) \citep{koh2017understanding, kong2021understanding}]\label{def: LBLI}
    We say $\mA$ is $\epsilon$-LBLI if for any set $Z$, $z\in Z$, and test sample $\hat{z}$, it holds that $p_{\mA(Z\setminus\{z\})}(\hat{z}) \leq e^{\epsilon} p_{\mA(Z)}(\hat{z})$.
\end{definition}

We discuss relationship among DP, US and LBLI algorithms in Remark \ref{remark: strong assumptions for feasibility}.

\begin{theorem}\label{thm: prac under LBLI}
    If $\mA$ is $\epsilon$-DP, $\epsilon$-US, or $\epsilon$-LBLI, then $\log\|\hat{\rho}\|_{\infty}\leq N'\epsilon$.
\end{theorem}

\begin{definition}[Ratio Stable (RS)]\label{def: RS}
    We say $\mA$ is $(\epsilon, \delta)$-RS if for any densities $\mu_1$, $\mu_2$ such that $\sup_x \mu_2(x)/\mu_1(x)<\infty$, with probability at least $1-\delta$, when i.i.d. samples $Z_i\sim \mu_i$ satisfy $|Z_1|=|Z_2|+1$, it holds that $\|\log\rho(\mu_1, p_{\mA(Z_1)}) - \log\rho(\mu_2, p_{\mA(Z_2)})\|_{\infty} \leq \epsilon$.
\end{definition}

\begin{theorem}\label{thm: RC RS}
    If $\mA$ is $(c_N,\delta_N)$-RC, then $\mA$ is $(2\log c_N,2\delta_N)$-RS.
\end{theorem}


\begin{theorem}\label{thm: prac under RS}
    If $\mA$ is $(\epsilon,\delta)$-RS, then with probability at least $1-N'\delta$, it holds that $\log\|\hat{\rho}\|_{\infty}\leq N'\epsilon+\log\|\rho_*\|_{\infty}$.
\end{theorem}

\begin{definition}[Error Stable (ES) \citep{bousquet2002stability}]\label{def: ES}
    We say $\mA$ is $(\epsilon,k)$-ES if for any set $Z$ and $z\in Z$, it holds that $|\mathbb{E}_{\hat{z}\sim p_{\mA(Z)}}\left[\log p_{\mA(Z\setminus\{z\})}(\hat{z}) - \log p_{\mA(Z)}(\hat{z})\right]^k| \leq \epsilon$.
\end{definition}

\begin{theorem}\label{thm: prac under ES}
    Let $N'=1$. If $\mA$ is $(\epsilon,2)$-ES, then $\KL{\hat{p}}{\hat{p}'}\leq\sqrt{\epsilon}$, and with probability at least $1-\delta$, it holds that $\log\hat{\rho}(x) \leq \sqrt{\epsilon(1-\delta)/\delta}$ for $x\sim\hat{p}$.
\end{theorem}

We prove these theorems by induction and central inequalities. See Appendix \ref{appendix: practical theory} for proofs. 

\section{Density Ratio Estimators for Fast Data Deletion}\label{sec: DRE}

A key step in the proposed framework is to train a density ratio estimator (DRE) $\hat{\rho}_{\mE}$ between $X$ and $X\setminus X'$. There is a rich literature of DRE techniques \citep{sugiyama2012density,nowozin2016f,moustakides2019training,khan2019deep,rhodes2020telescoping,kato2021non,choi2021featurized,choi2022density}. All of these methods are designed for settings with little prior information about the data, and the two set of samples can potentially be very separated. However, in the data deletion setting, we have strong prior information that one set ($X\setminus X'$) is a strict subset of the other ($X$).
We leverage this fact to design more focused DRE methods.
In Section \ref{sec: DRE classification}, we derive a simple DRE based on probabilistic classification, and compare it with standard methods \citep{sugiyama2012density}.
In Section \ref{sec: DRE VDM}, we use variational divergence minimization \citep{nowozin2016f} to train a DRE that is able to handle high dimensional real-world datasets. 

\subsection{Probabilistic Classification}\label{sec: DRE classification}

We derive a simple DRE based on probabilistic classification.
Let $f$ be a (soft) classifier for the task of distinguishing between $X\setminus X'$ and $X'$: $f(x)=\prob{x\in X'}$, and $1-f(x)=\prob{x\in X\setminus X'}$. By Bayes' rule, the resulting DRE between $X$ and $X\setminus X'$ is 
\begin{equation}
\begin{array}{rl}
    \hat{\rho}_{\mE}(x) = \hat{\rho}_f(x) 
    & \displaystyle = \frac{\prob{X}}{\prob{X\setminus X'}}\cdot\frac{\prob{X\setminus X'|x}}{\prob{X|x}} \\
    & \displaystyle = \frac{N}{N-N'}\cdot\frac{(1-f(x))/2}{f(x)+(1-f(x))/2} 
    = \frac{N}{N-N'}\cdot\frac{1-f(x)}{1+f(x)}.
\end{array}
\end{equation}

To provide intuition, we consider Kernel Density Estimation (KDE) \citep{murray1956remarks, parzen1962estimation}, a class of consistent algorithms which learn an explicit probability density.

\begin{example}[KDE]
\label{eg: KDE classifier}
    Let $\mA$ be KDE with Gaussian kernel function $K_{\sigma}(x) = \mN(x; 0,\sigma^2I)$. Then,
    \begin{align}
        \hat{\rho}(x) = \frac{\hat{p}_{\KDE}(x;X\setminus X')}{\hat{p}_{\KDE}(x;X)} = \frac{N}{N-N'}\cdot\frac{\sum_{i=N'+1}^N K_{\sigma}(x-x_i)}{\sum_{i=1}^N K_{\sigma}(x-x_i)}.
    \end{align}
    The following classifier $f$ recovers $\hat{\rho}_f=\hat{\rho}$:
    \begin{align}\label{eq: KDE classifier}
        f(x) = \frac{\sum_{i=1}^{N'} K_{\sigma}(x-x_i)}{\sum_{i=1}^{N'} K_{\sigma}(x-x_i)+2\sum_{i=N'+1}^N K_{\sigma}(x-x_i)}.
    \end{align}
    This is a weighted soft $N$-nearest-neighbour classifier. If $K_{\sigma}(x)=1\{\|x\|\leq\sigma\}$, then the classifier degrades to the majority votes of samples in $\mB_{\sigma}(x)$, where each sample in $X\setminus X'$ has two votes.
\end{example}

Example \ref{eg: KDE classifier} indicates that we need to duplicate $X\setminus X'$ when training the classifier. This observation is universal as the probability of $x$ belonging to $X\setminus X'$ is shared by two cases: $x\in X$ and $x\in X\setminus X'$.

\subsection{Variational Divergence Minimization}\label{sec: DRE VDM}

Note that KDE and classification-based DRE are especially friendly but may not be able to deal with complicated, high-dimensional datasets \citep{choi2022density}. Now, we consider the learner to be a Generative Adversarial Network (GAN) \citep{goodfellow2014generative}, a class of powerful implicit deep generative models. For these models, we derive a DRE based on variational divergence minimization (VDM), a technique used to analyze and train $f$-GAN \citep{nowozin2016f}. Because neural networks can have large capacity and VDM is designed to distinguish distributions, VDM-based DRE is more applicable with complicated data such as images compared to classification-based DRE. 
We begin with the definition of $\phi$-divergence (also called the $f$-divergence) below.
\begin{definition}[\citep{liese2006divergences}]\label{def: phi divergence}
    The $\phi$-divergence between distributions $\mu$ and $\nu$ is defined as 
    $D_{\phi}(\mu\|\nu) = \int_x \nu(x)\phi\left[\mu(x)/\nu(x)\right]dx$.
\end{definition}

We then apply VDM to $D_{\phi}(p_*'||p_*)$ in a similar way as $f$-GAN \citep{nowozin2016f}: 
\begin{align}\label{eq: VDM}
    D_{\phi}(p_*'||p_*) \geq \sup_T\  \left(\mathbb{E}_{x\sim p_*'}T(x) - \mathbb{E}_{x\sim p_*} \phi^*(T(x))\right).
\end{align}
where $\phi^*$ is the conjugate function of $\phi$ defined as $\phi^*(t):=\sup_u (ut-\phi(u))$. The optimal $T$ is obtained at $T(x)=\frac{d}{dt}\phi(p_*'(x)/p_*(x))=\frac{d}{dt}\phi(\rho_*(x))$ \citep{nguyen2010estimating}. To perform the actual training, we optimize the empirical version of \eqref{eq: VDM} based on the i.i.d. assumptions on $X$ and $X\setminus X'$:
\begin{align}
    T_{\phi} = \arg\max_T\ \mathbb{E}_{x\sim X\setminus X'}T(x) - \mathbb{E}_{x\sim X} \phi^*(T(x)),
\end{align}
and then solve the DRE through $\hat{\rho}_{\mE} := \hat{\rho}_{\phi} = (\frac{d}{dt}\phi)^{-1}(T_{\phi})$. We provide specific algorithms to train DRE for two $\phi$-divergences below. In both examples, $T$ is a neural network.

\begin{example}[Jensen-Shannon]\label{eg: VDM JS}
    Let $D_{\phi}$ be the Jason-Shannon divergence. With an additional $\log(\cdot)$ term, we recover the discriminator loss in GAN \citep{goodfellow2014generative}:
    \begin{align}
        T_{\phi} = \arg\max_T\ \mathbb{E}_{x\sim X\setminus X'}\log T(x) + \mathbb{E}_{x\sim X}\log (1-T(x)),~~
        \hat{\rho}_{\phi} = T_{\phi}/(1-T_{\phi}).
    \end{align}
\end{example}

\begin{example}[Kullback–Leibler]\label{eg: VDM KL}
    Let $D_{\phi}$ be the KL divergence. Then, we recover the discriminator loss in KL-GAN \citep{liu2018inductive}:
    \begin{align}\label{eq: KL-GAN}
        T_{\phi} = \arg\max_T\ \mathbb{E}_{x\sim X\setminus X'} T(x) -\mathbb{E}_{x\sim X}(\exp(T(x))-1),~~
        \hat{\rho}_{\phi} = \exp(T_{\phi}-1).
    \end{align}
\end{example}

Note that given enough capacity and data, we have $\hat{\rho}_{\phi}\approx\rho_*$ rather than $\hat{\rho}$, which may cause some bias. This bias can be alleviated when the learner $\mA$ is consistent and expressive enough, such as GAN \citep{liu2021towards}. 
We find KL divergence in Example \ref{eg: VDM KL} works well in practice. 

\section{Statistical Tests for Data Deletion}\label{sec: stat test}

We now turn our attention to the second main contribution of this work: the statistical deletion test.
In this section, we discuss statistical tests to distinguish whether a generative model has had particular points deleted.
Formally, we assume sample access to a distribution $q$, which is either the pre-trained model $\hat{p}$ or the re-trained model $\hat{p}'$. We consider the following hypothesis test:
\begin{align}\label{eq: test}
    H_0: q=\hat{p}; ~~
    H_1: q=\hat{p}'.
\end{align}
Several statistics for this test (not in the data deletion setting) have been proposed, including likelihood ratio ($\mathrm{LR}$) \citep{neyman1933ix}, $\mathrm{ASC}$ statistics \citep{kanamori2011f}, and maximum mean discrepancy ($\mathrm{MMD}$) \citep{gretton2012kernel}. In this section, we adapt $\mathrm{LR}$ and $\mathrm{ASC}$ to the data deletion setting, and discuss $\mathrm{MMD}$ in Appendix \ref{appendix: MMD}. 
In practice, we may not know $\hat{p}'$, so we use $H_1': q=\mD(\hat{p},X,X')$ to approximate $H_1$. We present theory on the approximation between $H_1$ and $H_1'$ when these statistics are used, thus providing an efficient way to test \eqref{eq: test} without re-training. \footnote{It is unclear how to test \eqref{eq: test} even with re-training if the learner $\mA$ (such as GAN) has no explicit likelihood.}

\subsection{Likelihood Ratio}
A common goodness-of-fit method is the likelihood ratio test.
In terms of having the smallest type-2 error, the likelihood ratio test is the most powerful of statistical tests (\citep{neyman1933ix}) and is performed as follows.
Given $m$ samples $Y\sim q$. The likelihood ratio statistic is defined as
\begin{align}
    \mathrm{LR}(Y,\hat{p},\hat{p}') = \frac1m\sum_{y\in Y}\log\frac{\hat{p}'(y)}{\hat{p}(y)} = \frac1m\sum_{y\in Y}\log\hat{\rho}(y).
\end{align}
As it is solely determined by $Y$ and $\hat{\rho}$, we abbreviate it as $\mathrm{LR}(Y,\hat{\rho})$. 
When we use $H_1'$ to approximate $H_1$ in practice, we compute $\mathrm{LR}(Y,\hat{\rho}_{\mE})$. In \textbf{Thm.} \ref{thm: lr approx under RC}, we prove it approximates $\mathrm{LR}(Y,\hat{\rho})$ with high probability under RC (\textbf{Def.} \ref{def: RC}), and in \textbf{Thm.} \ref{thm: lr approx under rho approx}, we show approximation when $\hat{\rho}_{\mE}$ is close to $\hat{\rho}$.

\begin{theorem}\label{thm: lr approx under RC}
    If $\mA$ is $(c_N,\delta_N)$-RC, then there exists a $\hat{\rho}_{\mE}$ such that with probability at least $1 - 2(\delta_N+\delta_{N-N'})$, it holds that $|\mathrm{LR}(Y,\hat{\rho}) - \mathrm{LR}(Y,\hat{\rho}_{\mE}) | \leq 2(\log c_N + \log c_{N-N'})$.
\end{theorem}

\begin{theorem}\label{thm: lr approx under rho approx}
    (1) If $\|\log\hat{\rho}-\log\hat{\rho}_{\mE}\|_{\infty}\leq\epsilon$, then $|\mathrm{LR}(Y,\hat{\rho}) - \mathrm{LR}(Y,\hat{\rho}_{\mE})| \leq \epsilon$. \\
    (2) If $\max(\|\log\hat{\rho}-\log\hat{\rho}_{\mE}\|_{1,\hat{p}},\|\log\hat{\rho}-\log\hat{\rho}_{\mE}\|_{1,\hat{p}'})\leq\epsilon$, then with probability at least $1-\delta$, it holds that $|\mathrm{LR}(Y,\hat{\rho}) - \mathrm{LR}(Y,\hat{\rho}_{\mE})| \leq \epsilon/\delta$.
\end{theorem}


Statistical properties of likelihood ratio and proofs to the above theorems are in Appendix \ref{appendix: LR}.

\subsection{ASC Statistics}
ASC statistics are used to estimate the $\phi$-divergence (\textbf{Def.} \ref{def: phi divergence}) \citep{kanamori2011f}. Because a broad family of $\phi$ functions can be used, these statistics include a wide range of statistics. Draw $m$ samples $Y\sim q$ and another $m$ samples $\hat{Y}$ from $\hat{p}$. The ASC statistic is defined as
\begin{align}
    \hat{\mathrm{ASC}}_{\phi}(\hat{Y},Y,\hat{\rho}) = \frac{1}{m}\sum_{y\in \hat{Y}}\frac{\phi(\hat{\rho}(y))}{1+\hat{\rho}(y)} + \frac{1}{m}\sum_{y\in Y}\frac{\phi(\hat{\rho}(y))}{1+\hat{\rho}(y)}.
\end{align}
When we use $H_1'$ to approximate $H_1$ in practice, we compute $\hat{\mathrm{ASC}}_{\phi}(\hat{Y},Y,\hat{\rho}_{\mE})$. In \textbf{Thm.} \ref{thm: asc approx under rho approx}, we show it approximates $\hat{\mathrm{ASC}}_{\phi}(\hat{Y},Y,\hat{\rho})$ when $\hat{\rho}_{\mE}$ is close to $\hat{\rho}$. 

\begin{theorem}\label{thm: asc approx under rho approx}
    If $\max(\|\psi(\hat{\rho})-\psi(\hat{\rho}_{\mE})\|_{1,\hat{p}},\|\psi(\hat{\rho})-\psi(\hat{\rho}_{\mE})\|_{1,\hat{p}'})\leq\epsilon$ where $\psi(t)=\phi(t)/(1+t)$, then with probability at least $1-\delta$, it holds that $|\hat{\mathrm{ASC}}_{\phi}(\hat{Y},Y,\hat{\rho}) - \hat{\mathrm{ASC}}_{\phi}(\hat{Y},Y,\hat{\rho}_{\mE})|\leq 2\epsilon/\delta$.
\end{theorem}

Statistical properties of ASC statistics and proof to the above theorem are in Appendix \ref{appendix: ASC}.

\section{Experiments}\label{sec: experiments}

\newcommand{\colorApprox}[1]{\textcolor{SkyBlue}{#1}}
\newcommand{\colorPre}[1]{\textcolor{BurntOrange}{#1}}
\newcommand{\colorRe}[1]{\textcolor{ForestGreen}{#1}}

In this section, we address  the following questions. 
\textbf{1) DRE Approximations}: do the methods in Section \ref{sec: DRE} produce ratios $\hat{\rho}_{\mE}$ that approximate the target ratio $\hat{\rho}$? 
\textbf{2) Fast Deletion}: is $\mD(\hat{p},X,X')=\hat{\rho}_{\mE}\cdot\hat{p}$ indistinguishable from the re-trained model $\hat{p}'$?
And \textbf{3) Hypothesis Test}: do the tests in Section \ref{sec: stat test} distinguish samples from pre-trained and re-trained models?

We first survey these questions in experiments on two-dimensional synthetic datasets. We then look at GANs trained on MNIST \citep{lecun2010mnist} and Fashion-MNIST \citep{xiao2017fashion}. All experiments were run on a single machine with one i9-9940X CPU (3.30GHz), one 2080Ti GPU, and 128GB memory. 

\subsection{Classification-based DRE for KDE on Synthetic Datasets}\label{sec: exp 2d}

\begin{figure}[!t]
\vspace{-0.5em} 
    	\begin{subfigure}[b]{0.15\textwidth}
		\centering 
		\includegraphics[trim=25 20 105 20, clip, width=0.99\textwidth]{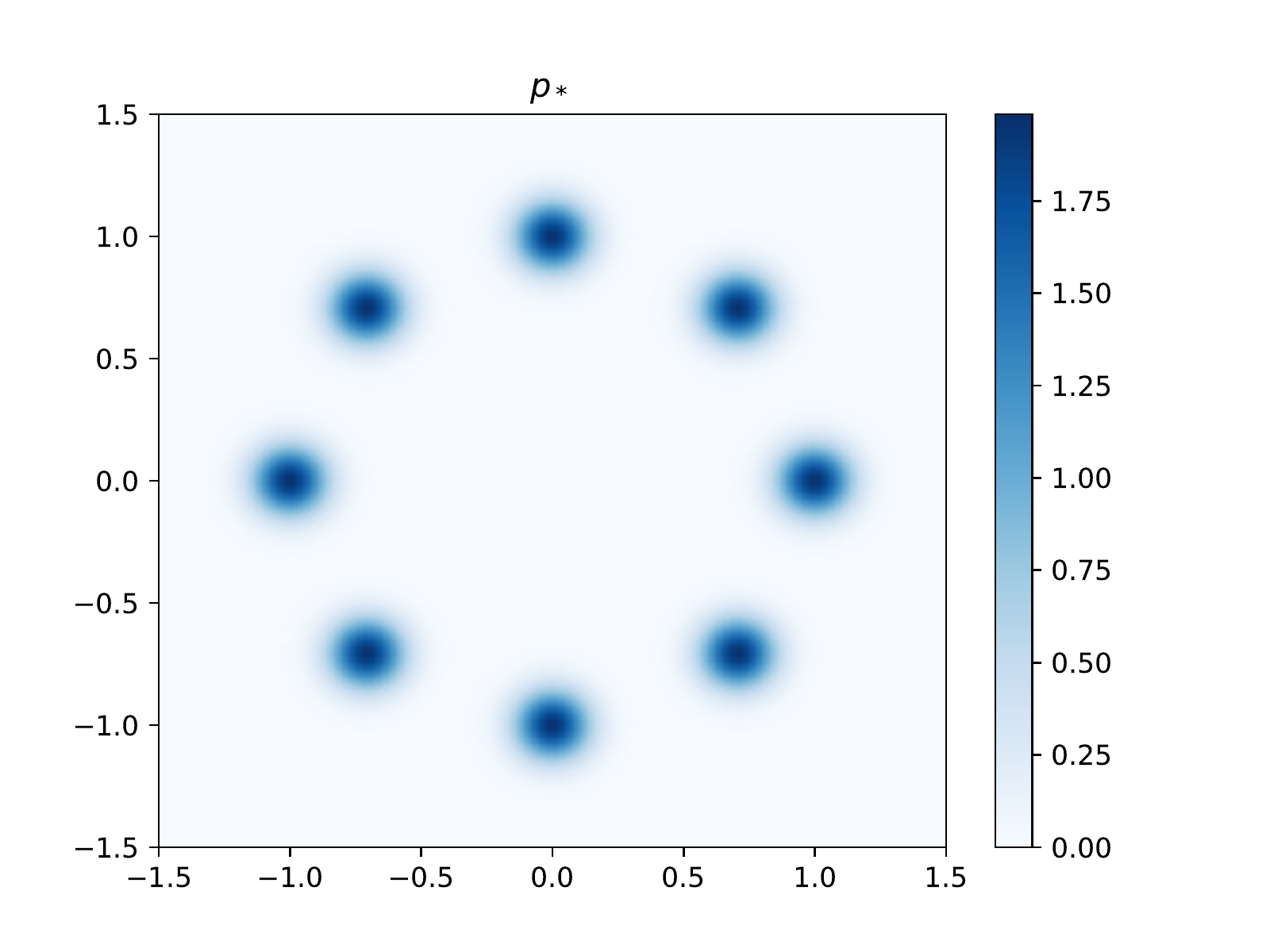}
		\caption{$p_*$}
		\label{fig: 2d setup p star}
	\end{subfigure}
	\begin{subfigure}[b]{0.15\textwidth}
		\centering 
		\includegraphics[trim=25 20 105 20, clip, width=0.99\textwidth]{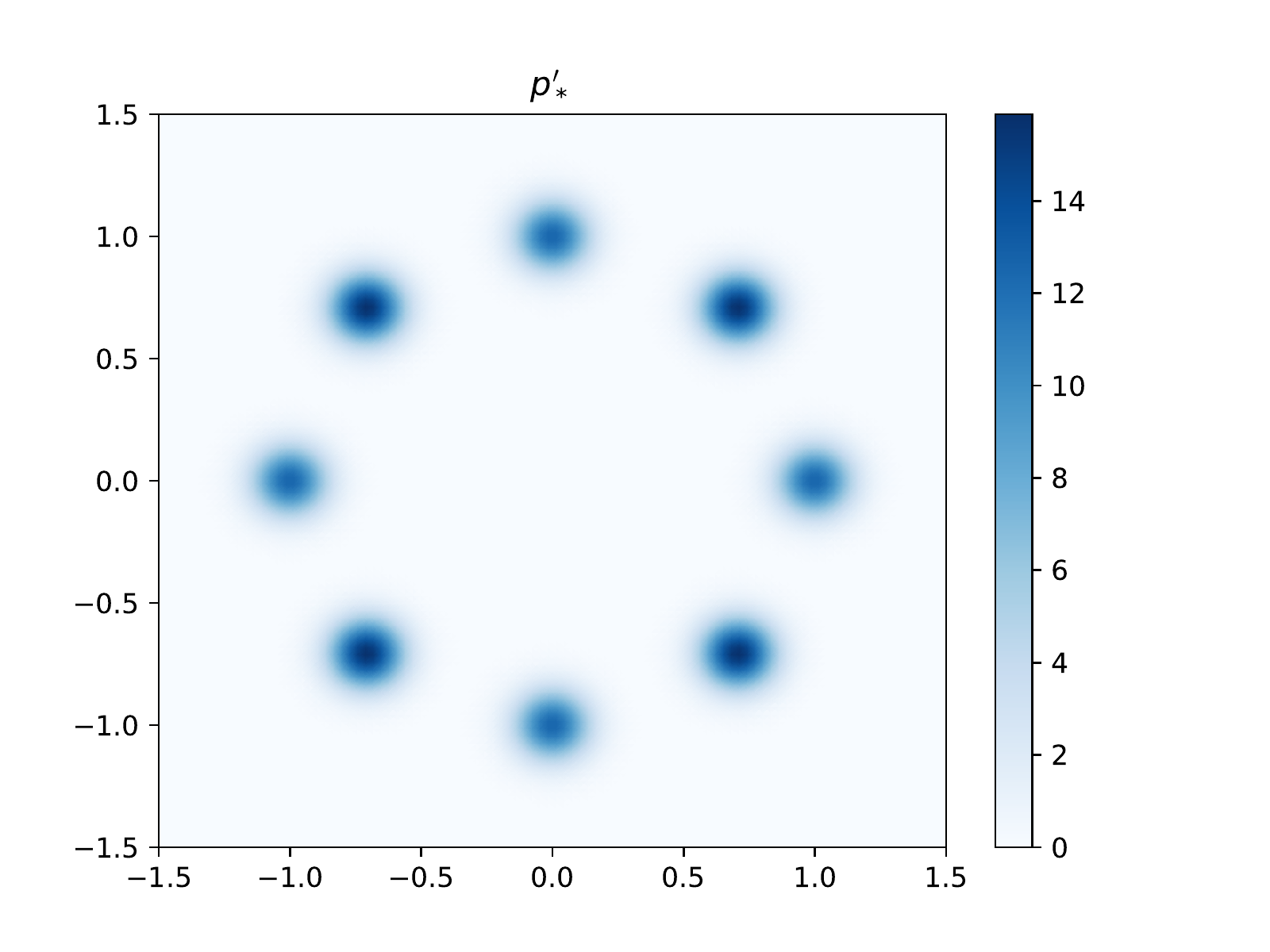}
		\caption{$p_*'$}
		\label{fig: 2d setup p star prime}
	\end{subfigure}
	\begin{subfigure}[b]{0.15\textwidth}
		\centering 
		\includegraphics[trim=25 20 105 20, clip, width=0.99\textwidth]{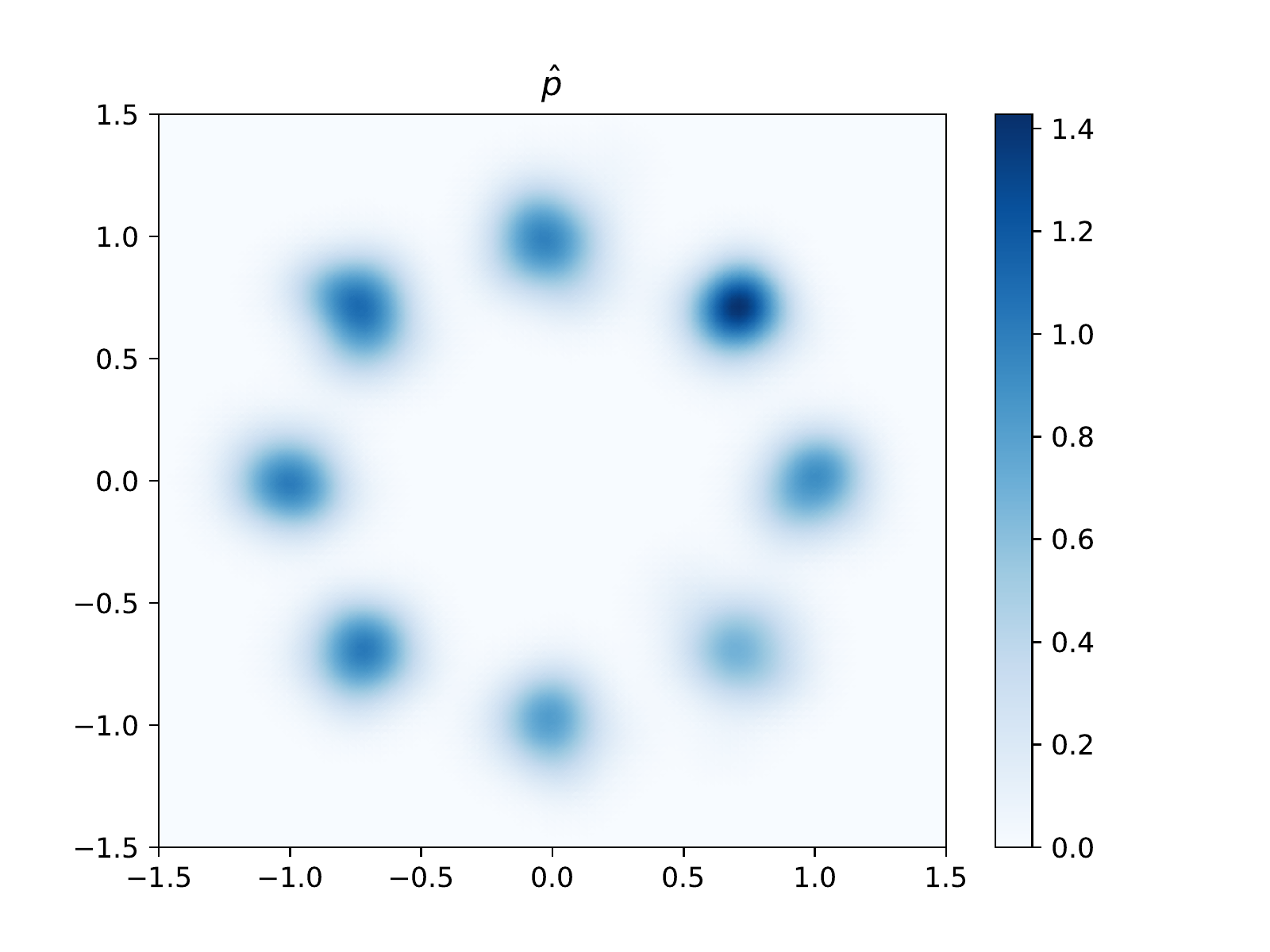}
		\caption{$\hat{p}$}
		\label{fig: 2d setup pre-trained}
	\end{subfigure}
	\begin{subfigure}[b]{0.15\textwidth}
		\centering 
		\includegraphics[trim=25 20 105 20, clip, width=0.99\textwidth]{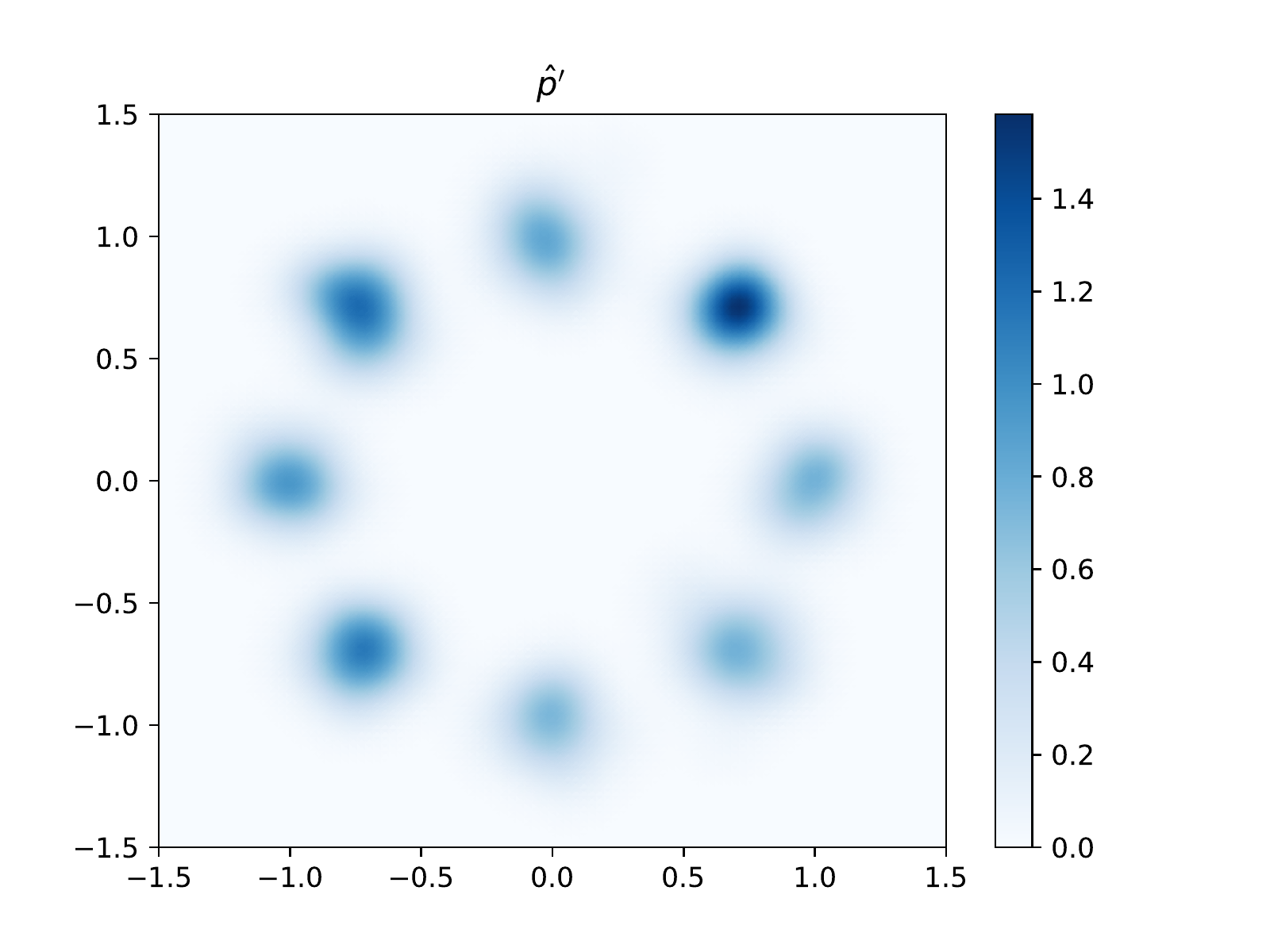}
		\caption{$\hat{p}'$}
		\label{fig: 2d setup re-trained}
	\end{subfigure}
	\begin{subfigure}[b]{0.175\textwidth}
		\centering 
		\includegraphics[trim=25 20 50 20, clip, width=0.99\textwidth]{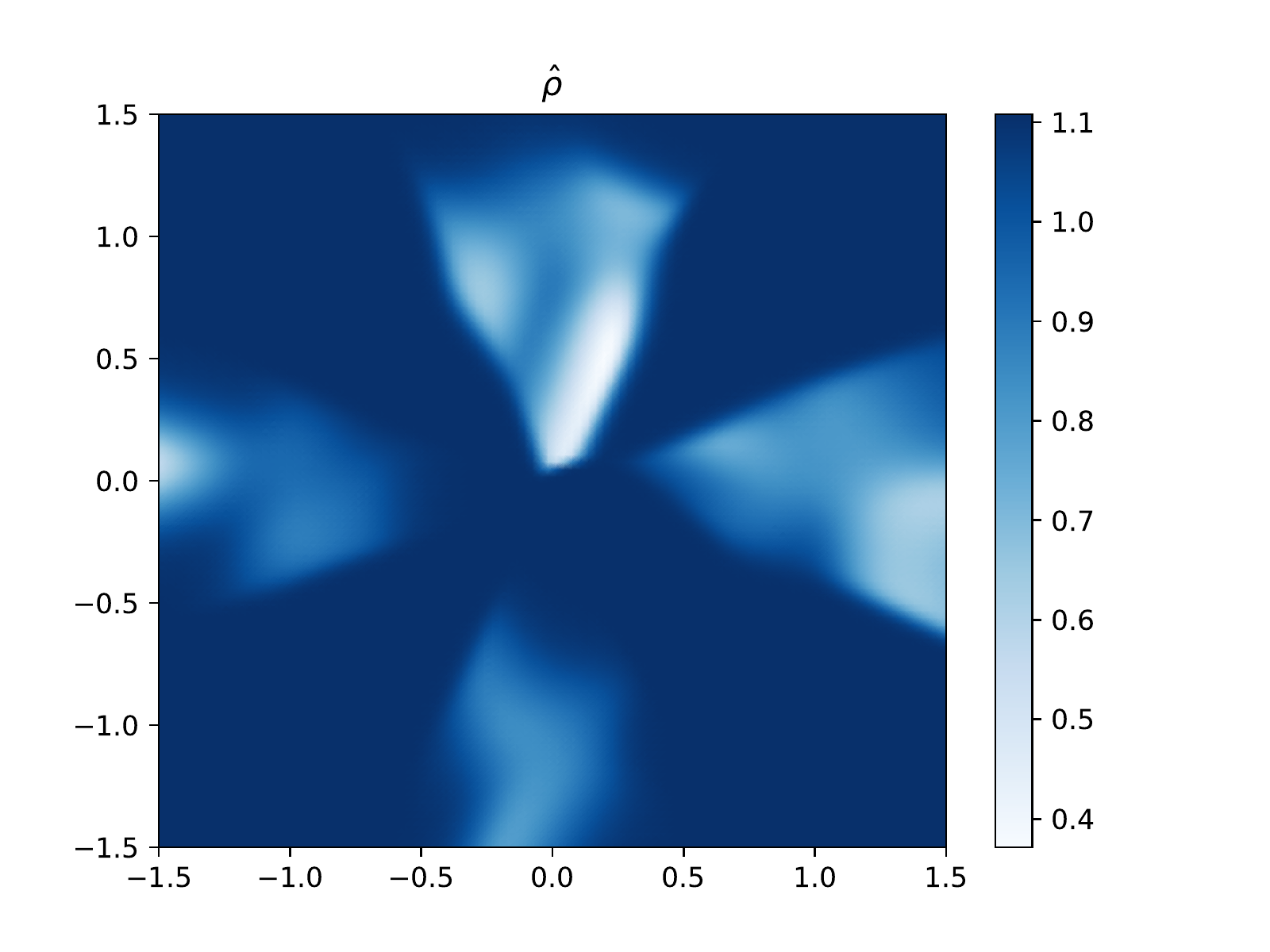}
		\caption{$\hat{\rho}$}
		\label{fig: 2d setup gt rho}
	\end{subfigure}
	\begin{subfigure}[b]{0.18\textwidth}
		\centering 
		\includegraphics[width=0.99\textwidth]{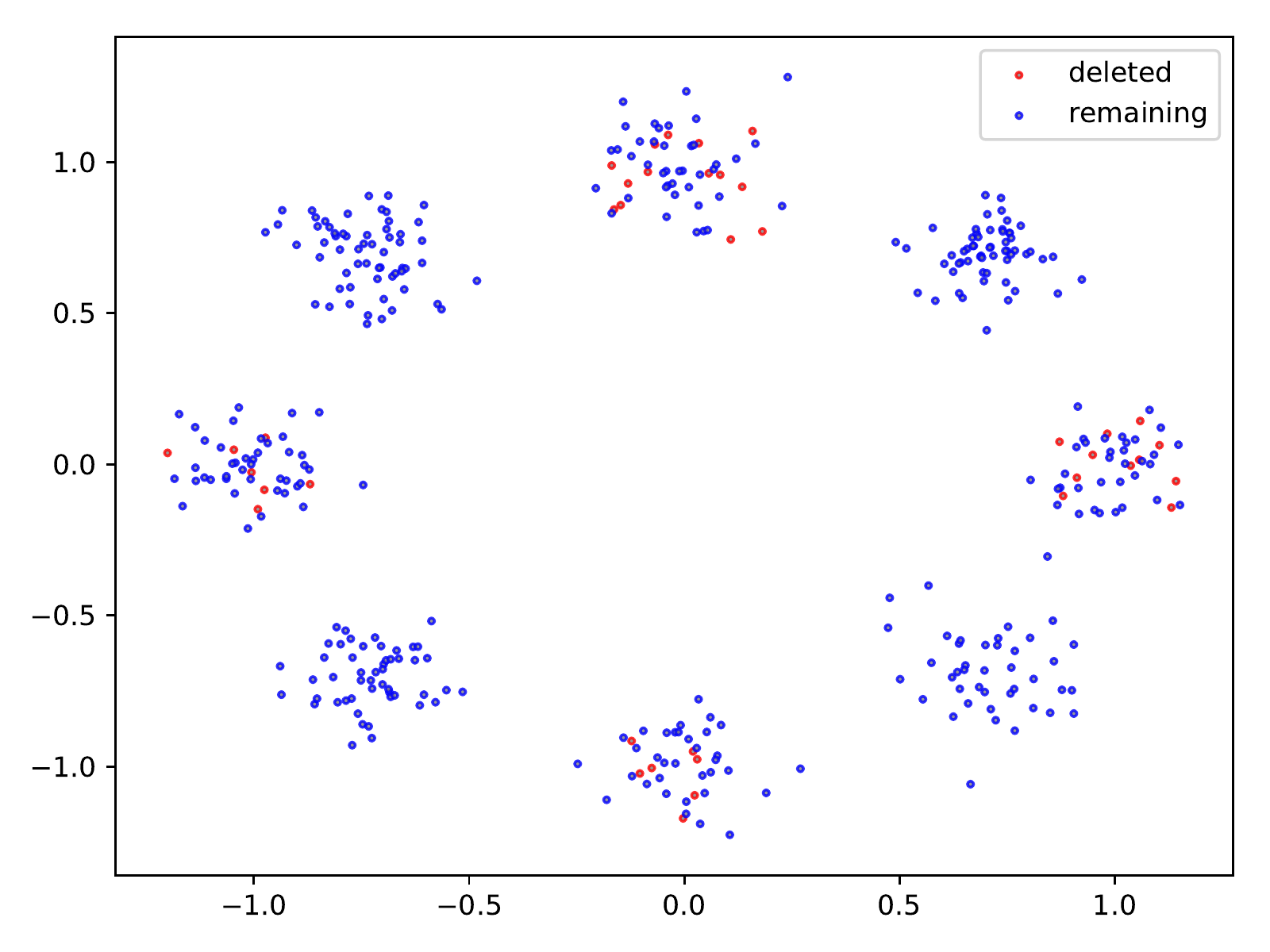}
		\caption{\textcolor{red}{$X'$} and \textcolor{blue}{$X\setminus X'$}}
		\label{fig: 2d setup samples}
	\end{subfigure}
	
	\vspace{-0.3em}
	\caption{Visualization of the experimental setup of MoG-8. (a) Data distribution $p_*$. (b) Distribution $p_*'$ with $\lambda=0.8$. (c) Pre-trained KDE $\hat{p}$ on $X$ with $\sigma_{\mA}=0.1$. (d) Re-trained KDE $\hat{p}'$ on $X\setminus X'$ with $\sigma_{\mA}=0.1$. (e) Density ratio $\hat{\rho}=\hat{p}'/\hat{p}$. (f) Deletion set $X'$ and the remaining set $X\setminus X'$.}
	\label{fig: 2d setup MoG-8}
	\vspace{-0.3em}
\end{figure}

\textbf{Experiment setup.} We generate two synthetic distributions ($p_*$) over $\mathbb{R}^2$ based on mixtures: a mixture of 8 Gaussian distributions (MoG-8) (Fig. \ref{fig: 2d setup p star}), and a checkerboard distribution with 8 squares on a $4\times4$ checkerboard (CKB-8) (Appendix \ref{appendix: exp 2d CKB}). We define $p_*'$ to be a weighted mixture version of $p_*$: 4 re-weighted clusters have weight $=\lambda\in(0,1)$ (for MoG-8, they are the clusters at 3, 6, 9, and 12 o'clock), and the other 4 have weight $=1$ (see Fig. \ref{fig: 2d setup p star prime}).
We draw $N=400$ samples from $p_*$ to form $X$, and randomly reject $1-\lambda$ fraction of samples in re-weighted clusters to form the deletion set $X'$ (see Fig. \ref{fig: 2d setup samples}).
We run KDE using a Gaussian kernel and $\sigma_{\mA}=0.1$ to obtain pre-trained models in Fig. \ref{fig: 2d setup pre-trained}, re-trained models in Fig. \ref{fig: 2d setup re-trained}, and their ratio $\hat{\rho}$ in Fig. \ref{fig: 2d setup gt rho}.
We use KDE because its learned density can be written explicitly and thus we are able to compute the exact likelihood ratio and examine the effectiveness of our DRE-based framework.

\textbf{Method and results.} We use the classification-based DRE described in Section \ref{sec: DRE classification}. We duplicate $X\setminus X'$ when training the classifiers according to Example \ref{eg: KDE classifier}. We consider two types of non-parametric classifiers: kernel-based classifiers (KBC) defined in \eqref{eq: KDE classifier} with potentially different $\sigma=\sigma_{\mC}$, and $k$-nearest-neighbour classifiers ($k$NN) defined as the fraction of positive votes in $k$ nearest neighbours. \footnote{We use non-parametric classifiers because the learning algorithm is non-parametric.
In preliminary experimentation we found that parametric classifiers such as logistic regression are less effective.
We conjecture that this may be due to class label imbalance, but leave a further investigation as future work.}
For each classifier, we draw 4 sets of i.i.d. samples (each of size $m$): (1) $\hat{Y}\sim\hat{p}$ (pre-trained model), (2) $Y_{\mD}\sim\hat{p}\cdot\hat{\rho}_{\mE}$ (approximated model) marked in \colorApprox{light blue}, (3) $Y_{H_0}\sim (q$ under $H_0)=\hat{p}$ marked in \colorPre{orange}, and (4) $Y_{H_1}\sim (q$ under $H_1)=\hat{p}'$ marked in \colorRe{green}. We compute $\mathrm{LR}$ and $\hat{\mathrm{ASC}}$ statistics for each set and for both density ratios $\{\hat{\rho},\hat{\rho}_{\mE}\}$. The above procedure is repeated for $R=250$ times and we report empirical distributions of these statistics.

Here we focus on MoG-8.
Additional results for MoG-8 with other parameters are in Appendix \ref{appendix: exp 2d MoG}. Results for CKB-8 are qualitatively similar to MoG-8 and are provided in Appendix \ref{appendix: exp 2d CKB}.

\begin{figure}[!t]
\vspace{-0.3em}
  	\begin{subfigure}[t!]{0.19\textwidth}
		\centering 
		\includegraphics[trim=25 20 50 37, clip, width=\textwidth]{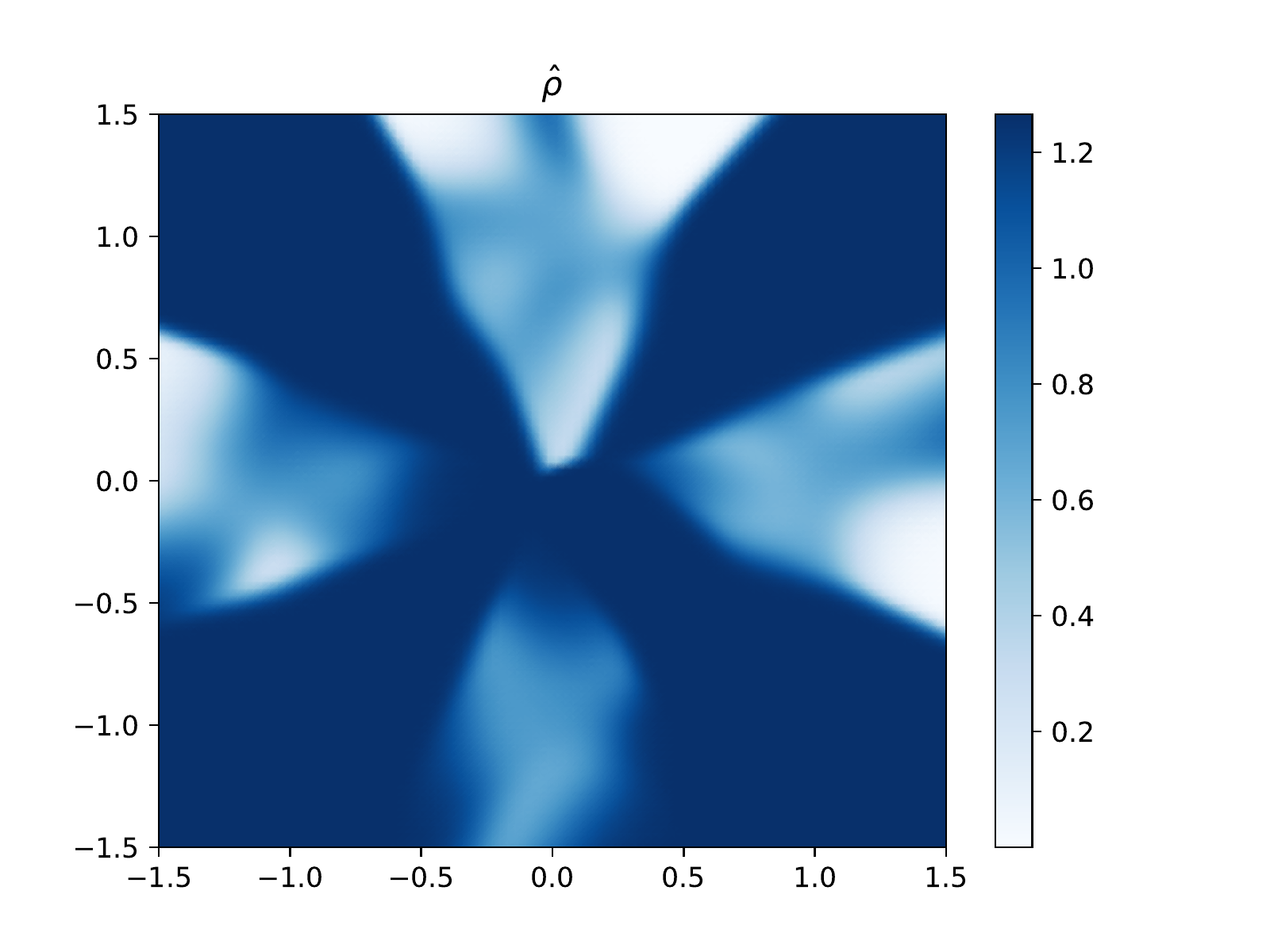}
		\caption{$\hat{\rho}$}
	\end{subfigure}
	\begin{subfigure}[t!]{0.19\textwidth}
		\centering 
		\includegraphics[trim=25 20 50 37, clip, width=\textwidth]{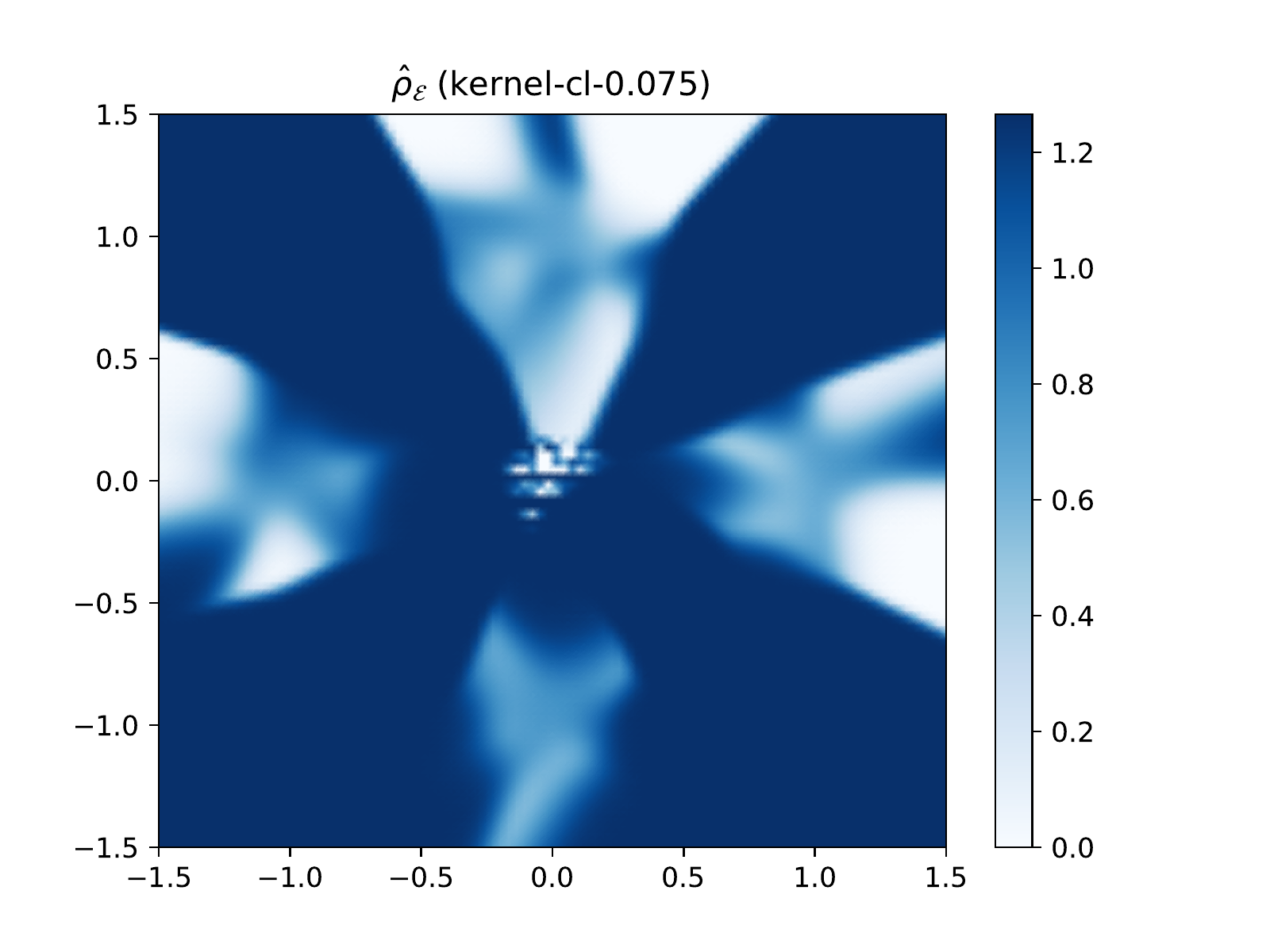}
		\caption{KBC ($\sigma_{\mC}$=$0.075$)}
	\end{subfigure}
	\begin{subfigure}[t!]{0.19\textwidth}
		\centering 
		\includegraphics[trim=25 20 50 37, clip, width=\textwidth]{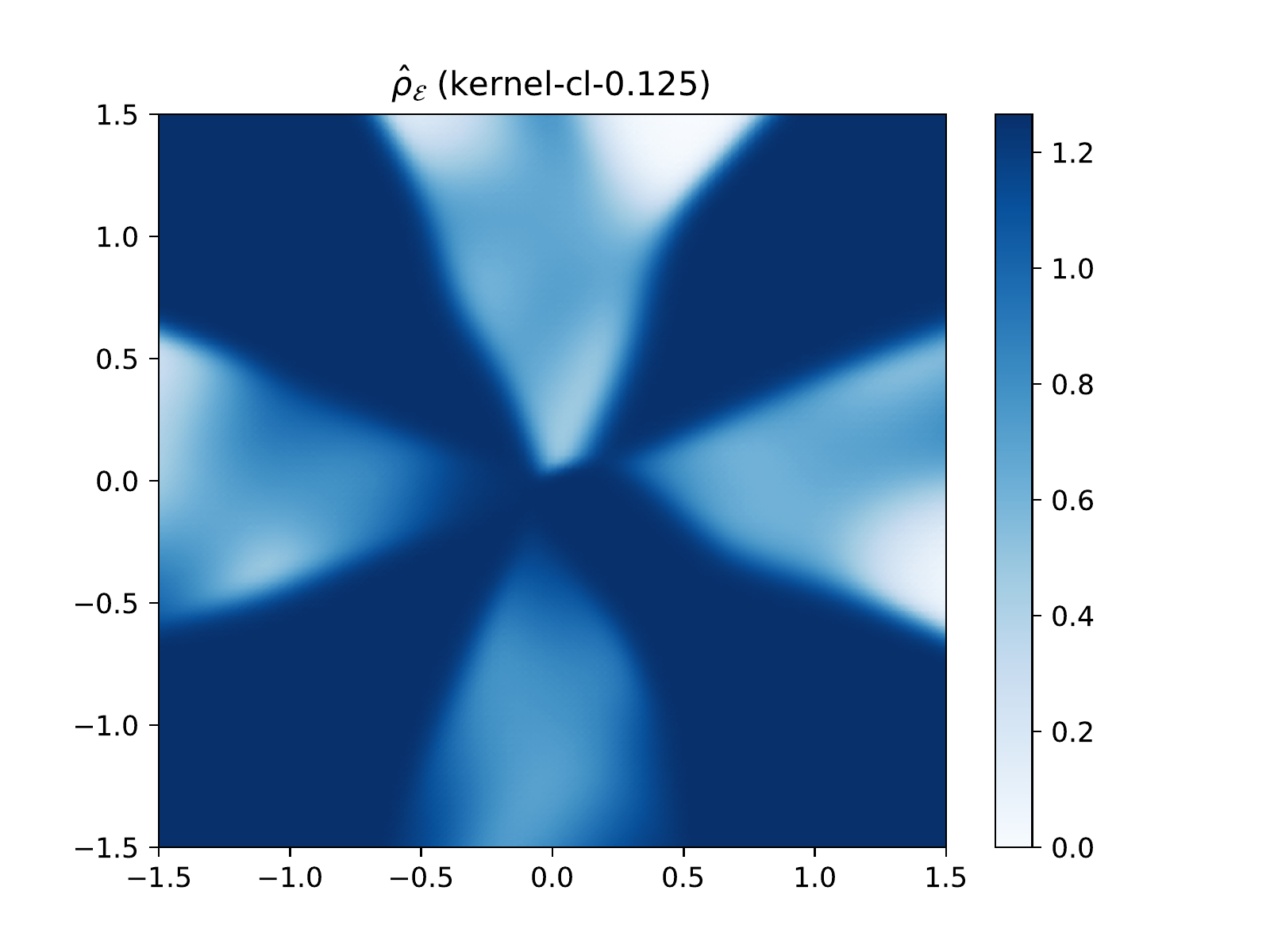}
		\caption{KBC ($\sigma_{\mC}$=$0.125$)}
	\end{subfigure}
	\begin{subfigure}[t!]{0.19\textwidth}
		\centering 
		\includegraphics[trim=25 20 50 37, clip, width=\textwidth]{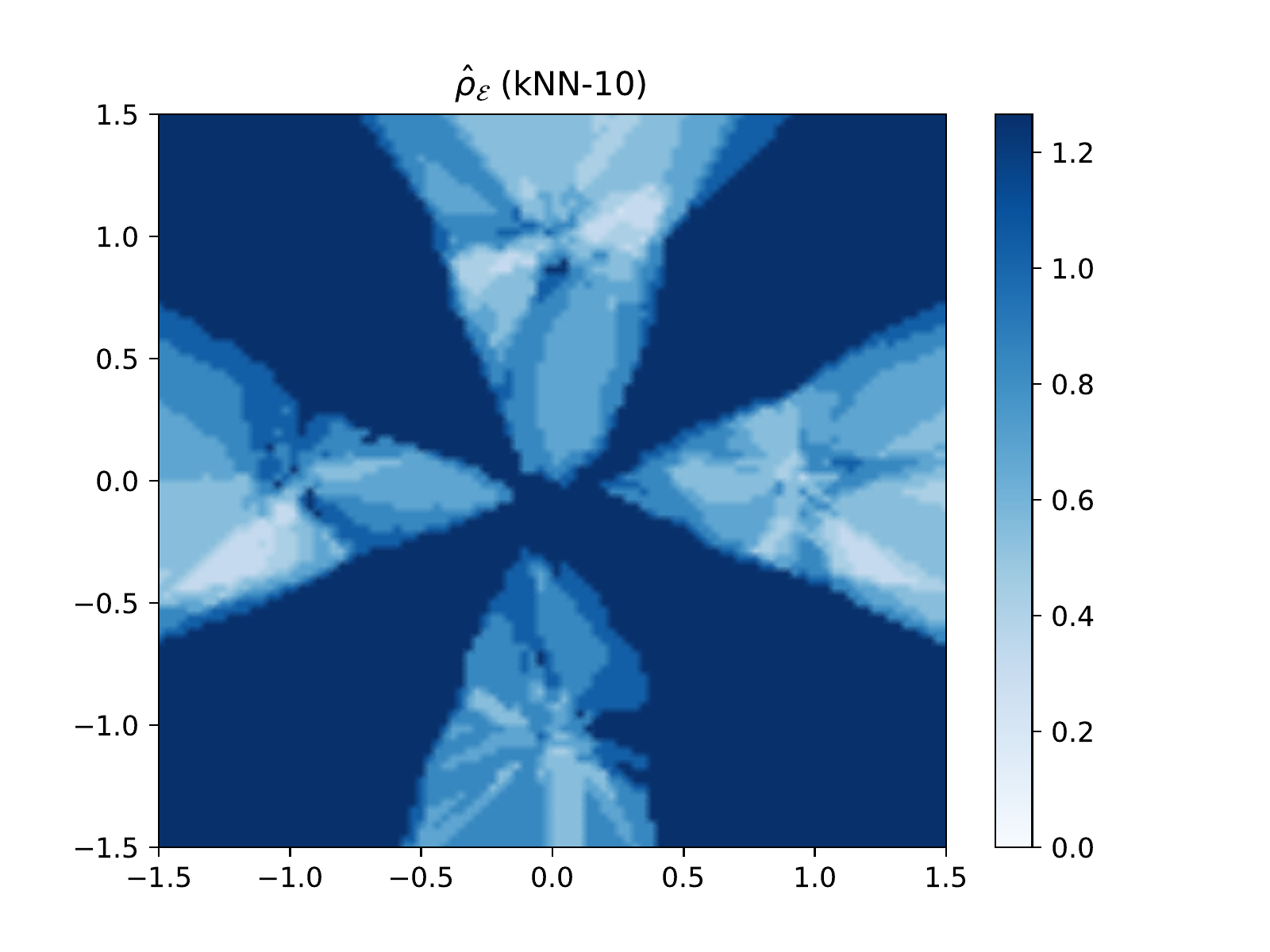}
		\caption{$k$-NN ($k$=$10$)}
	\end{subfigure}
	\begin{subfigure}[t!]{0.19\textwidth}
		\centering 
		\includegraphics[trim=25 20 50 37, clip, width=\textwidth]{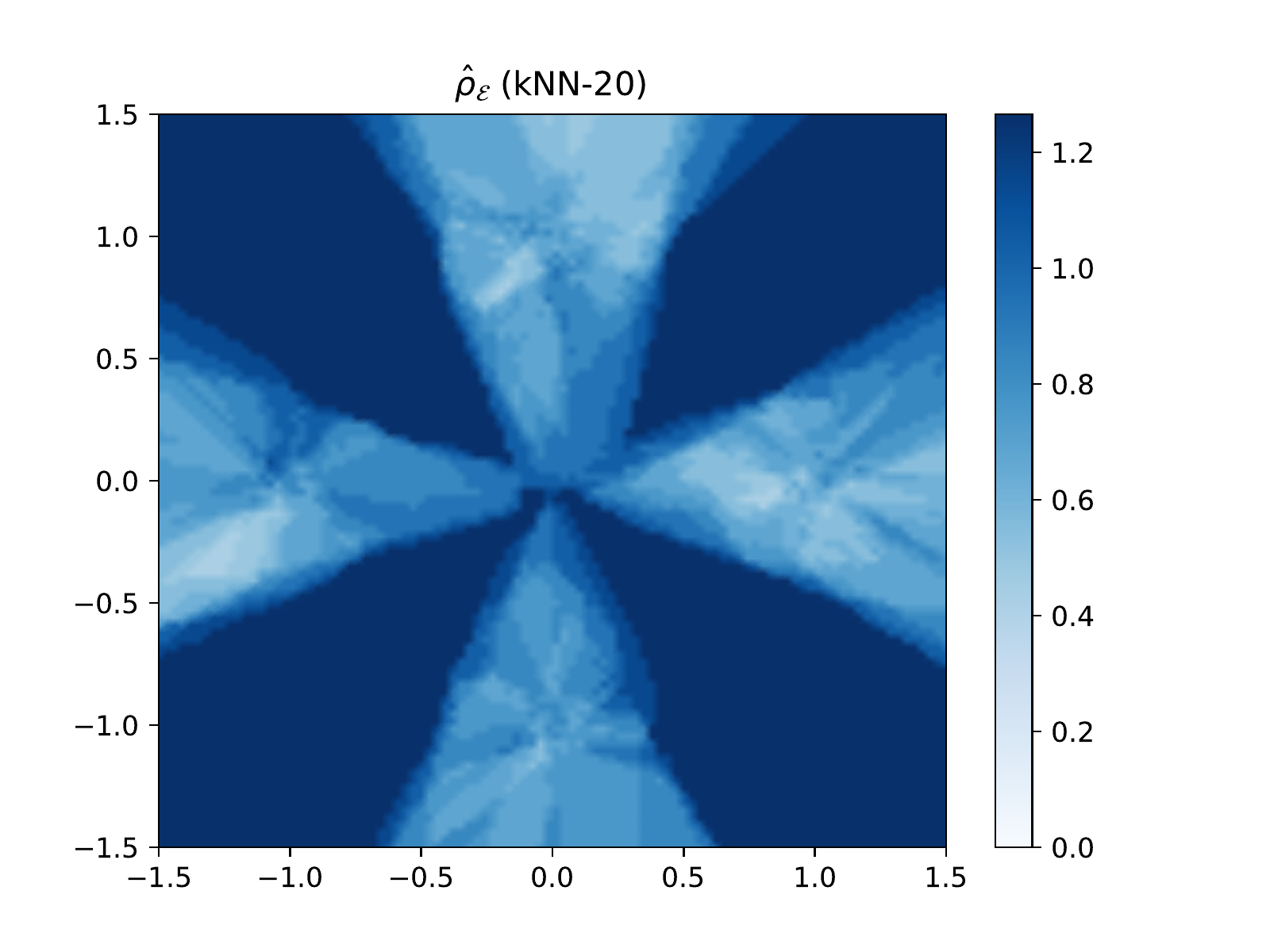}
		\caption{$k$-NN ($k$=$20$)}
	\end{subfigure}
	
	\vspace{-0.3em}
	\caption{Visualization of ratio $\hat{\rho}$ in (a) and $\hat{\rho}_{\mE}$ in (b)-(e) for different classifier-based DREs (MoG-8, $\lambda=0.6$, $\sigma_{\mA}=0.1$). These DREs are visually close to $\hat{\rho}$, thus qualitatively answering question 1.}
	\label{fig: 2d Q1 DRE}
	\vspace{-0.3em}
\end{figure}

\begin{figure}[!t]
\vspace{-0.3em}
  	\begin{subfigure}[t!]{0.3\textwidth}
	\centering 
	\includegraphics[trim=30 0 80 5, clip, height=1in]{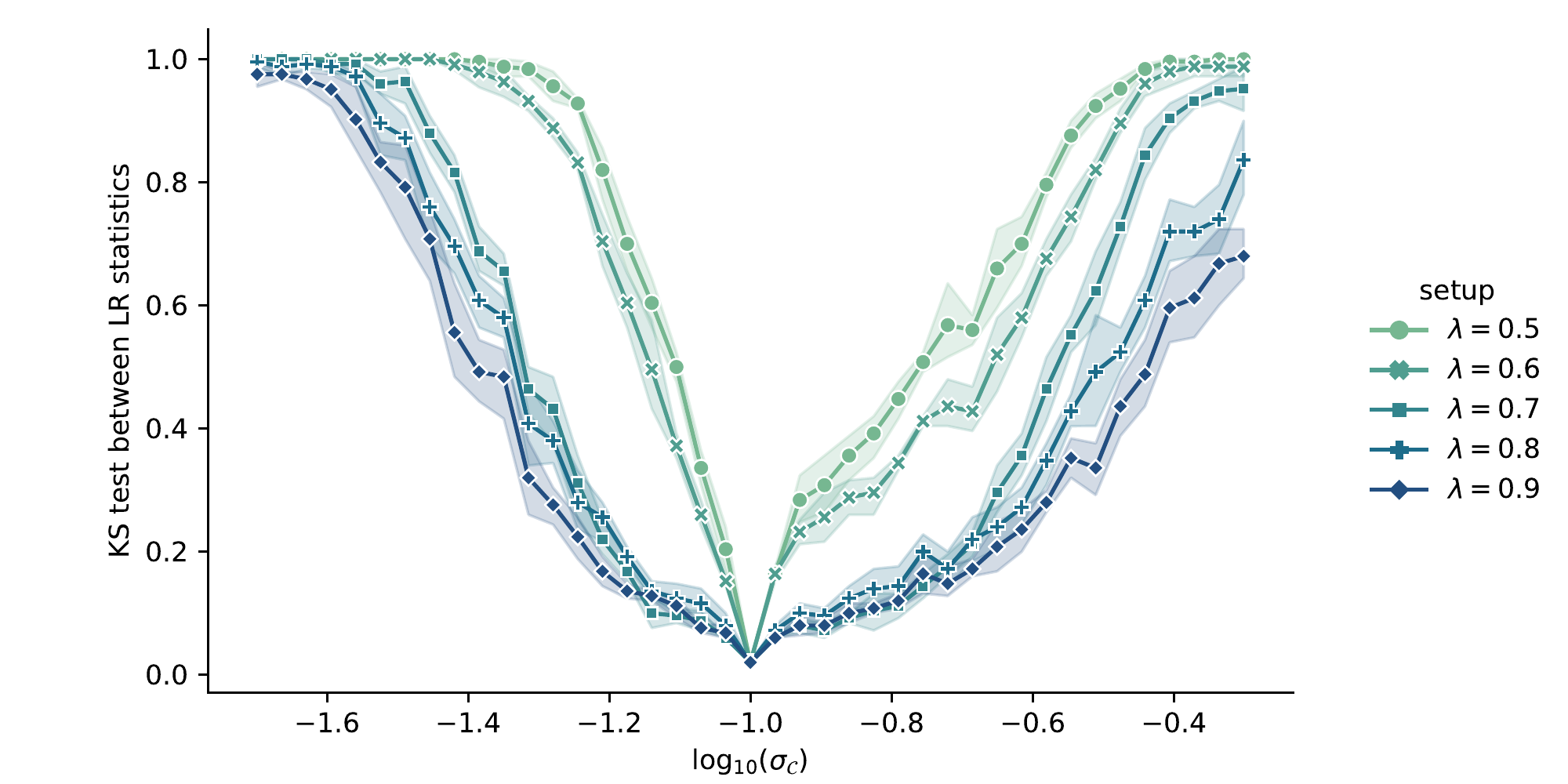}
	\caption{$\mathrm{LR}(Y_{H_0},\hat{\rho})$ vs $\mathrm{LR}(Y_{H_0},\hat{\rho}_{\mE})$}
	\label{fig: 2d Q1 KS}
	\end{subfigure}
	\begin{subfigure}[t!]{0.3\textwidth}
	\centering 
	\includegraphics[trim=30 0 80 5, clip, height=1in]{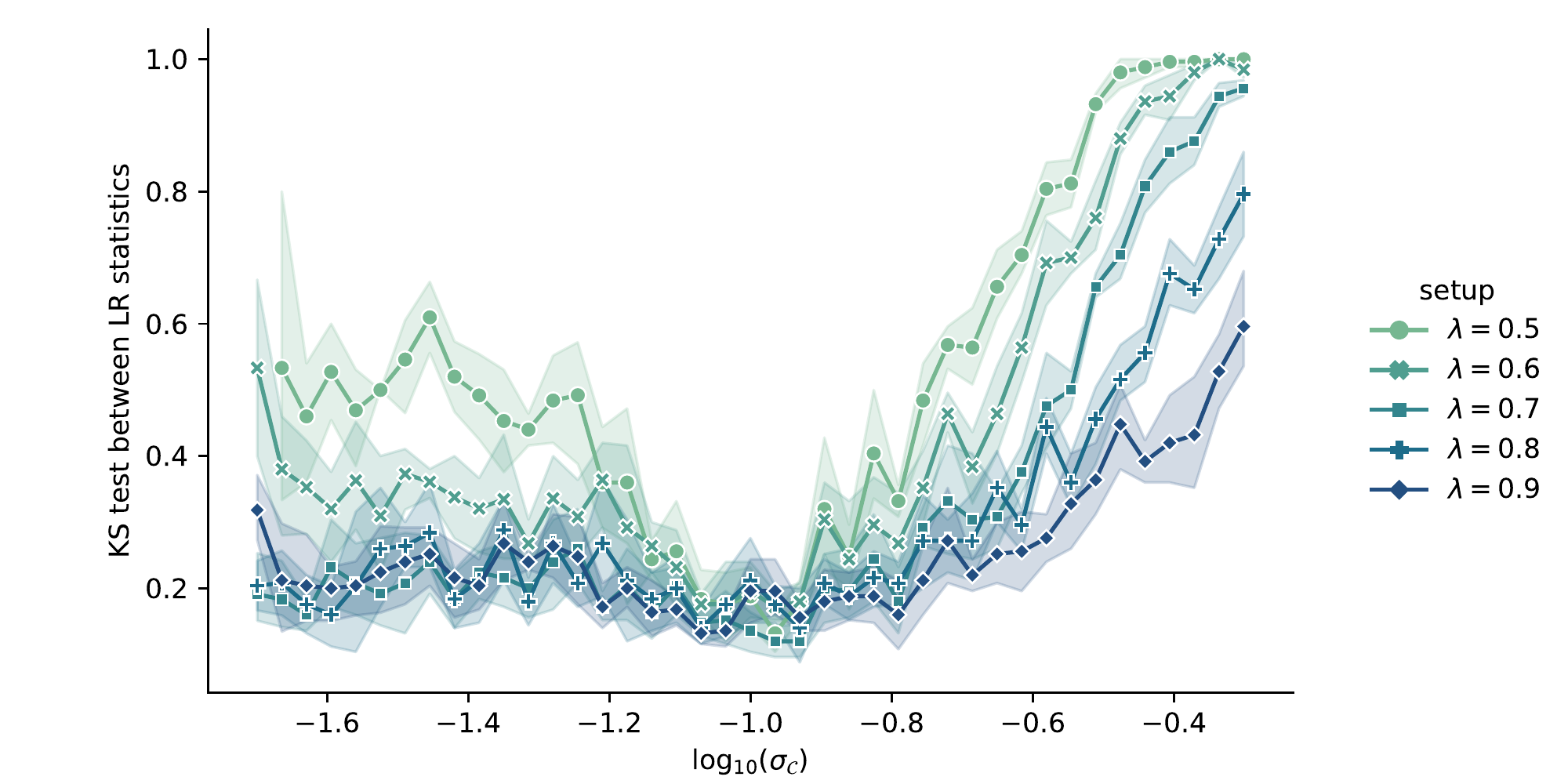}
	\caption{$\mathrm{LR}(Y_{H_1},\hat{\rho})$ vs $\mathrm{LR}(Y_{\mD},\hat{\rho})$}
	\label{fig: 2d Q2 KS}
	\end{subfigure}
	\begin{subfigure}[t!]{0.4\textwidth}
	\centering 
	\includegraphics[trim=30 0 0 5, clip, height=1in]{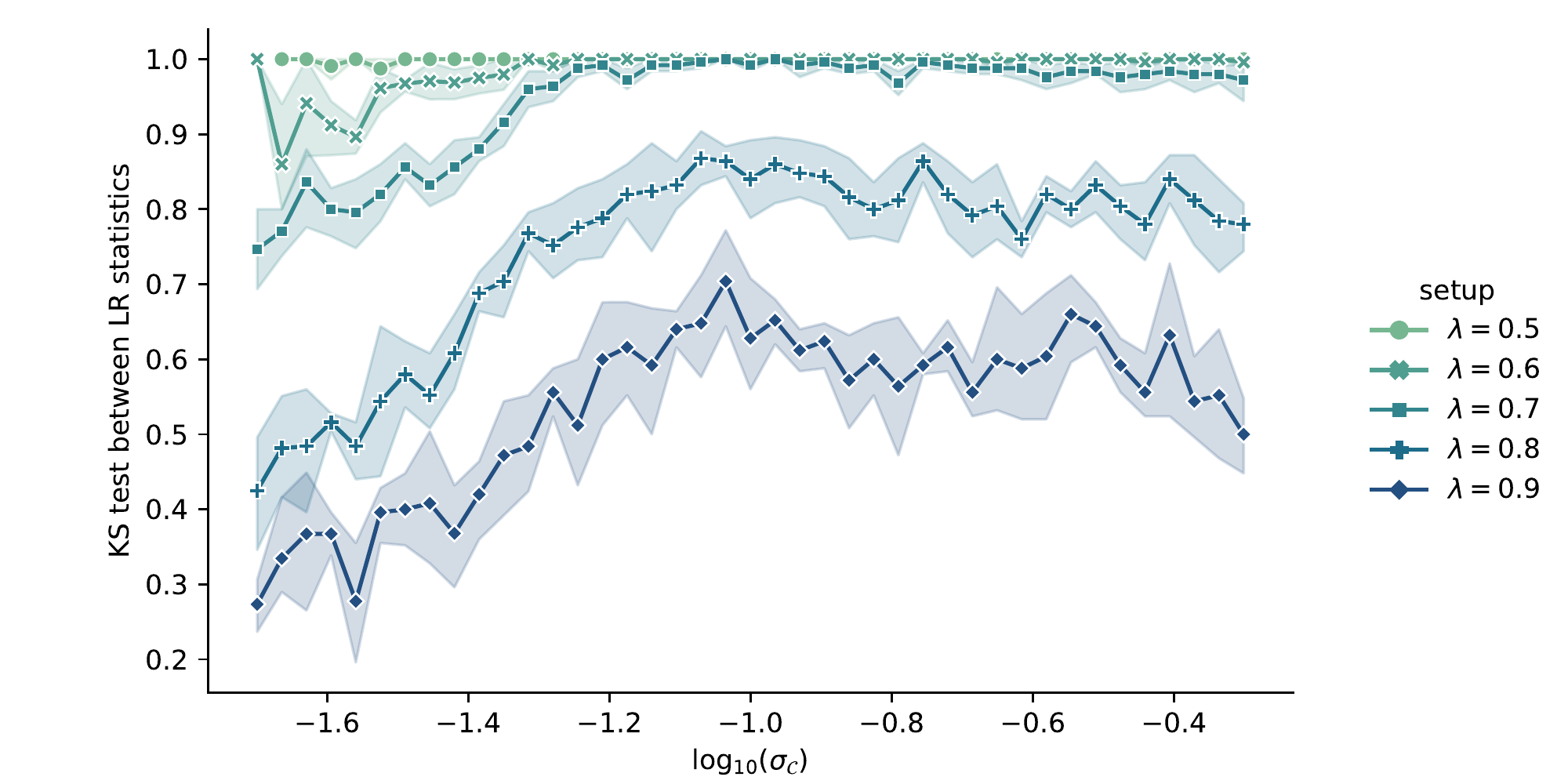}
	\caption{$\mathrm{LR}(Y_{H_0},\hat{\rho}_{\mE})$ vs $\mathrm{LR}(Y_{H_1},\hat{\rho}_{\mE})$}
	\label{fig: 2d Q3 KS}
	\end{subfigure}
	
	\vspace{-0.3em}
	\caption{KS tests between distributions of $\mathrm{LR}$ statistics for KBC with different $\sigma_{\mC}$. Smaller values indicate the two compared distributions are closer. Results for $\mathrm{ASC}$ statistics are in Appendix \ref{appendix: exp 2d} and are similar to $\mathrm{LR}$. When $\sigma_{\mC}\approx\sigma_{\mA}=0.1$, (a) answers question 1 (DRE Approximations)  by showing $\hat{\rho}_{\mE}\approx\hat{\rho}$ on the support of $\hat{p}$, (b) answers question 2 (Fast Deletion) by showing $Y_{H_1}$ (from $\hat{p}'$) and $Y_{\mD}$ (from the approximated model) cannot be distinguished by $\hat{\rho}$, and (c) answers question 3 (Hypothesis Test) by showing our DRE easily distinguishes $Y_{H_0}$ (from $\hat{p}$) and $Y_{H_1}$ (from $\hat{p}'$).}
	\vspace{-1.0em}
	\label{fig: 2d KS}
\end{figure}

We investigate \textbf{question 1} (DRE Approximations)  in two ways. First, we compare $\hat{\rho}_{\mE}$ and $\hat{\rho}$ in Fig. \ref{fig: 2d Q1 DRE}. We find that KBC with $\sigma_{\mC}$ close to $\sigma_{\mA}$  produces a very accurate approximation, and $k$NN with relatively small $k$ (e.g. $\leq20$)  performs similarly even though the estimated ratios are discrete.
We then conduct Kolmogorov–Smirnov (KS) tests between (1) the distributions of $\mathrm{LR}(Y_{H_0},\hat{\rho})$ versus $\mathrm{LR}(Y_{H_0},\hat{\rho}_{\mE})$, and (2) the distributions of $\hat{\mathrm{ASC}}_{\phi}(\hat{Y},Y_{H_0},\hat{\rho})$ versus $\hat{\mathrm{ASC}}_{\phi}(\hat{Y},Y_{H_0},\hat{\rho}_{\mE})$. If $\hat{\rho}\approx \hat{\rho}_{\mE}$ on the support of $\hat{p}$ then the KS statistics will be close to 0, meaning the two compared distributions are indistinguishable. In Fig. \ref{fig: 2d Q1 KS}, we plot KS statistics for KBC with different $\sigma_{\mC}$. The KS statistics are almost monotonically increasing with the difference between $\sigma_{\mC}$ and $\sigma_{\mA}=0.1$.
We further note that sometimes choosing a larger $\sigma_{\mC}$ gives a more accurate estimation than a smaller $\sigma_{\mC}$. We also find larger $\lambda$ (where less data are deleted) leads to better estimation, as expected. 

\begin{figure}[!t]
  	\begin{subfigure}[t!]{0.24\textwidth}
	\centering 
	\includegraphics[trim=0 0 0 0, clip, width=0.99\textwidth]{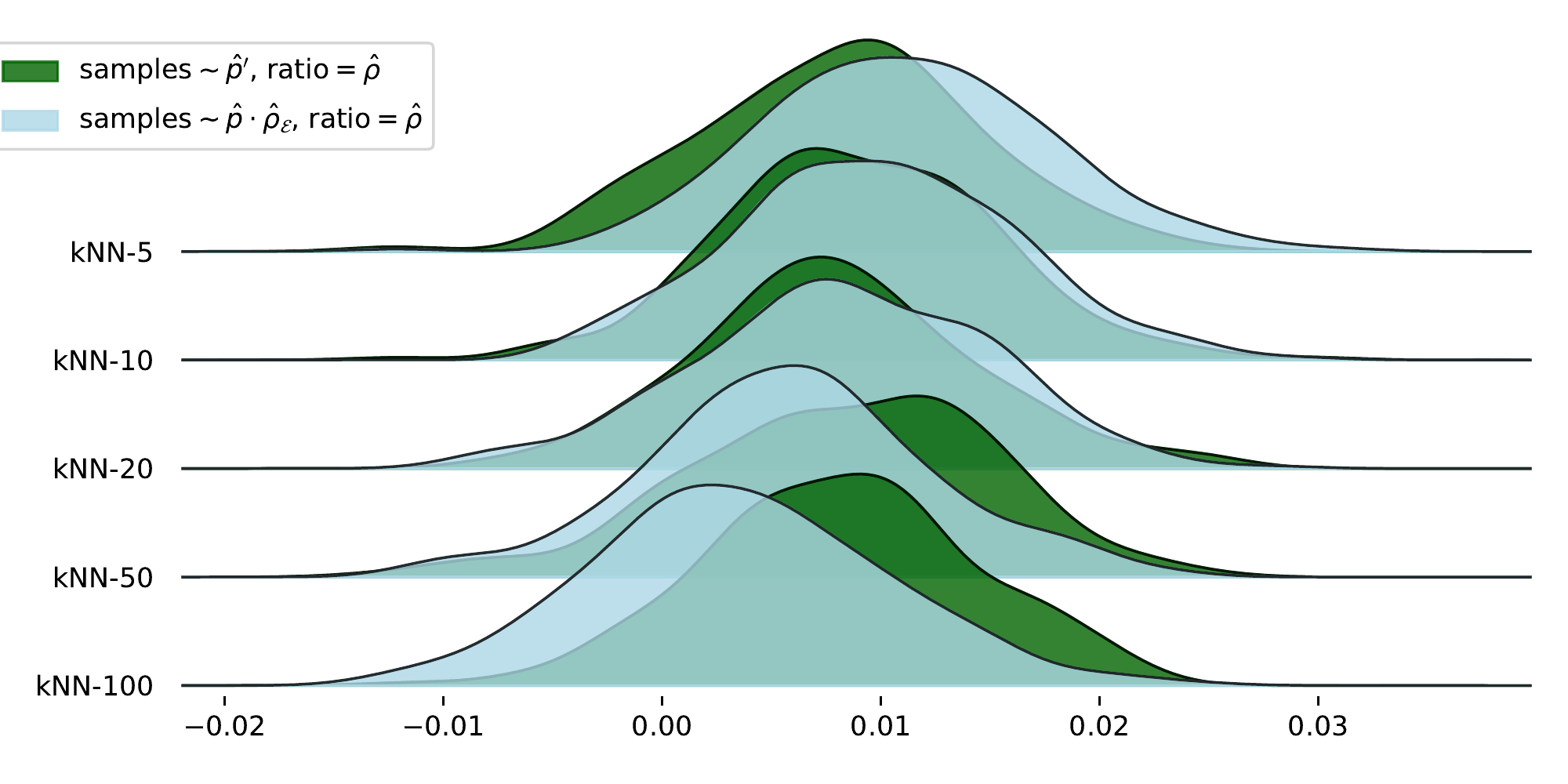}
	\caption{DRE based on $k$-NN}
	\end{subfigure}
	\begin{subfigure}[t!]{0.24\textwidth}
	\centering 
	\includegraphics[trim=0 0 0 0, clip, width=0.99\textwidth]{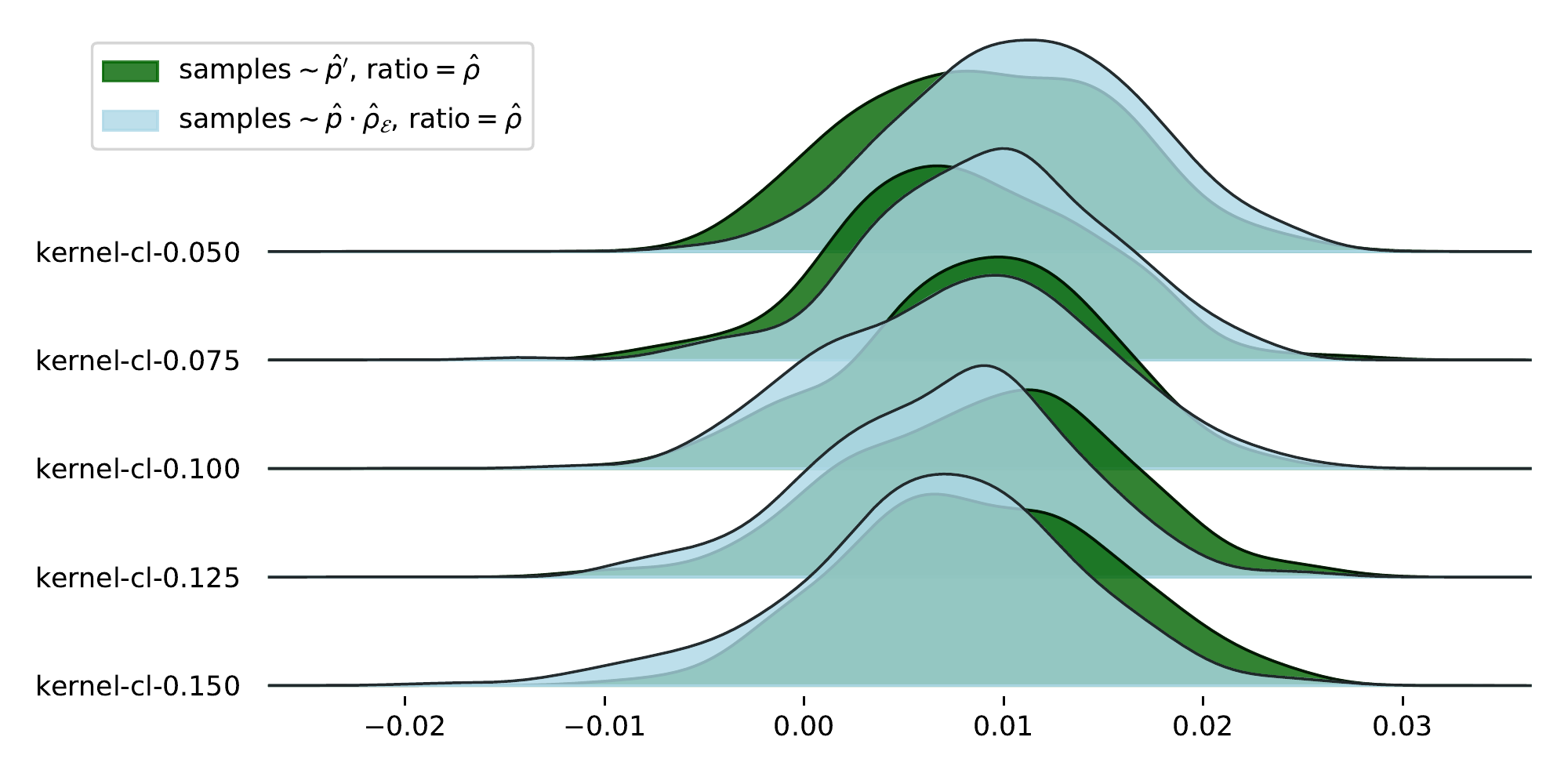}
	\caption{DRE based on KBC}
	\end{subfigure}
	\begin{subfigure}[t!]{0.24\textwidth}
	\centering 
	\includegraphics[trim=0 0 0 0, clip, width=0.99\textwidth]{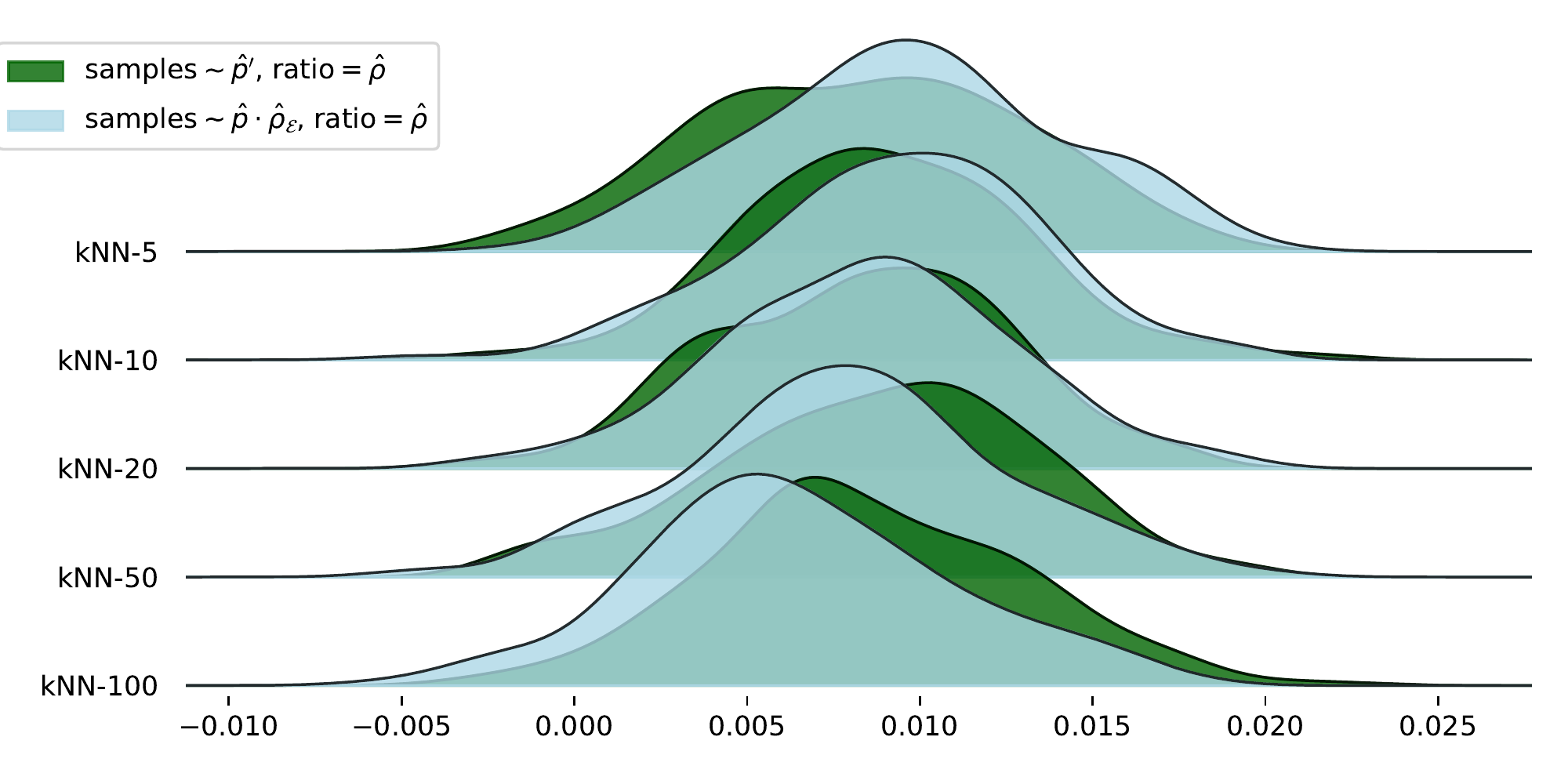}
	\caption{DRE based on $k$-NN}
	\end{subfigure}
	\begin{subfigure}[t!]{0.24\textwidth}
	\centering 
	\includegraphics[trim=0 0 0 0, clip, width=0.99\textwidth]{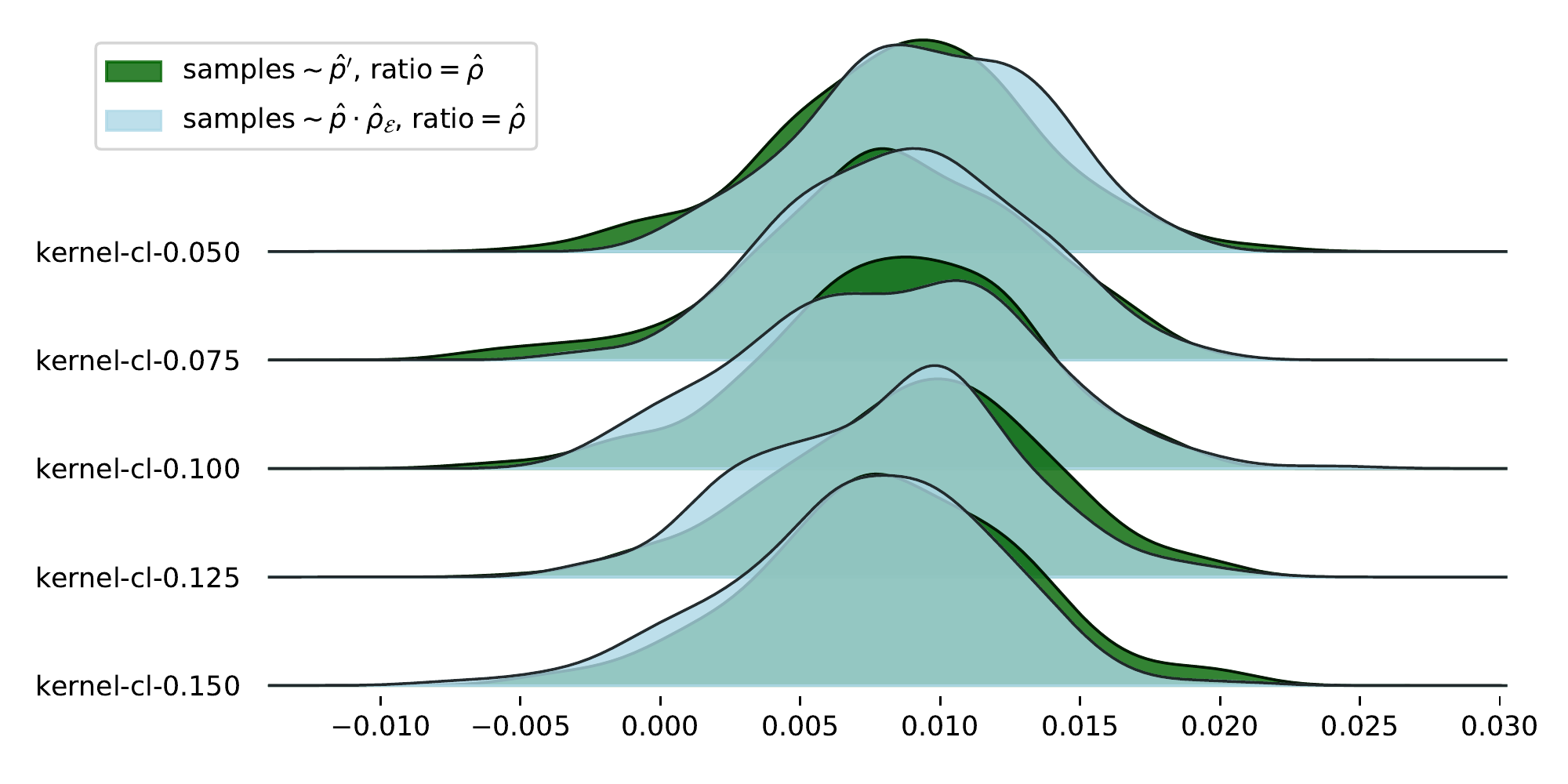}
	\caption{DRE based on KBC}
	\end{subfigure}
	
	\vspace{-0.3em}
	\caption{Distributions of $\mathrm{LR}$ and $\mathrm{ASC}$ statistics between samples $Y_{H_1}$ from $\hat{p}'$ and $Y_{\mD}$ from the approximated model for different classifier-based DRE (MoG-8, $\lambda=0.8$, $\sigma_{\mA}=0.1$). (a)-(b) $\mathrm{LR}(Y_{H_1},\hat{\rho})$ vs $\mathrm{LR}(Y_{\mD},\hat{\rho})$. (c)-(d) $\hat{\mathrm{ASC}}_{\phi}(\hat{Y},Y_{H_1},\hat{\rho})$ vs $\hat{\mathrm{ASC}}_{\phi}(\hat{Y},Y_{\mD},\hat{\rho})$ with $\phi(t)=t\log(t)$ (KL divergence). These distributions largely overlap with each other, indicating the approximated model cannot be distinguished from the re-trained model. This answers question 2 (Fast Deletion).}
	\vspace{-0.2em}
	\label{fig: 2d Q2 joy}
\end{figure}

We investigate \textbf{question 2} (Fast Deletion)  by asking whether the approximated model $\hat{\rho}_{\mE}\cdot\hat{p}$ and the re-trained model $\hat{p}'$ can be distinguished by the ground truth ratio $\hat{\rho}$. We do this by comparing (1) the distributions of $\mathrm{LR}(Y_{H_1},\hat{\rho})$ versus $\mathrm{LR}(Y_{\mD},\hat{\rho})$, and (2) the distributions of $\hat{\mathrm{ASC}}_{\phi}(\hat{Y},Y_{H_1},\hat{\rho})$ versus $\hat{\mathrm{ASC}}_{\phi}(\hat{Y},Y_{\mD},\hat{\rho})$. Qualitative comparisons are shown in Fig. \ref{fig: 2d Q2 joy}, and quantitative results (KS statistics for $\mathrm{LR}$) for KBC with different $\sigma_{\mC}$ are shown in Fig. \ref{fig: 2d Q2 KS}. We find for a wide range of classifiers, the approximate deletion cannot be distinguished from full re-training, indicating the classifier-based DRE is effective in this task. We also find they are less distinguishable when $\lambda$ is larger, as expected. 

\begin{figure}[!t]
  	\begin{subfigure}[t!]{0.24\textwidth}
	\centering 
	\includegraphics[trim=0 0 0 0, clip, width=0.99\textwidth]{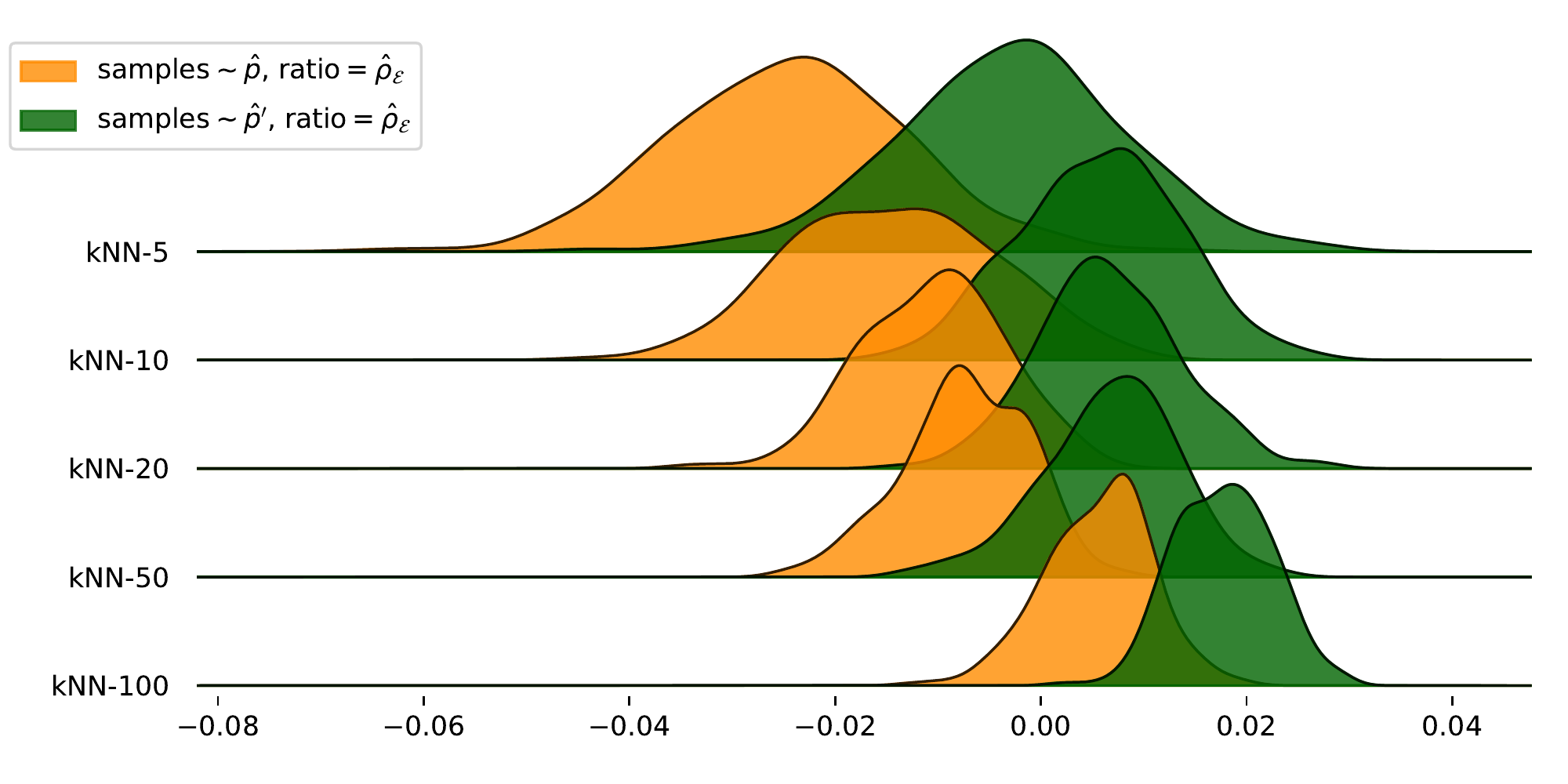}
	\caption{DRE based on $k$-NN}
	\end{subfigure}
	\begin{subfigure}[t!]{0.24\textwidth}
	\centering 
	\includegraphics[trim=0 0 0 0, clip, width=0.99\textwidth]{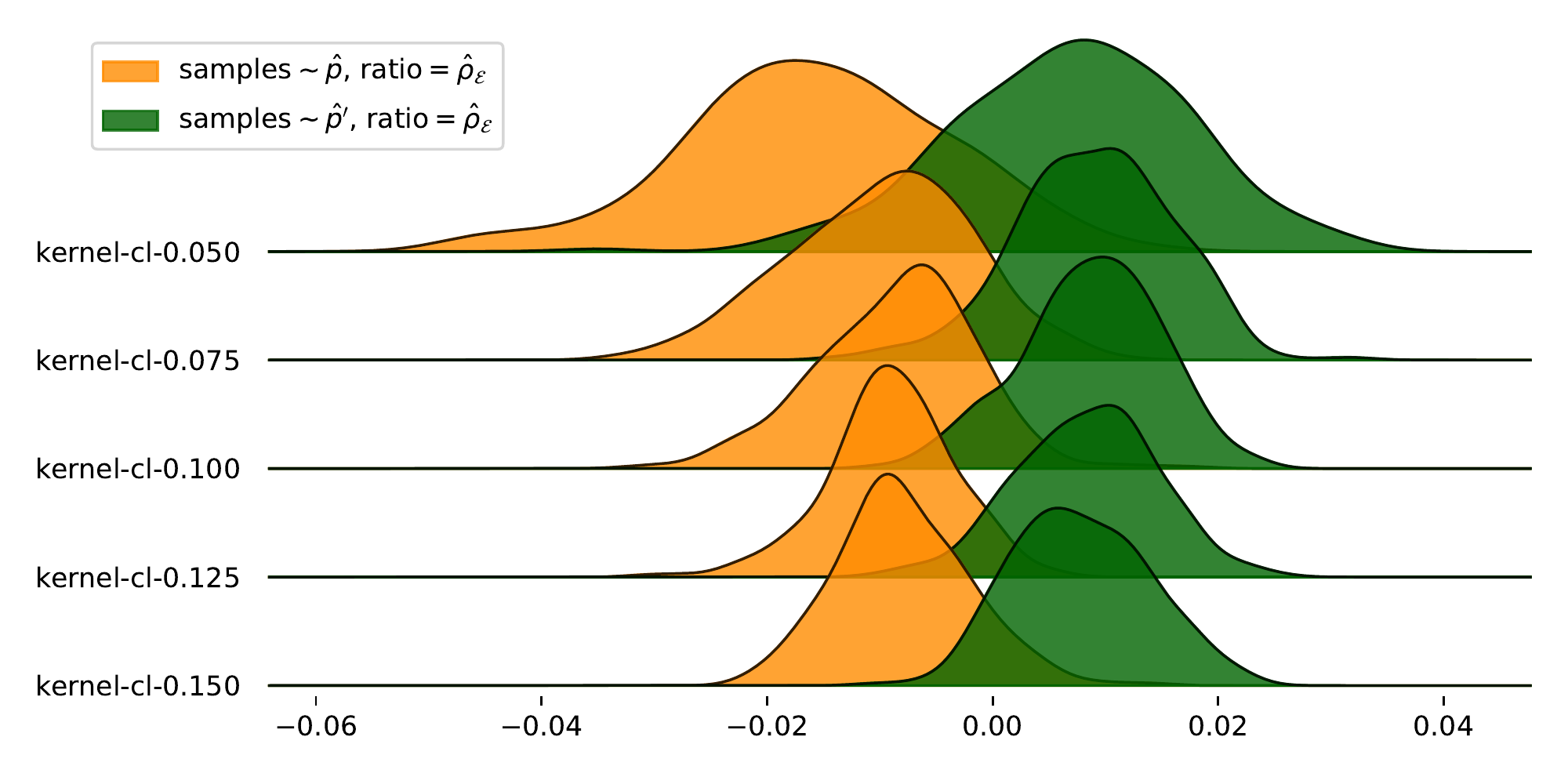}
	\caption{DRE based on KBC}
	\end{subfigure}
	\begin{subfigure}[t!]{0.24\textwidth}
	\centering 
	\includegraphics[trim=0 0 0 0, clip, width=0.99\textwidth]{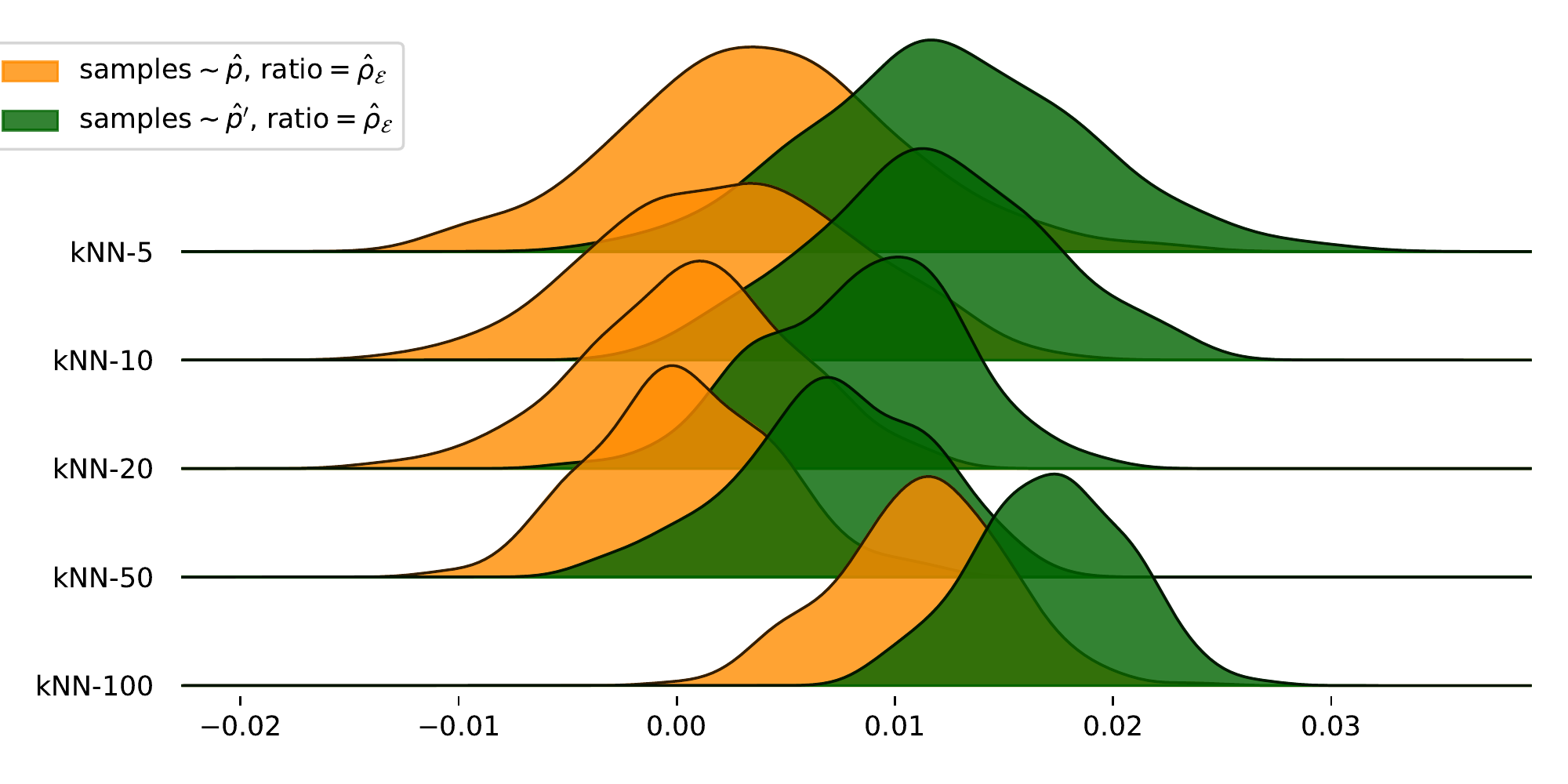}
	\caption{DRE based on $k$-NN}
	\end{subfigure}
	\begin{subfigure}[t!]{0.24\textwidth}
	\centering 
	\includegraphics[trim=0 0 0 0, clip, width=0.99\textwidth]{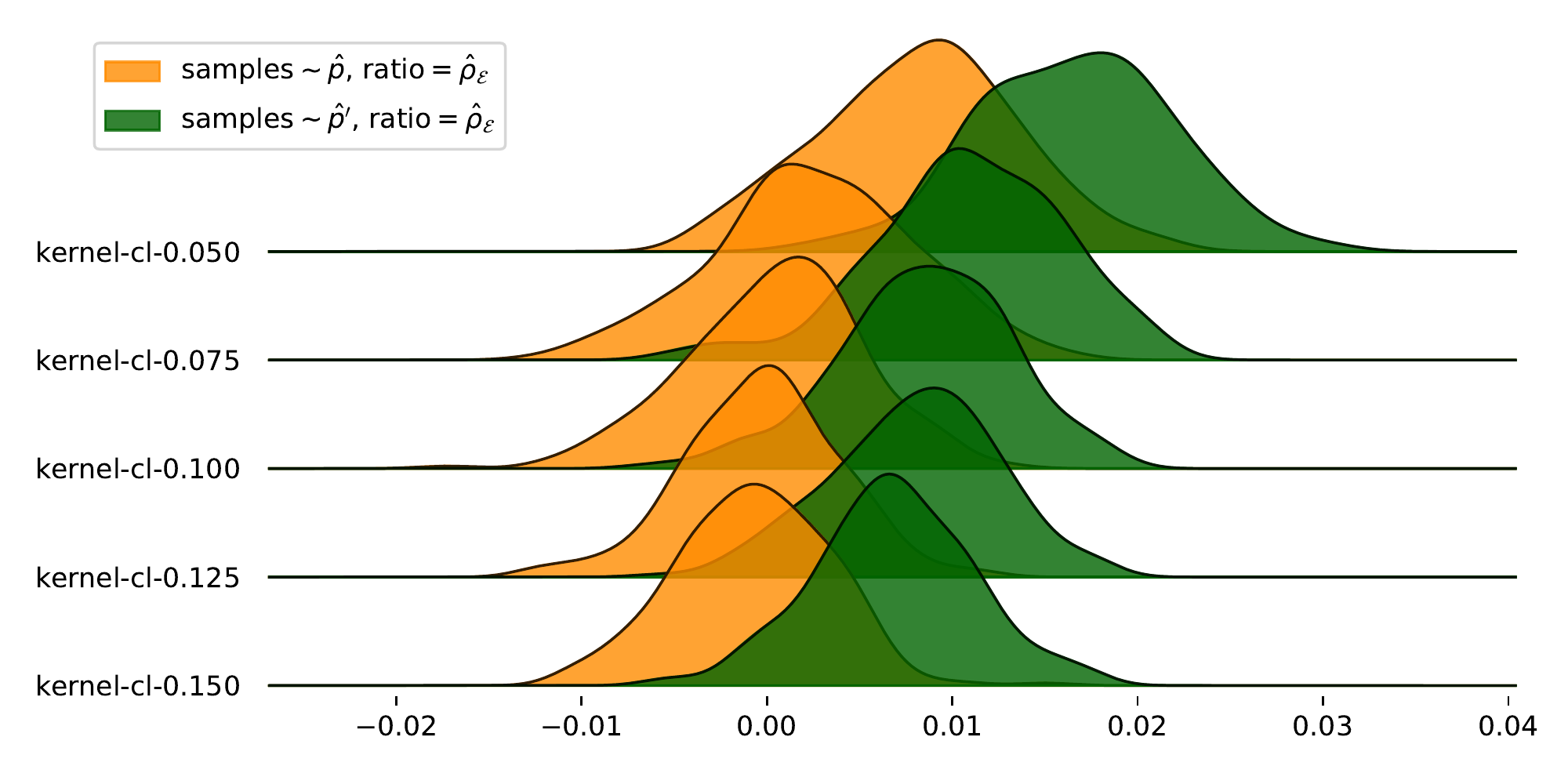}
	\caption{DRE based on KBC}
	\end{subfigure}
	
	\vspace{-0.3em}
	\caption{Distributions of $\mathrm{LR}$ and $\mathrm{ASC}$ statistics between samples $Y_{H_0}$ from $\hat{p}$ and $Y_{H_1}$ from $\hat{p}'$ for different classifier-based DRE (MoG-8, $\lambda=0.8$, $\sigma_{\mA}=0.1$). (a)-(b) $\mathrm{LR}(Y_{H_0},\hat{\rho}_{\mE})$ vs $\mathrm{LR}(Y_{H_1},\hat{\rho}_{\mE})$. (c)-(d) $\hat{\mathrm{ASC}}_{\phi}(\hat{Y},Y_{H_0},\hat{\rho}_{\mE})$ vs $\hat{\mathrm{ASC}}_{\phi}(\hat{Y},Y_{H_1},\hat{\rho}_{\mE})$ with $\phi(t)=t\log(t)$ (KL divergence). These distributions are separated from each other, indicating the DRE can distinguish between samples from pre-trained and re-trained models. This answers question 3 (Hypothesis Test).}
	\vspace{-1em}
	\label{fig: 2d Q3 joy}
\end{figure}

Finally, we answer \textbf{question 3} (Hypothesis Test)  by comparing (1) the distributions of $\mathrm{LR}(Y_{H_0},\hat{\rho}_{\mE})$ versus $\mathrm{LR}(Y_{H_1},\hat{\rho}_{\mE})$, and (2) the distributions of $\hat{\mathrm{ASC}}_{\phi}(\hat{Y},Y_{H_0},\hat{\rho}_{\mE})$ versus $\hat{\mathrm{ASC}}_{\phi}(\hat{Y},Y_{H_1},\hat{\rho}_{\mE})$. Qualitative comparisons are shown in Fig. \ref{fig: 2d Q3 joy}, and quantitative results (KS statistics for $\mathrm{LR}$) for KBC with different $\sigma_{\mC}$ are shown in Fig. \ref{fig: 2d Q3 KS}. We find $\hat{\rho}_{\mE}$ can distinguish samples between pre-trained and re-trained models for a wide range of classifiers. The likelihood ratio is slightly better than ASC statistics. In terms of the size of the deletion set, larger $\lambda$ makes the two models less distinguishable. 

\subsection{VDM-based DRE for GAN}\label{sec: exp GAN}

\textbf{Experimental setup.} 
The pre-trained model is a DCGAN \citep{radford2015unsupervised} on the full MNIST and Fashion-MNIST. Then, we let $\texttt{even-}\lambda$ be the subset with all odd labels and a $\lambda$ fraction of even labels randomly selected from the training set (so the rest $1-\lambda$ fraction of even labels form the deletion set $X'$), and similar for $\texttt{odd-}\lambda$. We re-train eight GAN models for $\lambda\in\{0.6,0.7,0.8,0.9\}$.

\textbf{Method and results.} We train VDM-based DRE introduced in Section \ref{sec: DRE VDM}. We optimize the KL-based loss function and obtain DRE described in \eqref{eq: KL-GAN}.

We investigate \textbf{question 2} (Fast Deletion)  by comparing label distribution of $m=50$K generated samples from the re-trained model and the approximated model. We run with five random seeds and report mean and standard errors in the bar plots. Results for $\texttt{even-}0.7$ are shown in Fig. \ref{fig: GAN Q2 MNIST} and \ref{fig: GAN Q2 FMNIST} , and more results for other deletion sets can be found in Appendix \ref{appendix: exp GAN}. We find the approximated model generates less labels some data with these labels are deleted from the training set. 

We investigate \textbf{question 3} (Hypothesis Test) in the same way as Section \ref{sec: exp 2d}. We draw i.i.d. samples $\hat{Y},Y_{H_0}\sim\hat{p}$, and $Y_{H_1}\sim\hat{p}'$, each of size $m=1000$. We then compute $\mathrm{LR}$ and $\hat{\mathrm{ASC}}$ statistics for each set and for density ratio $\hat{\rho}_{\mE}$. This procedure is repeated for $R=100$ times. We compare distributions of $\mathrm{LR}(Y_{H_0},\hat{\rho}_{\mE})$ versus $\mathrm{LR}(Y_{H_1}, \hat{\rho}_{\mE})$ for MNIST in Fig. \ref{fig: GAN Q3 LR MNIST}. Comparison of $\hat{\mathrm{ASC}}_{\phi}(\hat{Y},Y_{H_0},\hat{\rho}_{\mE})$ versus $\hat{\mathrm{ASC}}_{\phi}(\hat{Y},Y_{H_1},\hat{\rho}_{\mE})$, and results for Fashion-MNIST are in Appendix \ref{appendix: exp GAN}. We find in most cases, $\hat{\rho}_{\mE}$ can clearly distinguish samples between pre-trained and re-trained models.

\begin{figure}[!t]
\vspace{-0.5em}
  	\begin{subfigure}[t!]{0.33\textwidth}
	\centering 
	\includegraphics[trim=40 10 60 30, clip, width=0.95\textwidth]{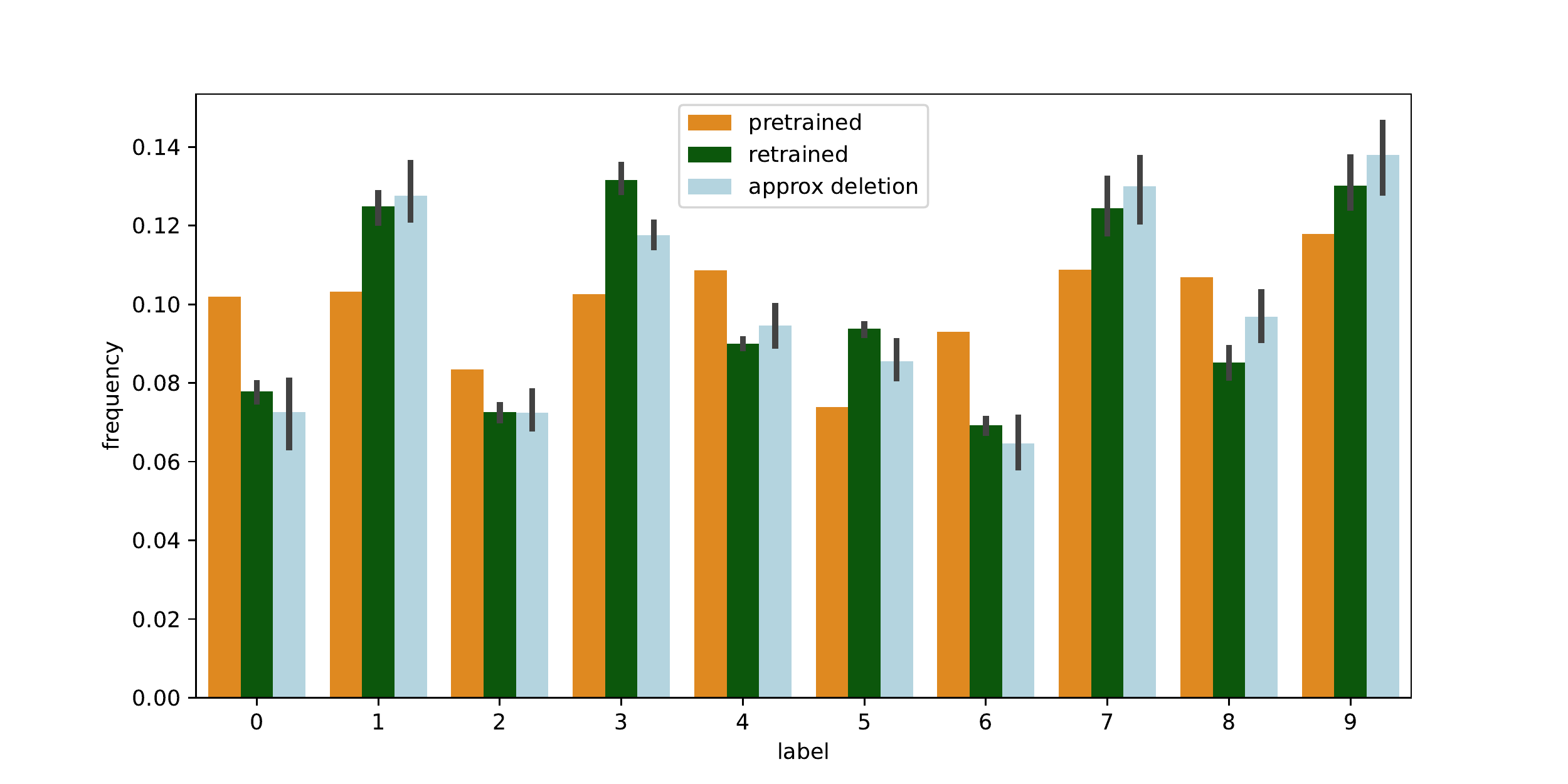}
	\caption{$\texttt{even-}0.7$ (MNIST)}
	\label{fig: GAN Q2 MNIST}
	\end{subfigure}
	\begin{subfigure}[t!]{0.33\textwidth}
	\centering 
	\includegraphics[trim=40 10 60 30, clip, width=0.95\textwidth]{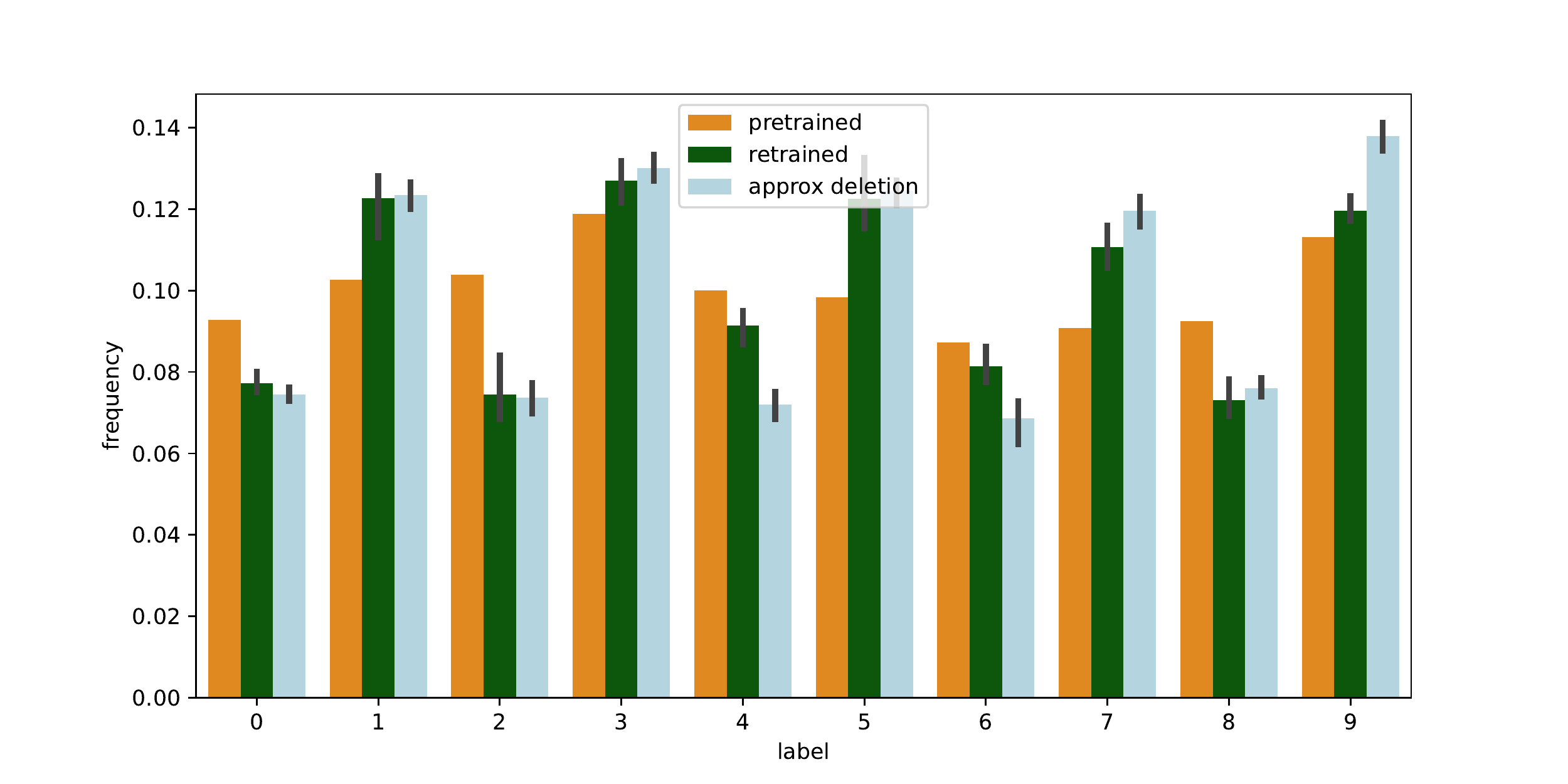}
	\caption{$\texttt{even-}0.7$ (FMNIST)}
	\label{fig: GAN Q2 FMNIST}
	\end{subfigure}
	\begin{subfigure}[t!]{0.33\textwidth}
	\centering 
	\includegraphics[trim=0 10 0 0, clip, width=0.95\textwidth]{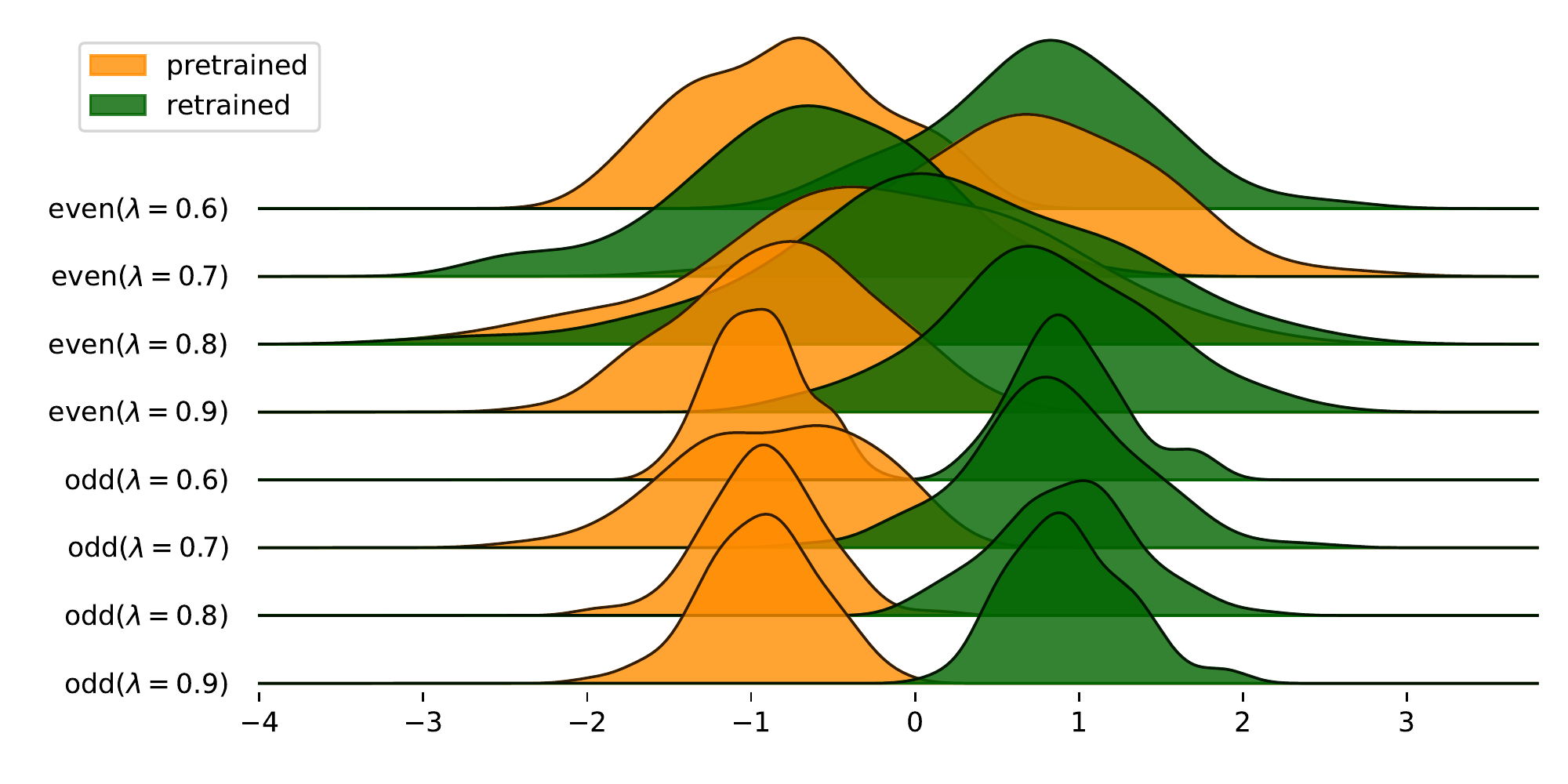}
	\caption{$\mathrm{LR}(Y_{H_0},\hat{\rho}_{\mE})$ vs $\mathrm{LR}(Y_{H_1},\hat{\rho}_{\mE})$}
	\label{fig: GAN Q3 LR MNIST}
	\end{subfigure}
	
	\caption{(a)-(b) Label distributions of 50K generated samples from the pre-trained model, re-trained model, and the approximated model. The distribution from the approximated model is close to the re-trained model, and much fewer even digits are generated by the approximated model. (c) Distributions of $\mathrm{LR}$ statistics between samples $Y_{H_0}$ from $\hat{p}$ and $Y_{H_1}$ from $\hat{p}'$. These distributions are separated from each other, indicating the DRE can distinguish between $Y_{H_0}$ and $Y_{H_1}$.}
	\vspace{-1.5em}
	\label{fig: exp GAN}
\end{figure}

\section{Related Work}\label{sec: related work}

Exact data deletion from learned models (where the altered model is identical to the re-trained model) was introduced as \textit{machine unlearning} \citep{cao2015towards}. Such deletion can be performed efficiently for relatively simpler learners such as linear regression \citep{chambers1971regression} and $k$-nearest neighbors \citep{schelter2020amnesia}. Machine unlearning for convex risk minimization was shown theoretically possible under total variation stability \citep{ullah2021machine}. There is a relaxed notion of machine unlearning where the altered model is statistically indistinguishable from the re-trained model \citep{ginart2019making}. Others have introduced further definitions of approximate data deletion \citep{guo2019certified, neel2021descent,sekhari2021remember,izzo2021approximate} and developed methods for performing approximate deletion efficiently for supervised learning algorithms.

The unsupervised setting has received substantially less attention with respect to data deletion. A notable exception is clustering: there are efficient deletion algorithms for $k$-clustering \citep{ginart2019making,borassi2020sliding}. Our work instead focuses on generative models where the goal is to learn distribution from data rather than doing clustering. Potential avenues for future work include forging a deeper connection between approximate data deletion for generative models and differential privacy \citep{dwork2006calibrating} and using recent advances in \textit{certified removal} \citep{guo2019certified} for generative models.

Outside of the context of data deletion, density ration estimation seeks to estimate the ratio between two densities from samples. For example, ratio can be estimated via probabilistic classification \citep{sugiyama2012density} or variational divergence minimization \citep{nowozin2016f}. There are also many other techniques in the literature \citep{yamada2011relative,sugiyama2012density,nowozin2016f,moustakides2019training,khan2019deep,rhodes2020telescoping,kato2021non,choi2021featurized,choi2022density}. All of these methods are designed for settings with little prior information about the data, and the two set of samples can potentially be very separated. In contrast, we consider a setting where we have strong prior information (namely that the only two possibilities are that \(X'\) was or was not deleted). We adapt probabilistic classification \citep{sugiyama2012density} and variational divergence minimization \citep{nowozin2016f} for our setting as they lend themselves naturally to incorporating the knowledge that training data is being deleted. 
An avenue of future work is incorporating such knowledge into other density ratio estimation methods, any of which can be used within our general framework in Fig. \ref{fig: framework}.

\section{Conclusions and Future Work}\label{sec: conclusion}

In this paper, we propose a density-ratio-based framework for data deletion in generative modeling.
Using this framework, we introduce our two main contributions: a fast method for approximate data deletion and a statistical test for estimating whether or not training points have been deleted.
We provide formal guarantees for both contributions under various learner assumptions.
In addition, we investigate our approximate deletion method and statistical test on real and synthetic datasets for various generative models.
One limitation and important future direction of this work is that it may be challenging to apply the density-ratio-based framework to modern large datasets, as density ratio estimation on these datasets can be hard.

We conclude with a discussion of our contributions in relationship to three related areas: differential privacy, membership inference, and influence functions.
First, we note that if the learner is differentially private \citep{dwork2006calibrating}, then the re-trained model is close to the pre-trained model.
This means that there is no need to perform data deletion, and it is definitionally impossible to test whether training data have been deleted.
Second, membership inference attackers query whether a particular sample is used for training \citep{shokri2017membership}.
This is akin to when the deletion set $X'=\{x'\}$ contains only one sample and membership inference is performed to test whether the training set contains $x'$ or not.
In contrast, our deletion test is based on additional prior knowledge and tests whether the training set is $X$ or $X\setminus\{x'\}$.
Therefore, membership inference is stronger but harder than the deletion test.
Finally, we highlight that influence functions \citep{koh2017understanding,koh2019accuracy,basu2020second,kong2021understanding} designed for likelihood in generative models can potentially be used to estimate density ratio in our framework.

\section*{Acknowledgement}
We thank Kamalika Chaudhuri for discussion and helpful feedback. This work was supported by NSF under CNS 1804829 and ARO MURI W911NF2110317.

\bibliographystyle{abbrvnat}
\bibliography{ms}

\newpage 
\appendix
\input{appendix}

\end{document}

%% file: header.tex
\usepackage{amsthm}

\newtheorem{theorem}{Theorem}

\newtheorem{definition}[]{Definition}
\newtheorem{example}[]{Example}
\newtheorem{remark}[]{Remark}

\newcommand{\mA}{\mathcal{A}}
\newcommand{\mB}{\mathcal{B}}
\newcommand{\mC}{\mathcal{C}}
\newcommand{\mD}{\mathcal{D}}
\newcommand{\mE}{\mathcal{E}}

\newcommand{\mN}{\mathcal{N}}

\newcommand{\MMD}{\mathrm{MMD}}
\newcommand{\KDE}{\mathrm{KDE}}
\newcommand{\E}{\mathbb{E}}
\newcommand{\VAR}{\mathbb{VAR}}

\newcommand{\KL}[2]{\mathbb{KL}\left(#1\|#2\right)}
\newcommand{\prob}[1]{\mathrm{Prob}\left(#1\right)}

%% file: appendix.tex
\section{Notation Table}\label{appendix: notation}

\begin{table}[!h]
    \centering
    \caption{Notations used in this paper.}
    \begin{tabular}{c|l}
    \toprule
        $p_*$ & data distribution \\
        $X$ & training set: $N$ i.i.d. samples from $p_*$ \\
        $X'$ & deletion set: $N'$ samples from $X$ \\
        $\mA$ & algorithm of the generative model \\
        $\hat{p}$ & pretrained generative model on $X$ \\
        $\hat{p}'$ & retrained generative model on $X\setminus X'$ \\ \hline 
        $p_*'$ & distribution s.t. $X\setminus X'$ are i.i.d. samples from $p_*'$ \\
        $\rho_*$ & density ratio $p_*'/p_*$ \\
        $\hat{\rho}$ & density ratio $\hat{p}'/\hat{p}$ \\
        DRE & density ratio estimator \\
        $\hat{\rho}_{\mE}$ & abbreviation for $\hat{\rho}_{\mE}(X,X\setminus X')$; DRE between $X$ and $X\setminus X'$ \\
        $\mD(\hat{p},X,X')$ & approximate deletion $\hat{\rho}_{\mE}\cdot\hat{p}$ \\ \hline 
        $q$ & the distribution to be tested \\
        $m$ & number of samples drawn from models \\
        $Y$ & $m$ i.i.d. samples from $q$ \\
        $Y_{H_i}$ & $m$ i.i.d. samples from $q$ under $H_i$, $i=1,2$ \\
        $\hat{Y}$ & $m$ i.i.d. samples from the pretrained model $\hat{p}$ \\
        $Y_{\mD}$ & $m$ i.i.d. samples from the approximate deletion $\mD(\hat{p},X,X')$ \\ \hline
        $R$ & number of repeats for each statistic \\
        $\MMD^2$ & squared MMD metric \\
        $\hat{\MMD}_u^2$ & unbiased MMD estimator \\
        $\mathrm{LR}$ & likelihood ratio \\
        $D_{\phi}$ & the $\phi$-divergence \\
        $\hat{\mathrm{ASC}}_{\phi}$ & ASC statistic, or the $\phi$-divergence estimator \\
        $\mathrm{IF}$ & influence functions \\
        $\tilde{\mathrm{IF}}$ & influence function estimator \\
        $C$ & number of checkpoints to compute $\tilde{\mathrm{IF}}$ \\
        $\eta$ & learning rate to compute $\tilde{\mathrm{IF}}$ \\ \hline 
        $\lambda$ & parameter used to define $\rho_*$ in 2d experiments \\ 
        KDE & kernel density estimator \\
        KBC & kernel-based classifier \\
        $k$NN & $k$ nearest neighbour classifier \\
        $\sigma$ & bandwidth used to define kernel $\mN(0,\sigma^2I)$ in KDE \\
        $\sigma_{\mA}$ & bandwidth of the learning algorithm in 2d experiments \\
        $\sigma_{\mC}$ & bandwidth of KBC in 2d experiments \\
    \bottomrule
    \end{tabular}    
\end{table}

\newpage
\section{Theory for the Framework in Section \ref{sec: framework}}\label{appendix: theory}

\subsection{Omitted Proofs in Section \ref{sec: approx theory}}\label{appendix: approx theory}

\paragraph{Proof of \textbf{Thm.} \ref{thm: approx under RC}}
\begin{proof}
    Notice that
    \begin{align*}
        \hat{\rho} = \frac{\hat{p}'}{\hat{p}} = \frac{\hat{p}'}{p_*'}\cdot\frac{p_*'}{p_*}\cdot\frac{p_*}{\hat{p}}.
    \end{align*}
    With probability at least $1-\delta_N$
    \begin{align*}
        \frac{1}{c_N} \leq \frac{p_*}{\hat{p}} \leq c_N.
    \end{align*}
    With probability at least $1-\delta_{N-N'}$
    \begin{align*}
         \frac{1}{c_{N-N'}} \leq \frac{\hat{p}'}{p_*'} \leq c_{N-N'}.
    \end{align*}
    Therefore, with probability at least $1-\delta_N-\delta_{N-N'}$,
    \begin{align*} 
        \int_{\mathbb{R}^d}\hat{p}\left|\hat{\rho}-\rho_*\right| dx 
        &= \int_{\mathbb{R}^d} p_*'\left|\frac{\hat{p}'}{p_*'}-\frac{\hat{p}}{p_*}\right| dx \\
        & \leq \max\left(c_N-\frac{1}{c_{N-N'}}, c_{N-N'}-\frac{1}{c_N}\right) \\
        & \leq 2(c_N+c_{N-N'}-2).
    \end{align*}
    
    Now, we choose a fixed RC algorithm $\mA_0$, and define $\hat{\rho}_{\mE}(Z_1,Z_2) = \rho(p_{\mA_0(Z_1)}, p_{\mA_0(Z_2)})$. Then, with probability at least $1-\delta_N-\delta_{N-N'}$,
    \begin{align*}
        \int_{\mathbb{R}^d}\hat{p}\left|\hat{\rho}_{\mE}-\rho_*\right| dx \leq 2(c_N+c_{N-N'}-2).
    \end{align*}
    Therefore, with probability at least $1-2\delta_N-2\delta_{N-N'}$,
    \begin{align*}
        \|\hat{\rho}_{\mE}\cdot\hat{p}-\hat{p}'\|_1 = \int_{\mathbb{R}^d}\hat{p}\left|\hat{\rho}_{\mE}-\hat{\rho}\right| dx \leq 4(c_N+c_{N-N'}-2).
    \end{align*}
\end{proof}

\paragraph{Proof of \textbf{Thm.} \ref{thm: approx under TVC}}
\begin{proof}
    Notice that
    \begin{align*}
        \int_{\mathbb{R}^d}\hat{p}^2\left|\hat{\rho}-\rho_*\right| dx 
        &= \int_{\mathbb{R}^d} \hat{p}\left|\hat{p}'-\rho_*\hat{p}\right| dx \\
        &= \int_{\mathbb{R}^d} \hat{p}\left|\hat{p}'-p_*'+p_*'-\rho_*(\hat{p}-p_*+p_*)\right| dx \\
        &= \int_{\mathbb{R}^d} \hat{p}\left|\hat{p}'-p_*'-\rho_*(\hat{p}-p_*)\right| dx.
    \end{align*}
    With probability at least $1-\delta_N$, $|\hat{p}-p_*| \leq \epsilon_N$, and with probability at least $1-\delta_{N-N'}$, $|\hat{p}'-p_*'| \leq \epsilon_{N-N'}$. Therefore, with probability at least $1-\delta_N-\delta_{N-N'}$, 
    \begin{align*}
        \int_{\mathbb{R}^d}\hat{p}^2\left|\hat{\rho}-\rho_*\right| dx \leq \epsilon_{N-N'} + \|\rho_*\|_{\infty}\epsilon_N.
    \end{align*}
    Now, we choose a fixed TVC algorithm $\mA_0$, and define $\hat{\rho}_{\mE}(Z_1,Z_2) = \rho(p_{\mA_0(Z_1)}, p_{\mA_0(Z_2)})$. Then, with probability at least $1-\delta_N-\delta_{N-N'}$,
    \begin{align*}
        \int_{\mathbb{R}^d}\hat{p}^2\left|\hat{\rho}_{\mE}-\rho_*\right| dx \leq \epsilon_{N-N'} + \|\rho_*\|_{\infty}\epsilon_N.
    \end{align*}
    Therefore, with probability at least $1-2\delta_N-2\delta_{N-N'}$,
    \begin{align*}
        \|\hat{\rho}_{\mE}\cdot\hat{p}-\hat{p}'\|_{1,\hat{p}}= \int_{\mathbb{R}^d}\hat{p}^2\left|\hat{\rho}-\hat{\rho}_{\mE}\right| dx 
        \leq 2\left(\epsilon_{N-N'} + \|\rho_*\|_{\infty}\epsilon_N\right).
    \end{align*}
\end{proof}

\subsection{Omitted Proofs in Section \ref{sec: practical theory}}\label{appendix: practical theory}

\paragraph{Proof of \textbf{Thm.} \ref{thm: prac under LBLI}}
\begin{proof}
    By taking $Z_0=Z$, $Z_1=Z\setminus\{z\}$, and $\hat{Z}=\{\hat{z}\}$, we conclude $\epsilon$-DP implies $\epsilon$-US. By taking one side of the $\epsilon$-US bound, we conclude $\epsilon$-US implies $\epsilon$-LBLI.
    
    Define 
    \begin{align*}
        \hat{\rho}_k = \frac{p_{\mA(X\setminus X'_{1:k-1})}}{p_{\mA(X\setminus X'_{1:k})}}
    \end{align*}
    for $k=1,\cdots,N'$. Then, $\epsilon$-LBLI indicates $\log\|\hat{\rho}_k\|_{\infty}\leq\epsilon$. Notice that 
    \begin{align*}
        \hat{\rho} = \prod_{k=1}^{N'}\hat{\rho}_k.
    \end{align*}
    Therefore, we have $\log\|\hat{\rho}\|_{\infty}\leq N'\epsilon$. 
\end{proof}

\paragraph{Proof of \textbf{Thm.} \ref{thm: RC RS}}
\begin{proof}
    With probability at least $1-\delta_N$, 
    \begin{align*}
        -\log c_N \leq \log\rho(\mu_i,p_{\mA(Z_i)}) \leq \log c_N. 
    \end{align*}
    Therefore, with probability at least $1-2\delta_N$, 
    \begin{align*}  
        \left\|\log\rho(\mu_1,p_{\mA(Z_1)})-\log\rho(\mu_2,p_{\mA(Z_2)})\right\|_{\infty} \leq 2\log c_N. 
    \end{align*}
\end{proof}

\paragraph{Proof of \textbf{Thm.} \ref{thm: prac under RS}}
\begin{proof}
    Define $Z_k = X\setminus X'_{1:k}$ and $\mu_k$ be the distribution such that $Z_k$ contains i.i.d. samples from $\mu_k$. Specifically, $\mu_0=p_*$ and $\mu_{N'}=p_*'$. Then, we have 
    \begin{align*}
        \log\rho_* - \log\hat{\rho} &= \log\frac{\mu_{N'}}{\mu_0} - \log\frac{p_{\mA(Z_{N'})}}{p_{\mA(Z_0)}} \\
        &= \sum_{k=1}^{N'}\left(\log\frac{\mu_k}{\mu_{k-1}}-\log\frac{p_{\mA(Z_k)}}{p_{\mA(Z_{k-1})}}\right) \\
        &= \sum_{k=1}^{N'}\left(\log\rho(\mu_{k-1}, p_{\mA(Z_{k-1})})-\log\rho(\mu_k, p_{\mA(Z_k)})\right).
    \end{align*}
    Therefore, with probability at least $1-N'\delta$, we have
    \begin{align*}
        \|\log\rho_* - \log\hat{\rho}\|_{\infty} \leq N'\epsilon,
    \end{align*}
    which indicates $\log\|\hat{\rho}\|_{\infty}\leq N'\epsilon+\log\|\rho_*\|_{\infty}$.
\end{proof}

\paragraph{Proof of \textbf{Thm.} \ref{thm: prac under ES}}
\begin{proof}
    By rewriting ES for $\hat{p}$ and $\hat{p}'$, we have
    \begin{align*}
        \mathbb{E}_{x\sim\hat{p}}\log\hat{\rho} &= -\KL{\hat{p}}{\hat{p}'},
    \end{align*}
    \begin{align*}    
        \mathbb{E}_{x\sim\hat{p}}(\log\hat{\rho})^2 &\leq \epsilon.
    \end{align*}
    Because 
    \begin{align*}
        \mathbb{E}_{x\sim\hat{p}}(\log\hat{\rho})^2 \geq \left(\mathbb{E}_{x\sim\hat{p}}\log\hat{\rho}\right)^2,
    \end{align*}
    we have $\KL{\hat{p}}{\hat{p}'}\leq\sqrt{\epsilon}$. Then, according to Cantelli's inequality \citep{cantelli1910intorno}, for any positive $a$,
    \begin{align*}
        \prob{\log\hat{\rho}\geq-\KL{\hat{p}}{\hat{p}'}+a} \leq \frac{\VAR(\log\hat{\rho})}{\VAR(\log\hat{\rho}) + a^2}.
    \end{align*}
    By letting
    \begin{align*}
        a = \sqrt{\frac{1-\delta}{\delta}\cdot\VAR(\log\hat{\rho})},
    \end{align*}
    we have with probability at least $1-\delta$ for samples $x\sim\hat{p}$, 
    \begin{align*}
        \log\hat{\rho}(x) & \leq \sqrt{\frac{1-\delta}{\delta}\cdot\left(\mathbb{E}_{x\sim\hat{p}}(\log\hat{\rho})^2 - \KL{\hat{p}}{\hat{p}'}^2\right)} - \KL{\hat{p}}{\hat{p}'} \\
        & \leq \sqrt{\frac{\epsilon(1-\delta)}{\delta}}.
    \end{align*}
\end{proof}

\begin{remark}\label{remark: strong assumptions for feasibility}
    Note that $\epsilon$-DP implies $\epsilon$-US and $\epsilon$-US implies $\epsilon$-LBLI. If $\mA$ is $\epsilon$-DP or $\epsilon$-US, the re-trained model satisfies $\hat{p}'\approx\hat{p}$, and there is no need to perform deletion. If $\mA$ is $\epsilon$-LBLI but not $\epsilon$-US, then there exists a sample $\hat{z}$ such that $\hat{p}'(\hat{z}) \ll \hat{p}(\hat{z})$. Intuitively, in non-parametric methods, $\hat{z}$ can be samples near $X'$. $\epsilon$-LBLI can be achieved under some regulatory assumptions on the loss function and the Hessian matrix with respect to parameters \citep{giordano2019higher,giordano2019swiss, basu2020second}.
\end{remark}

\newpage
\section{Statistical Tests in Section \ref{sec: stat test}}\label{appendix: stat test}

\subsection{Likelihood Ratio Tests}\label{appendix: LR}

\paragraph{Proof of \textbf{Thm.} \ref{thm: lr approx under RC}}

\begin{proof}
	By definition of RC, we have with probability at least $1-\delta_N$, 
	\begin{align*}
		|\log\hat{p}-\log p_*|\leq\log c_N,
	\end{align*}
	and with probability at least $1-\delta_{N-N'}$, 
	\begin{align*}
		|\log\hat{p}'-\log p_*'|\leq\log c_{N-N'}.
	\end{align*}
	Therefore, with probability at least $1-\delta_N-\delta_{N-N'}$,
	\begin{align*}
		|\log\hat{\rho}-\log \rho_*|\leq\log c_{N-N'}+\log c_N.
	\end{align*}
	Now, we choose a fixed RC algorithm $\mA_0$, and define $\hat{\rho}_{\mE}(Z_1,Z_2) = \rho(p_{\mA_0(Z_1)}, p_{\mA_0(Z_2)})$. Then, we also have with probability at least $1-\delta_N-\delta_{N-N'}$,
	\begin{align*}
		|\log\hat{\rho}_{\mE}-\log \rho_*|\leq\log c_{N-N'}+\log c_N.
	\end{align*}
	Therefore, with probability at least $1-2(\delta_N+\delta_{N-N'})$,
	\begin{align*}
		|\log\hat{\rho}-\log\hat{\rho}_{\mE}|\leq 2(\log c_{N-N'}+\log c_N),
	\end{align*}
	and the conclusion follows.
\end{proof}

\paragraph{Proof of \textbf{Thm.} \ref{thm: lr approx under rho approx}}

\begin{proof}
	(1) Notice that 
	\begin{align*}
		|\mathrm{LR}(Y,\hat{p},\hat{p}') - \mathrm{LR}(Y,\hat{p},\hat{\rho}_{\mE}\cdot\hat{p})|
		& = \frac1m\sum_{y\in Y} |\log\hat{\rho}(y)-\log\hat{\rho}_{\mE}(y)| \\
		& \leq\frac1m\cdot m\epsilon \\
		& = \epsilon.
	\end{align*}
	
	(2) If $H_0$ is true, then $Y\sim\hat{p}$. Then, 
	\begin{align*}
		\mathbb{E}_Y |\mathrm{LR}(Y,\hat{p},\hat{p}') - \mathrm{LR}(Y,\hat{p},\hat{\rho}_{\mE}\cdot\hat{p})|
		& = \mathbb{E}_Y \left|\frac1m\sum_{y\in Y}(\log\hat{\rho}(y)-\log\hat{\rho}_{\mE}(y))\right| \\
		& \leq \mathbb{E}_Y \left(\frac1m \sum_{y\in Y}\left|\log\hat{\rho}(y)-\log\hat{\rho}_{\mE}(y)\right| \right)\\
		& = \mathbb{E}_{y\sim\hat{p}} \left|\log\hat{\rho}(y)-\log\hat{\rho}_{\mE}(y)\right| \\
		& = \|\log\hat{\rho}-\log\hat{\rho}_{\mE}\|_{1,\hat{p}} \\
		& \leq \epsilon.
	\end{align*}
	By Markov's inequality, we have with probability at least $1-\delta$, $|\mathrm{LR}(Y,\hat{p},\hat{p}') - \mathrm{LR}(Y,\hat{p},\hat{\rho}_{\mE}\cdot\hat{p})|\leq\epsilon/\delta$. The proof for $H_1$ is similar. 
\end{proof}

\paragraph{Statistical properties of $\mathrm{LR}$ statistics.}
Let $\phi(t)=\log(t)^2$. When $H_0$ is true, we have
\begin{align*}
    \mathbb{E}_{Y\sim\hat{p}}~~\mathrm{LR}(Y,\hat{p},\hat{p}') &= \mathbb{E}_{\hat{p}}\log\frac{\hat{p}'}{\hat{p}} = -\KL{\hat{p}}{\hat{p}'}, \\
    \VAR_{Y\sim\hat{p}}~~\mathrm{LR}(Y,\hat{p},\hat{p}') &= \frac1m\left(\mathbb{E}_{\hat{p}}\left(\log\frac{\hat{p}'}{\hat{p}}\right)^2 - \left(\mathbb{E}_{\hat{p}}\log\frac{\hat{p}'}{\hat{p}}\right)^2\right) \\
    &= \frac1m\left(D_{\log^2}(\hat{p}\|\hat{p}')-\KL{\hat{p}}{\hat{p}'}^2\right).
\end{align*}

When $H_1$ is true, we have
\begin{align*}
    \mathbb{E}_{Y\sim\hat{p}'}~~\mathrm{LR}(Y,\hat{p},\hat{p}') &= \mathbb{E}_{\hat{p}'}\log\frac{\hat{p}'}{\hat{p}} = \KL{\hat{p}'}{\hat{p}}, \\
    \VAR_{Y\sim\hat{p}'}~~\mathrm{LR}(Y,\hat{p},\hat{p}') &= \frac1m\left(\mathbb{E}_{\hat{p}'}\left(\log\frac{\hat{p}'}{\hat{p}}\right)^2 - \left(\mathbb{E}_{\hat{p}'}\log\frac{\hat{p}'}{\hat{p}}\right)^2\right) \\
    &= \frac1m\left(D_{\log^2}(\hat{p}'\|\hat{p})-\KL{\hat{p}'}{\hat{p}}^2\right).
\end{align*}

\subsection{ASC Tests}\label{appendix: ASC}

\paragraph{Proof of \textbf{Thm.} \ref{thm: asc approx under rho approx}}

\begin{proof}
	Take expectations $Y\sim q$ and $\hat{Y}\sim\hat{p}$. Then, we have
	\begin{align*}
		\mathbb{E} |\hat{\mathrm{ASC}}_{\phi}(\hat{Y},Y,\hat{\rho}) - \hat{\mathrm{ASC}}_{\phi}(\hat{Y},Y,\hat{\rho}_{\mE})|
		& = \mathbb{E} \left| \frac1m\left(\sum_{y\in\hat{Y}}+\sum_{y\in Y}\right)(\psi(\hat{\rho}(y))-\psi(\hat{\rho}_{\mE}(y))) \right| \\
		& \leq \mathbb{E} \left( \frac1m\sum_{y\in\hat{Y}}|\psi(\hat{\rho}(y))-\psi(\hat{\rho}_{\mE}(y))| \right) \\
		& +  \mathbb{E} \left( \frac1m\sum_{y\in Y}|\psi(\hat{\rho}(y))-\psi(\hat{\rho}_{\mE}(y))| \right) \\
		& = \mathbb{E}_{y\sim\hat{p}} |\psi(\hat{\rho}(y))-\psi(\hat{\rho}_{\mE}(y))| +  \mathbb{E}_{y\sim q}|\psi(\hat{\rho}(y))-\psi(\hat{\rho}_{\mE}(y))| \\
		& = \|\psi(\hat{\rho})-\psi(\hat{\rho}_{\mE})\|_{1,\hat{p}} + \|\psi(\hat{\rho})-\psi(\hat{\rho}_{\mE})\|_{1,q} \\
		& \leq 2\epsilon.
	\end{align*}
	By Markov's inequality, we have with probability at least $1-\delta$, it holds that $|\hat{\mathrm{ASC}}_{\phi}(\hat{Y},Y,\hat{\rho}) - \hat{\mathrm{ASC}}_{\phi}(\hat{Y},Y,\hat{\rho}_{\mE})|\leq 2\epsilon/\delta$.

\end{proof}

\paragraph{Statistical properties of $\mathrm{ASC}$ statistics.}
When $H_0$ is true, we have
\begin{align*}
    \mathbb{E}_{Y\sim\hat{p},\hat{Y}\sim\hat{p}}~~\hat{\mathrm{ASC}}_{\phi}(\hat{Y},Y,\hat{\rho}) 
    = \mathbb{E}_{\hat{p}}\left(\frac{2\phi(\hat{\rho}(y))}{1+\hat{\rho}(y)}\right).
\end{align*}

When $H_1$ is true, we have
\begin{align*}
    \mathbb{E}_{Y\sim\hat{p}',\hat{Y}\sim\hat{p}}~~\hat{\mathrm{ASC}}_{\phi}(\hat{Y},Y,\hat{\rho}) 
    &= (\mathbb{E}_{\hat{p}}+\mathbb{E}_{\hat{p}'}) \frac{\phi(\hat{\rho})}{1+\hat{\rho}} \\
    &= \mathbb{E}_{\hat{p}} (1+\hat{\rho})\cdot\frac{\phi(\hat{\rho})}{1+\hat{\rho}} \\
    &= \mathbb{E}_{\hat{p}}\left(\phi(\hat{\rho}(y))\right).
\end{align*}

\subsection{MMD Tests}\label{appendix: MMD}

\paragraph{Definition of MMD.}  Let $K_{\MMD}(\cdot,\cdot)$ be a kernel function. The Maximum Mean Discrepancy (MMD) \citep{gretton2012kernel} between $\hat{p}$ and $q$ is defined as 
\begin{align*}
    \MMD^2(q,\hat{p}) = \left(\mathbb{E}_{x,y\sim\hat{p}}-2\mathbb{E}_{x\sim\hat{p},y\sim q}+\mathbb{E}_{x,y\sim q}\right)K_{\MMD}(x,y).
\end{align*}

Given $m$ i.i.d. samples $\hat{Y}\sim\hat{p}$ and $m$ i.i.d. samples $Y\sim q$, an unbiased estimator of $\MMD^2$ is
\begin{align*}
    \hat{\MMD}_u^2(Y,\hat{Y}) = \frac{1}{m(m-1)}\sum_{i\neq j}(K_{\MMD}(y_i,y_j)+K_{\MMD}(\hat{y}_i,\hat{y}_j)) - \frac{2}{m^2}\sum_{i,j}K_{\MMD}(y_i,\hat{y}_j).
\end{align*}

\paragraph{Asymptotic and concentration properties\citep{serfling2009approximation,gretton2012kernel}.}
Define
\begin{align*}
    h((y_i,\hat{y}_i), (y_j,\hat{y}_j)) &= K_{\MMD}(y_i,y_j) + K_{\MMD}(\hat{y}_i,\hat{y}_j) - K_{\MMD}(y_i,\hat{y}_j) - K_{\MMD}(y_j,\hat{y}_i).
\end{align*}
Then, we have
\begin{align*}
    \hat{\MMD}_u^2(Y,\hat{Y}) 
    &= \frac{1}{m(m-1)}\sum_{i\neq j}^{m} h((y_i,\hat{y}_i), (y_j,\hat{y}_j)).
\end{align*}
Define
\begin{align*}
    \sigma_u^2 &= 4\left(\E_{\substack{y\sim q\\\hat{y}\sim \hat{p}}} \left[\E_{\substack{y'\sim q\\\hat{y}'\sim \hat{p}}}h((y,\hat{y}),(y',\hat{y}'))\right]^2 - \left[\E_{\substack{y,y'\sim q\\\hat{y},\hat{y}'\sim \hat{p}}} h((y,\hat{y}),(y',\hat{y}'))\right]^2\right) \\
    &= 4\cdot\E_{\substack{y\sim q\\\hat{y}\sim \hat{p}}}\VAR_{\substack{y'\sim q\\\hat{y}'\sim \hat{p}}}~~h((y,\hat{y}),(y',\hat{y}')).
\end{align*}
Then, it holds that
\begin{align*}
    \sqrt{m}\left(\hat{\MMD}_u^2(Y,\hat{Y}) - \MMD^2(q,\hat{p})\right) \rightarrow \mathcal{N}(0,\sigma_u^2) \text{ in distribution}
\end{align*}

As for concentration properties, with probability at least $1-\delta$, it holds that
\begin{align*}
    \MMD_u^2(Y,\hat{Y}) - \MMD^2(q,\hat{p}) \leq 4\sqrt{\frac1m\log\frac1\delta}\cdot \sup_{x,y}K_{\MMD}(x,y),
\end{align*}
with have the same bound on the other side.

\paragraph{Asymptotic and concentration properties in the context of deletion test.} Now, we look at these properties in the context of deletion test. If $H_0$ is true, 
\begin{align*}
    \mathbb{E}_{Y\sim\hat{p}}~~\hat{\MMD}_u^2(Y, \hat{Y}) &= 0, \\
    \VAR_{Y\sim\hat{p}}~~\hat{\MMD}_u^2(Y, \hat{Y}) &= \frac4m\cdot\E_{\substack{y\sim \hat{p}\\\hat{y}\sim \hat{p}}}\VAR_{\substack{y'\sim \hat{p}\\\hat{y}'\sim \hat{p}}}~~h((y,\hat{y}),(y',\hat{y}')).
\end{align*}
And with probability at least $1-\delta$, 
\begin{align*}
    \left|\hat{\MMD}_u^2(Y,\hat{Y})\right| \leq 4\sqrt{\frac1m\log\frac2\delta}\cdot\sup_{x,y}K_{\MMD}(x,y).
\end{align*}
If $H_1$ is true, 
\begin{align*}
    \mathbb{E}_{Y\sim\hat{p}'}~~\hat{\MMD}_u^2(Y, \hat{Y}) &= \MMD^2(\hat{p}',\hat{p}), \\
    \VAR_{Y\sim\hat{p}}~~\hat{\MMD}_u^2(Y, \hat{Y}) &= \frac4m\cdot\E_{\substack{y\sim q\\\hat{y}\sim \hat{p}}}\VAR_{\substack{y'\sim q\\\hat{y}'\sim \hat{p}}}~~h((y,\hat{y}),(y',\hat{y}')).
\end{align*}
And with probability at least $1-\delta$, 
\begin{align*}
    \left|\hat{\MMD}_u^2(Y,\hat{Y})-\MMD^2(\hat{p}',\hat{p})\right| \leq 4\sqrt{\frac1m\log\frac2\delta}\cdot\sup_{x,y}K_{\MMD}(x,y).
\end{align*}

\begin{example}[KDE] Now, we compute $\MMD(\hat{p}',\hat{p})^2$ for KDE with the standard Gaussian kernel. We let $K_{\MMD}$ be the standard RBF kernel: $K_{\MMD}(x,y)=\exp(-\|x-y\|^2/2)$. Let $x,x'\sim q$, $y,y'\sim \hat{p}$, and $z_i,z_i'\sim \mN(x_i,I)$. Then,
\begin{align*}
    \E_{x,x'}K_{\MMD}(x,x') & = \frac{1}{N^2}\sum_{i=1}^N\sum_{j=1}^N\E_{z_i,z_j'}K_{\MMD}(z_i,z_j') \\
    \E_{y,y'}K_{\MMD}(y,y') & = \frac{1}{(N-N')^2}\sum_{i=N'+1}^N\sum_{j=N'+1}^N\E_{z_i,z_j'}K_{\MMD}(z_i,z_j') \\
    \E_{x,y}K_{\MMD}(x,y) & = \frac{1}{N(N-N')}\sum_{i=1}^N\sum_{j=N'+1}^N\E_{z_i,z_j'}K_{\MMD}(z_i,z_j').
\end{align*}
By rearranging, we have
\begin{align*}
    \MMD^2(\hat{p}',\hat{p}) = \left(\frac{N'^2}{N^2(N-N')^2}\sum_{i=N'+1}^N\sum_{j=N'+1}^N - \frac{N+N'}{N^2(N-N')}\sum_{i=1}^{N'}\sum_{j=N'+1}^N + \frac{1}{N^2}\sum_{i=1}^N\sum_{j=1}^{N'}\right) \E_{z_i,z_j'}K_{\MMD}(z_i,z_j').
\end{align*}

We then compute $\E_{z_i,z_j'}K_{\MMD}(z_i,z_j')$.
\begin{align*}
    \E_{z_i,z_j'}K_{\MMD}(z_i,z_j') 
    &= \int_{\mathbb{R}^d}\int_{\mathbb{R}^d} \mN(z_i;x_i,I)\mN(z_j';x_j,I) K_{\MMD}(z_i,z_j') dz_i dz_j' \\
    & = \int_{\mathbb{R}^d}\int_{\mathbb{R}^d} \frac{1}{(2\pi)^{d}} \exp\left(-\frac{\|z_i-x_i\|^2 + \|z_j'-x_j\|^2}{2} - \frac{\|z_i-z_j'\|^2}{2}\right) dz_i dz_j'.
\end{align*}
We apply a change-of-variable formula:
\begin{align*}
    z_i &= -\frac{v_i}{\sqrt{2}} - \frac{v_j'}{\sqrt{6}} + \frac23x_i+\frac13x_j, \\
    z_j &= -\frac{v_i}{\sqrt{2}} + \frac{v_j'}{\sqrt{6}} + \frac13x_i+\frac23x_j. \\
\end{align*}
Then, 
\begin{align*}
    \frac{\|z_i-x_i\|^2 + \|z_j'-x_j\|^2}{2} + \frac{\|z_i-z_j'\|^2}{2} 
    = \frac12\left(\|v_i\|^2 + \|v_j'\|^2 + \frac{\|x_i-x_j\|^2}{3}\right).
\end{align*}
Therefore,
\begin{align*}
    \E_{z_i,z_j'}K_{\MMD}(z_i,z_j') &= \int_{\mathbb{R}^d}\int_{\mathbb{R}^d} \frac{1}{(2\pi)^{d}} \exp\left(-\frac{\|v_i\|^2 + \|v_j'\|^2}{2} - \frac{\|x_i-x_j\|^2}{6}\right) \left|\det\left(\frac{\partial(z_i,z_j')}{\partial(v_i,v_j')}\right)\right|^d dv_i dv_j' \\
    & = 3^{-\frac d2} \exp\left(-\frac{\|x_i-x_j\|^2}{6}\right).
\end{align*}

Summing up, we have
\begin{align*}
    \MMD^2(\hat{p}',\hat{p}) = 3^{-\frac d2} \left(\frac{N'^2}{N^2(N-N')^2}\sum_{i=N'+1}^N\sum_{j=N'+1}^N - \frac{N+N'}{N^2(N-N')}\sum_{i=1}^{N'}\sum_{j=N'+1}^N + \frac{1}{N^2}\sum_{i=1}^N\sum_{j=1}^{N'}\right) \exp\left(-\frac{\|x_i-x_j\|^2}{6}\right).
\end{align*}

\end{example}

\newpage
\section{Experiments on Two-Dimensional Synthetic Datasets}\label{appendix: exp 2d}

\subsection{MoG-8}\label{appendix: exp 2d MoG}

\paragraph{Setup.}

\begin{figure}[!h]
\vspace{-0.2em} 
    	\begin{subfigure}[b]{0.15\textwidth}
		\centering 
		\includegraphics[trim=25 20 105 20, clip, width=0.99\textwidth]{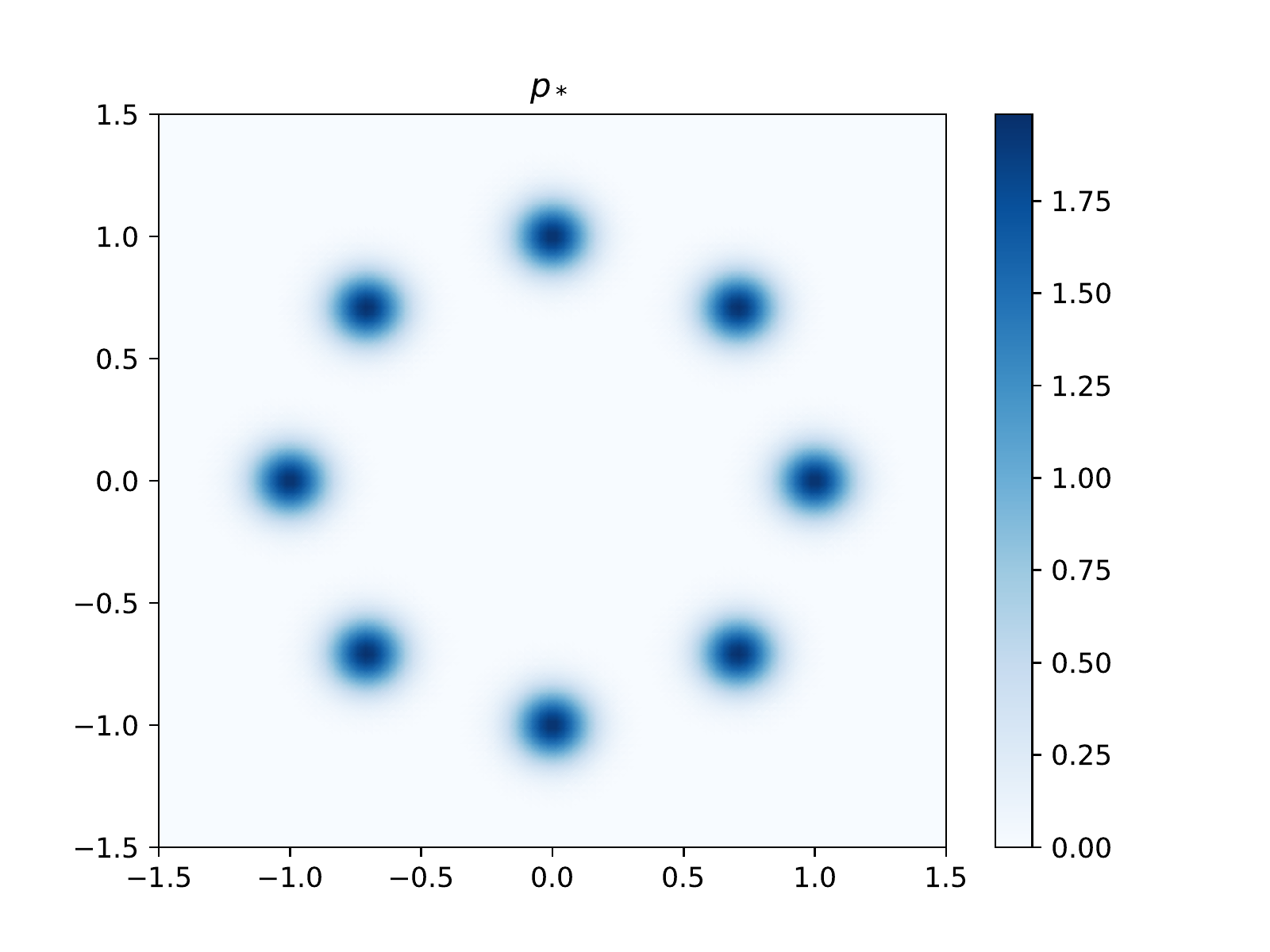}
		\caption{$p_*$}
	\end{subfigure}
	\begin{subfigure}[b]{0.15\textwidth}
		\centering 
		\includegraphics[trim=25 20 105 20, clip, width=0.99\textwidth]{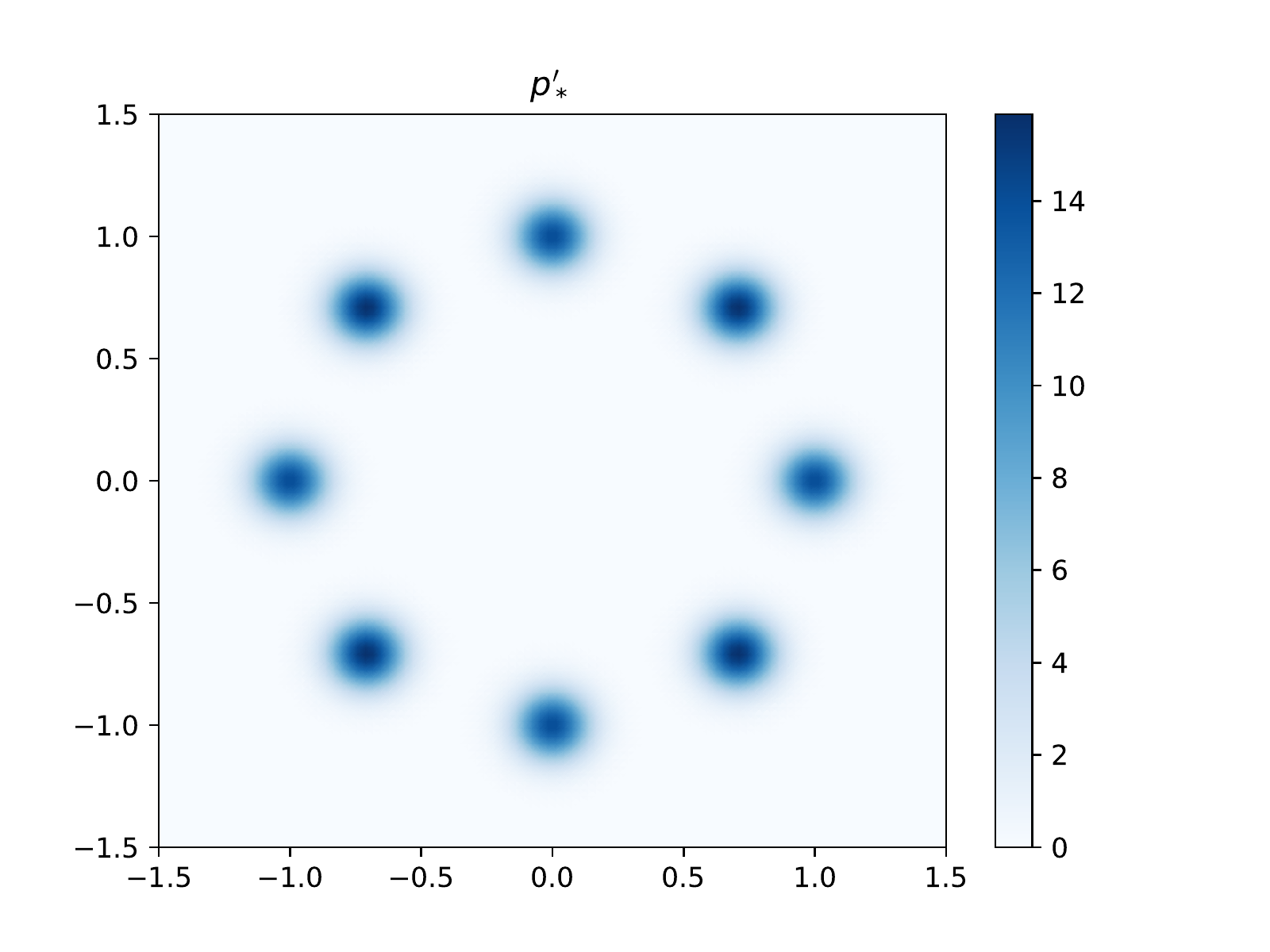}
		\caption{$p_*'(\lambda=0.9)$}
	\end{subfigure}
	\begin{subfigure}[b]{0.15\textwidth}
		\centering 
		\includegraphics[trim=25 20 105 20, clip, width=0.99\textwidth]{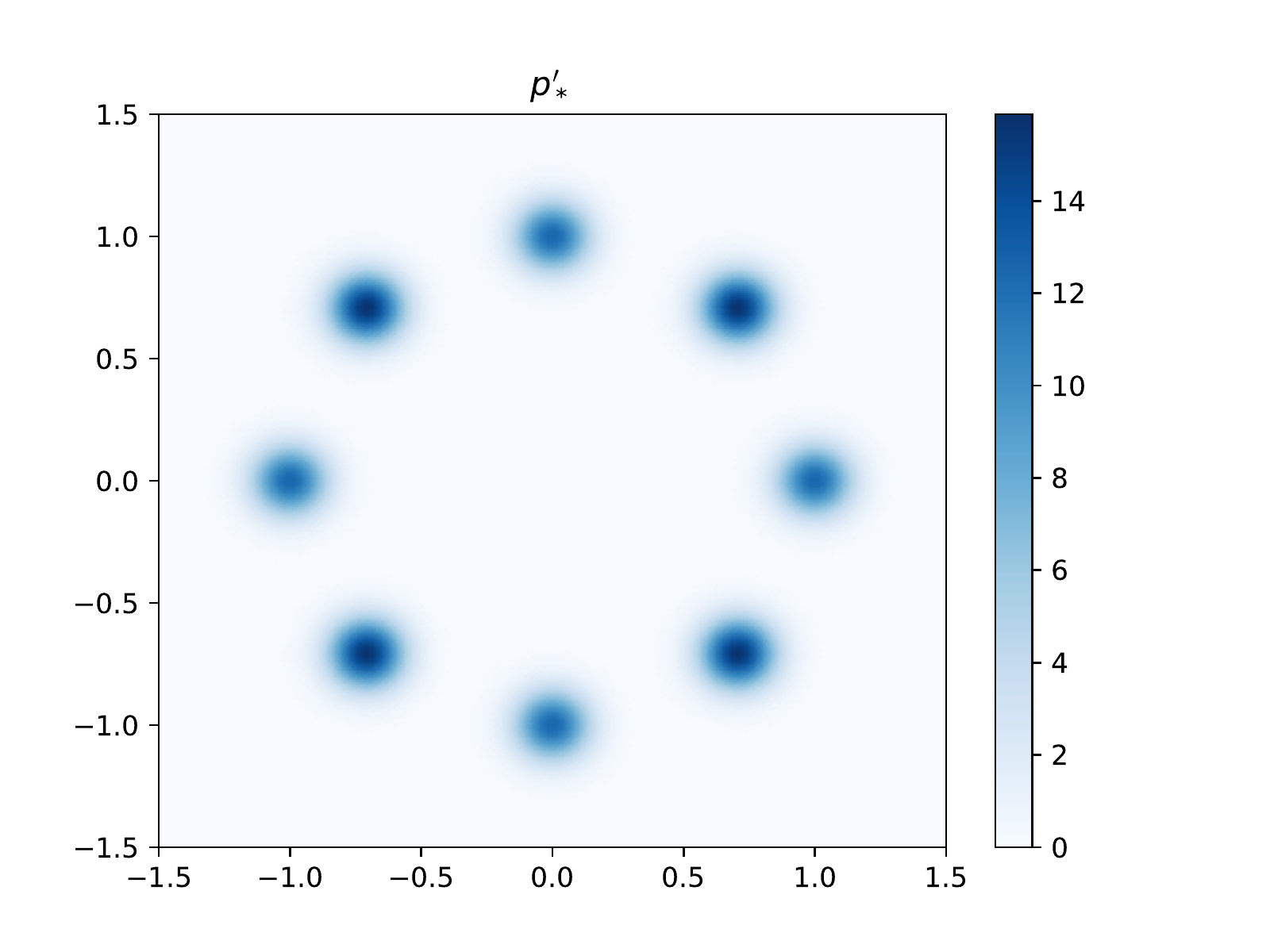}
		\caption{$p_*'(\lambda=0.8)$}
	\end{subfigure}
	\begin{subfigure}[b]{0.15\textwidth}
		\centering 
		\includegraphics[trim=25 20 105 20, clip, width=0.99\textwidth]{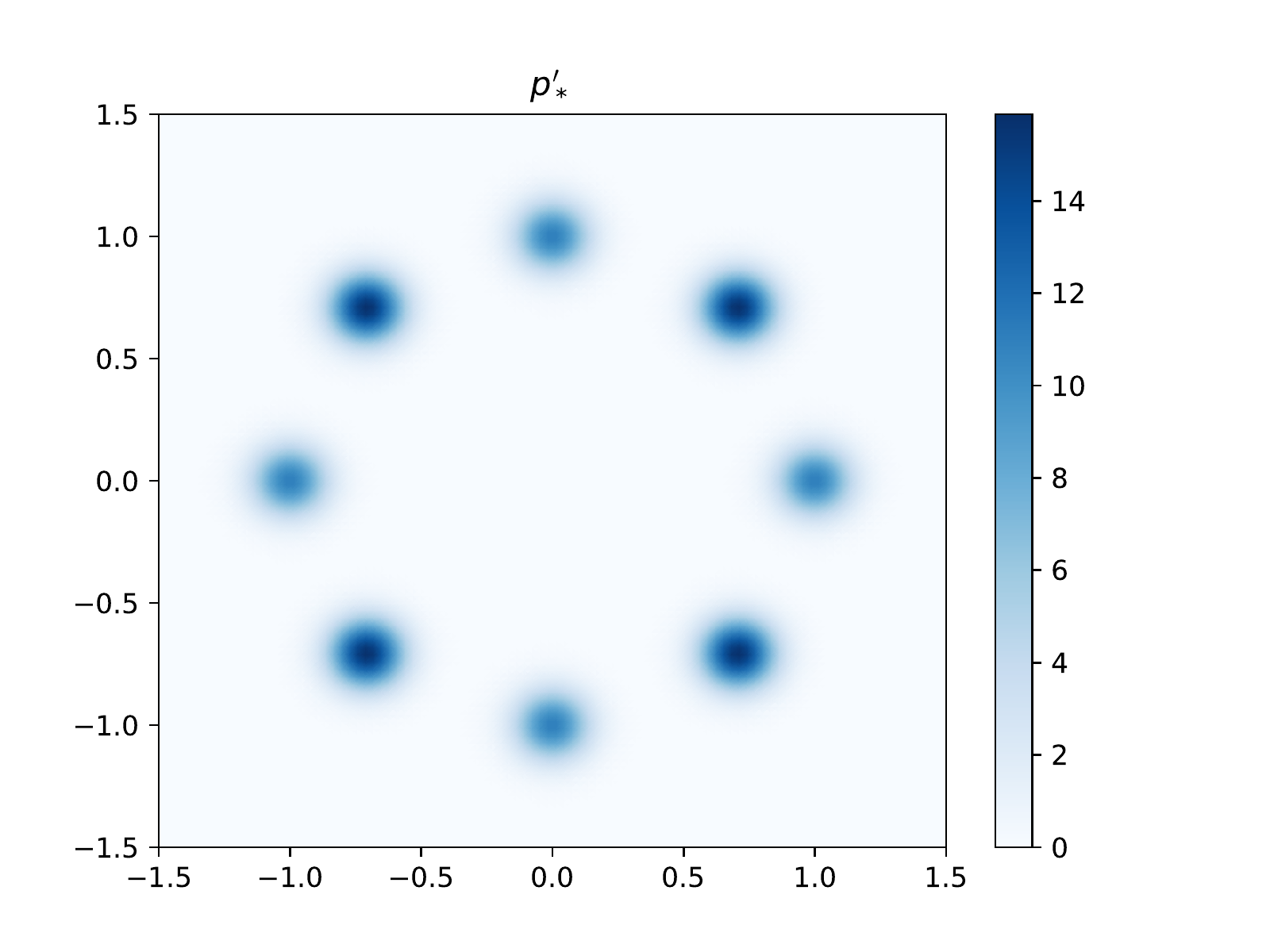}
		\caption{$p_*'(\lambda=0.7)$}
	\end{subfigure}
	\begin{subfigure}[b]{0.15\textwidth}
		\centering 
		\includegraphics[trim=25 20 105 20, clip, width=0.99\textwidth]{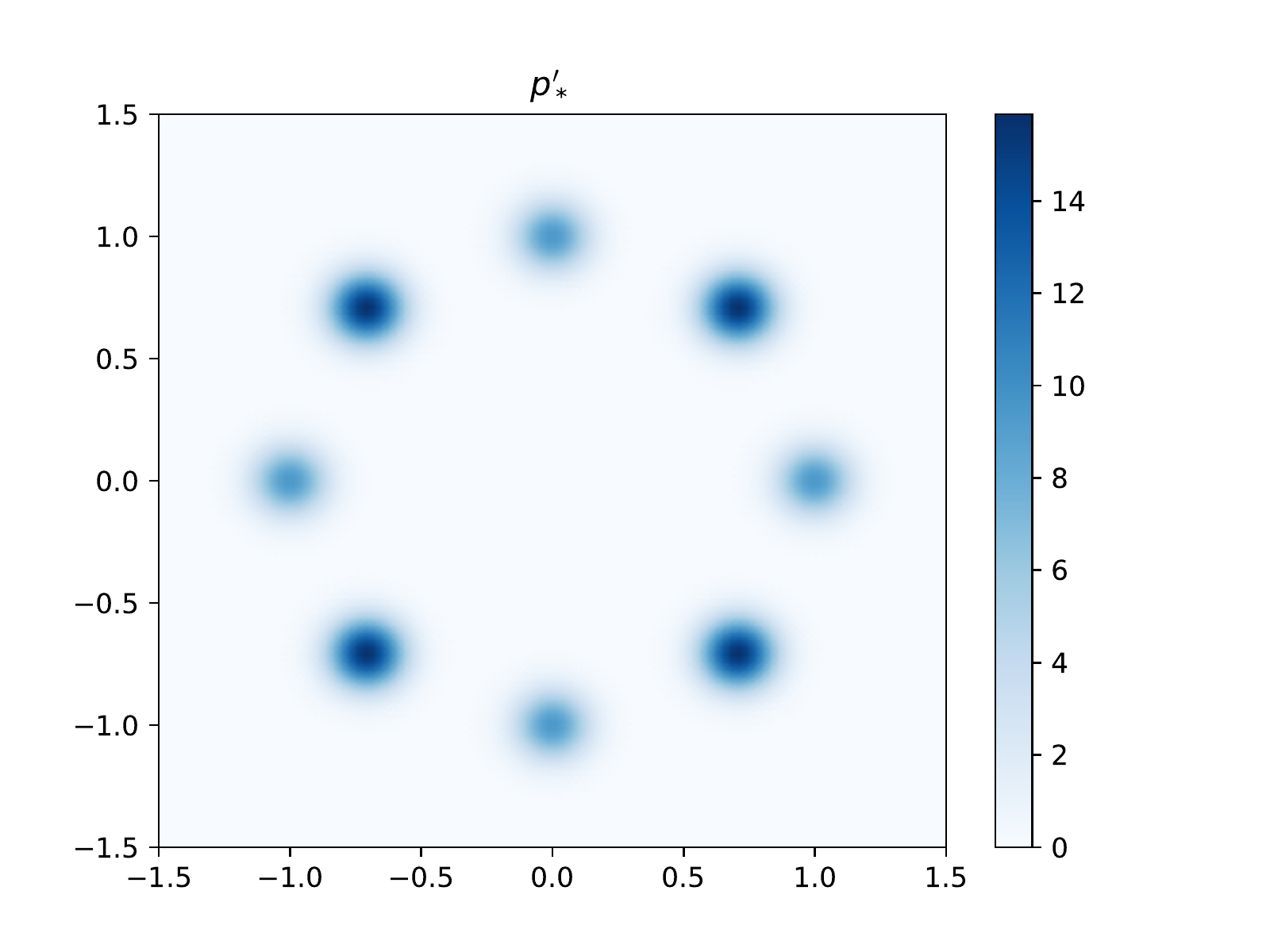}
		\caption{$p_*'(\lambda=0.6)$}
	\end{subfigure}
	\begin{subfigure}[b]{0.15\textwidth}
		\centering 
		\includegraphics[trim=25 20 105 20, clip, width=0.99\textwidth]{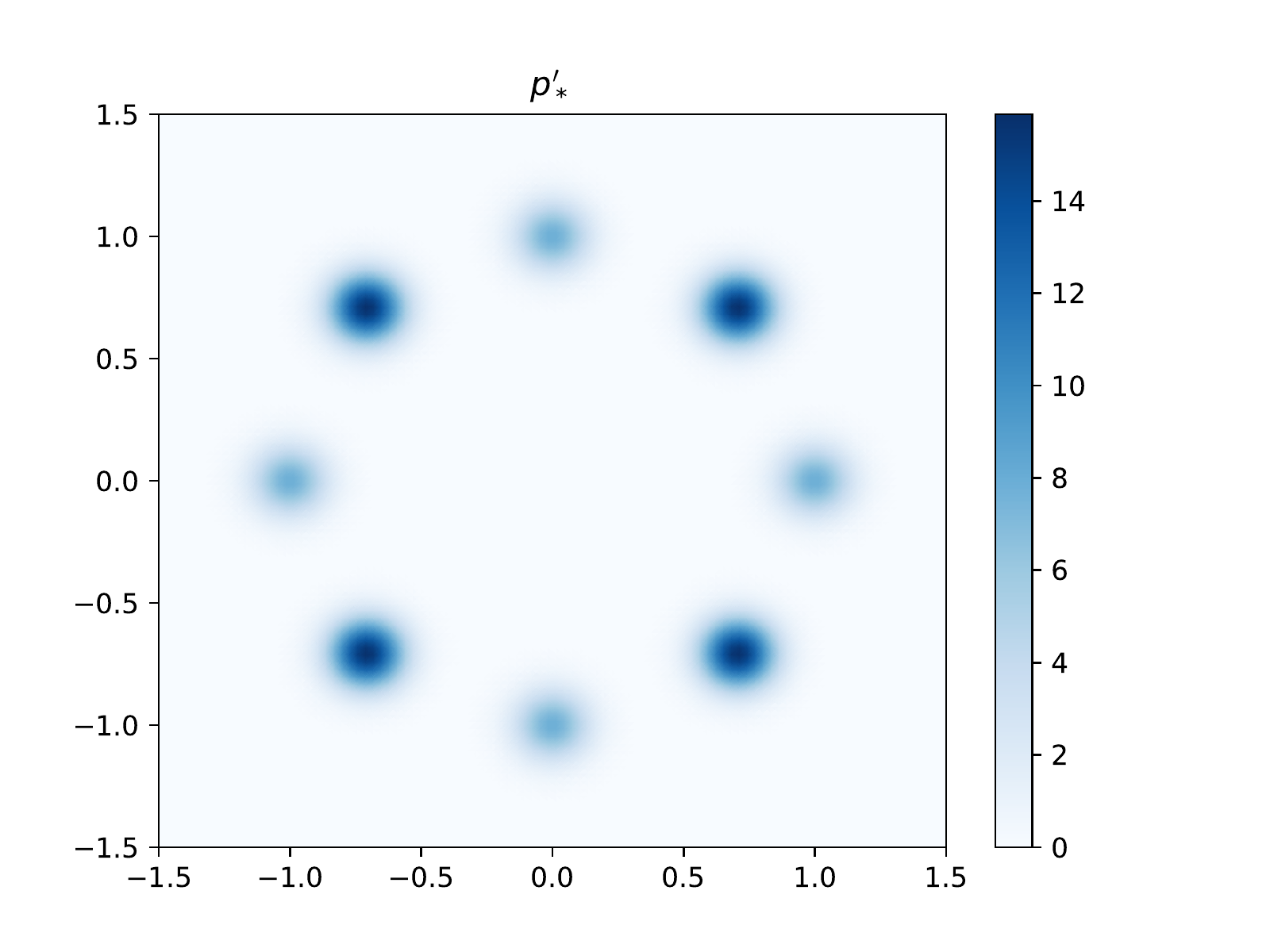}
		\caption{$p_*'(\lambda=0.5)$}
	\end{subfigure}

	\vspace{-0.3em}
	\caption{Visualization of the experimental setup of MoG-8. (a) Data distribution $p_*$. (b) - (f) $p_*'$ with different $\lambda$ values. A larger $\lambda$ means less data is deleted.}
	\label{fig: 2d setup MoG-8 appendix}
	\vspace{-0.2em}
\end{figure}

The data distribution is defined as
\[p_*(x) = \frac18\sum_{i=1}^8 \mN(x;(\cos\theta_i,\sin\theta_i),0.1I),\]
where $\theta_i=\frac{2\pi i}{8}$. The modified distribution $p_*'$ with weight $\lambda$ is defined as 
\[p_*'(x) = \frac{1}{4(1+\lambda)}\sum_{i=1}^8 w_i\mN(x;(\cos\theta_i,\sin\theta_i),0.1I),\]
where $w_i=1$ for even $i$ and $\lambda$ for odd $i$. The construction algorithm for $X$ is randomly sampling a cluster id between 1 and 8 and randomly drawing a sample from the corresponding Gaussian distribution. The construction algorithm for $X'$ is to include a sample $x\in X$ with probability $1-\lambda$ if $x$ is from $i$-th Gaussian for odd $i$. The distributions and data with different $\lambda$ are shown in Fig. \ref{fig: 2d setup MoG-8 appendix}. 

Other hyperparameters are set as follows. The number of training samples $N=400$ unless specified. The number of samples for the deletion test $m=400$ unless specified. The number of repeats for each setup is $R=250$ unless specified. The learning algorithm KDE has bandwidth $\sigma_{\mA}=0.1$ unless specified.

\newpage
\paragraph{Question 1 (DRE Approximations).}

\begin{figure}[!h]
\vspace{-0.3em}
  	\begin{subfigure}[t!]{0.19\textwidth}
		\centering 
		\begin{overpic}[trim=25 20 50 37, clip, width=\textwidth]{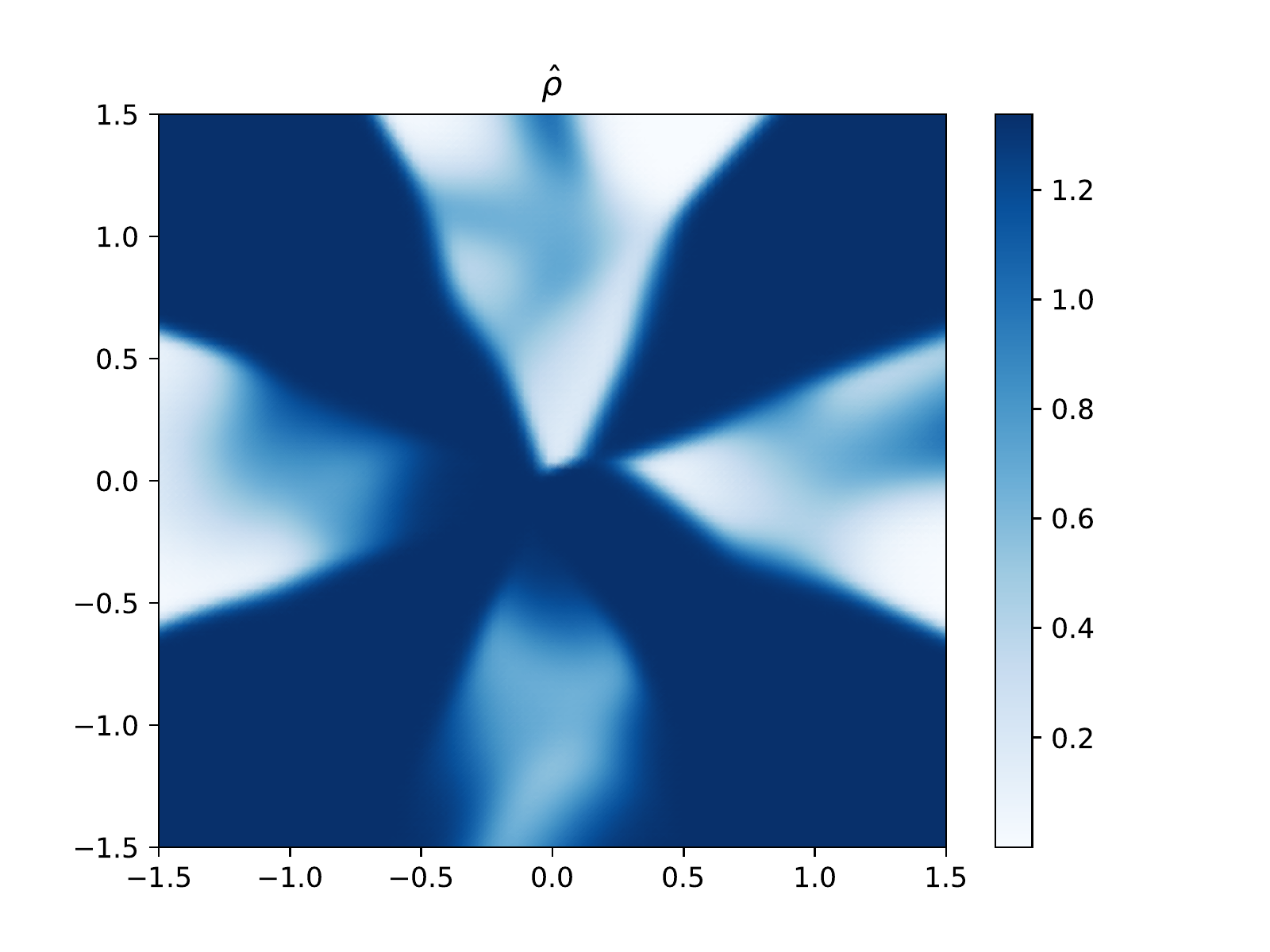} \put(-40,34){$\lambda=0.5$}\end{overpic}\\
		\begin{overpic}[trim=25 20 50 37, clip, width=\textwidth]{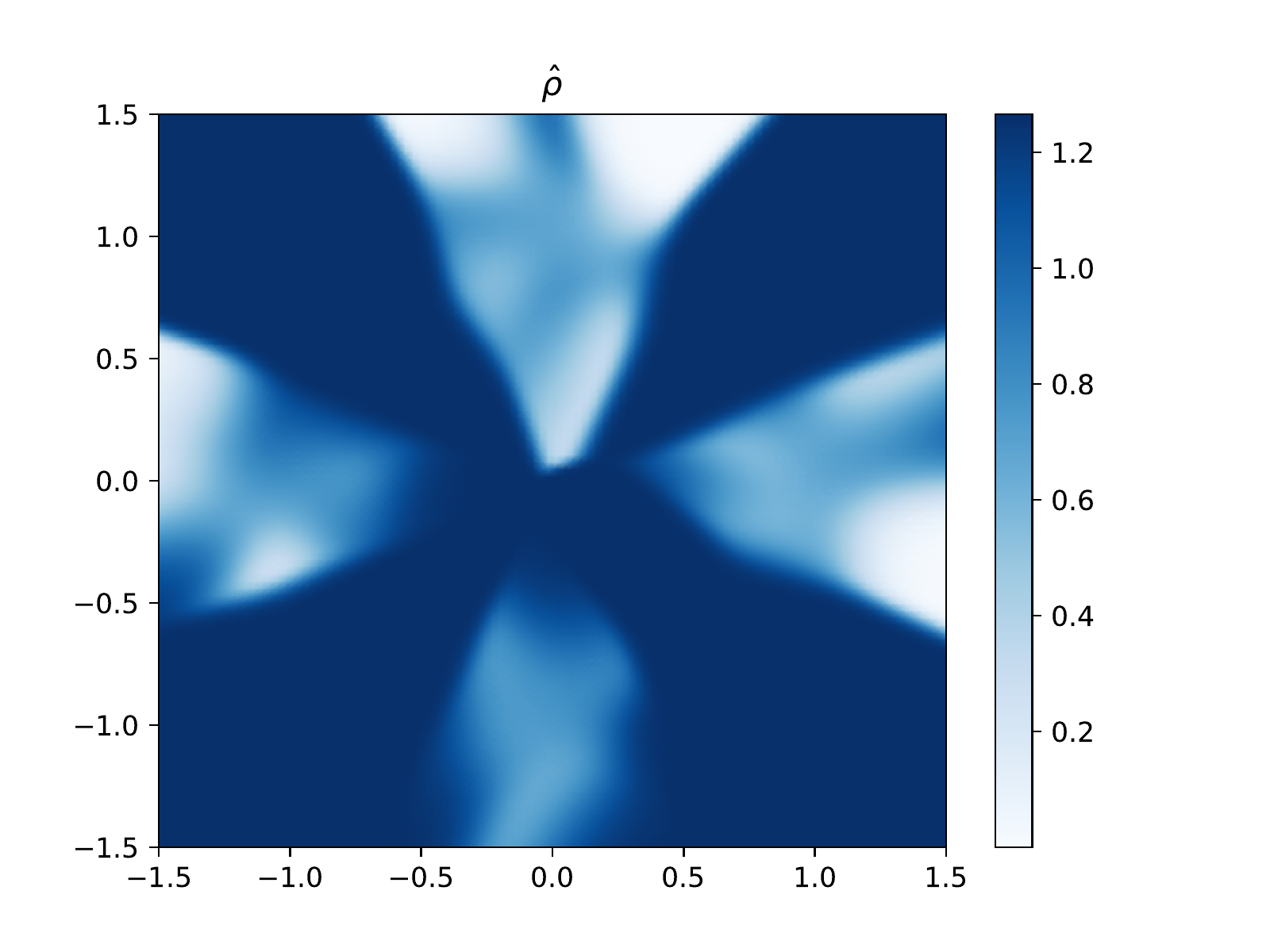} \put(-40,34){$\lambda=0.6$}\end{overpic}\\
		\begin{overpic}[trim=25 20 50 37, clip, width=\textwidth]{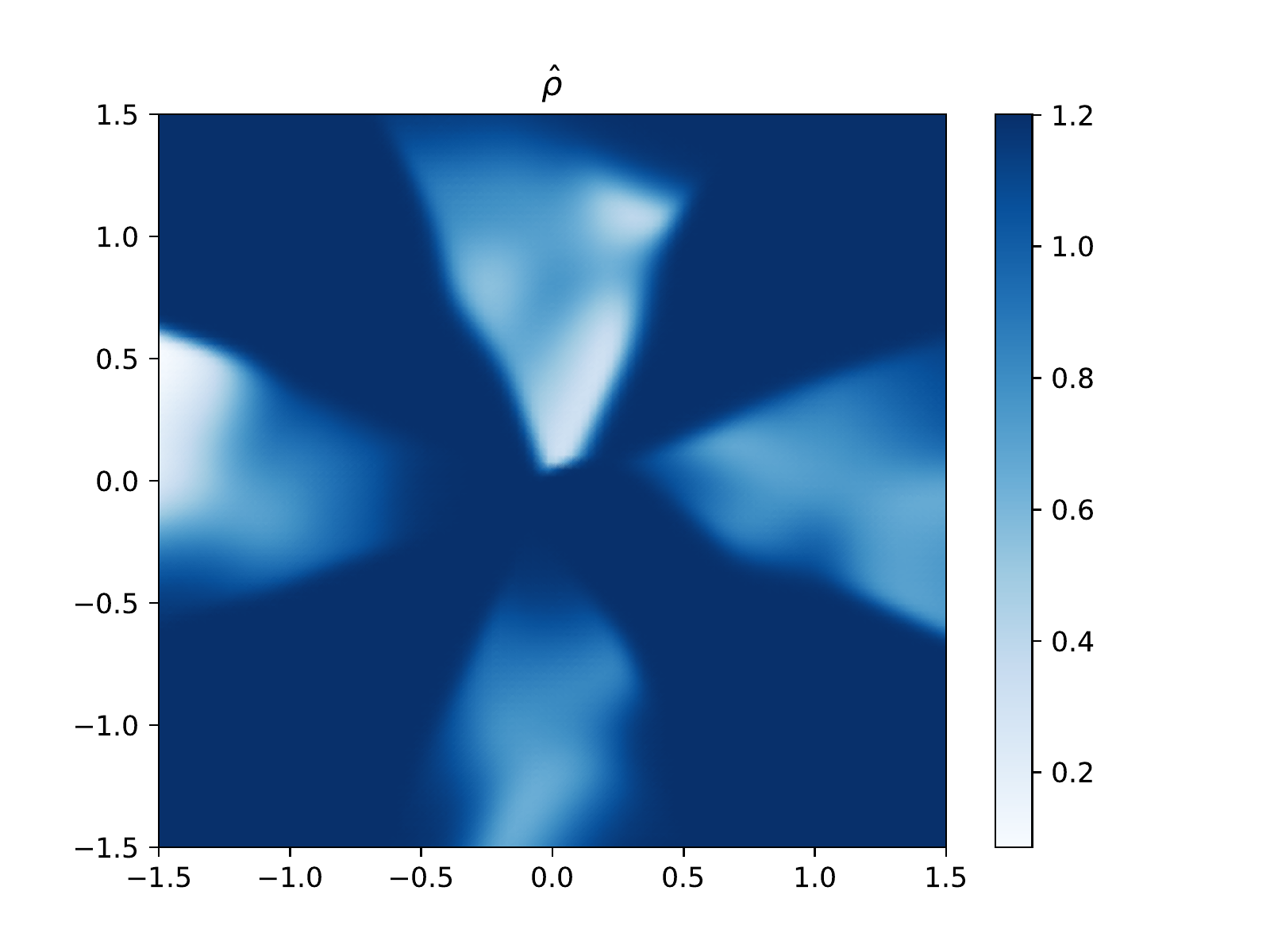} \put(-40,34){$\lambda=0.7$}\end{overpic}\\
		\begin{overpic}[trim=25 20 50 37, clip, width=\textwidth]{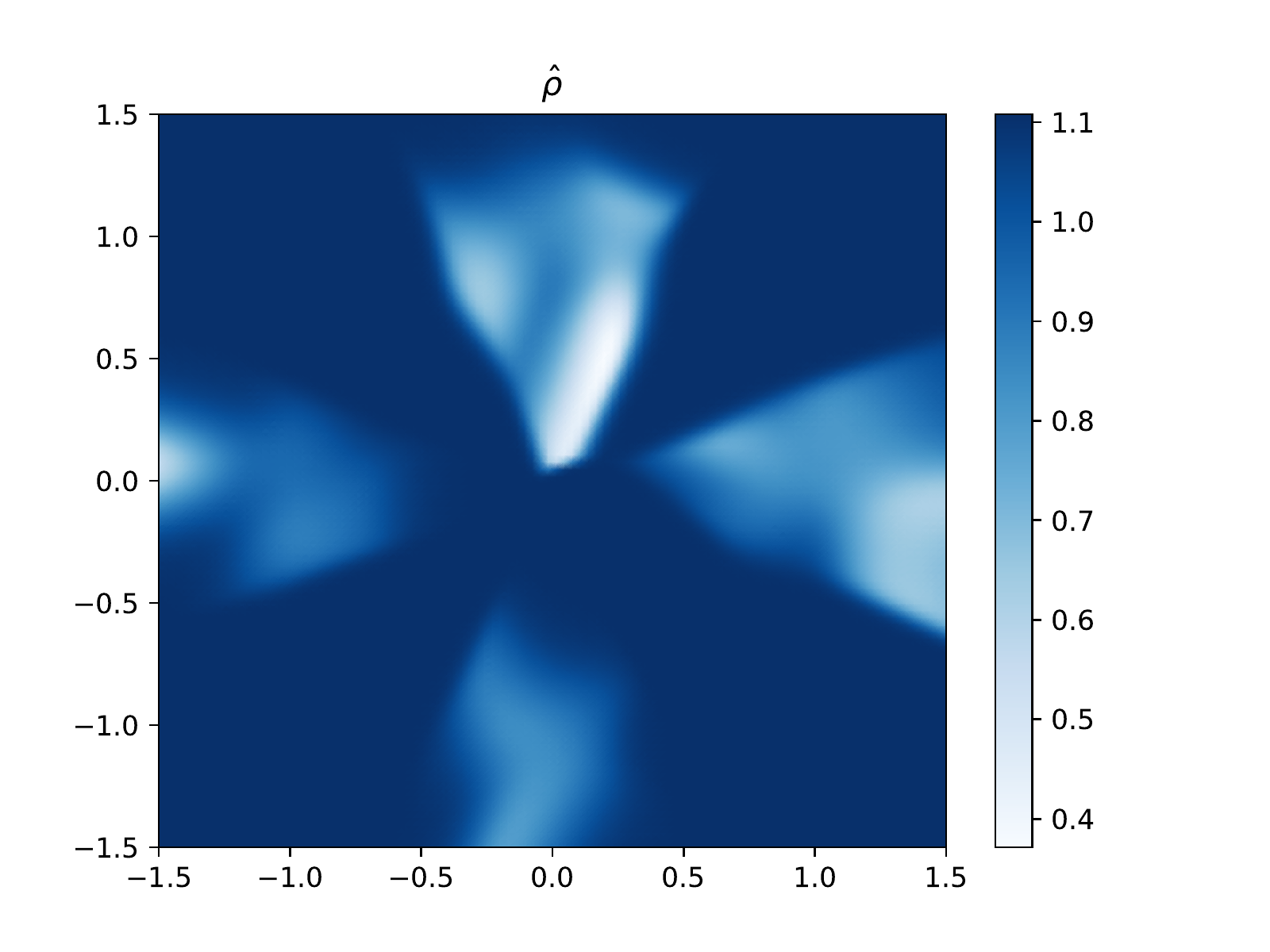} \put(-40,34){$\lambda=0.8$}\end{overpic}\\
		\begin{overpic}[trim=25 20 50 37, clip, width=\textwidth]{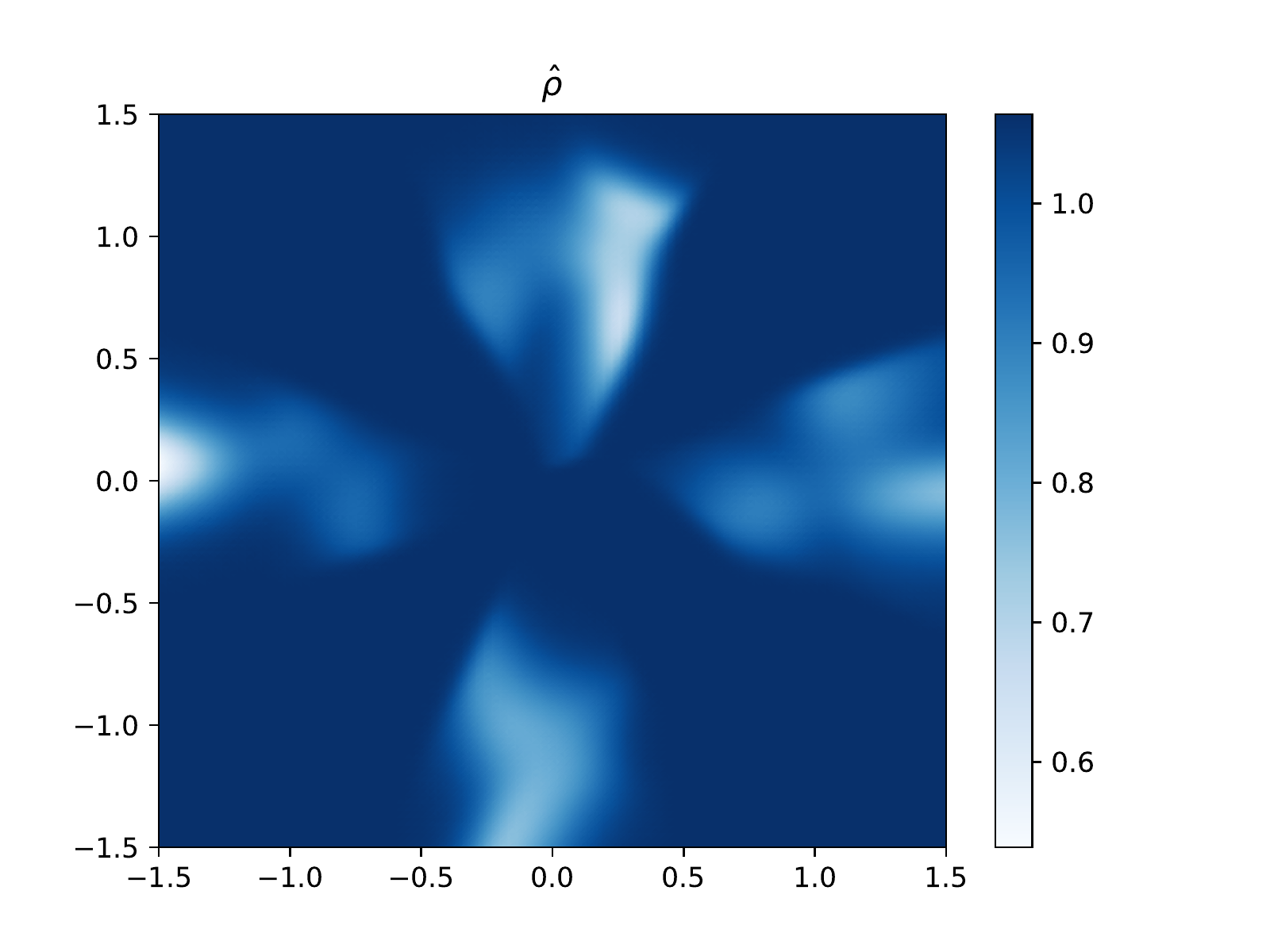} \put(-40,34){$\lambda=0.9$}\end{overpic}\\
		\caption{$\hat{\rho}$}
	\end{subfigure}
	\begin{subfigure}[t!]{0.19\textwidth}
		\centering 
		\includegraphics[trim=25 20 50 37, clip, width=\textwidth]{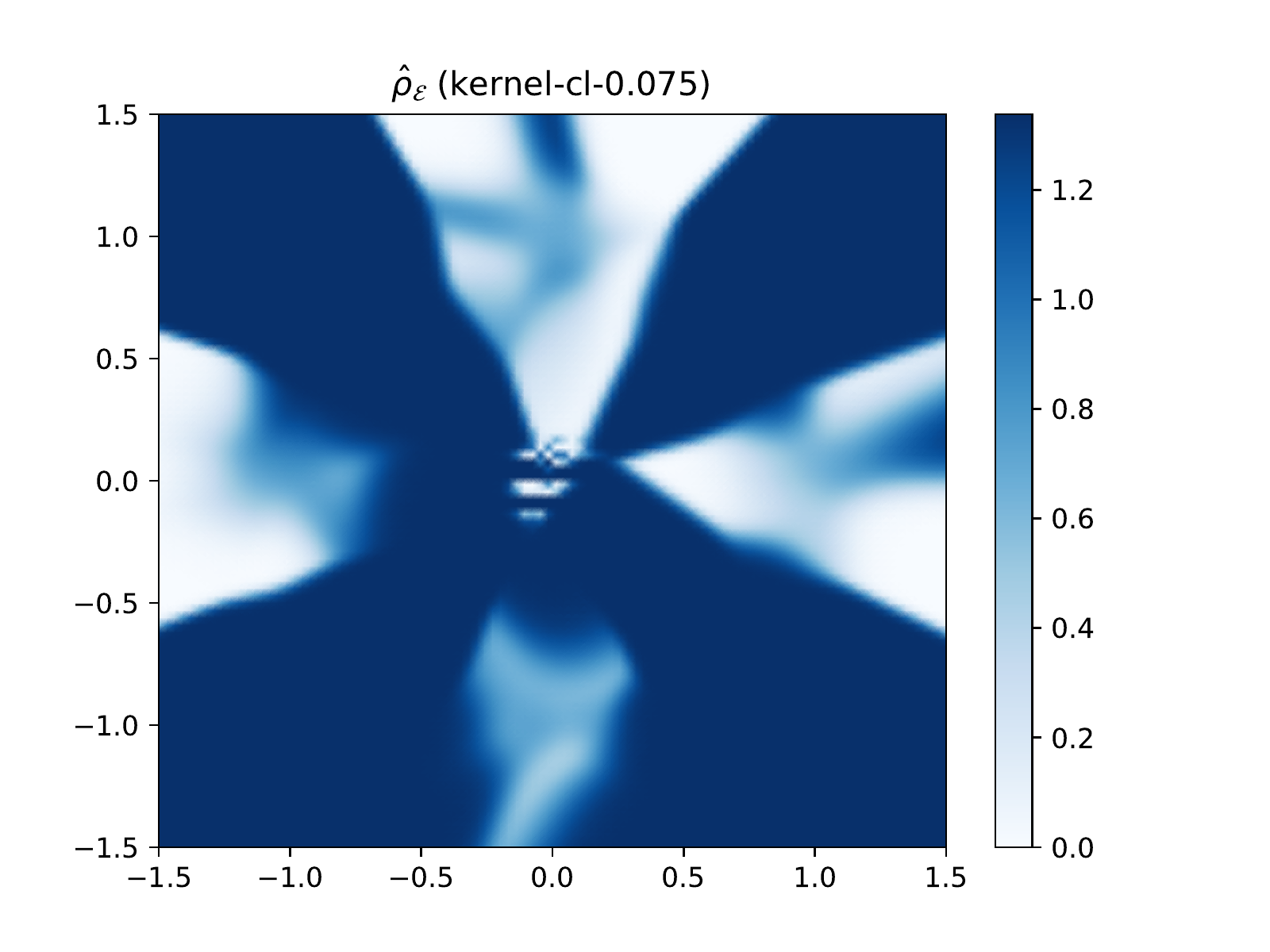}\\
		\includegraphics[trim=25 20 50 37, clip, width=\textwidth]{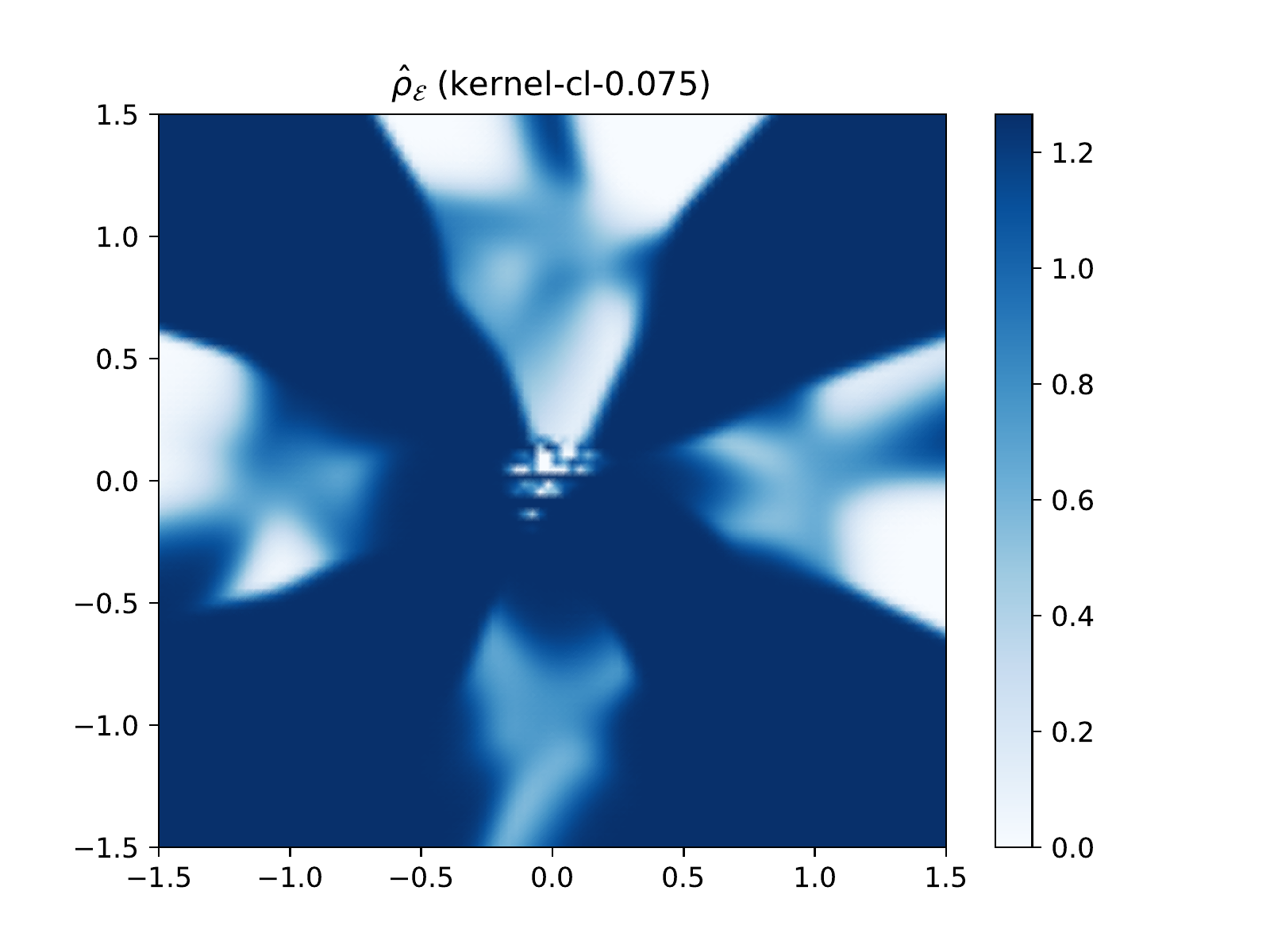}\\
		\includegraphics[trim=25 20 50 37, clip, width=\textwidth]{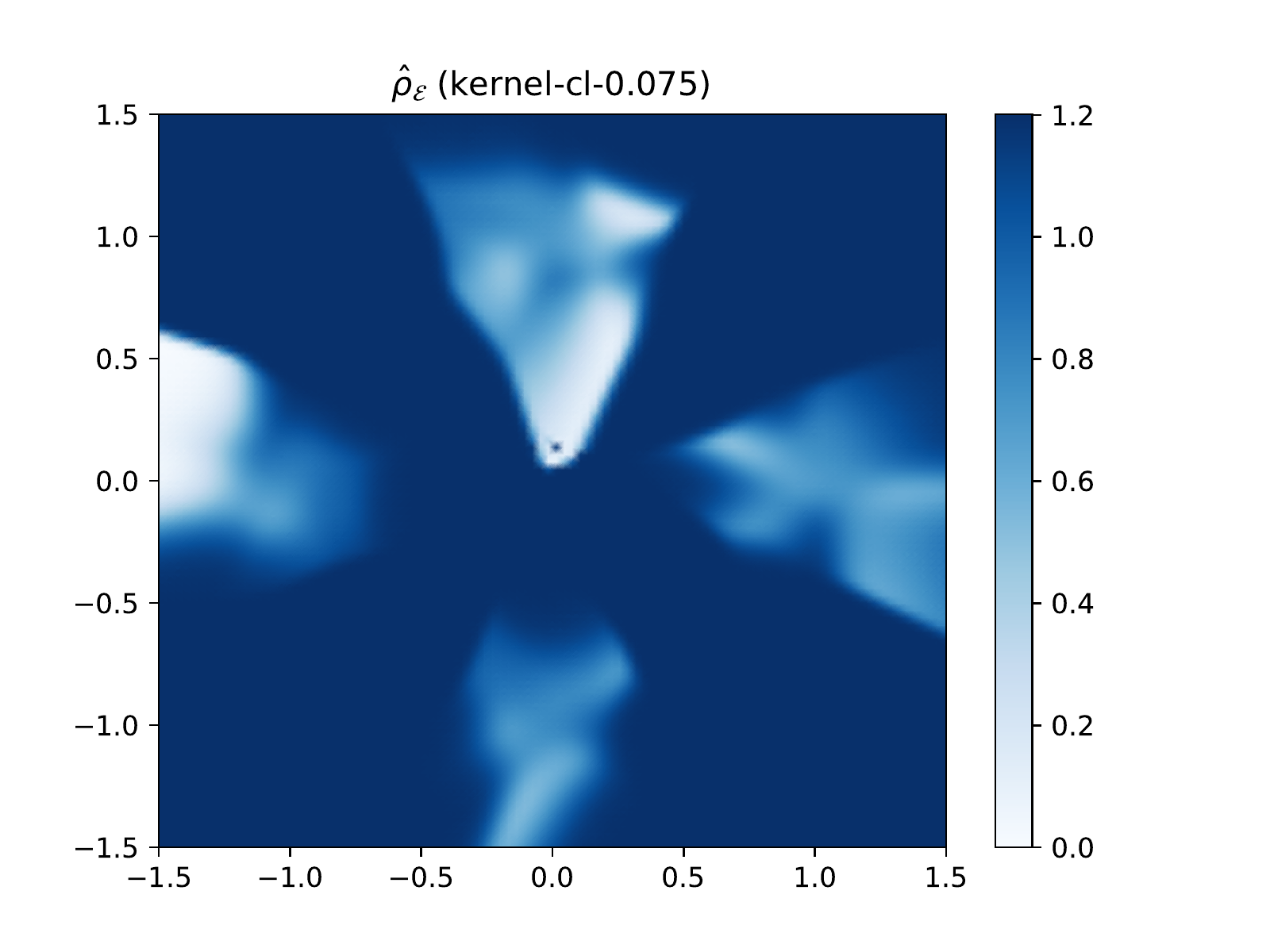}\\
		\includegraphics[trim=25 20 50 37, clip, width=\textwidth]{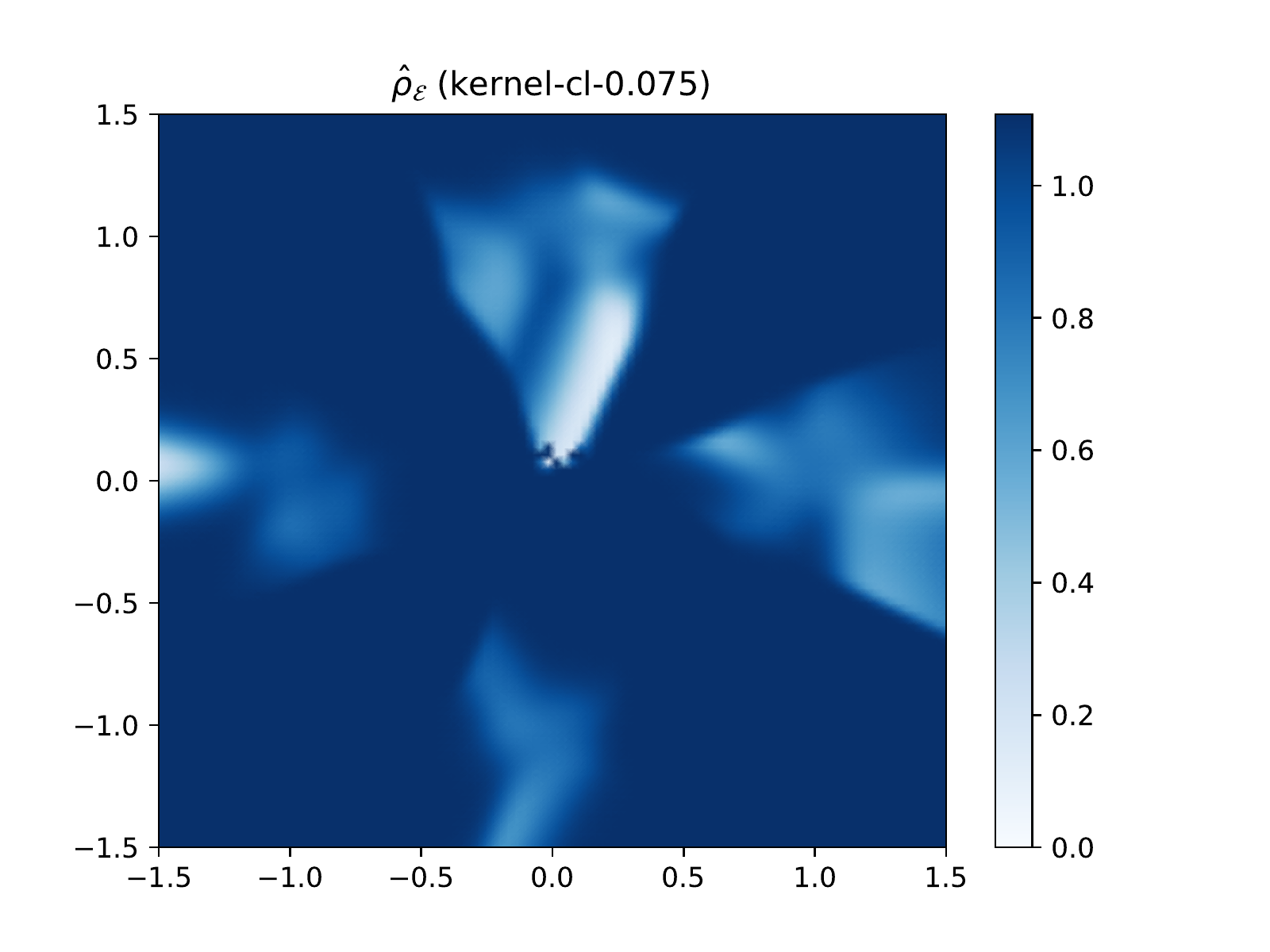}\\
		\includegraphics[trim=25 20 50 37, clip, width=\textwidth]{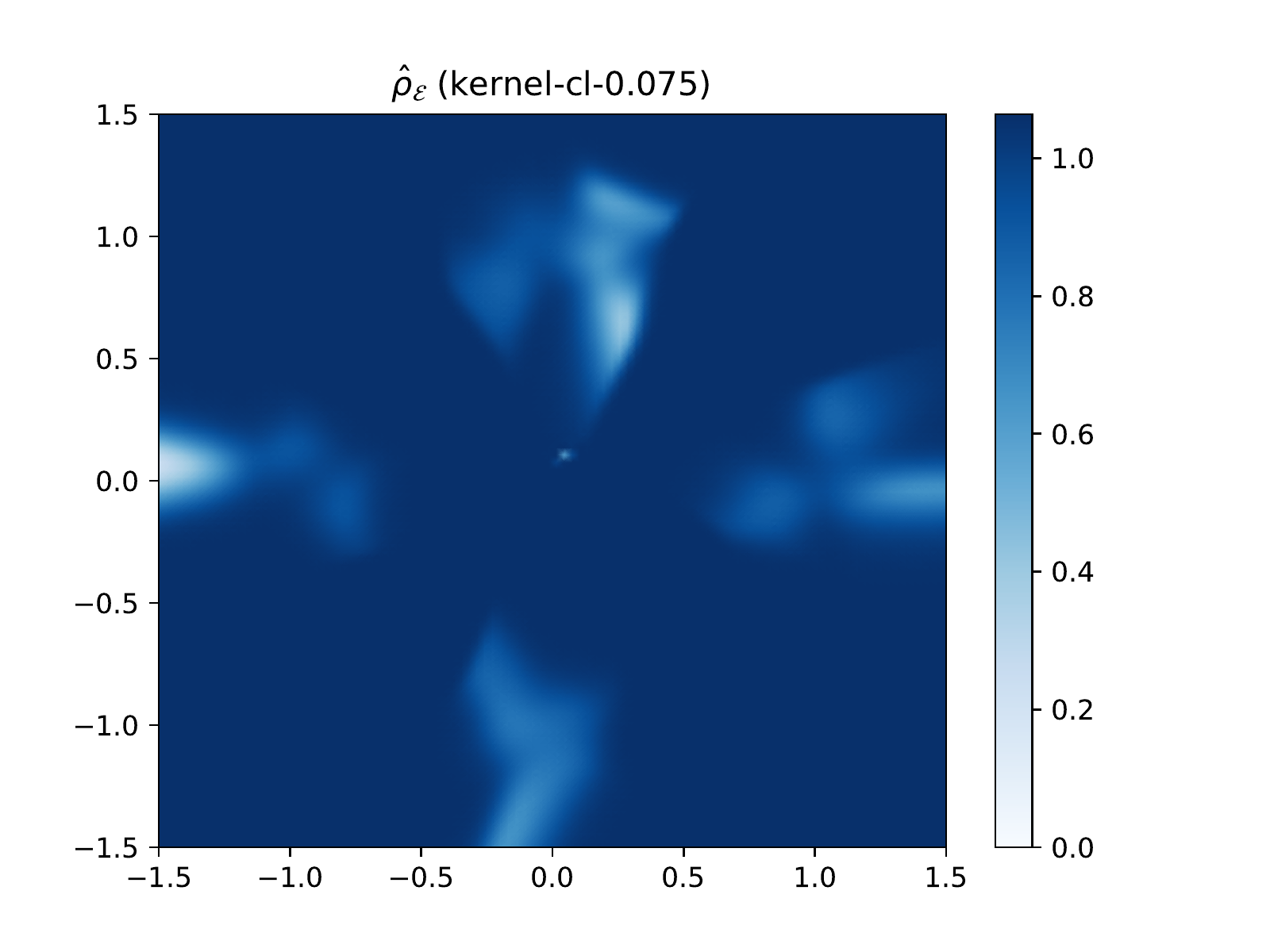}\\
		\caption{KBC ($\sigma_{\mC}$=$0.075$)}
	\end{subfigure}
	\begin{subfigure}[t!]{0.19\textwidth}
		\centering 
		\includegraphics[trim=25 20 50 37, clip, width=\textwidth]{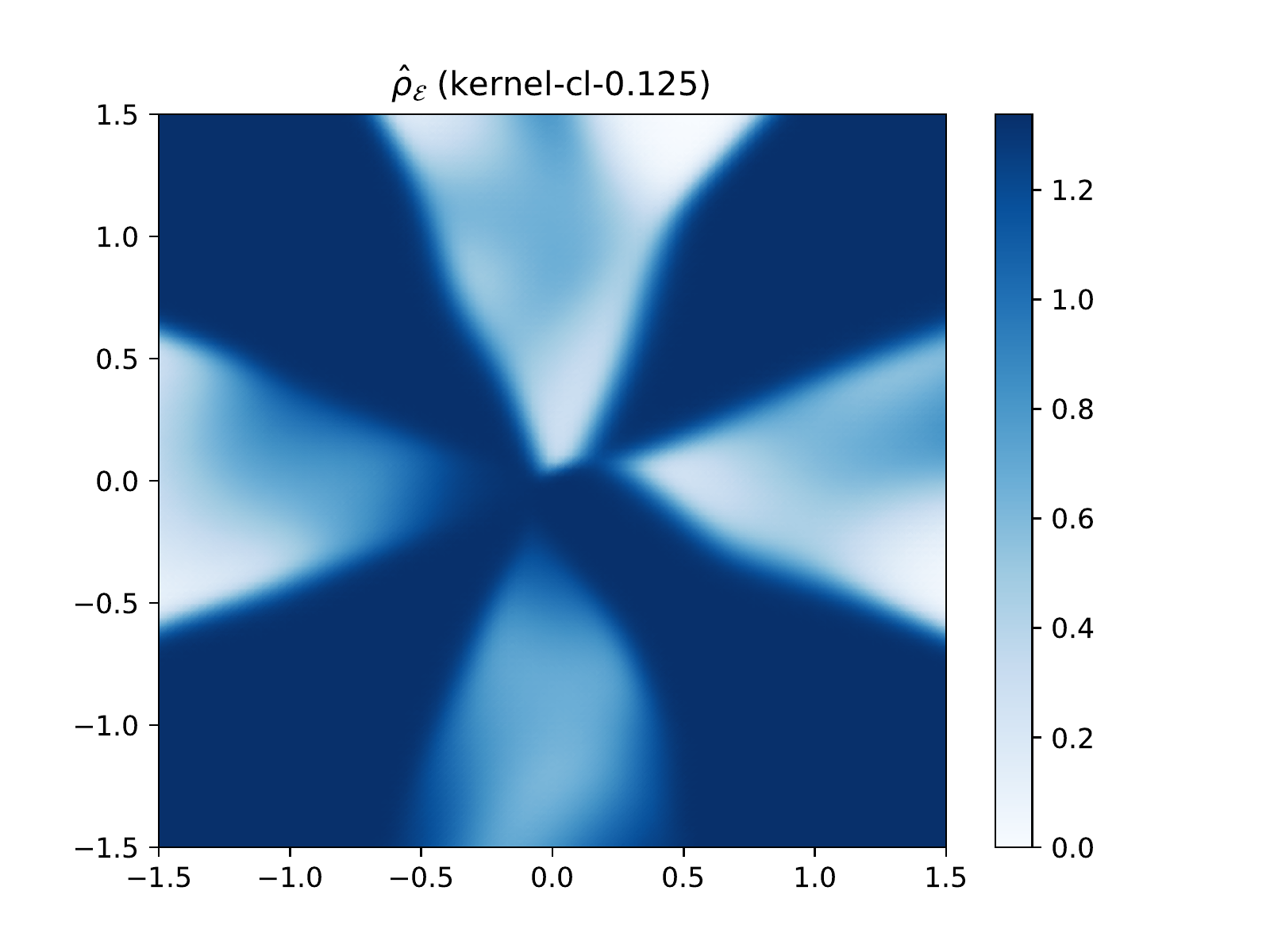}\\
		\includegraphics[trim=25 20 50 37, clip, width=\textwidth]{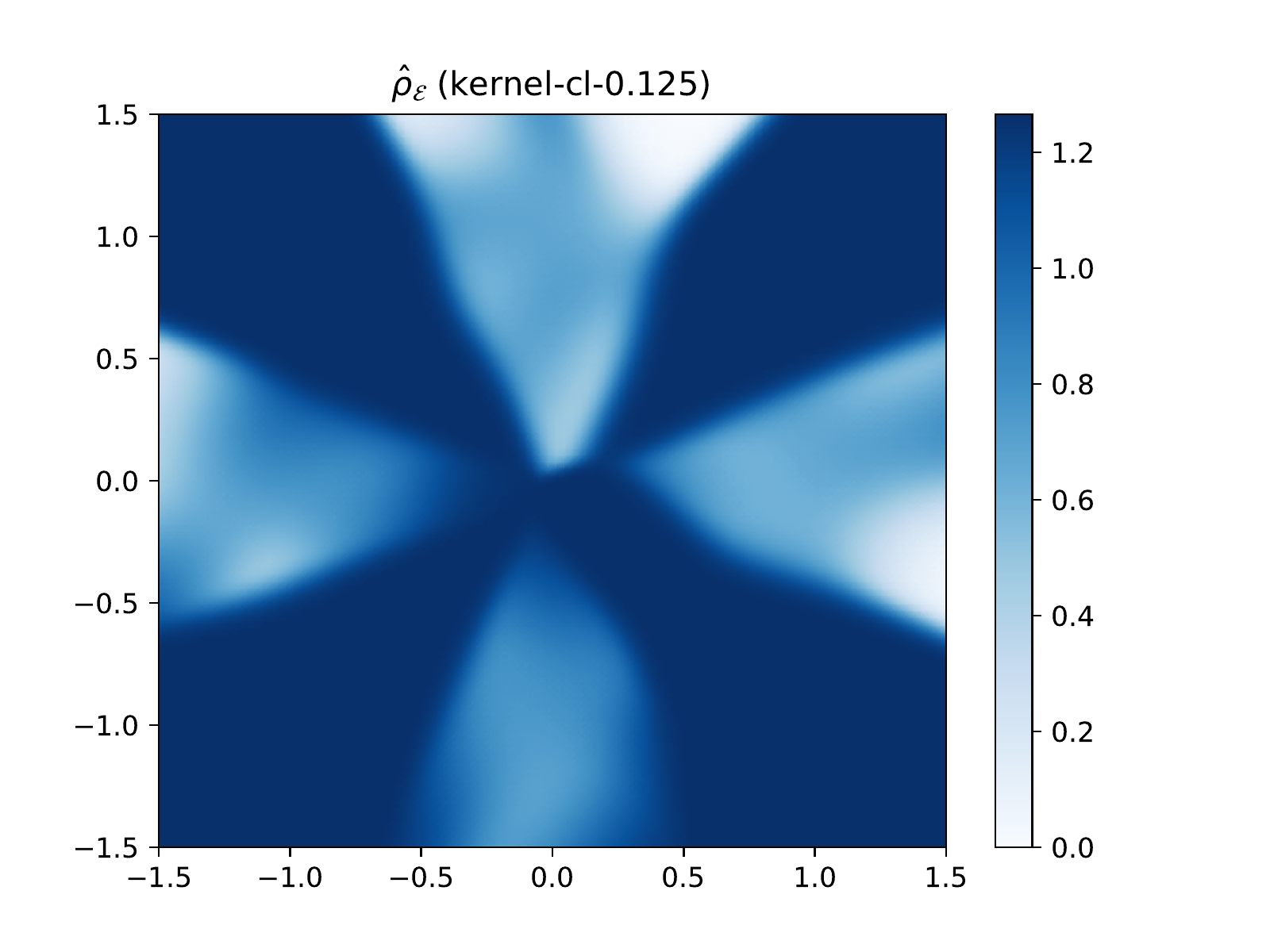}\\
		\includegraphics[trim=25 20 50 37, clip, width=\textwidth]{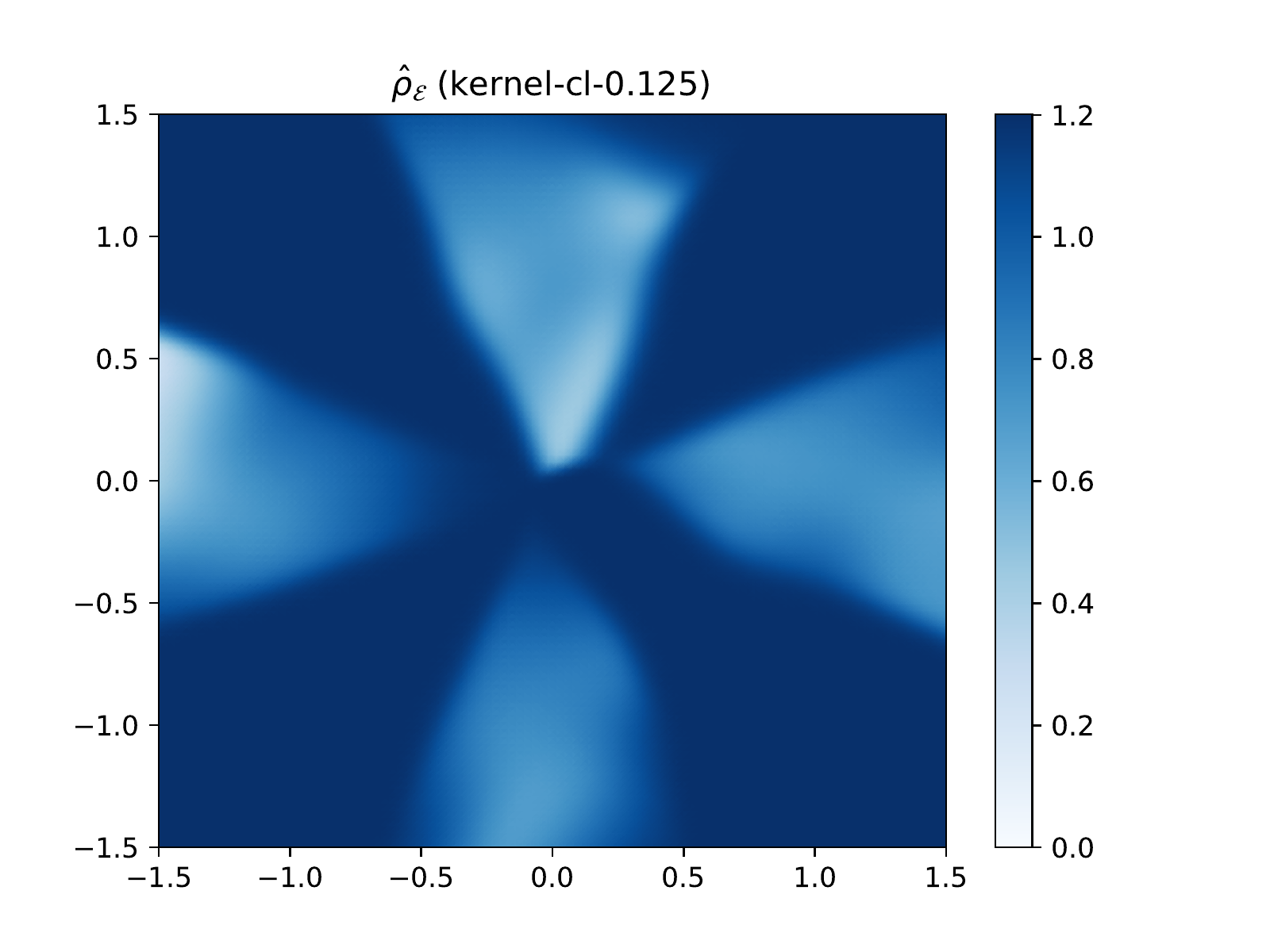}\\
		\includegraphics[trim=25 20 50 37, clip, width=\textwidth]{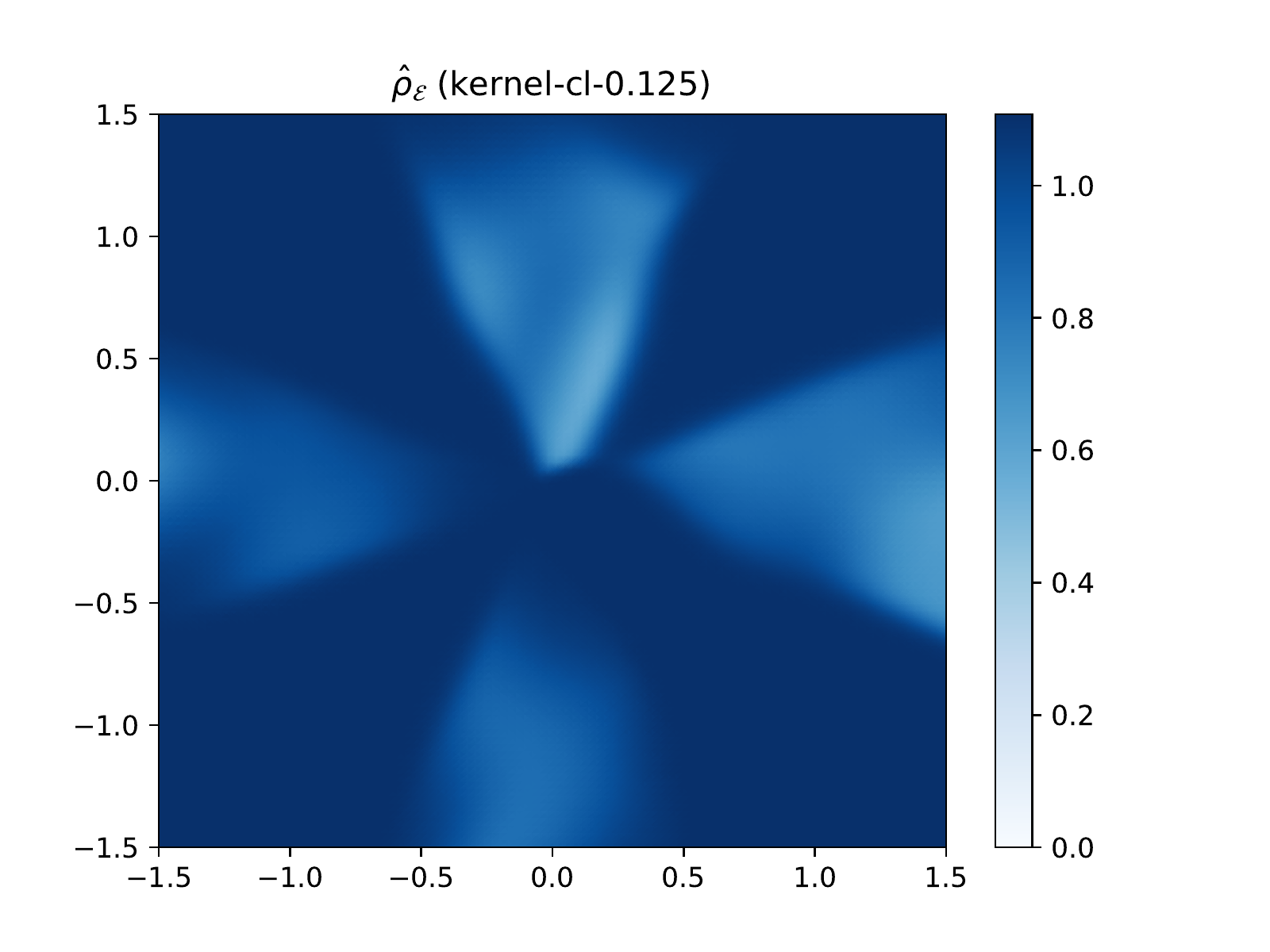}\\
		\includegraphics[trim=25 20 50 37, clip, width=\textwidth]{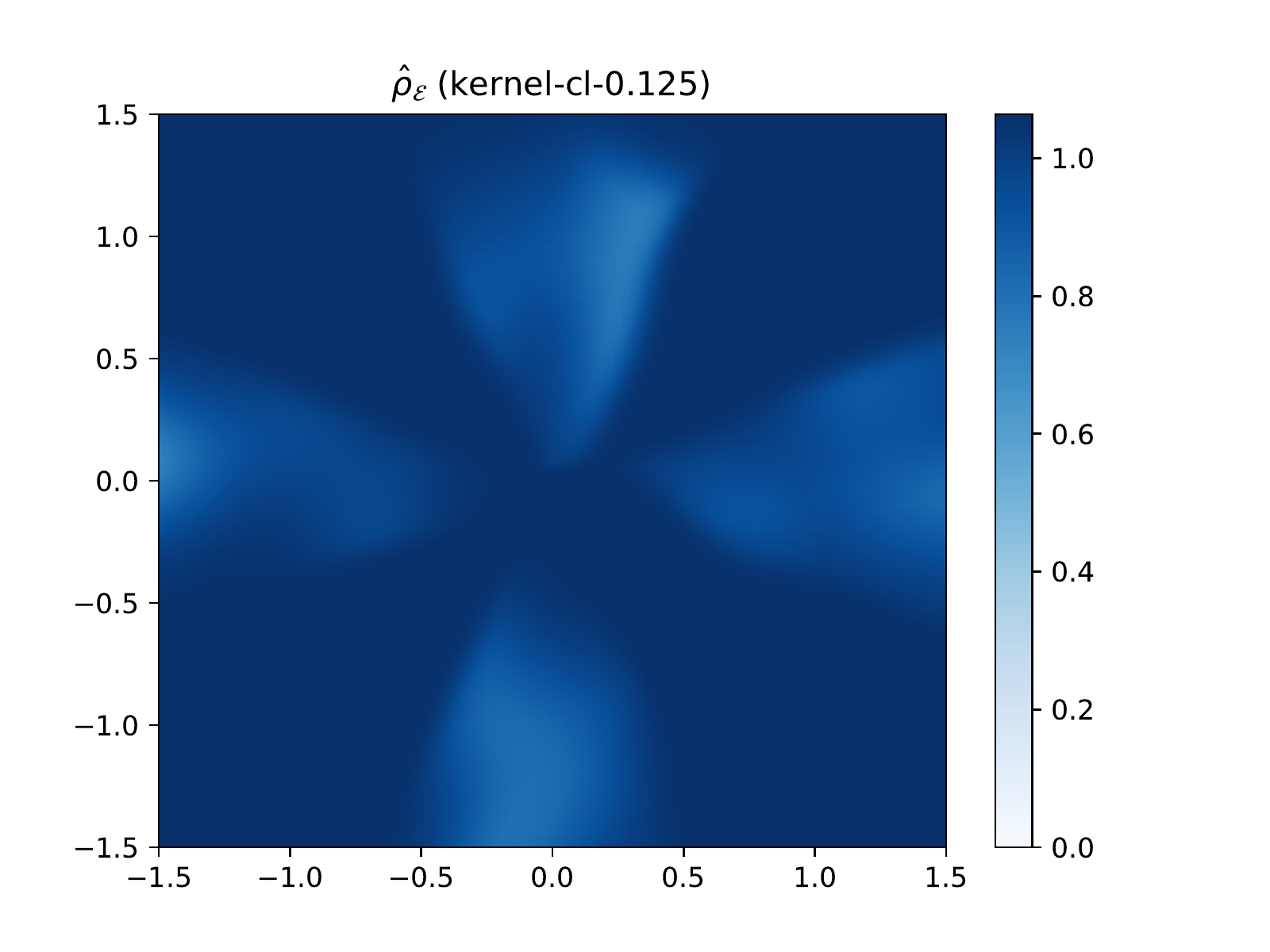}\\
		\caption{KBC ($\sigma_{\mC}$=$0.125$)}
	\end{subfigure}
	\begin{subfigure}[t!]{0.19\textwidth}
		\centering 
		\includegraphics[trim=25 20 50 37, clip, width=\textwidth]{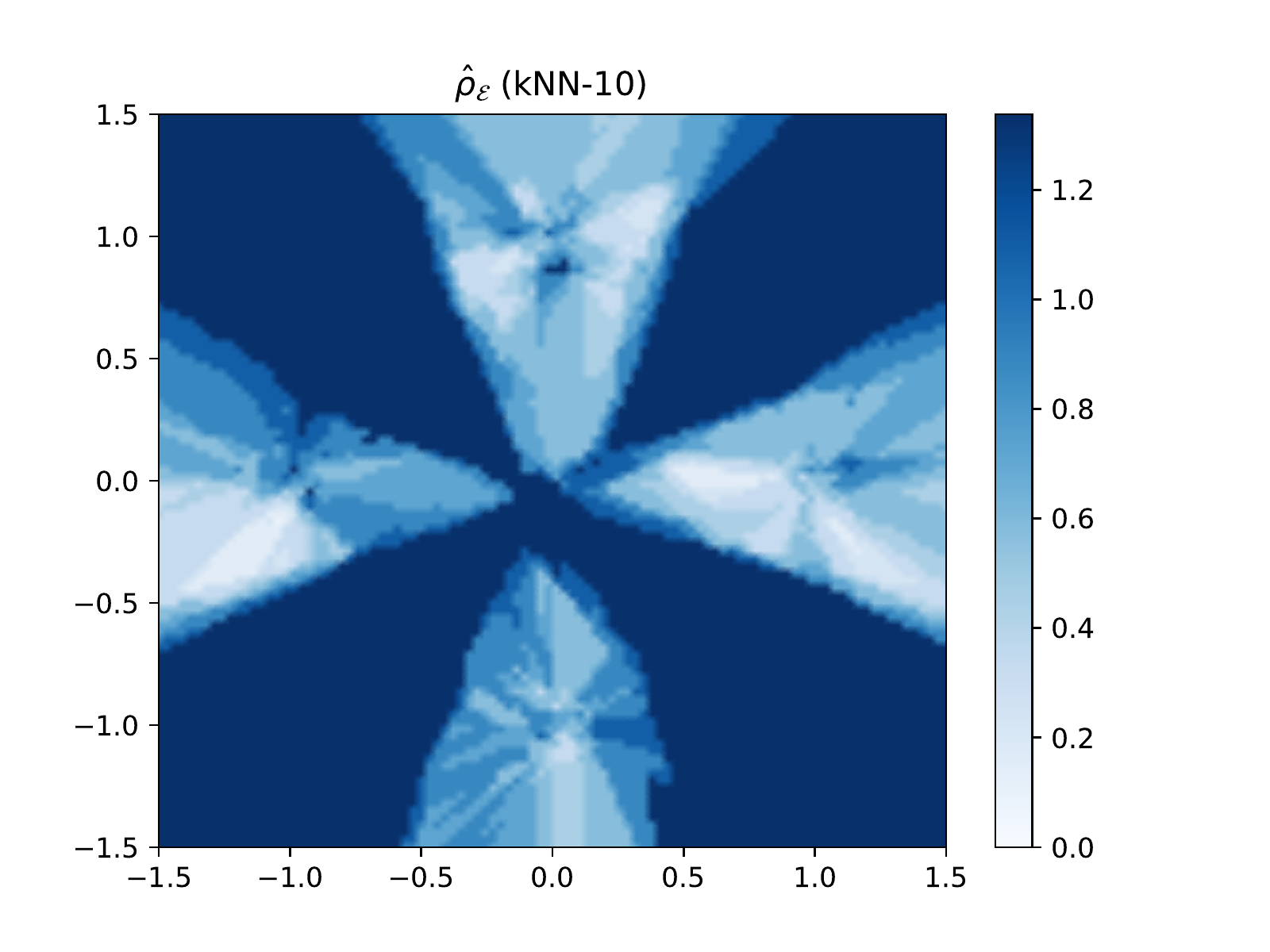}\\
		\includegraphics[trim=25 20 50 37, clip, width=\textwidth]{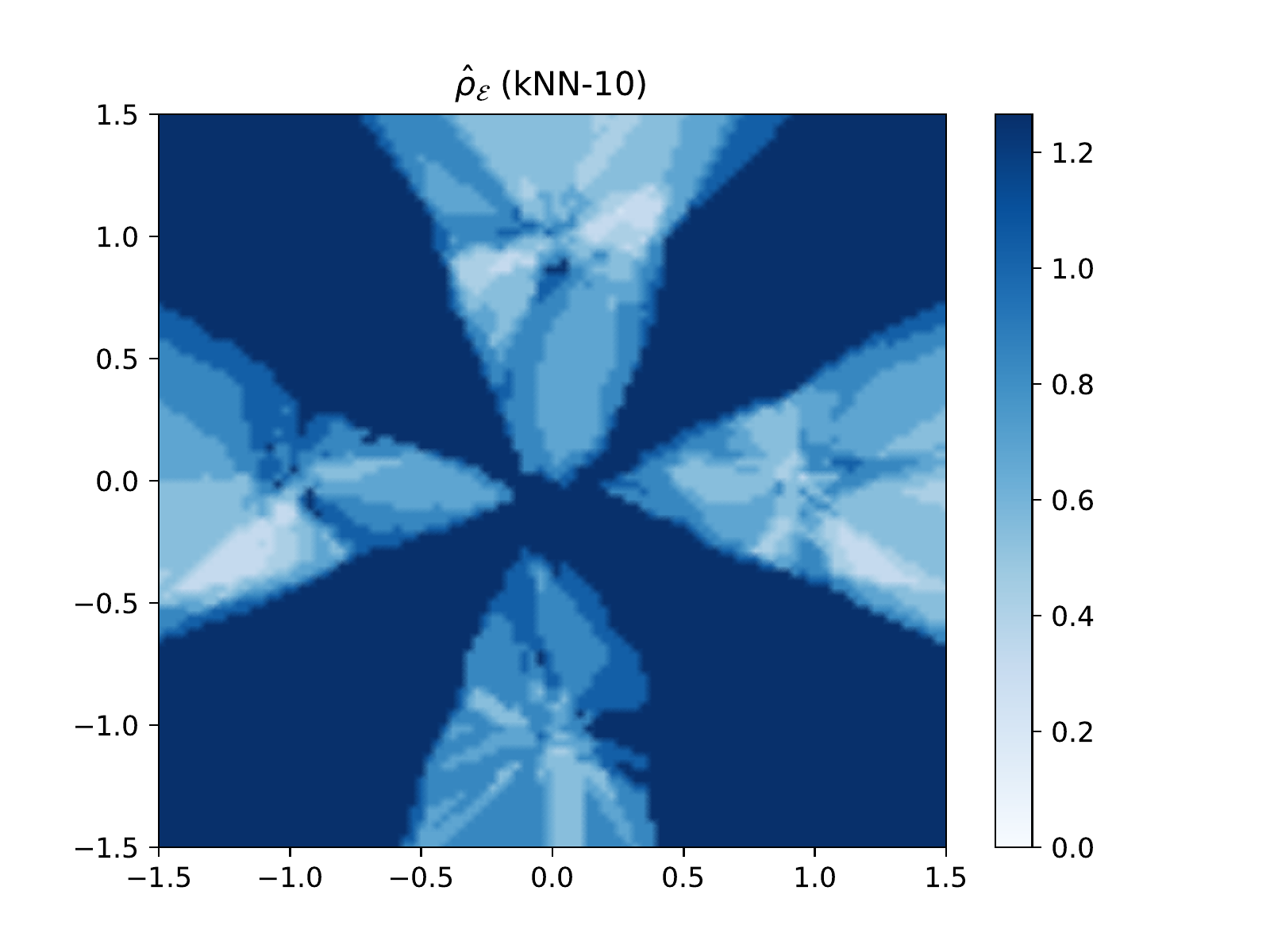}\\
		\includegraphics[trim=25 20 50 37, clip, width=\textwidth]{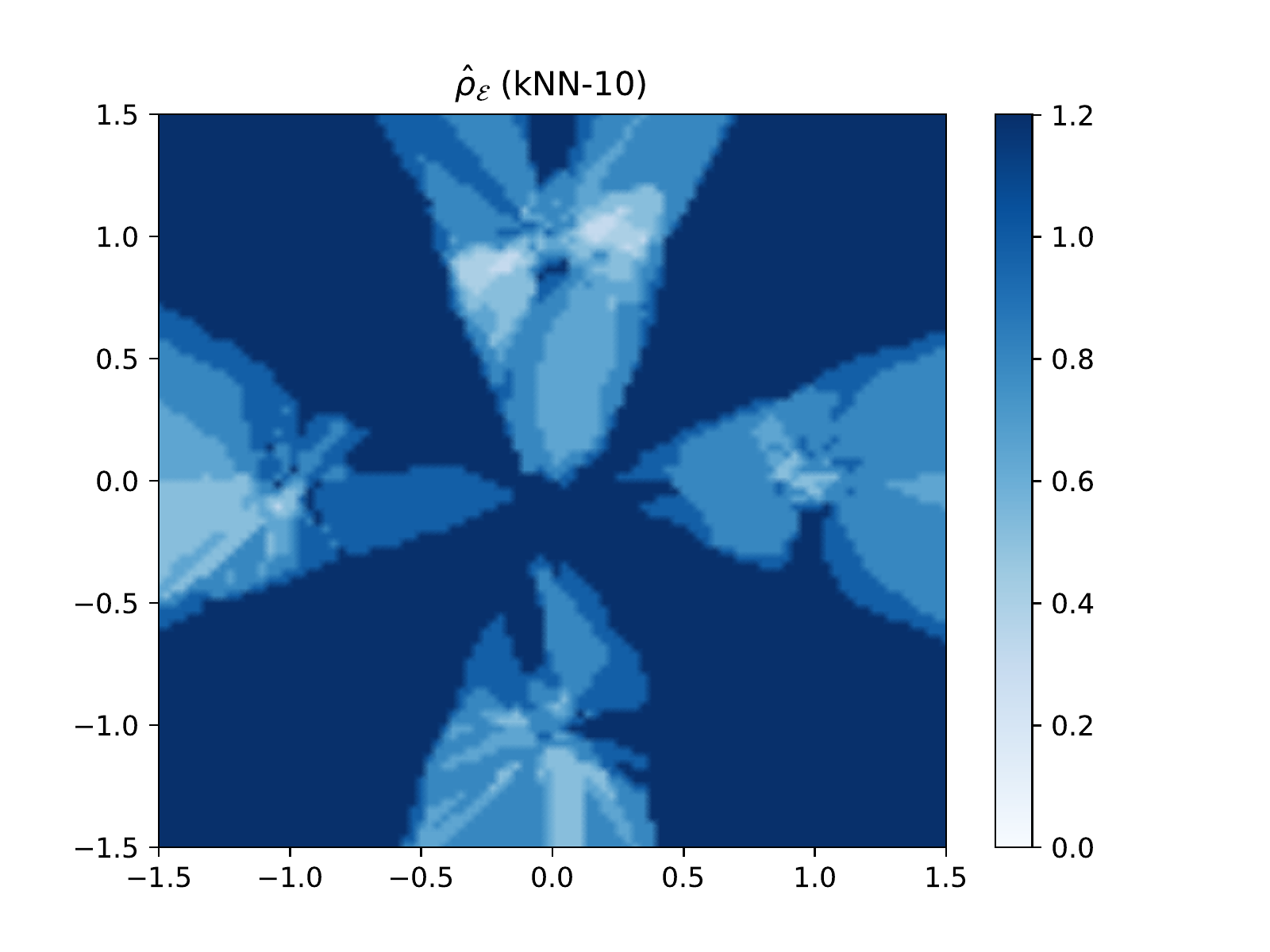}\\
		\includegraphics[trim=25 20 50 37, clip, width=\textwidth]{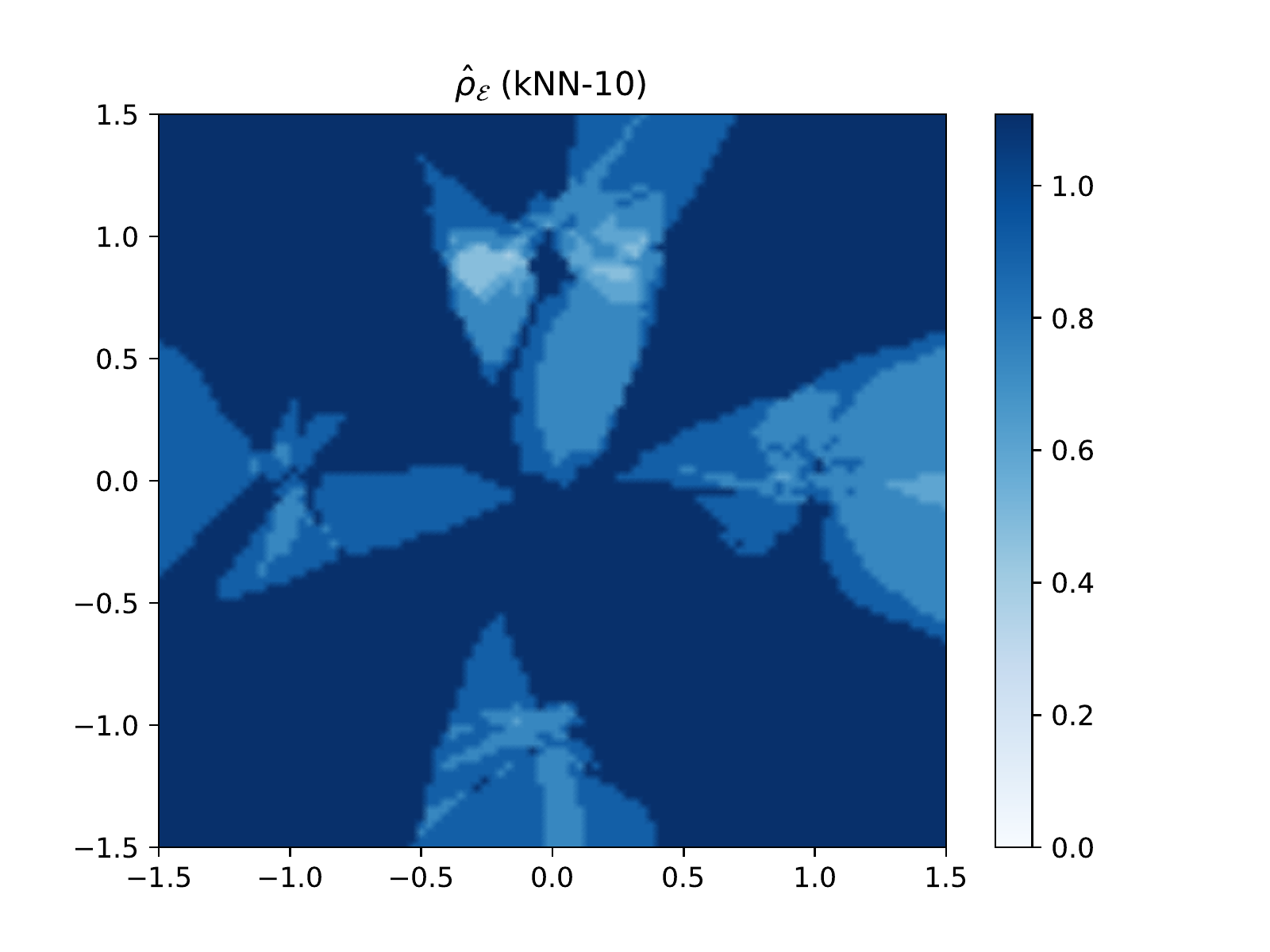}\\
		\includegraphics[trim=25 20 50 37, clip, width=\textwidth]{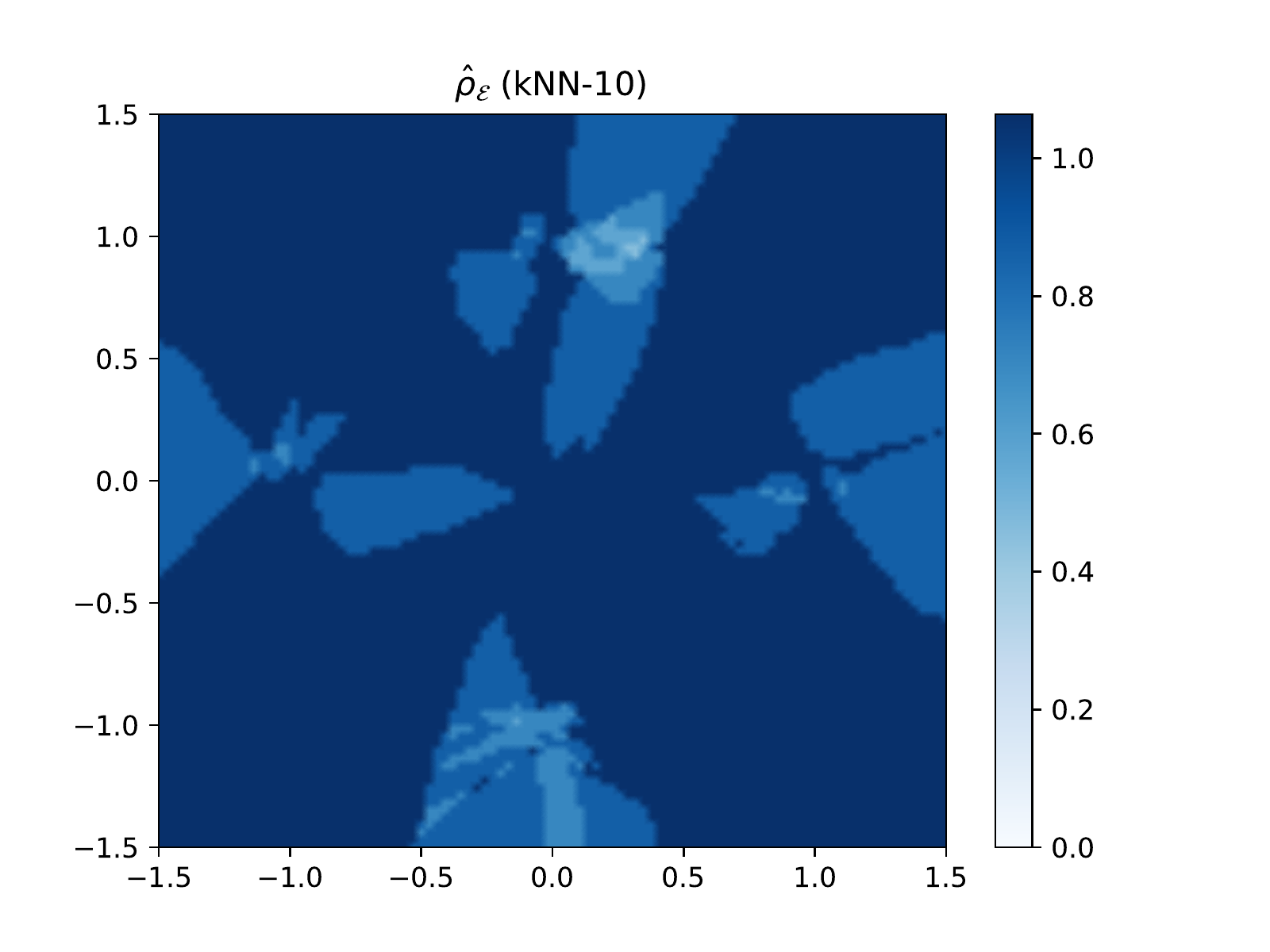}\\
		\caption{$k$-NN ($k$=$10$)}
	\end{subfigure}
	\begin{subfigure}[t!]{0.19\textwidth}
		\centering 
		\includegraphics[trim=25 20 50 37, clip, width=\textwidth]{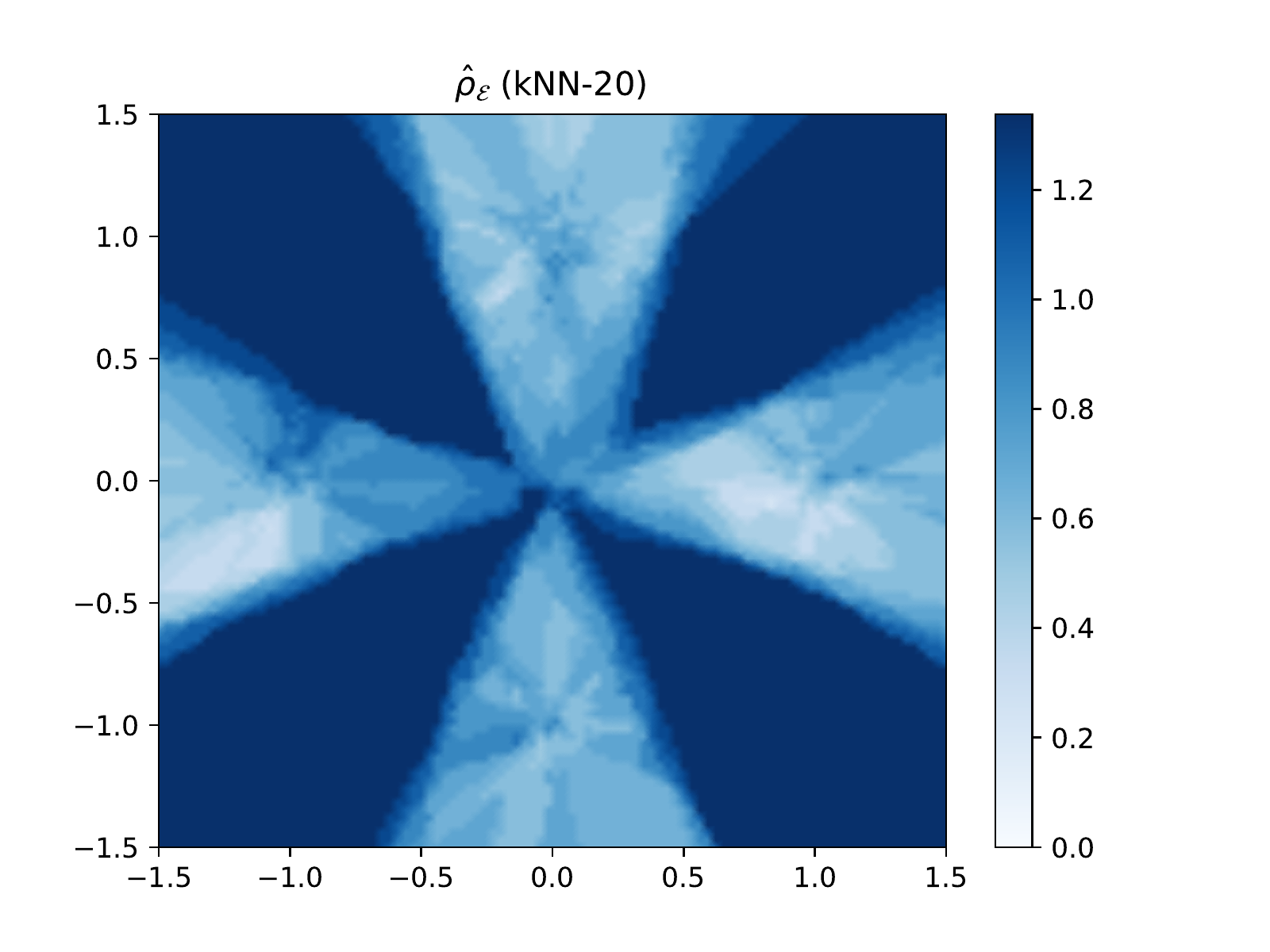}\\
		\includegraphics[trim=25 20 50 37, clip, width=\textwidth]{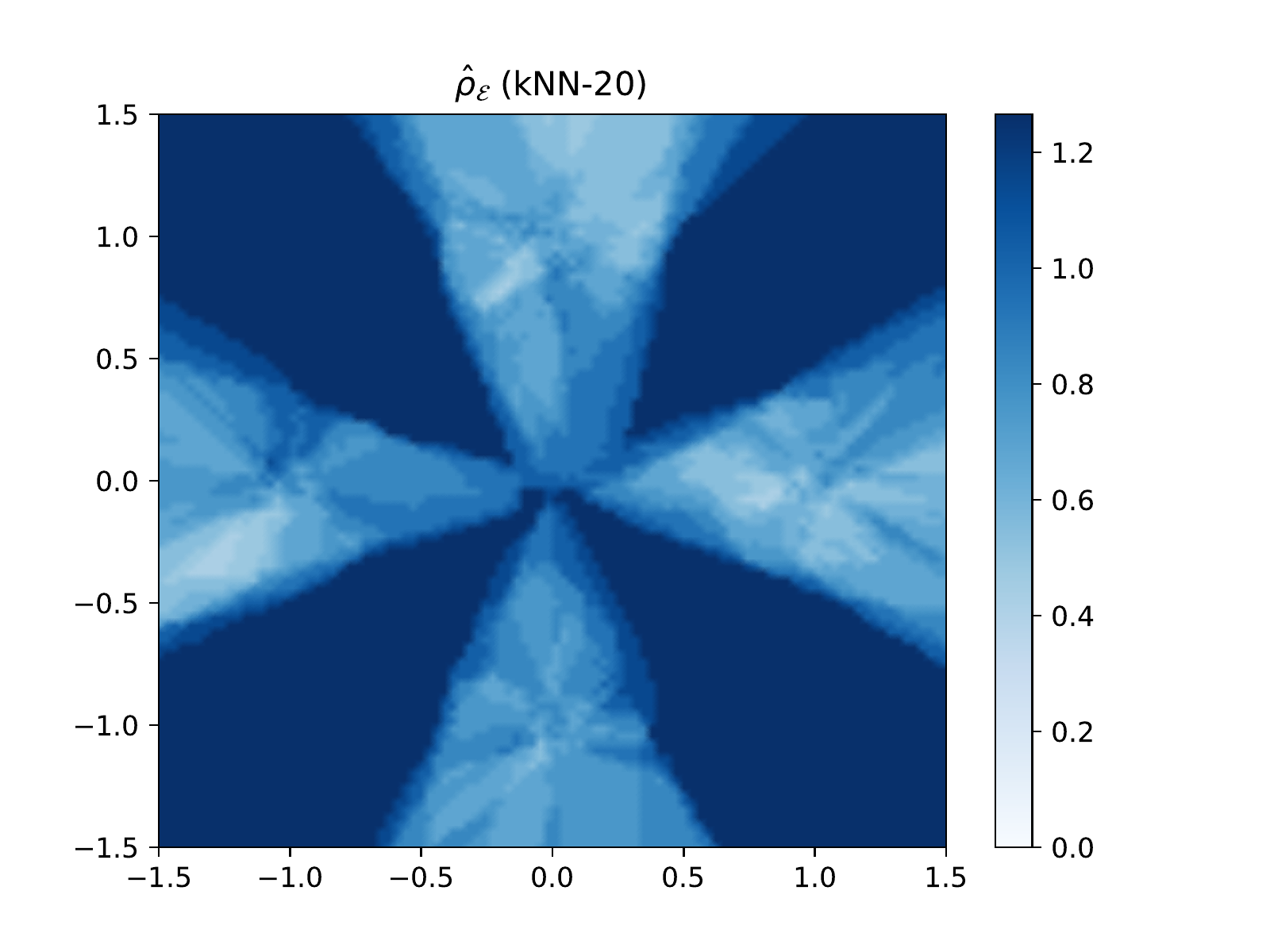}\\
		\includegraphics[trim=25 20 50 37, clip, width=\textwidth]{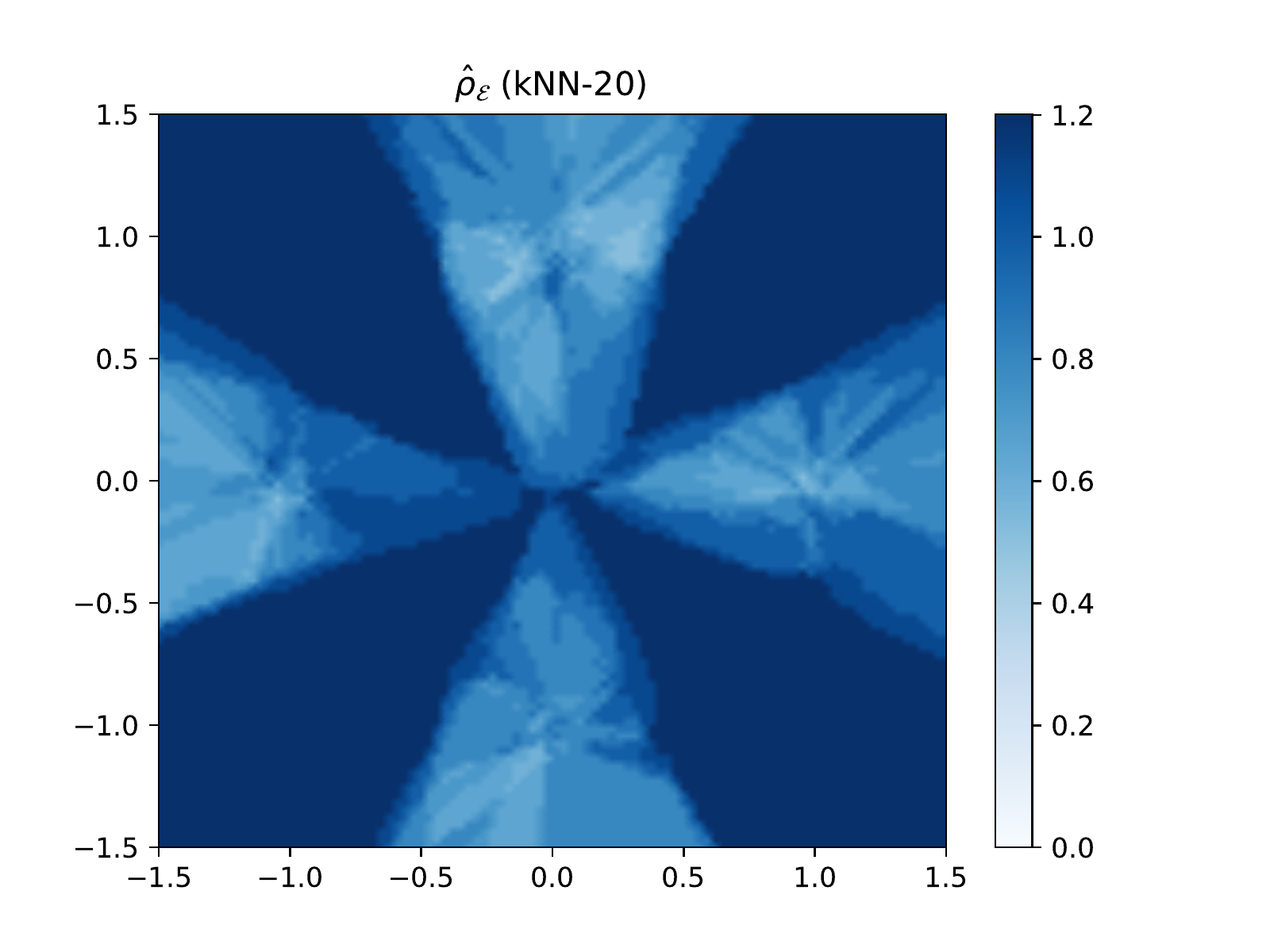}\\
		\includegraphics[trim=25 20 50 37, clip, width=\textwidth]{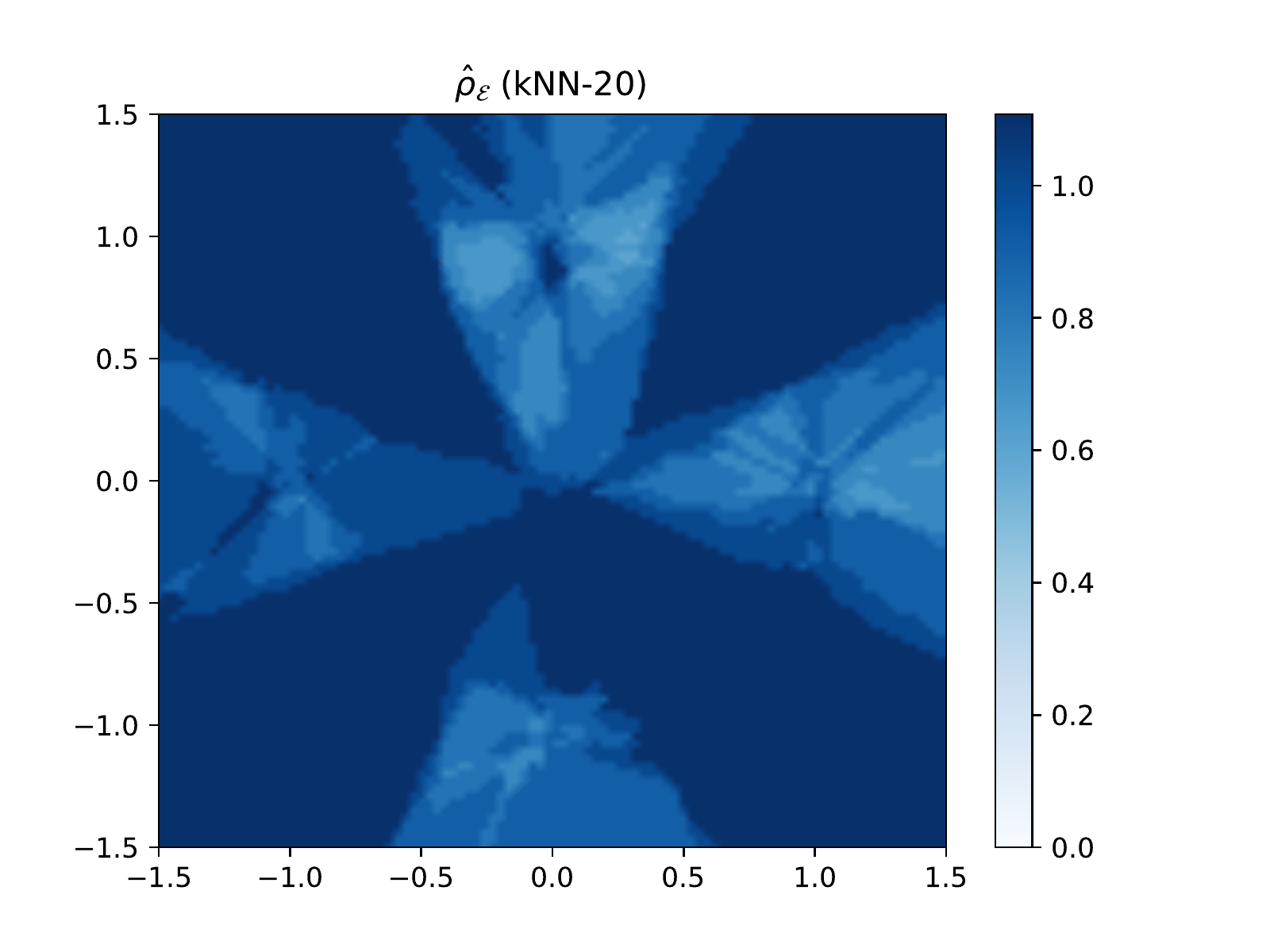}\\
		\includegraphics[trim=25 20 50 37, clip, width=\textwidth]{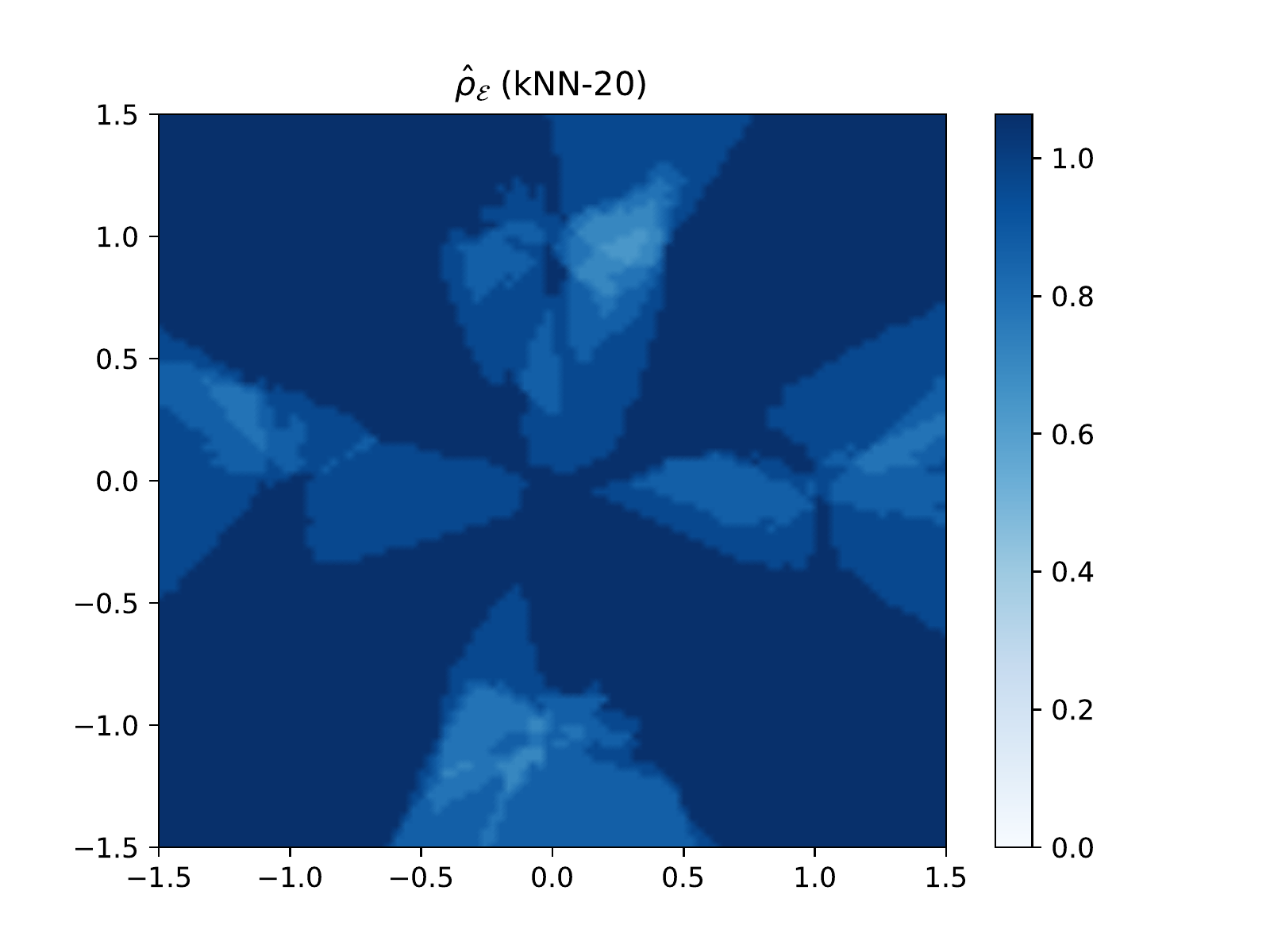}\\
		\caption{$k$-NN ($k$=$20$)}
	\end{subfigure}
	
	\vspace{-0.3em}
	\caption{Visualization of ratio $\hat{\rho}$ in (a) and $\hat{\rho}_{\mE}$ in (b)-(e) for different classifier-based DREs.}
	\label{fig: 2d Q1 DRE MoG-8 appendix}
	\vspace{-0.3em}
\end{figure}

We visualize $\hat{\rho}$ and $\hat{\rho}_{\mE}$ in Fig. \ref{fig: 2d Q1 DRE MoG-8 appendix} (extension of Fig. \ref{fig: 2d Q1 DRE}). These figures give qualitative answers to question 1. 

\begin{figure}[!h]
\vspace{-0.3em}
  	\begin{subfigure}[t!]{0.5\textwidth}
	\centering 
	\includegraphics[trim=30 0 0 5, clip, width=0.95\textwidth]{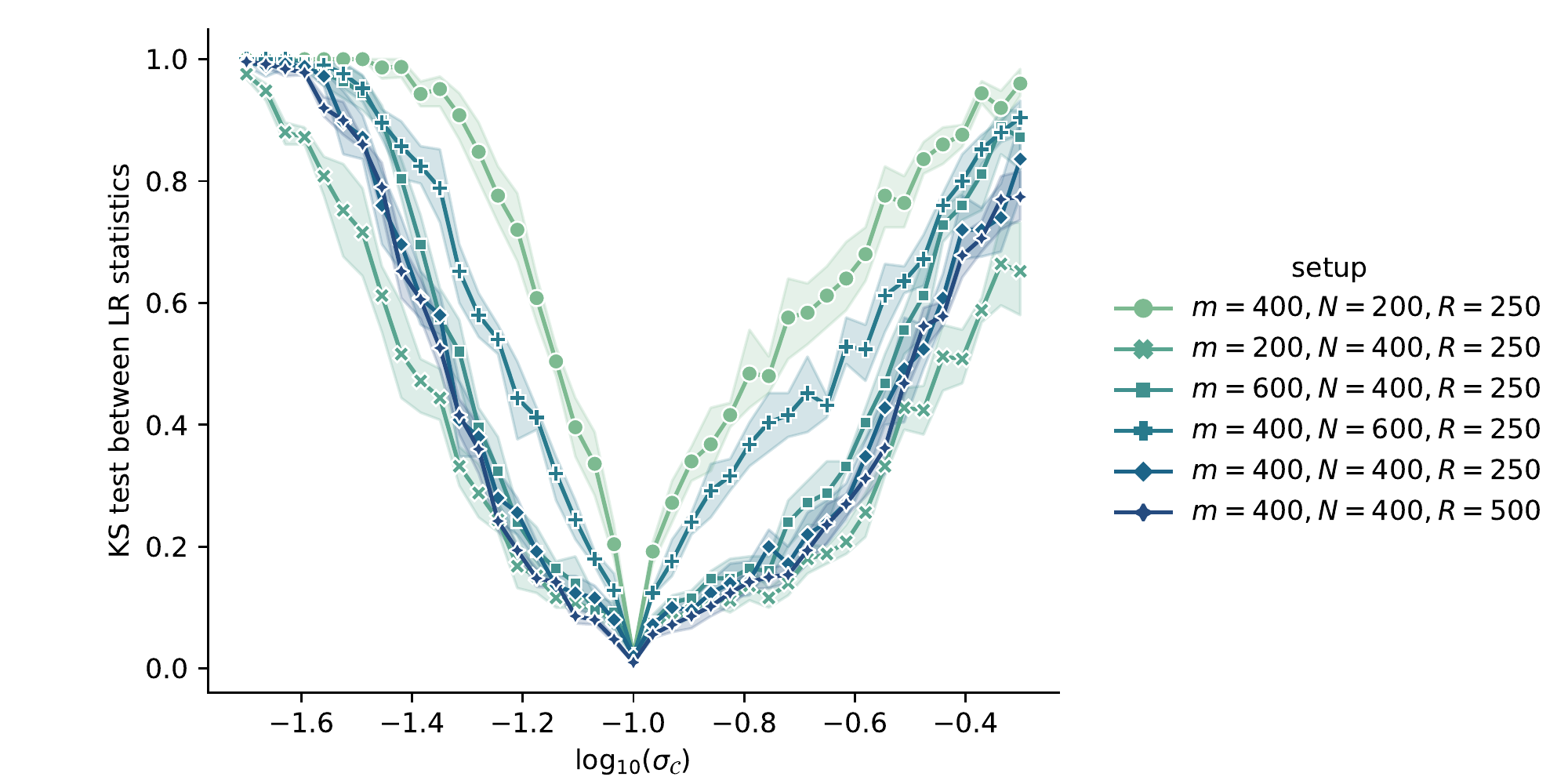}
	\caption{$\mathrm{LR}$ statistics with $\lambda=0.8$ and different $m,N,R$}
	\end{subfigure}
	\begin{subfigure}[t!]{0.5\textwidth}
	\centering 
	\includegraphics[trim=30 0 0 5, clip, width=0.95\textwidth]{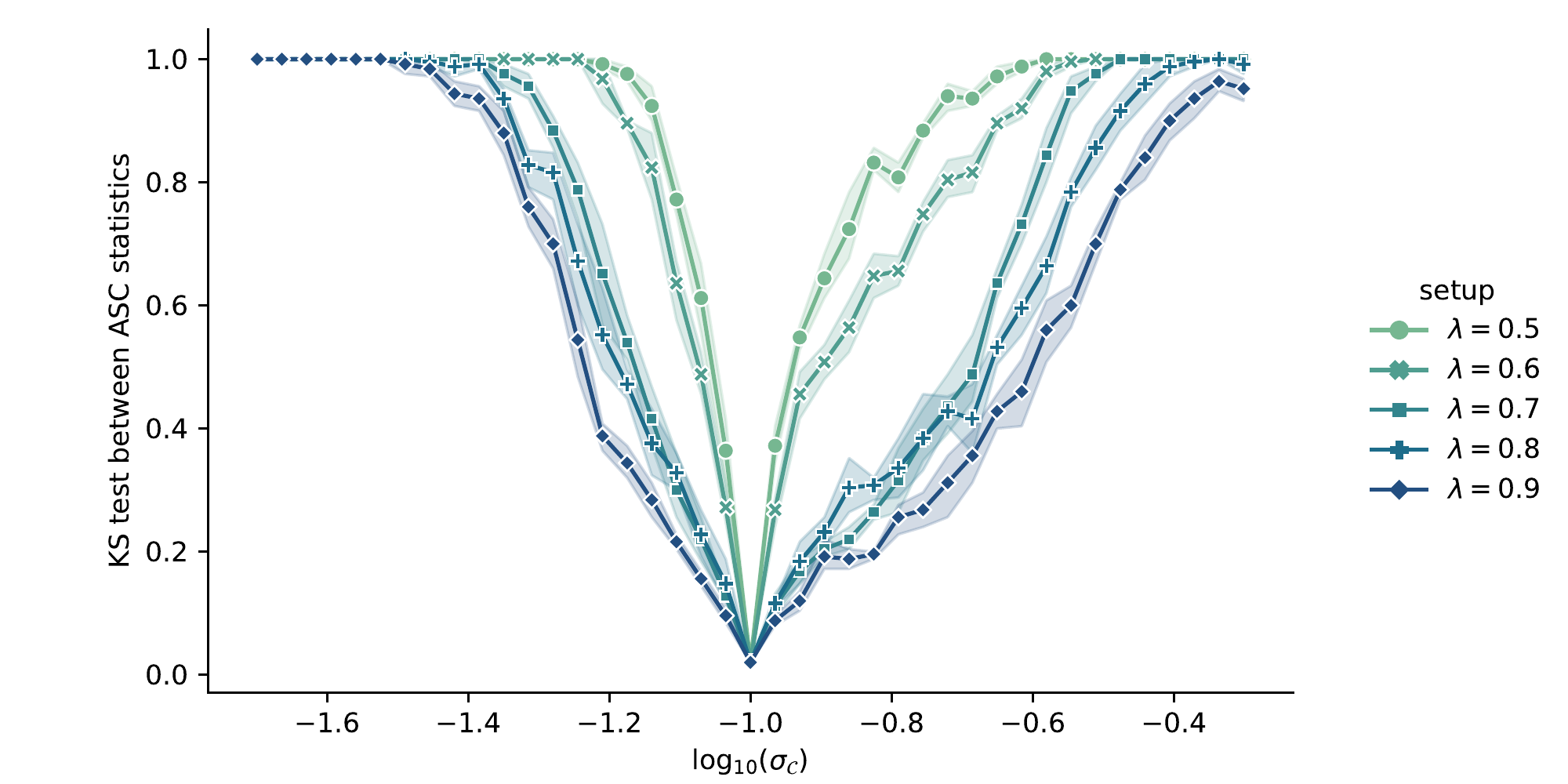}
	\caption{$\mathrm{ASC}$ statistics with $\phi(t)=\log(t)$}
	\end{subfigure}\\
	\begin{subfigure}[t!]{0.5\textwidth}
	\centering 
	\includegraphics[trim=30 0 0 5, clip, width=0.95\textwidth]{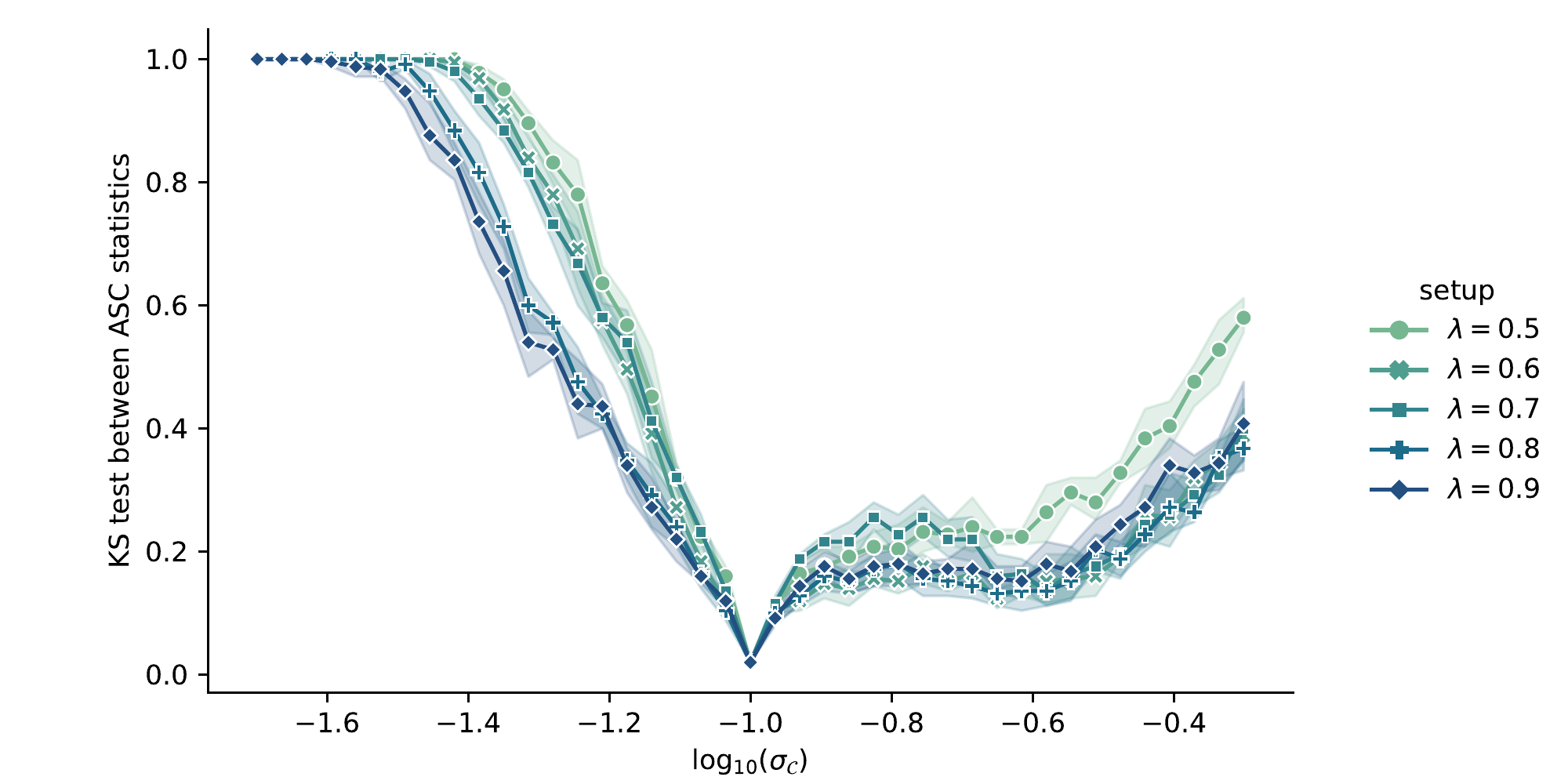}
	\caption{$\mathrm{ASC}$ statistics with $\phi(t)=t\log(t)$ (KL)}
	\end{subfigure}
	\begin{subfigure}[t!]{0.5\textwidth}
	\centering 
	\includegraphics[trim=30 0 0 5, clip, width=0.95\textwidth]{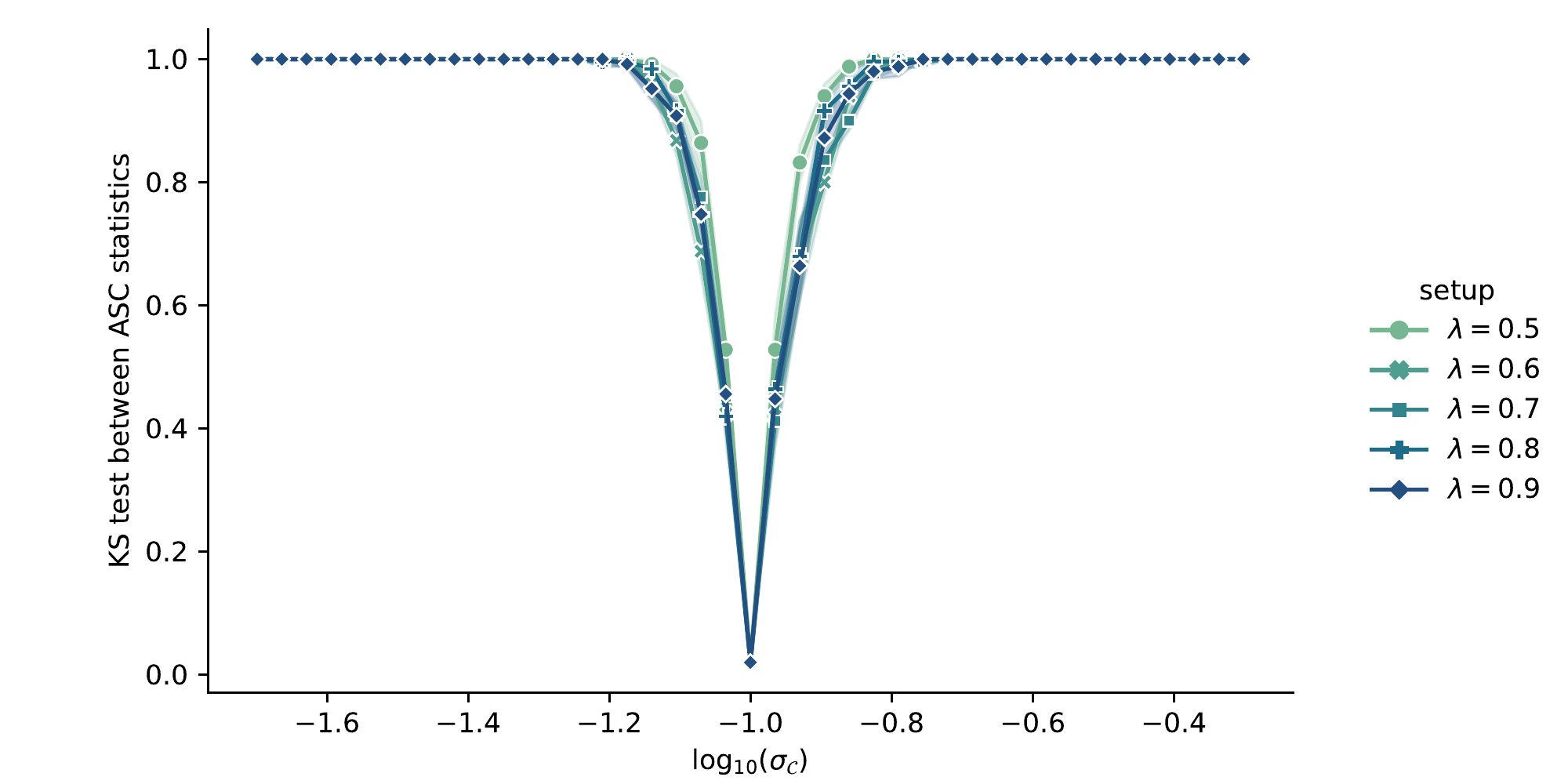}
	\caption{$\mathrm{ASC}$ statistics with $\phi(t)=(\sqrt{t}-1)^2$ (Hellinger)}
	\end{subfigure}
	
	\vspace{-0.3em}
	\caption{KS tests between distributions of statistics for KBC with different $\sigma_{\mC}$. (a) $\mathrm{LR}(Y_{H_0},\hat{\rho})$ vs $\mathrm{LR}(Y_{H_0},\hat{\rho}_{\mE})$ with $\lambda=0.8$ and different $m,N,R$, complementary to Fig. \ref{fig: 2d Q1 KS}. (b)-(d) $\hat{\mathrm{ASC}}_{\phi}(\hat{Y},Y_{H_0},\hat{\rho})$ vs $\hat{\mathrm{ASC}}_{\phi}(\hat{Y},Y_{H_0},\hat{\rho}_{\mE})$ for different $\phi$. Smaller values indicate the two compared distributions are closer. }
	\vspace{-0.3em}
	\label{fig: 2d Q1 KS MoG-8 appendix}
\end{figure}

\newpage
We visualize KS test results for KBC with different bandwidth $\sigma_{\mC}$ in Fig. \ref{fig: 2d Q1 KS MoG-8 appendix} (extension of Fig. \ref{fig: 2d Q1 KS}). When $\sigma_{\mC}\approx\sigma_{\mA}=0.1$, the KS values are small, indicating KBC with these $\sigma_{\mC}$ can lead to classifier-based DRE $\hat{\rho}_{\mE}$ that is close to $\hat{\rho}$.  In terms of statistics, the estimation is most accurate under KL and least accurate under Hellinger distance. In terms of $\lambda$, the estimation is more accurate for larger $\lambda$, where less data are deleted, as expected.

\begin{figure}[!h]
\vspace{-0.5em}
  	\begin{subfigure}[t!]{0.5\textwidth}
	\centering 
	\includegraphics[width=0.8\textwidth]{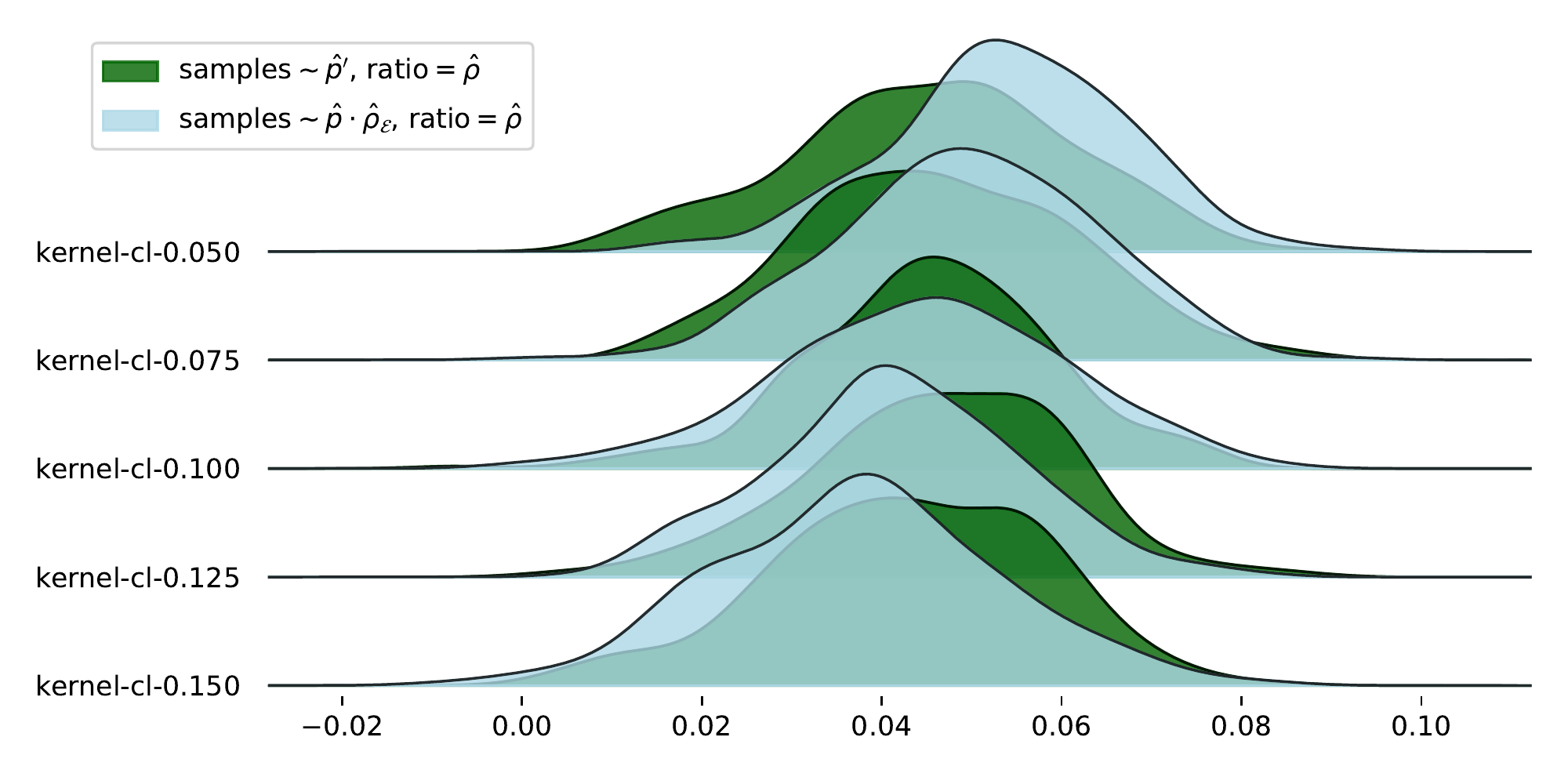}
	\caption{$\mathrm{LR}$ for KBC-based DRE ($\lambda=0.6$)}
	\end{subfigure}
	\begin{subfigure}[t!]{0.5\textwidth}
	\centering 
	\includegraphics[width=0.8\textwidth]{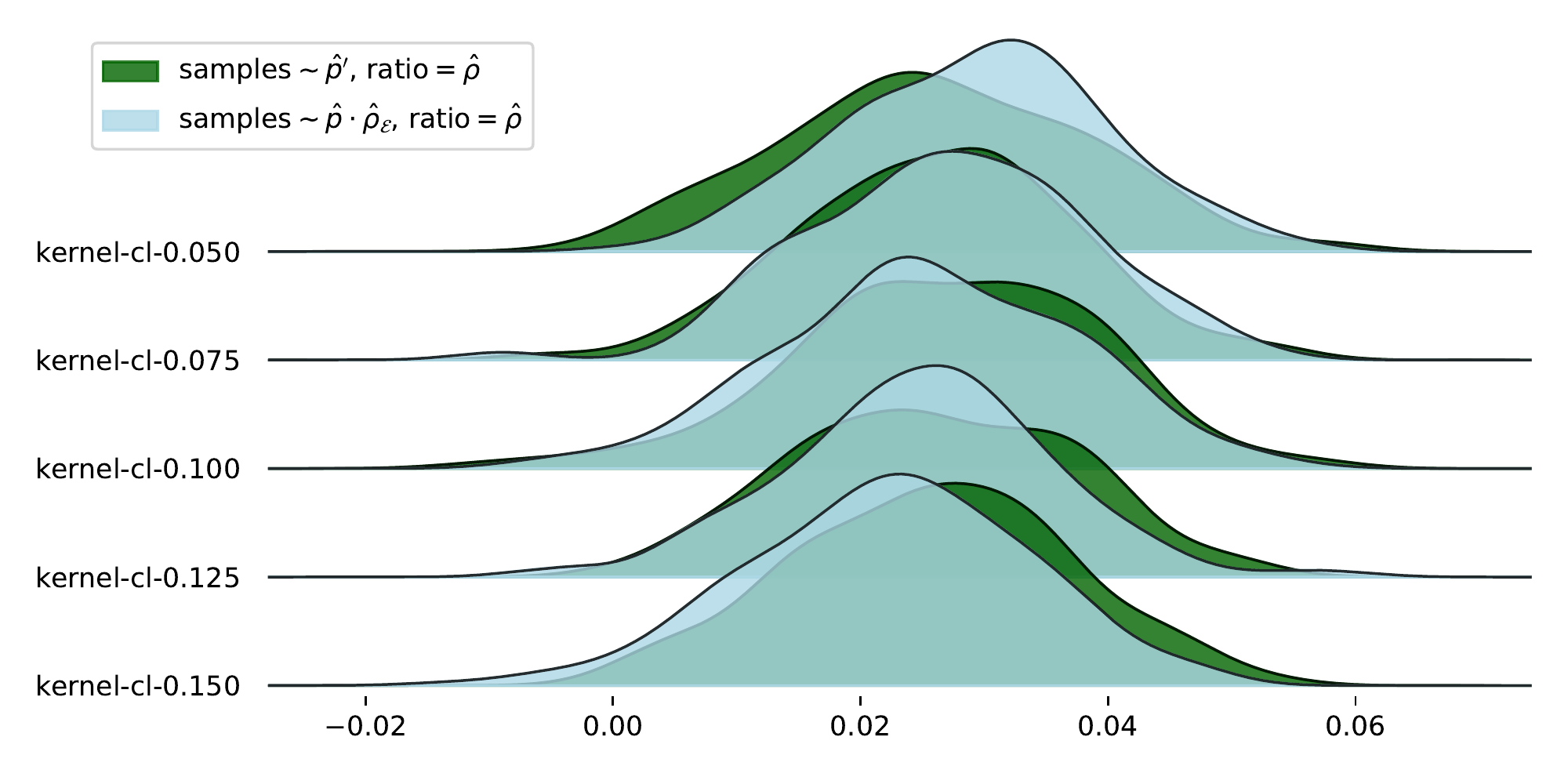}
	\caption{$\mathrm{LR}$ for KBC-based DRE ($\lambda=0.7$)}
	\end{subfigure}\\
	\begin{subfigure}[t!]{0.5\textwidth}
	\centering 
	\includegraphics[width=0.8\textwidth]{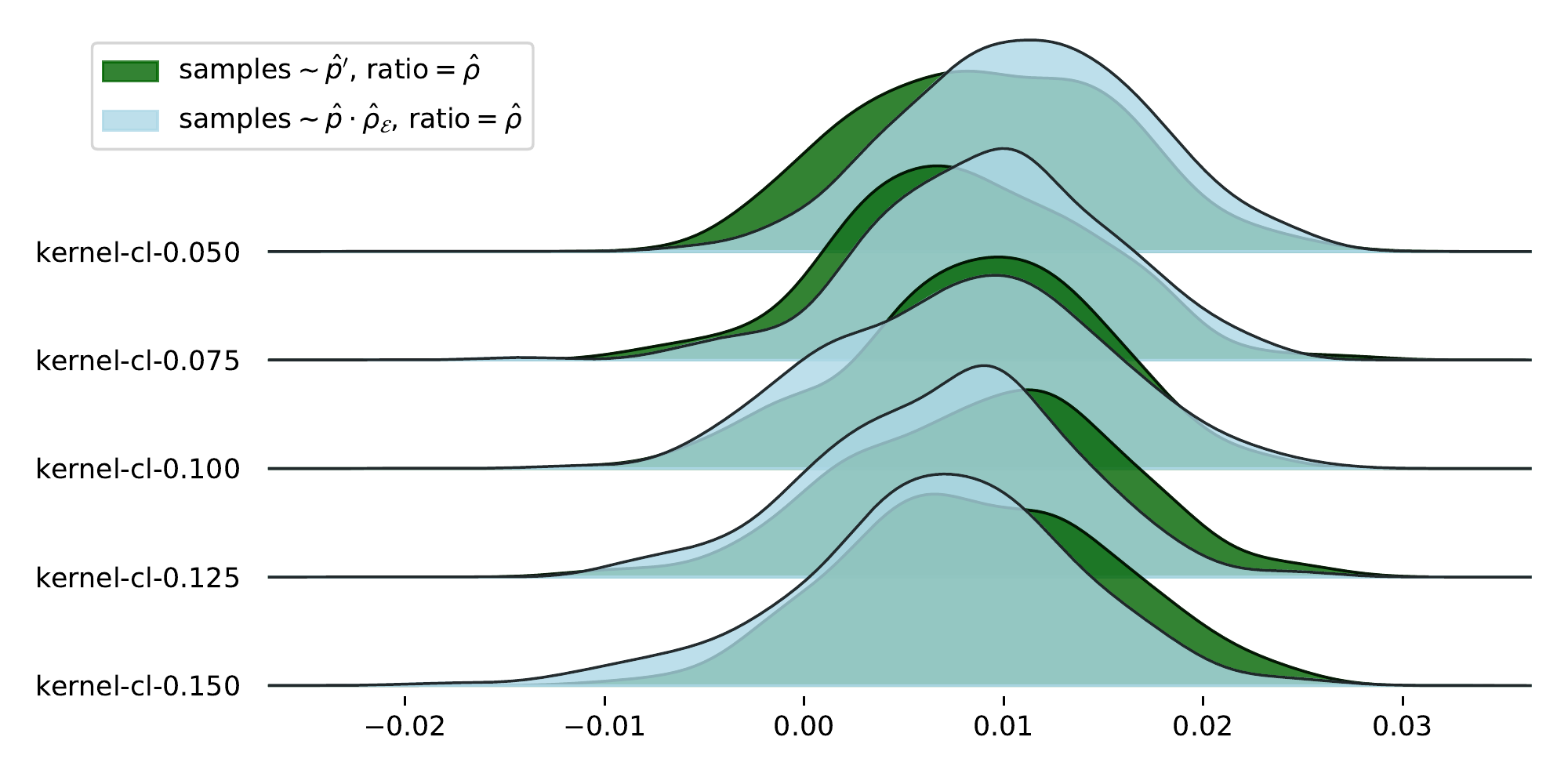}
	\caption{$\mathrm{LR}$ for KBC-based DRE ($\lambda=0.8$)}
	\end{subfigure}
	\begin{subfigure}[t!]{0.5\textwidth}
	\centering 
	\includegraphics[width=0.8\textwidth]{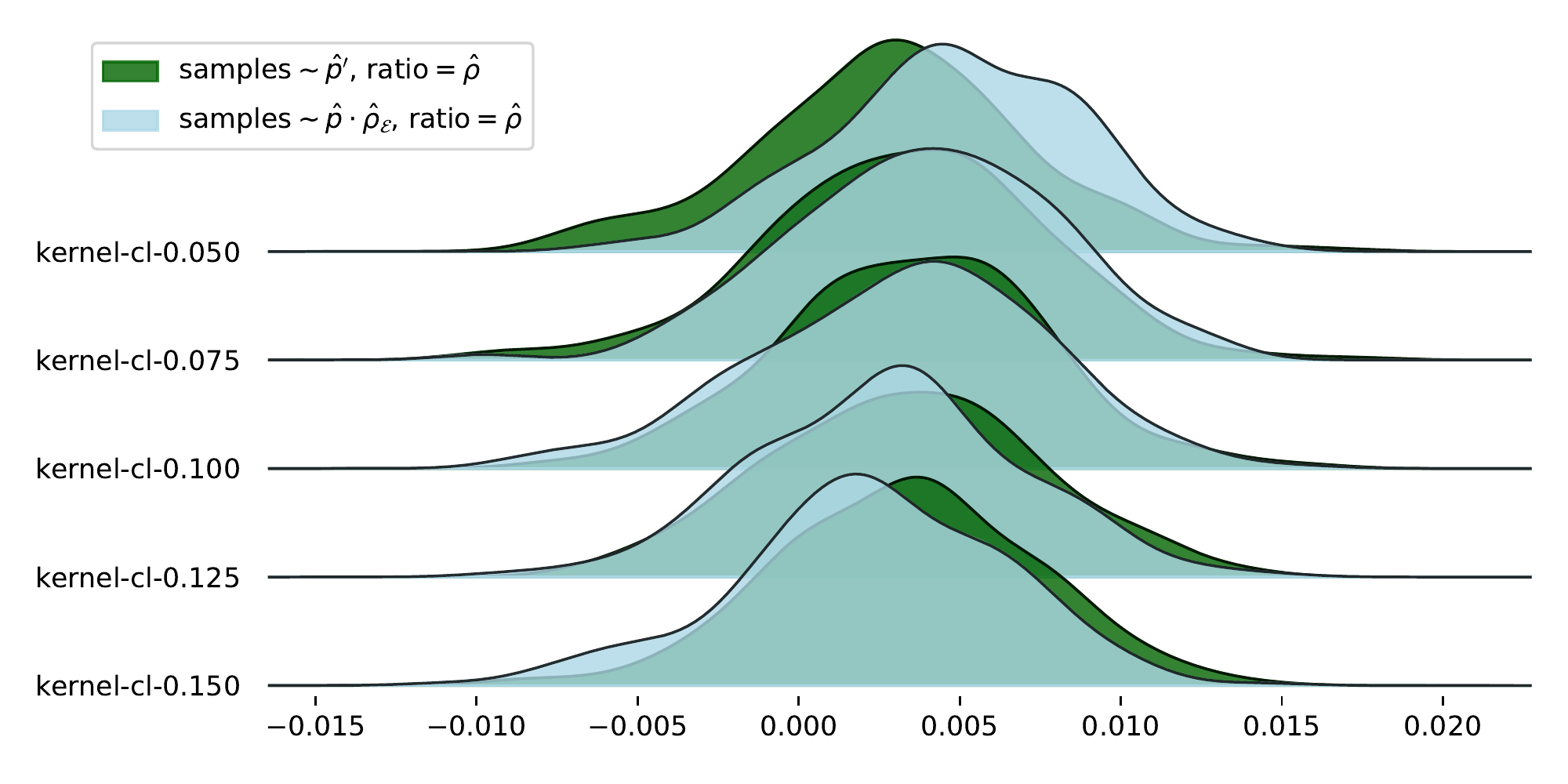}
	\caption{$\mathrm{LR}$ for KBC-based DRE ($\lambda=0.9$)}
	\end{subfigure}\\
	\begin{subfigure}[t!]{0.5\textwidth}
	\centering 
	\includegraphics[width=0.8\textwidth]{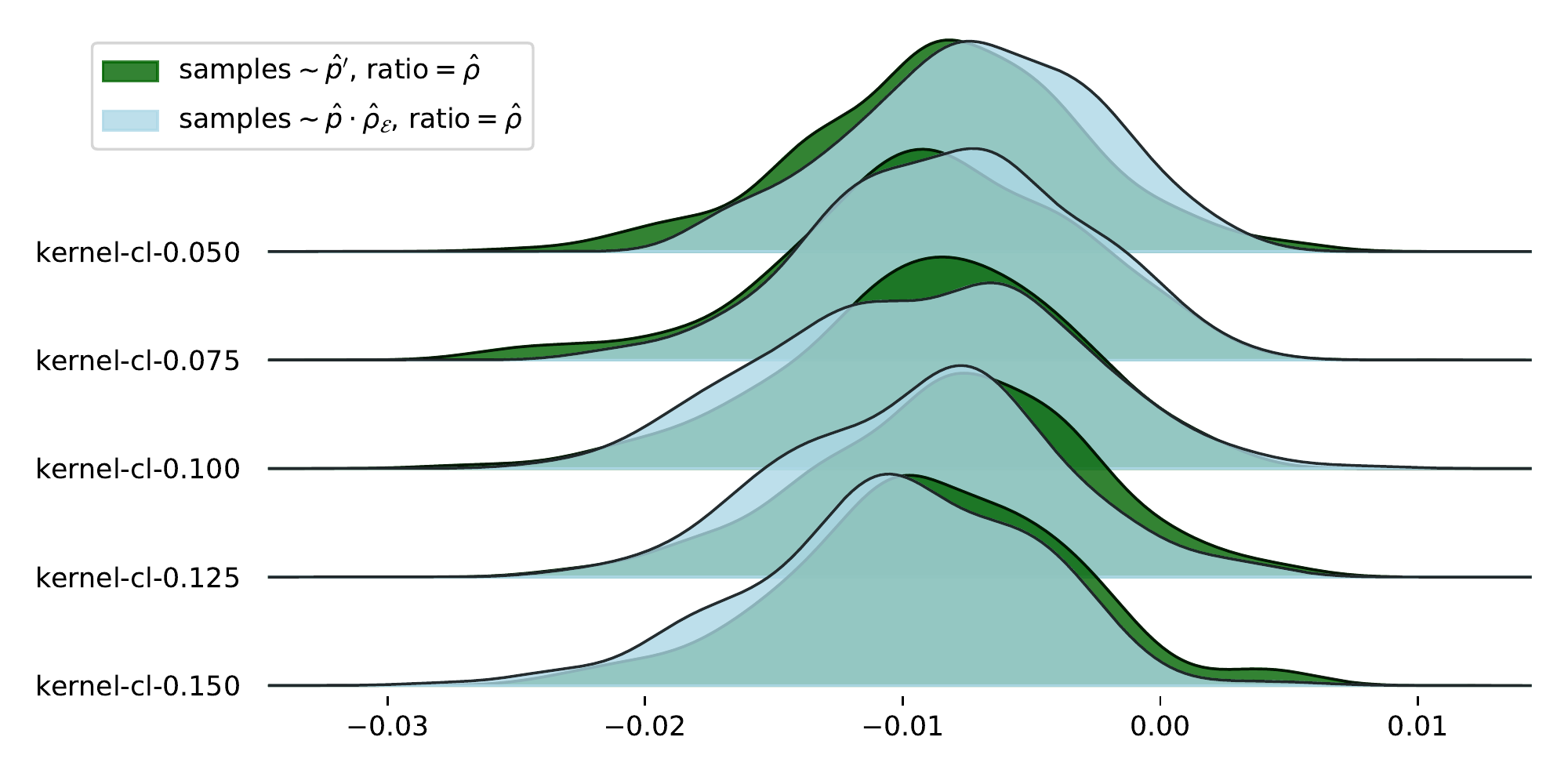}
	\caption{$\mathrm{ASC}$ for KBC-based DRE ($\phi(t)=\log(t)$)}
	\end{subfigure}
	\begin{subfigure}[t!]{0.5\textwidth}
	\centering 
	\includegraphics[width=0.8\textwidth]{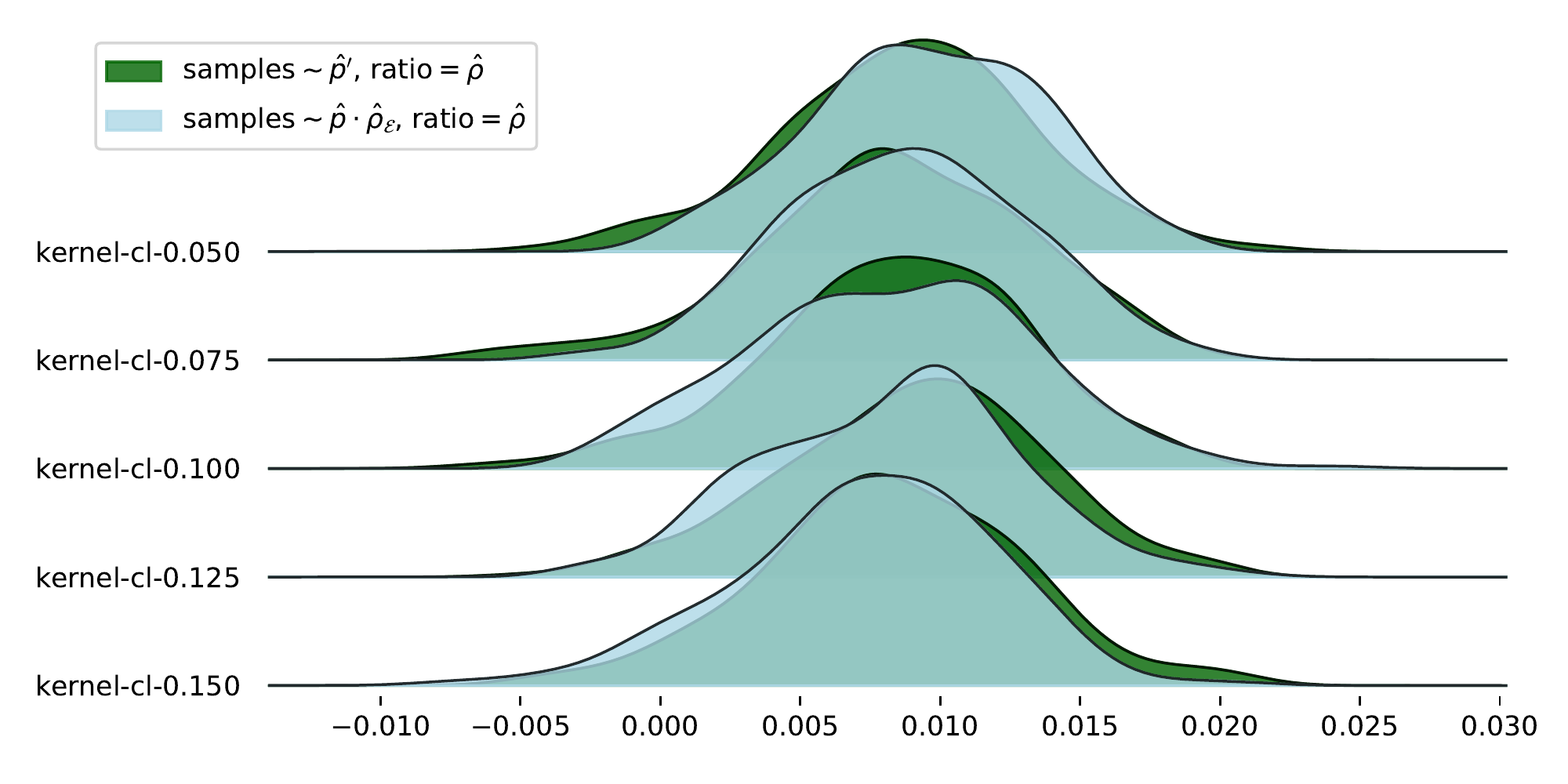}
	\caption{$\mathrm{ASC}$ for KBC-based DRE ($\phi(t)=t\log(t)$)}
	\end{subfigure}\\
	\begin{subfigure}[t!]{0.5\textwidth}
	\centering 
	\includegraphics[width=0.8\textwidth]{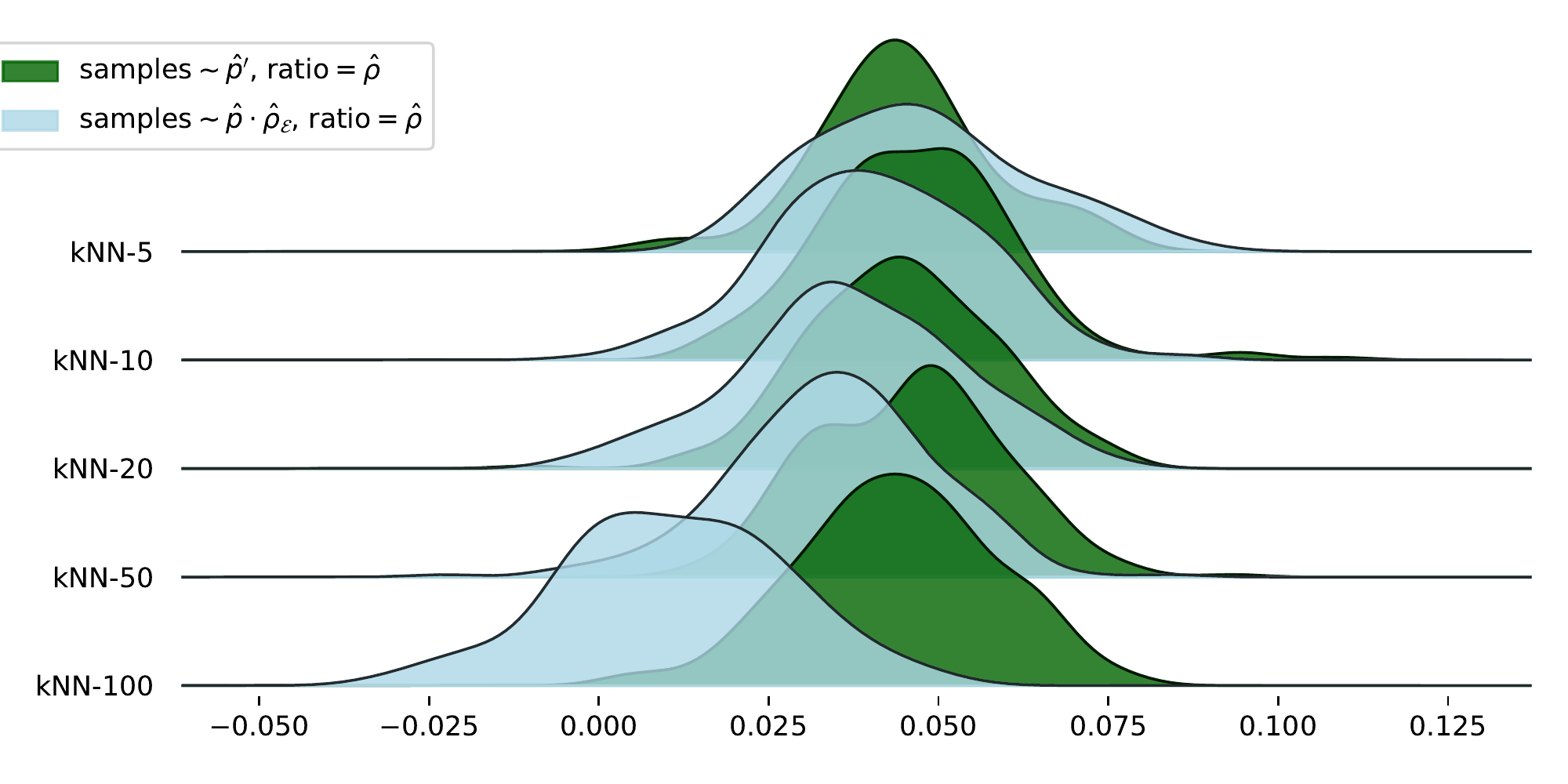}
	\caption{$\mathrm{LR}$ for $k$NN-based DRE ($\lambda=0.6$)}
	\end{subfigure}
	\begin{subfigure}[t!]{0.5\textwidth}
	\centering 
	\includegraphics[width=0.8\textwidth]{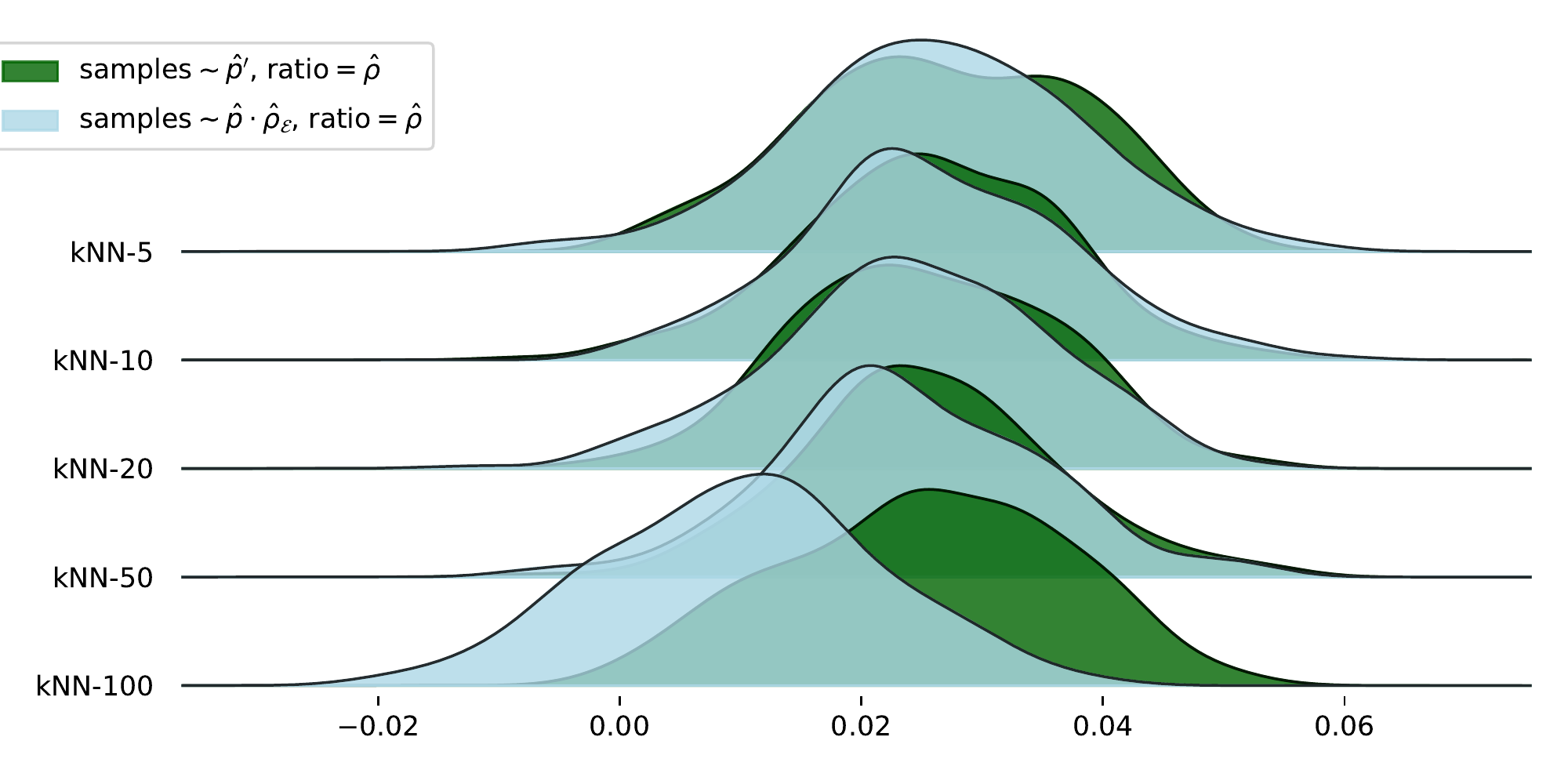}
	\caption{$\mathrm{LR}$ for $k$NN-based DRE ($\lambda=0.7$)}
	\end{subfigure}\\
	\begin{subfigure}[t!]{0.5\textwidth}
	\centering 
	\includegraphics[width=0.8\textwidth]{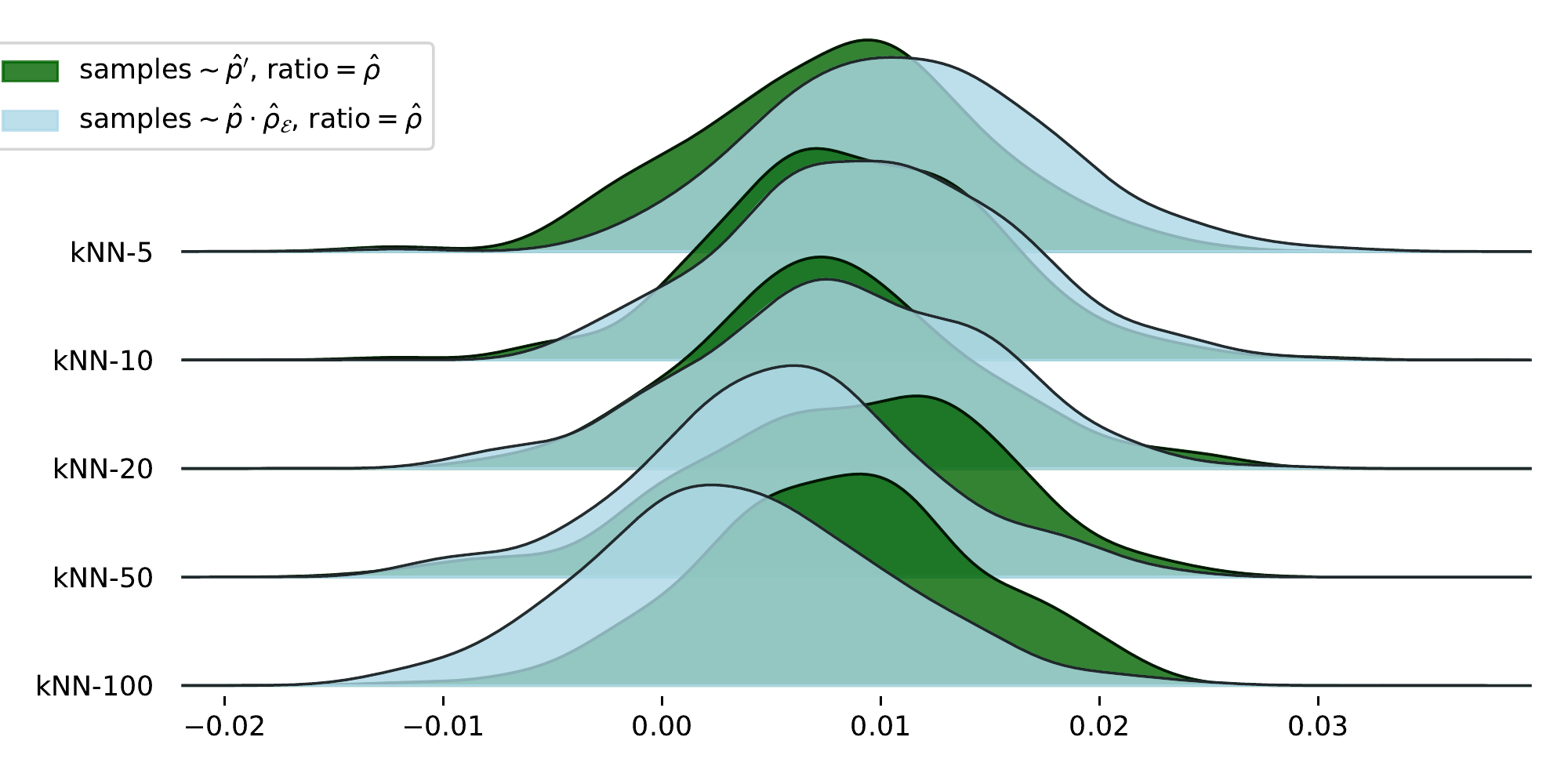}
	\caption{$\mathrm{LR}$ for $k$NN-based DRE ($\lambda=0.8$)}
	\end{subfigure}
	\begin{subfigure}[t!]{0.5\textwidth}
	\centering 
	\includegraphics[width=0.8\textwidth]{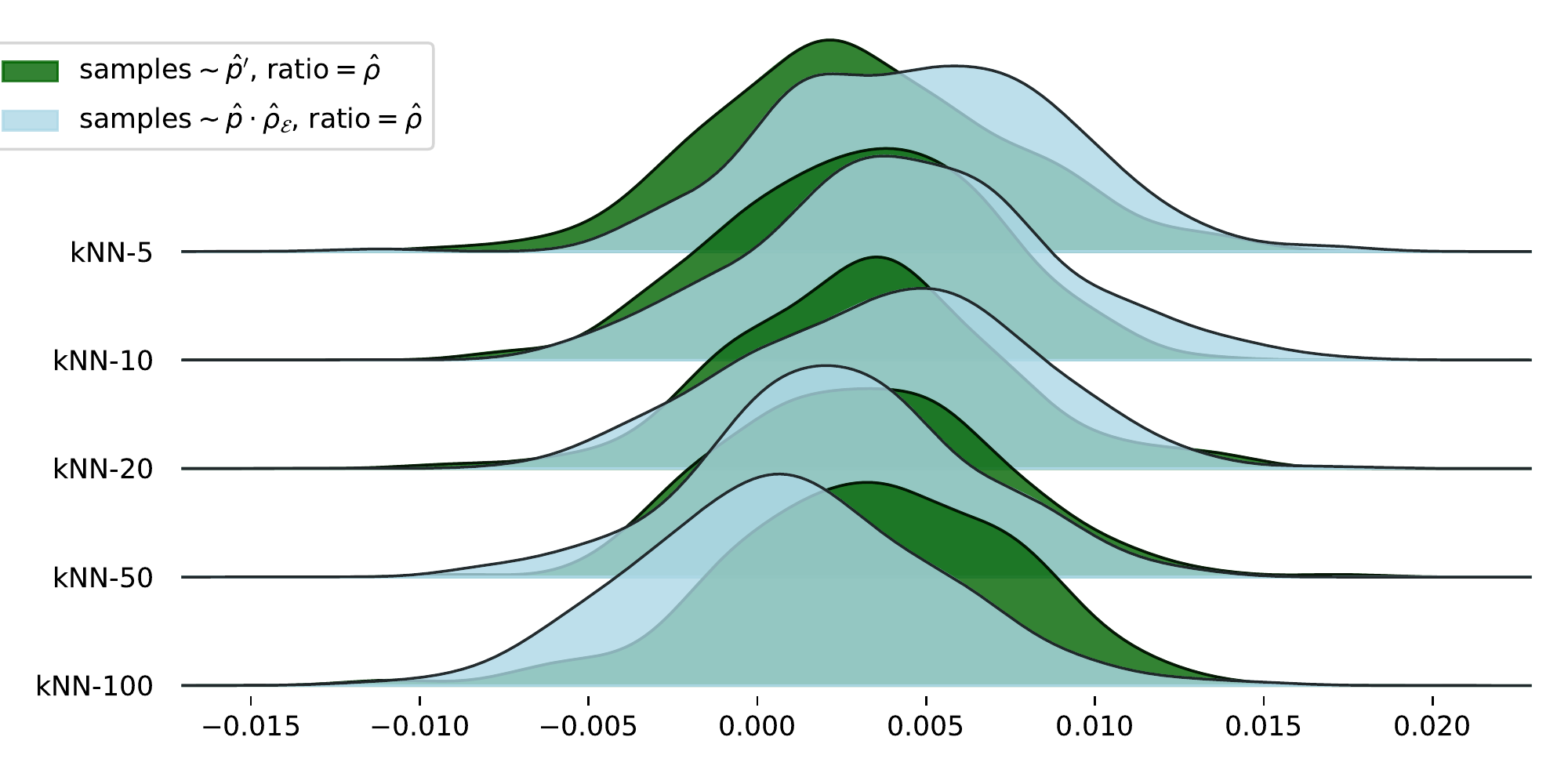}
	\caption{$\mathrm{LR}$ for $k$NN-based DRE ($\lambda=0.9$)}
	\end{subfigure}\\
	\begin{subfigure}[t!]{0.5\textwidth}
	\centering 
	\includegraphics[width=0.8\textwidth]{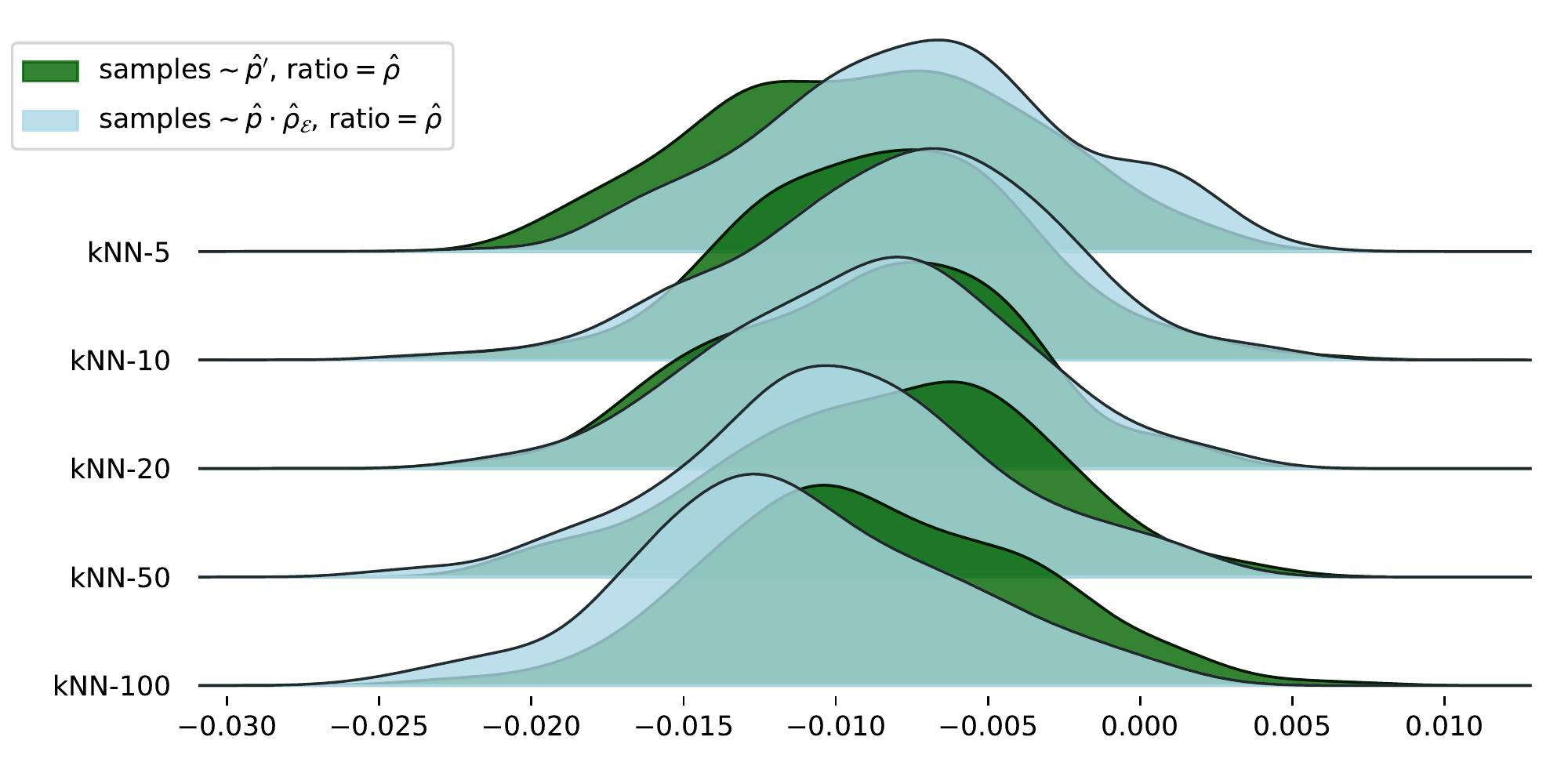}
	\caption{$\mathrm{ASC}$ for $k$NN-based DRE ($\phi(t)=\log(t)$)}
	\end{subfigure}
	\begin{subfigure}[t!]{0.5\textwidth}
	\centering 
	\includegraphics[width=0.8\textwidth]{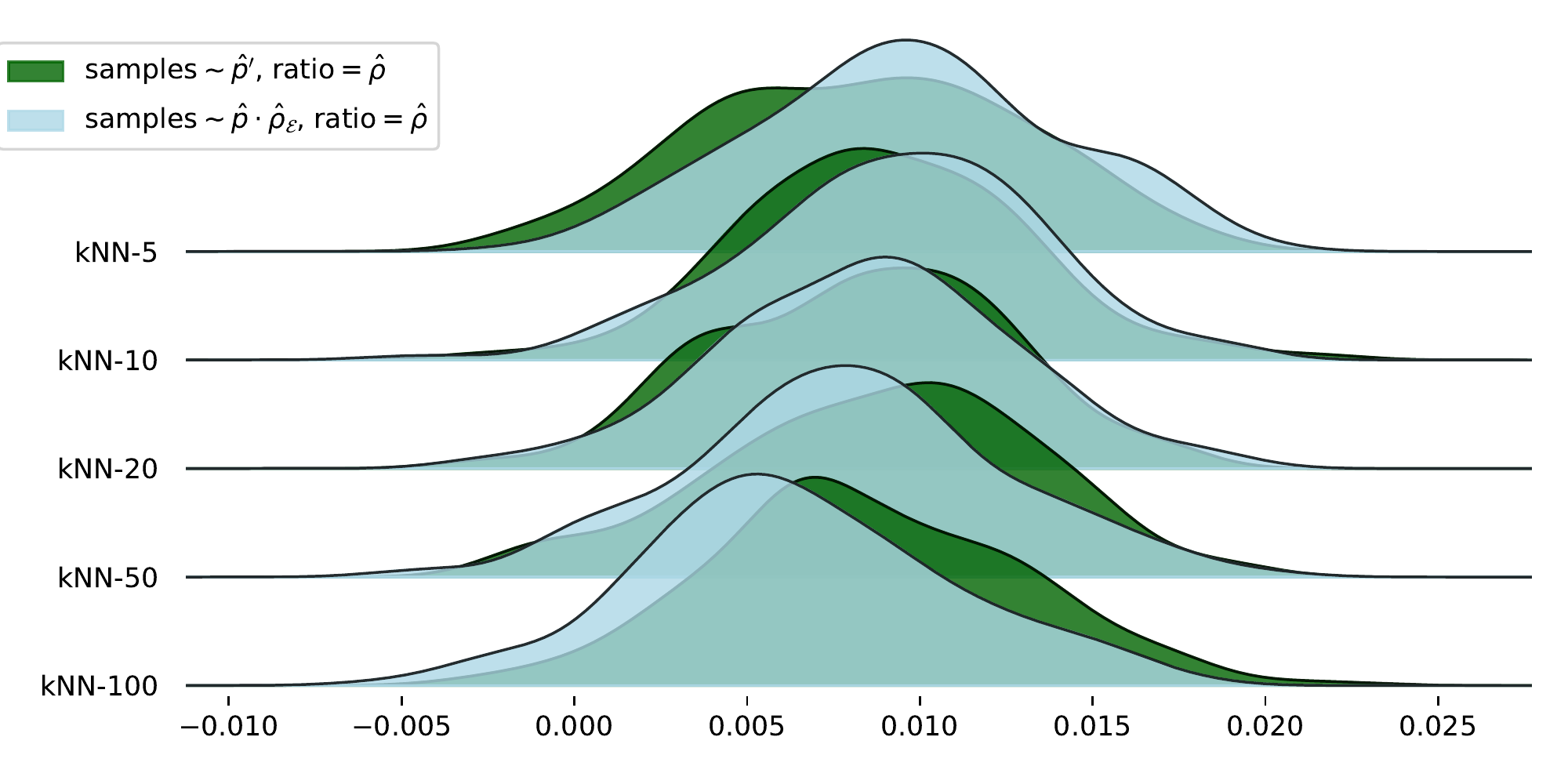}
	\caption{$\mathrm{ASC}$ for $k$NN-based DRE ($\phi(t)=t\log(t)$)}
	\end{subfigure}\\
	
	\vspace{-0.3em}
	\caption{(a)-(f) KBC-based DRE. (g)-(l) $k$NN-based DRE. (a)-(d)\&(g)-(j) $\mathrm{LR}(Y_{H_1},\hat{\rho})$ vs $\mathrm{LR}(Y_{\mD},\hat{\rho})$. (e)-(f)\&(k)-(l) $\hat{\mathrm{ASC}}_{\phi}(\hat{Y},Y_{H_1},\hat{\rho})$ vs $\hat{\mathrm{ASC}}_{\phi}(\hat{Y},Y_{\mD},\hat{\rho})$.}
	\vspace{-1em}
	\label{fig: 2d Q2 joy MoG-8 appendix}
\end{figure}

\newpage
\paragraph{Question 2 (Fast Deletion).}
We visualize distributions of LR and ASC statistics between $Y_{H_1}$ and $Y_{\mD}$ in Fig. \ref{fig: 2d Q2 joy MoG-8 appendix} (extension of Fig. \ref{fig: 2d Q2 joy}). The more overlapping between the distributions, the less distinguishable between the approximated and re-trained models. KBC is generally better than $k$NN. For $k$NN a moderate $k$ (e.g. between 10 and 50) has better overlapping. 

\begin{figure}[!h]
\vspace{-0.3em}
  	\begin{subfigure}[t!]{0.5\textwidth}
	\centering 
	\includegraphics[trim=30 0 0 5, clip, width=0.95\textwidth]{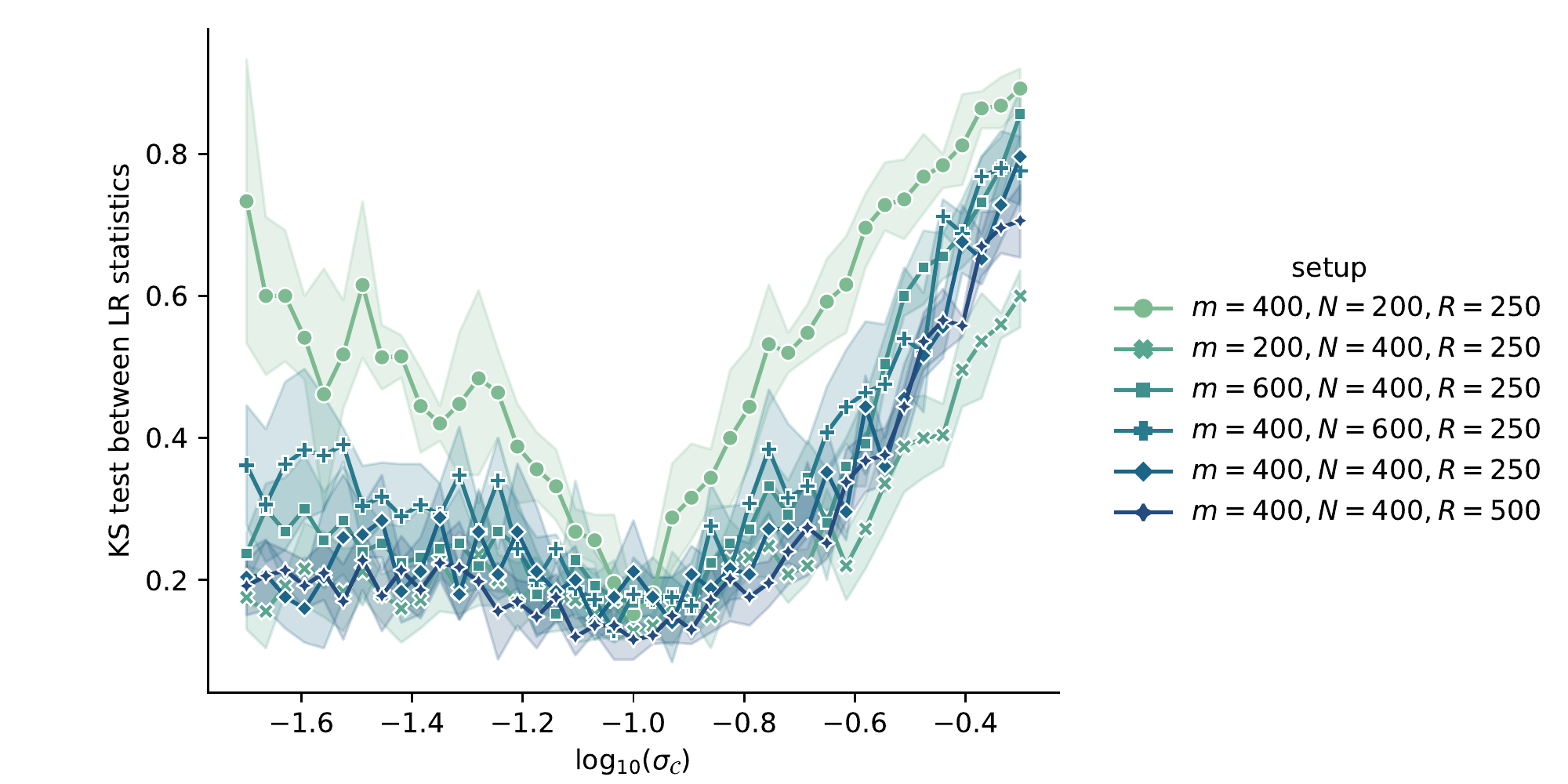}
	\caption{$\mathrm{LR}$ statistics with $\lambda=0.8$ and different $m,N,R$}
	\end{subfigure}
	\begin{subfigure}[t!]{0.5\textwidth}
	\centering 
	\includegraphics[trim=30 0 0 5, clip, width=0.95\textwidth]{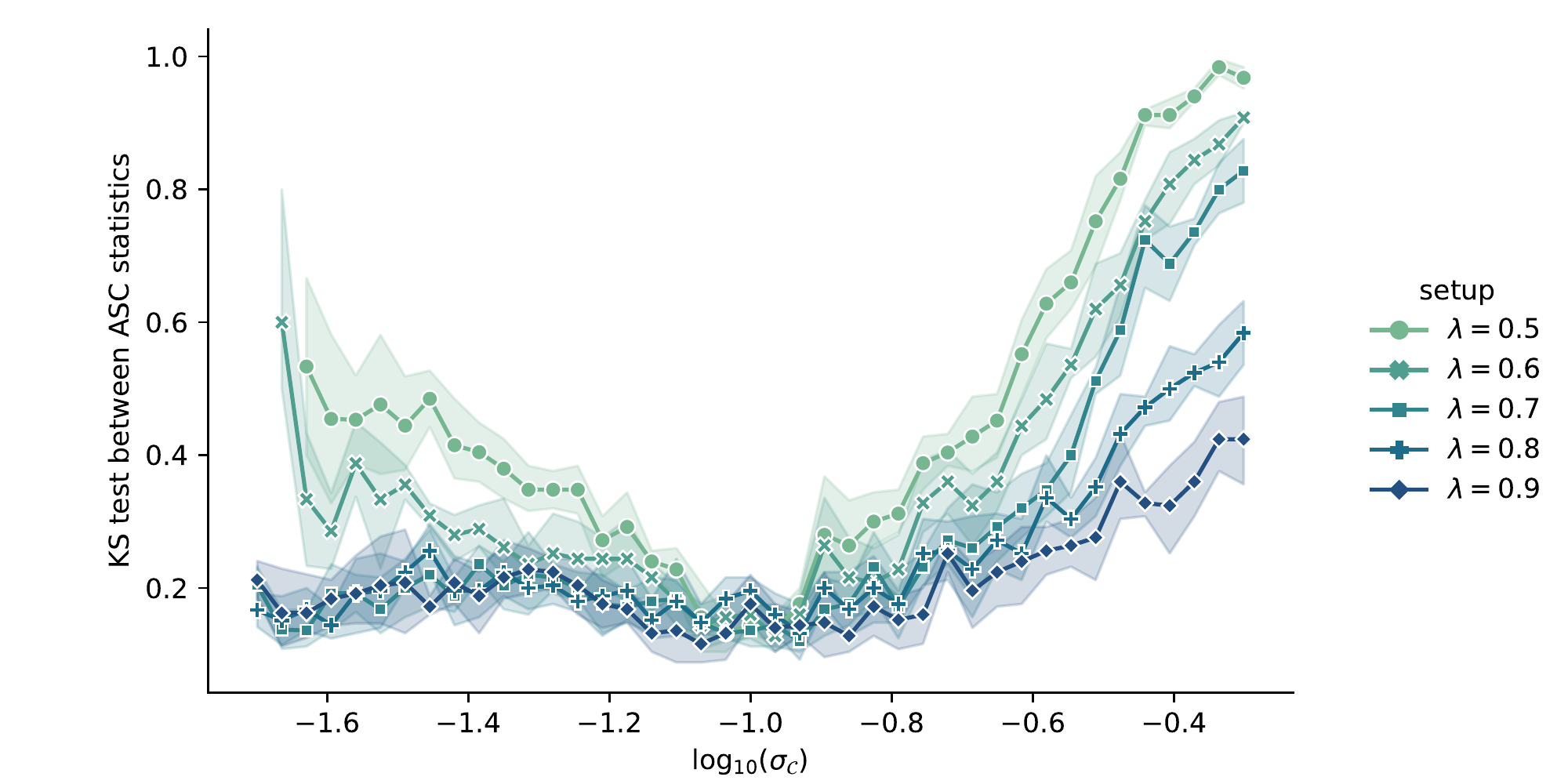}
	\caption{$\mathrm{ASC}$ statistics with $\phi(t)=\log(t)$}
	\end{subfigure}\\
	\begin{subfigure}[t!]{0.5\textwidth}
	\centering 
	\includegraphics[trim=30 0 0 5, clip, width=0.95\textwidth]{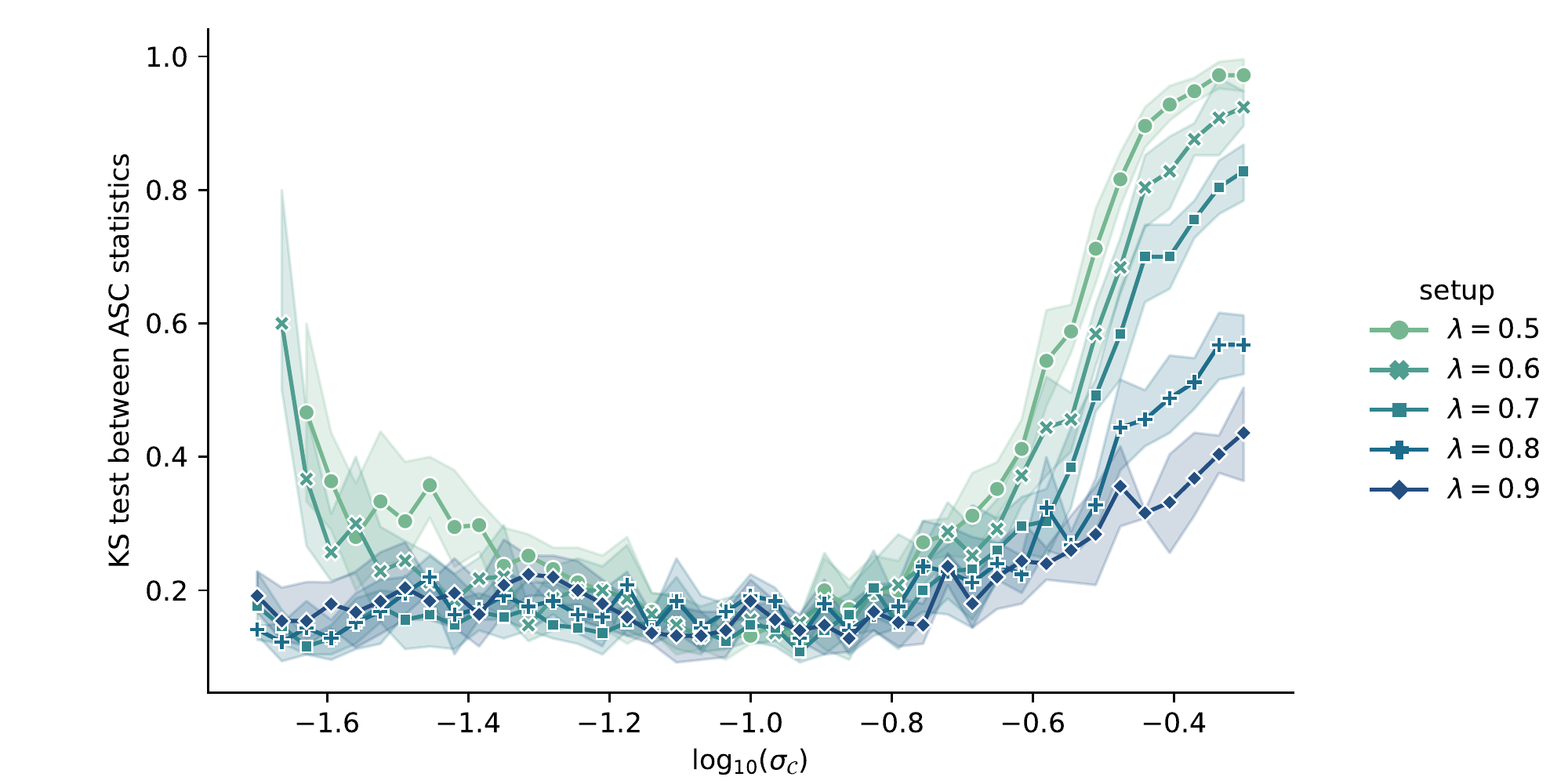}
	\caption{$\mathrm{ASC}$ statistics with $\phi(t)=t\log(t)$}
	\end{subfigure}
	\begin{subfigure}[t!]{0.5\textwidth}
	\centering 
	\includegraphics[trim=30 0 0 5, clip, width=0.95\textwidth]{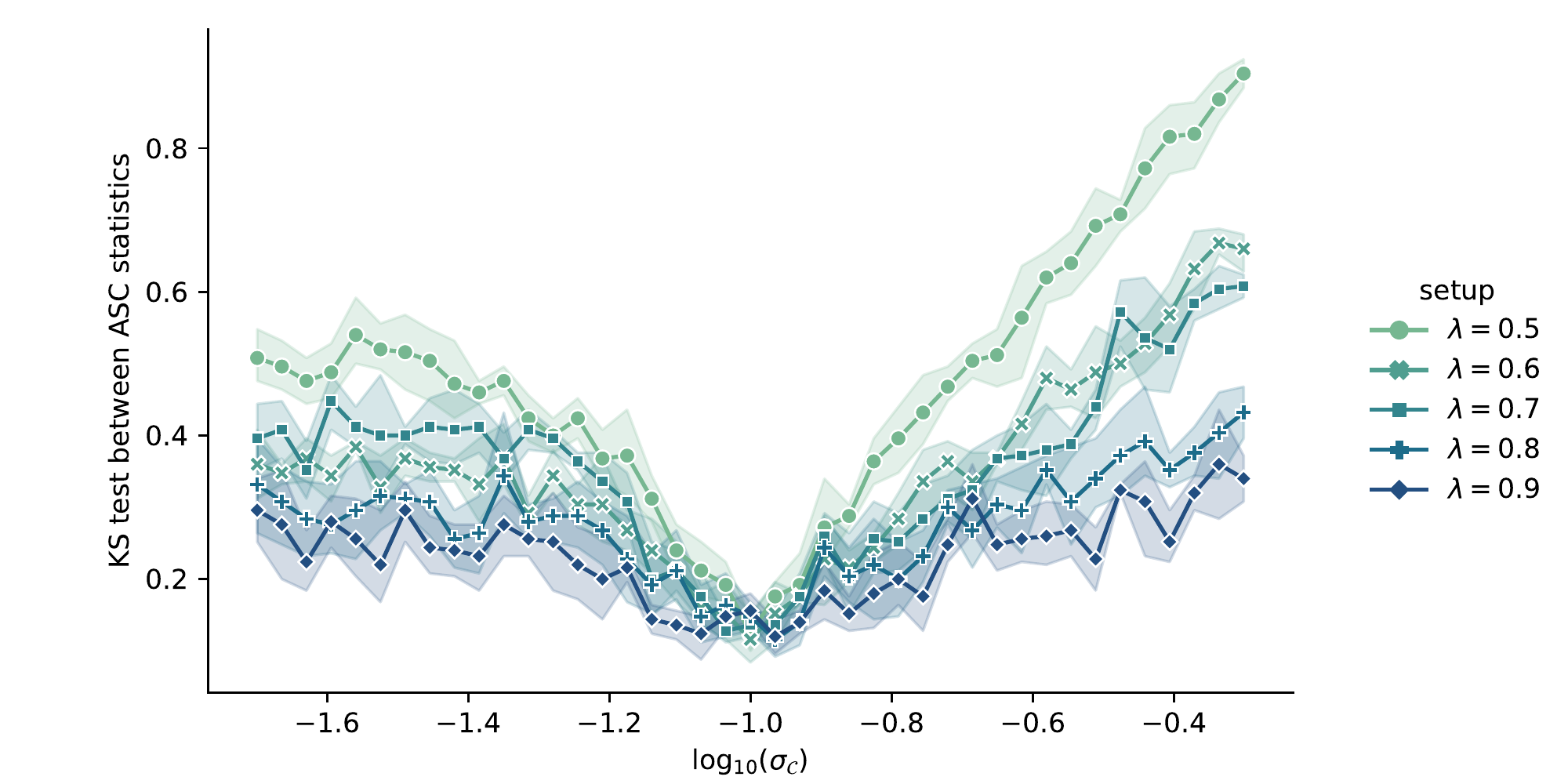}
	\caption{$\mathrm{ASC}$ statistics with $\phi(t)=(\sqrt{t}-1)^2$}
	\end{subfigure}
	
	\vspace{-0.3em}
	\caption{KS tests between distributions of statistics for KBC with different $\sigma_{\mC}$. (a) $\mathrm{LR}(Y_{H_1},\hat{\rho})$ vs $\mathrm{LR}(Y_{\mD},\hat{\rho})$ with $\lambda=0.8$ and different $m,N,R$, complementary to Fig. \ref{fig: 2d Q2 KS}. (b)-(d) $\hat{\mathrm{ASC}}_{\phi}(\hat{Y},Y_{H_1},\hat{\rho})$ vs $\hat{\mathrm{ASC}}_{\phi}(\hat{Y},Y_{\mD},\hat{\rho})$ for different $\phi$. Smaller values indicate the two compared distributions are closer. }
	\vspace{-0.3em}
	\label{fig: 2d Q2 KS MoG-8 appendix}
\end{figure}

\newpage
We visualize KS test results for KBC with different bandwidth $\sigma_{\mC}$ in Fig. \ref{fig: 2d Q2 KS MoG-8 appendix} (extension of Fig. \ref{fig: 2d Q2 KS}). The KS values are small for a wide range of $\sigma_{\mC}$, indicating KBC with these $\sigma_{\mC}$ can lead to approximated models indistinguishable from the re-trained model.  There is no clear difference between LR and ASC statistics. In terms of $\lambda$, the models are less distinguishable when $\lambda$ is larger, as expected.

\newpage
\paragraph{Question 3 (Hypothesis Test).}

\begin{figure}[!h]
\vspace{-0.5em}
  	\begin{subfigure}[t!]{0.5\textwidth}
	\centering 
	\includegraphics[width=0.8\textwidth]{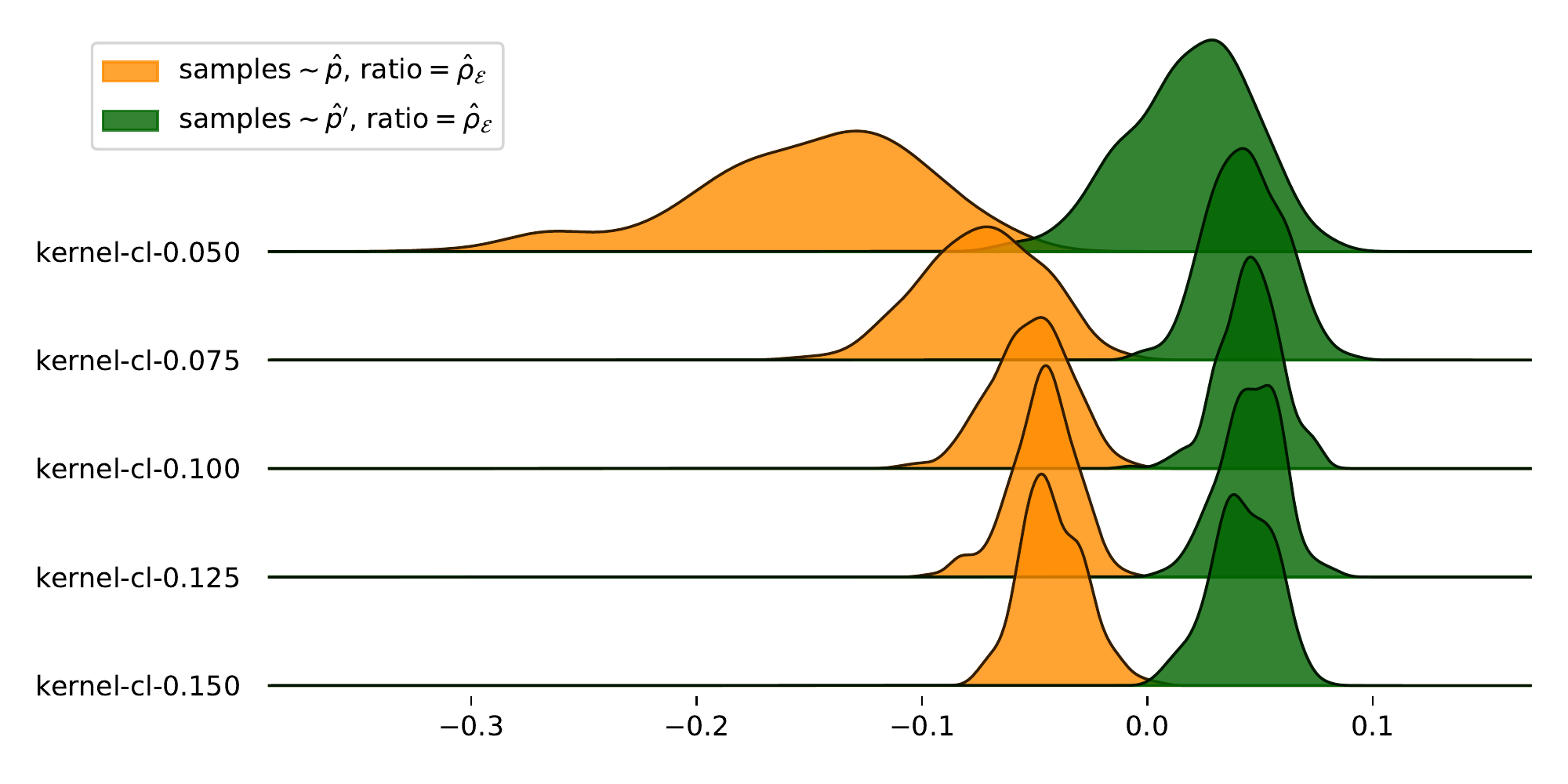}
	\caption{$\mathrm{LR}$ for KBC-based DRE ($\lambda=0.6$)}
	\end{subfigure}
	\begin{subfigure}[t!]{0.5\textwidth}
	\centering 
	\includegraphics[width=0.8\textwidth]{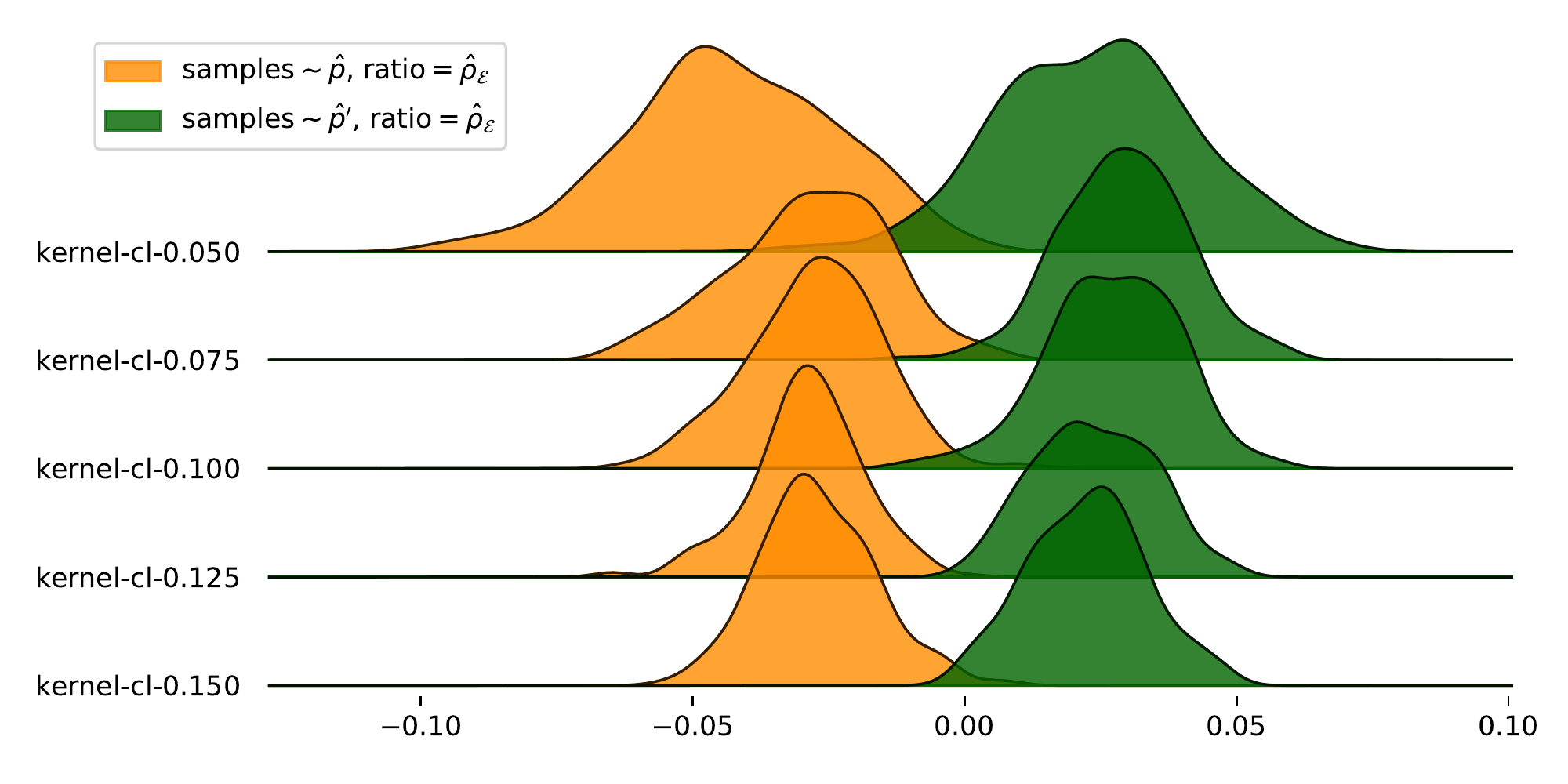}
	\caption{$\mathrm{LR}$ for KBC-based DRE ($\lambda=0.7$)}
	\end{subfigure}\\
	\begin{subfigure}[t!]{0.5\textwidth}
	\centering 
	\includegraphics[width=0.8\textwidth]{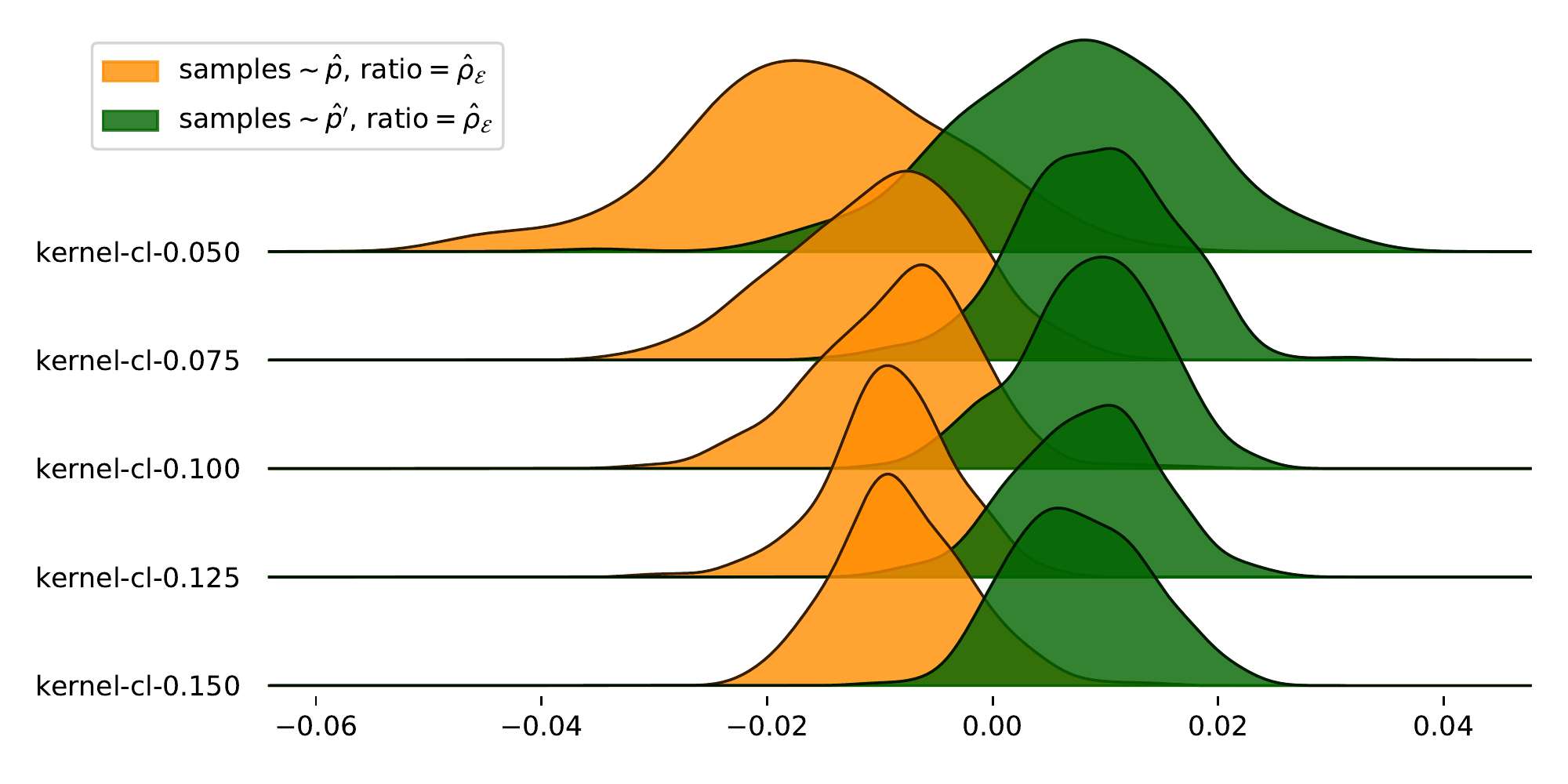}
	\caption{$\mathrm{LR}$ for KBC-based DRE ($\lambda=0.8$)}
	\end{subfigure}
	\begin{subfigure}[t!]{0.5\textwidth}
	\centering 
	\includegraphics[width=0.8\textwidth]{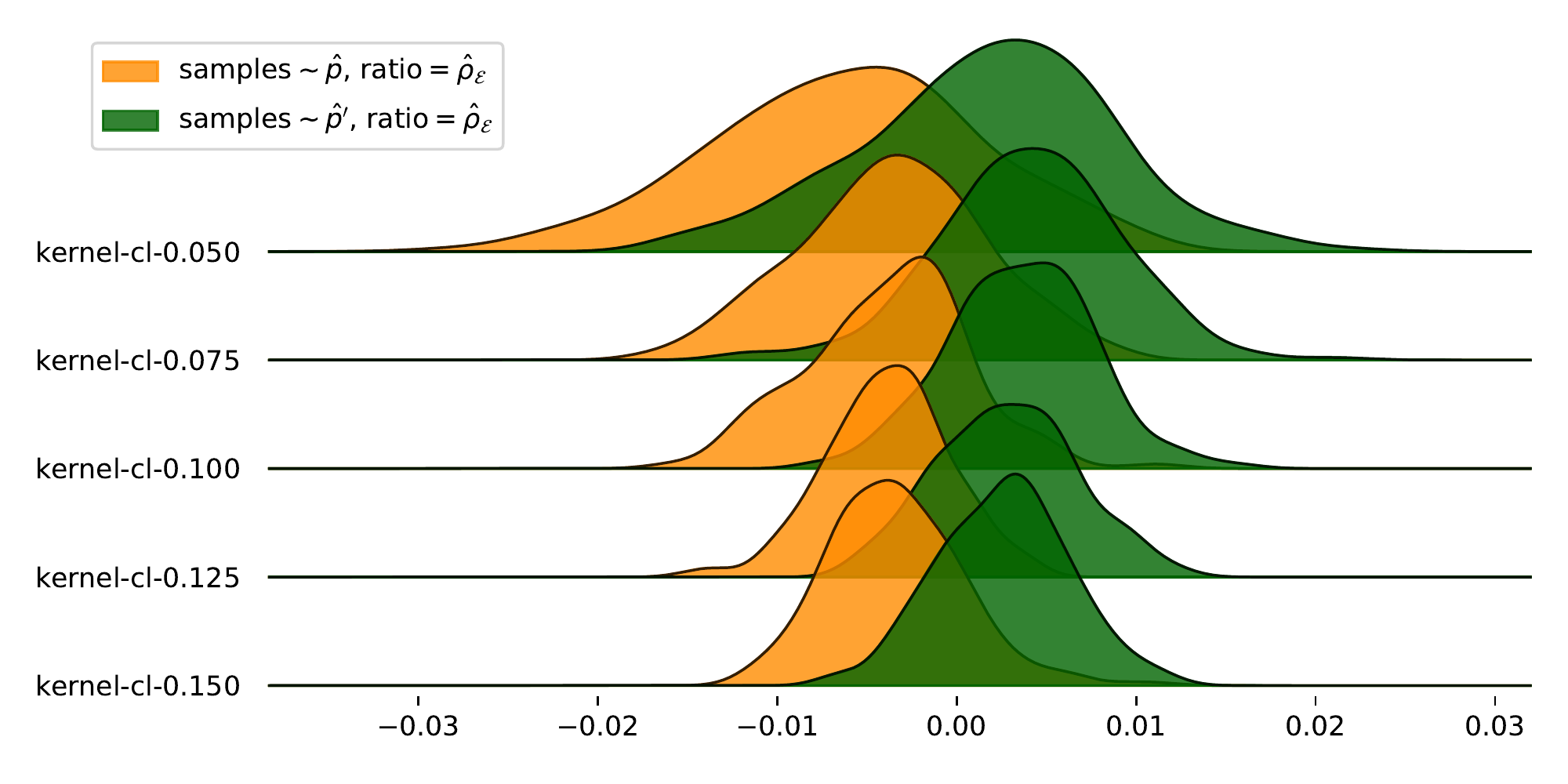}
	\caption{$\mathrm{LR}$ for KBC-based DRE ($\lambda=0.9$)}
	\end{subfigure}\\
	\begin{subfigure}[t!]{0.5\textwidth}
	\centering 
	\includegraphics[width=0.8\textwidth]{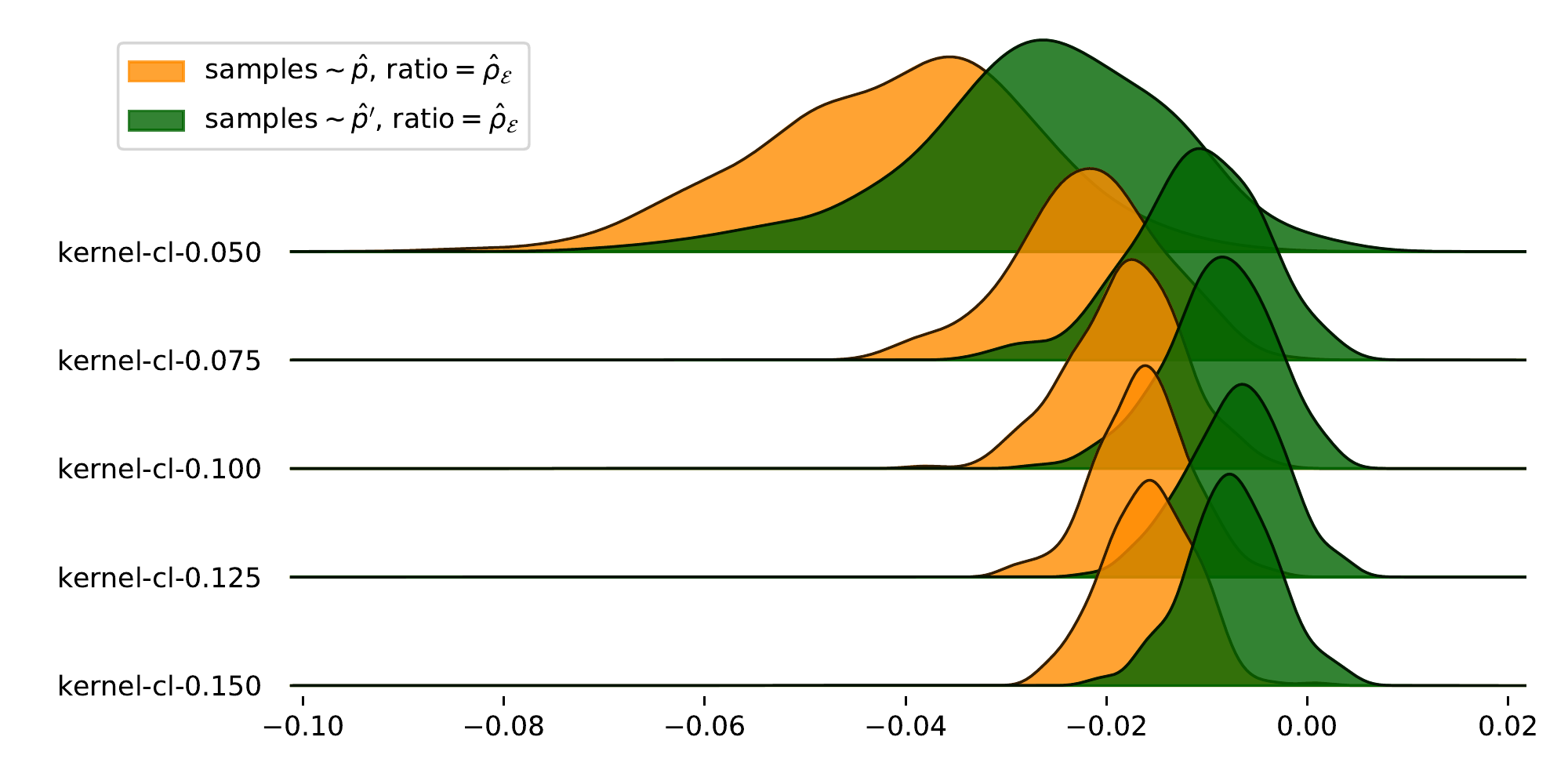}
	\caption{$\mathrm{ASC}$ for KBC-based DRE ($\phi(t)=\log(t)$)}
	\end{subfigure}
	\begin{subfigure}[t!]{0.5\textwidth}
	\centering 
	\includegraphics[width=0.8\textwidth]{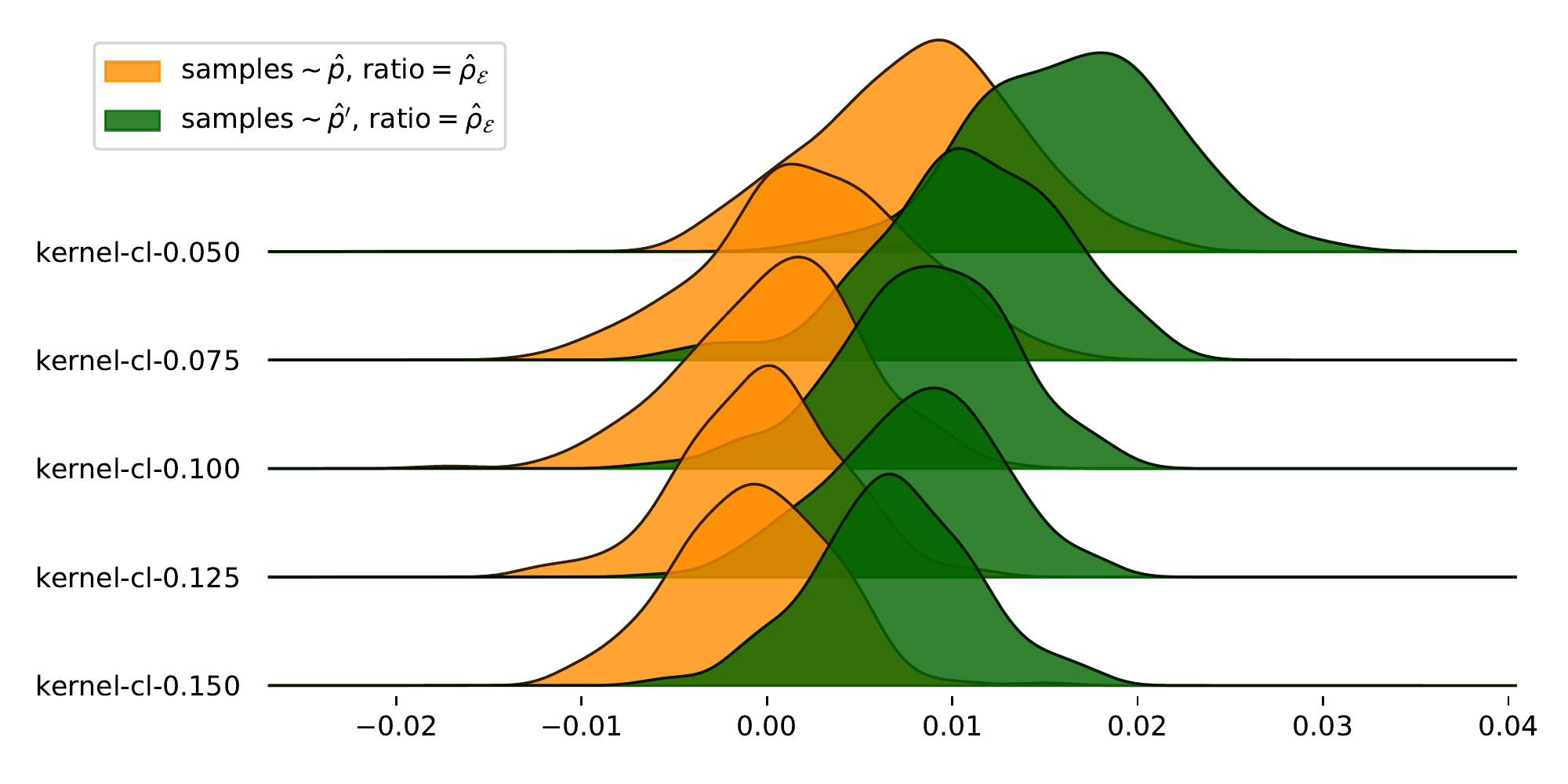}
	\caption{$\mathrm{ASC}$ for KBC-based DRE ($\phi(t)=t\log(t)$)}
	\end{subfigure}\\
	\begin{subfigure}[t!]{0.5\textwidth}
	\centering 
	\includegraphics[width=0.8\textwidth]{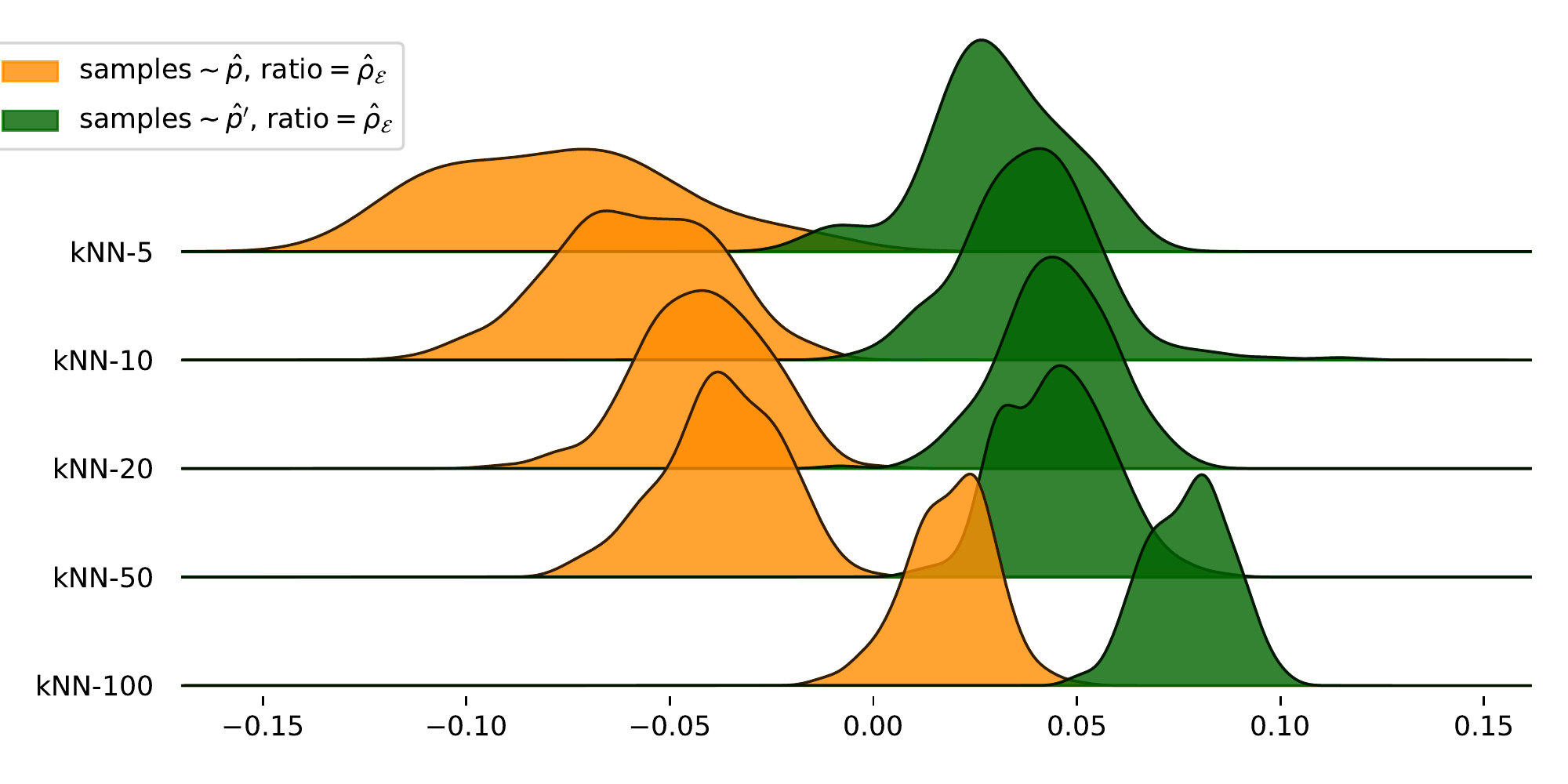}
	\caption{$\mathrm{LR}$ for $k$NN-based DRE ($\lambda=0.6$)}
	\end{subfigure}
	\begin{subfigure}[t!]{0.5\textwidth}
	\centering 
	\includegraphics[width=0.8\textwidth]{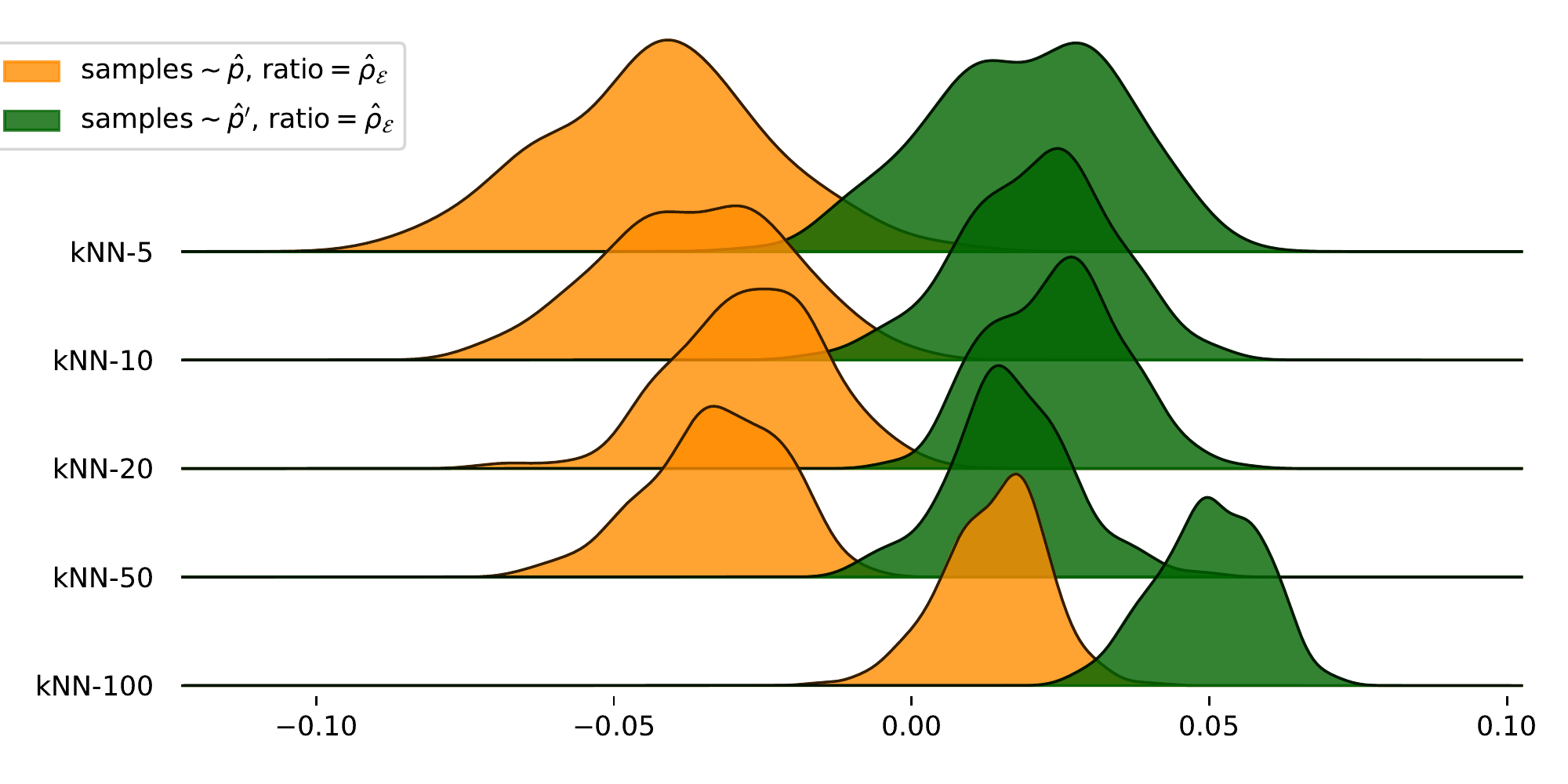}
	\caption{$\mathrm{LR}$ for $k$NN-based DRE ($\lambda=0.7$)}
	\end{subfigure}\\
	\begin{subfigure}[t!]{0.5\textwidth}
	\centering 
	\includegraphics[width=0.8\textwidth]{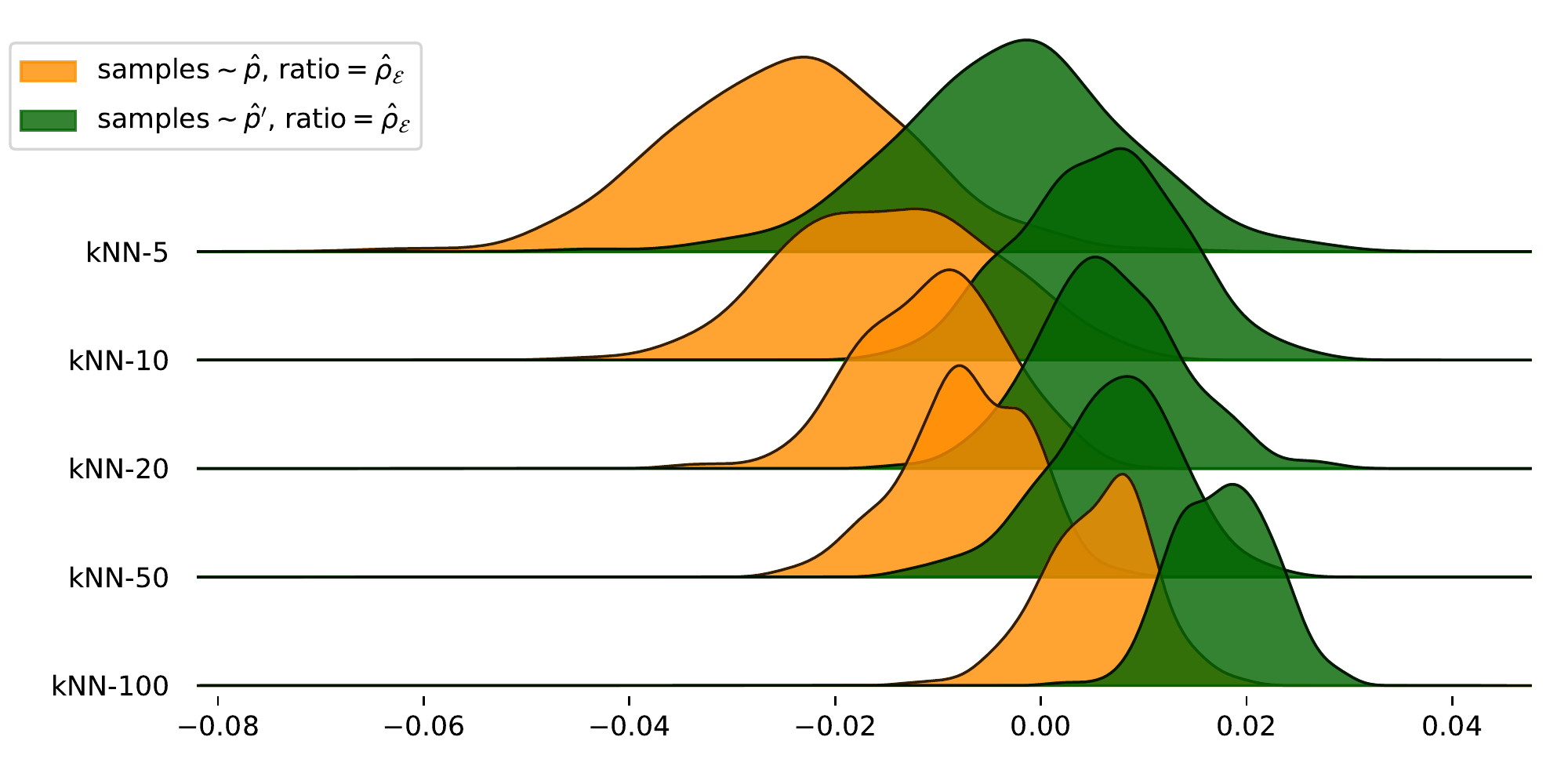}
	\caption{$\mathrm{LR}$ for $k$NN-based DRE ($\lambda=0.8$)}
	\end{subfigure}
	\begin{subfigure}[t!]{0.5\textwidth}
	\centering 
	\includegraphics[width=0.8\textwidth]{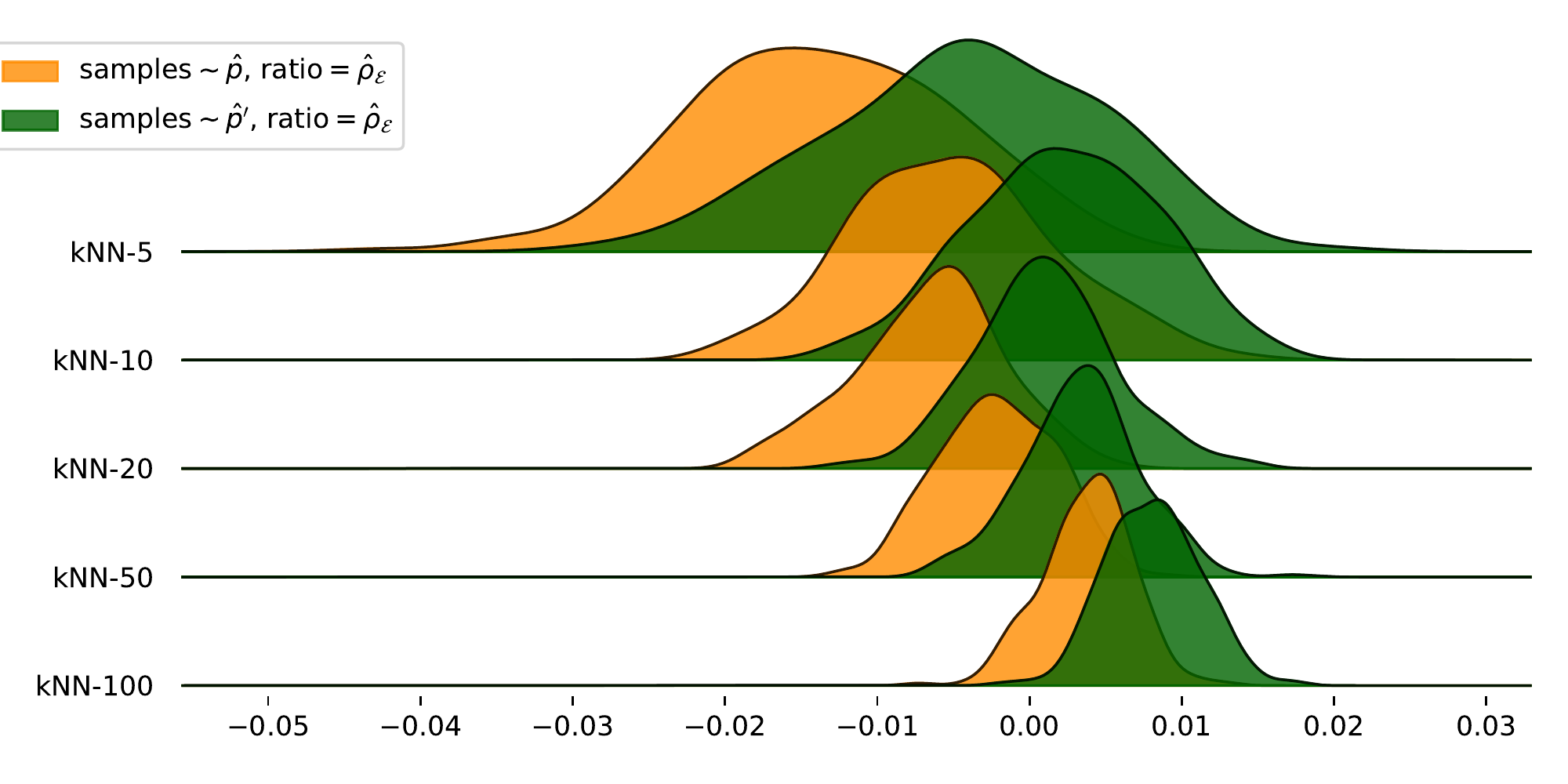}
	\caption{$\mathrm{LR}$ for $k$NN-based DRE ($\lambda=0.9$)}
	\end{subfigure}\\
	\begin{subfigure}[t!]{0.5\textwidth}
	\centering 
	\includegraphics[width=0.8\textwidth]{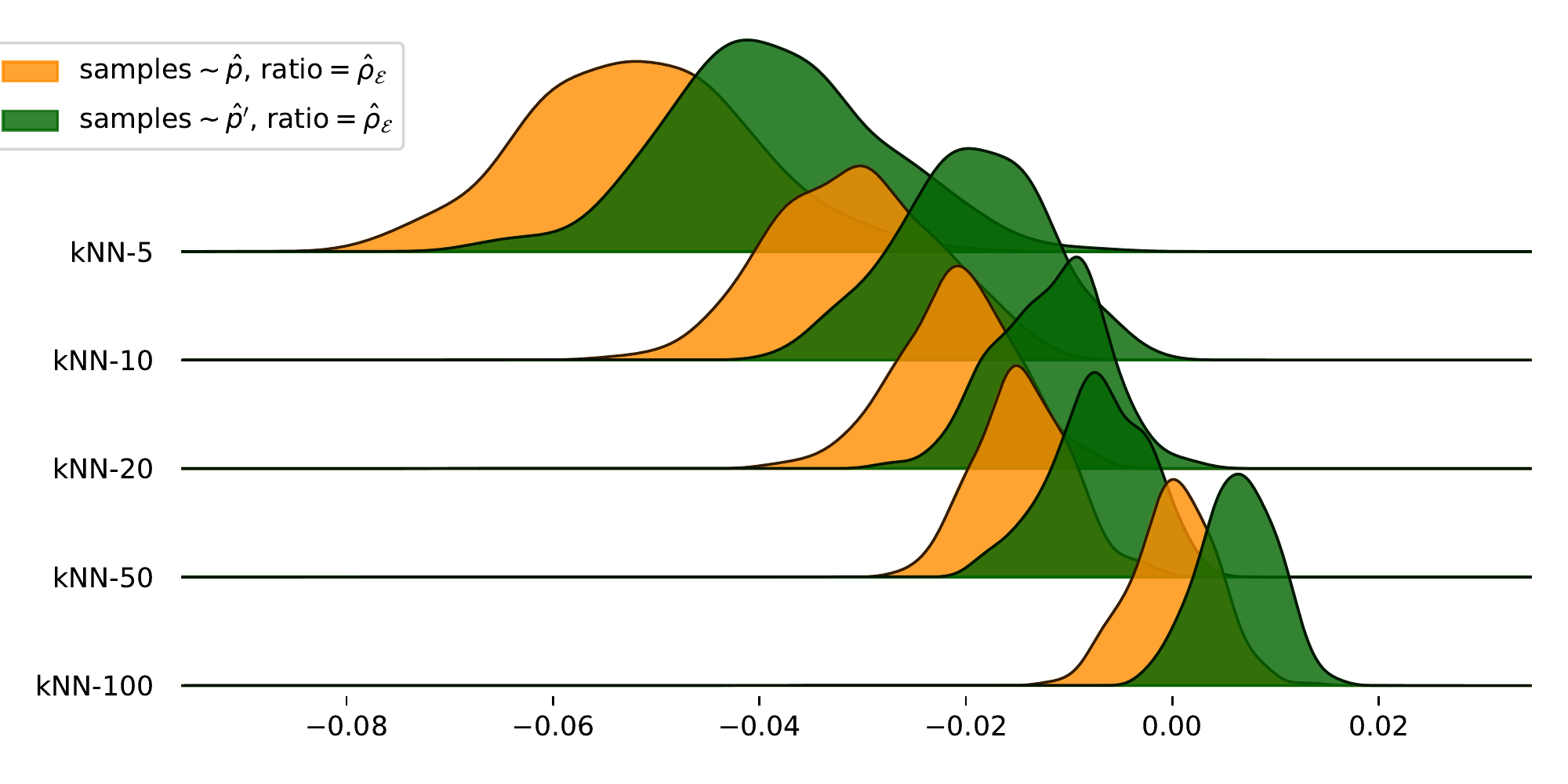}
	\caption{$\mathrm{ASC}$ for $k$NN-based DRE ($\phi(t)=\log(t)$)}
	\end{subfigure}
	\begin{subfigure}[t!]{0.5\textwidth}
	\centering 
	\includegraphics[width=0.8\textwidth]{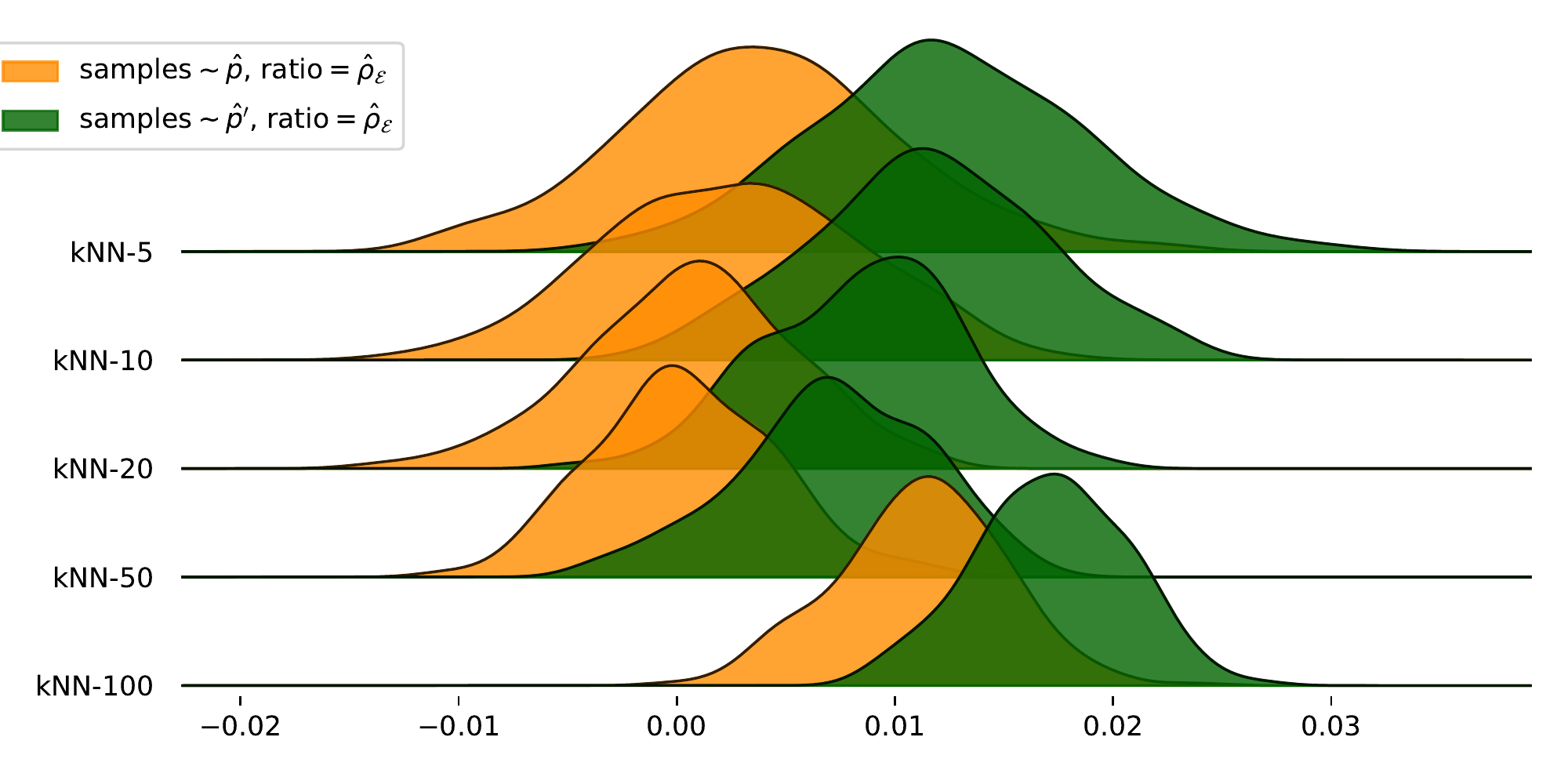}
	\caption{$\mathrm{ASC}$ for $k$NN-based DRE ($\phi(t)=t\log(t)$)}
	\end{subfigure}\\
	
	\vspace{-0.3em}
	\caption{(a)-(f) KBC-based DRE. (g)-(l) $k$NN-based DRE. (a)-(d)\&(g)-(j) $\mathrm{LR}(Y_{H_0},\hat{\rho})$ vs $\mathrm{LR}(Y_{H_1},\hat{\rho})$. (e)-(f)\&(k)-(l) $\hat{\mathrm{ASC}}_{\phi}(\hat{Y},Y_{H_0},\hat{\rho})$ vs $\hat{\mathrm{ASC}}_{\phi}(\hat{Y},Y_{H_1},\hat{\rho})$.}
	\vspace{-2em}
	\label{fig: 2d Q3 joy MoG-8 appendix}
\end{figure}

We visualize distributions of LR and ASC statistics between $Y_{H_0}$ and $Y_{H_1}$ in Fig. \ref{fig: 2d Q3 joy MoG-8 appendix} (extension of Fig. \ref{fig: 2d Q3 joy}). The separation between the distributions indicates how the DRE can distinguish samples between pre-trained and re-trained models. We observe separation for a wide range of classifiers, and KBC is generally comparable to $k$NN. In terms of statistics, the LR is better than ASC. In terms of $\lambda$, larger $\lambda$ makes the two models less distinguishable.

\begin{figure}[!h]
\vspace{-0.3em}
  	\begin{subfigure}[t!]{0.5\textwidth}
	\centering 
	\includegraphics[trim=30 0 0 5, clip, width=0.95\textwidth]{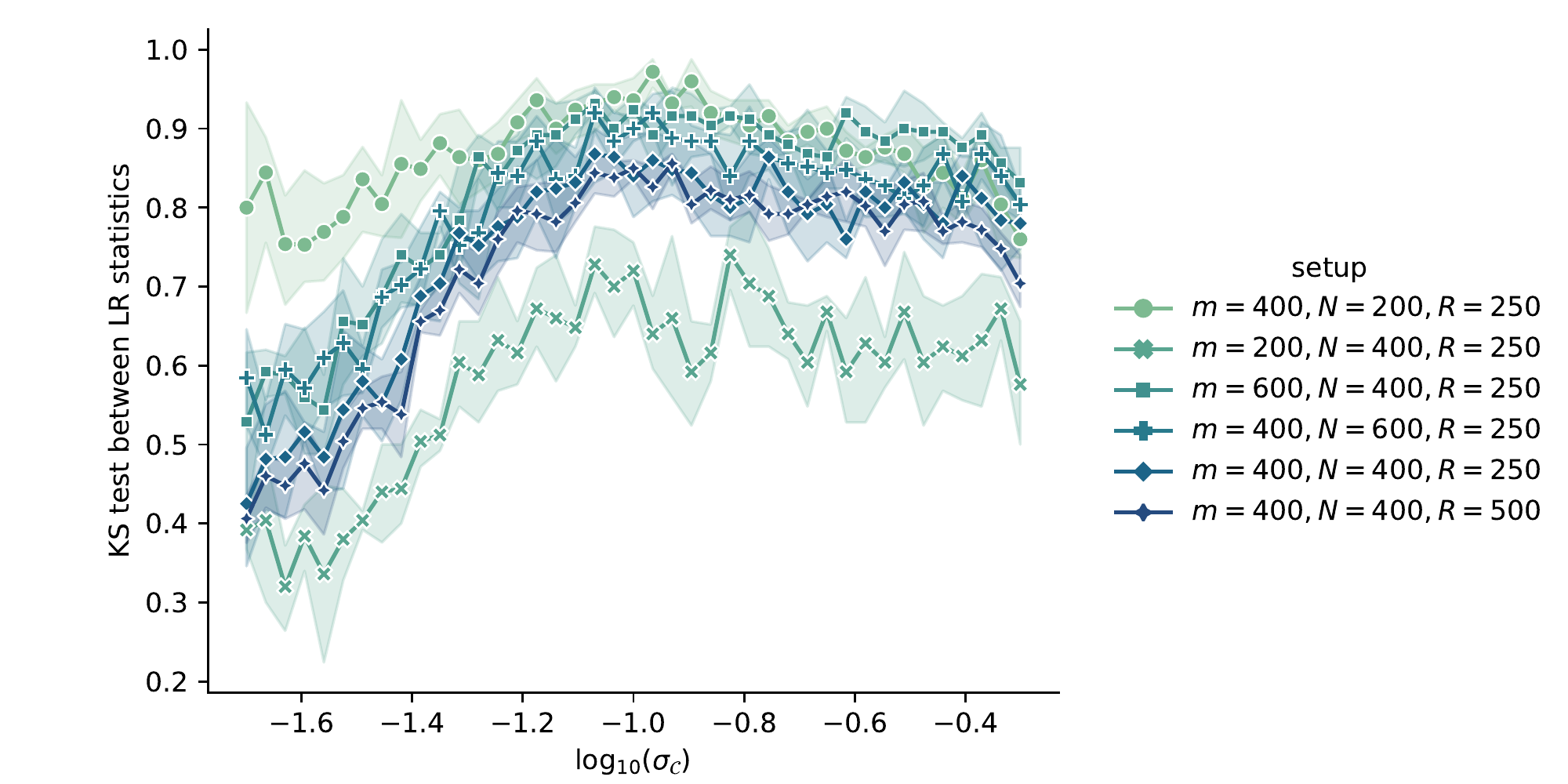}
	\caption{$\mathrm{LR}$ statistics with $\lambda=0.8$ and different $m,N,R$}
	\end{subfigure}
	\begin{subfigure}[t!]{0.5\textwidth}
	\centering 
	\includegraphics[trim=30 0 0 5, clip, width=0.95\textwidth]{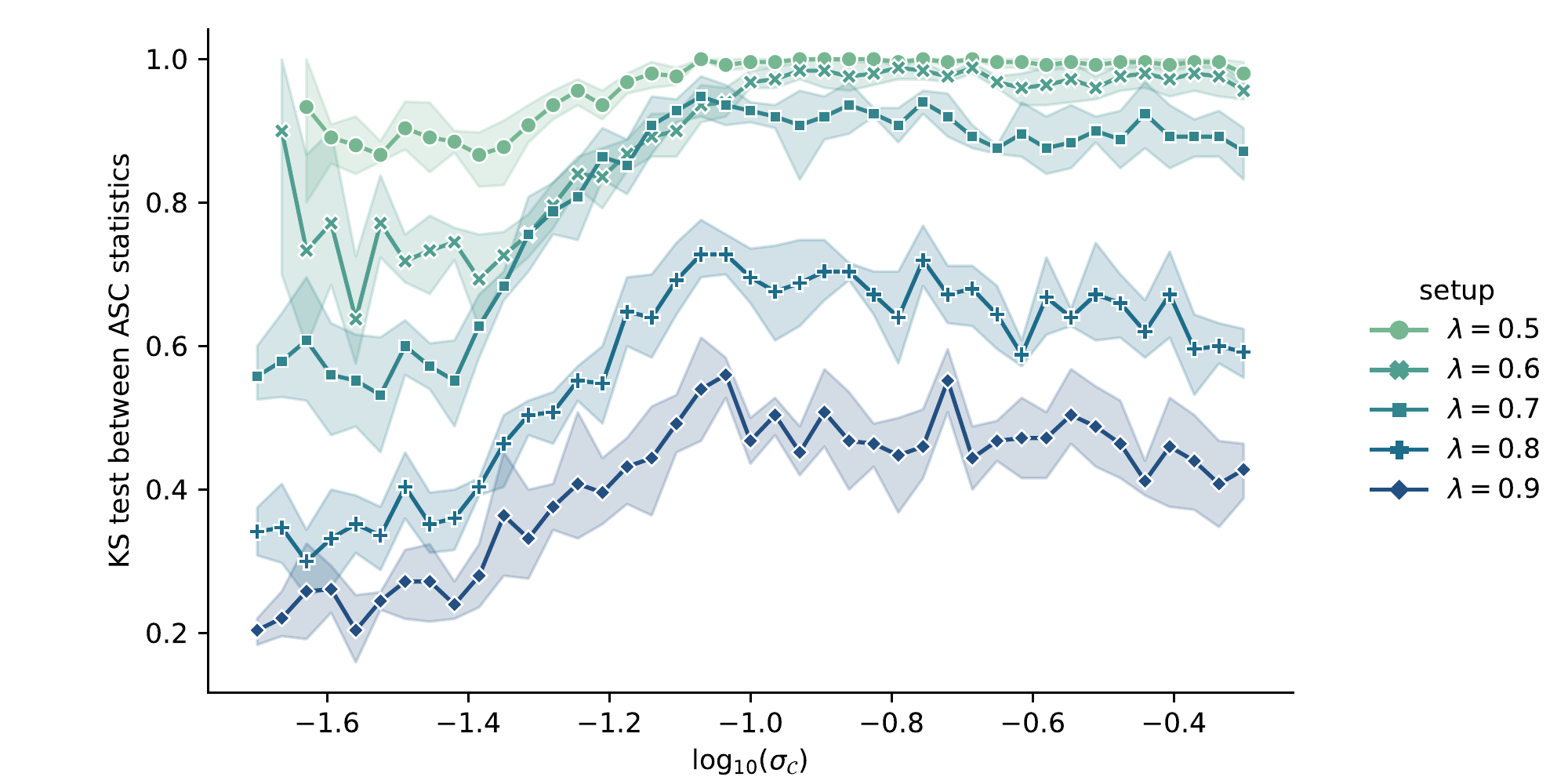}
	\caption{$\mathrm{ASC}$ statistics with $\phi(t)=\log(t)$}
	\end{subfigure}\\
	\begin{subfigure}[t!]{0.5\textwidth}
	\centering 
	\includegraphics[trim=30 0 0 5, clip, width=0.95\textwidth]{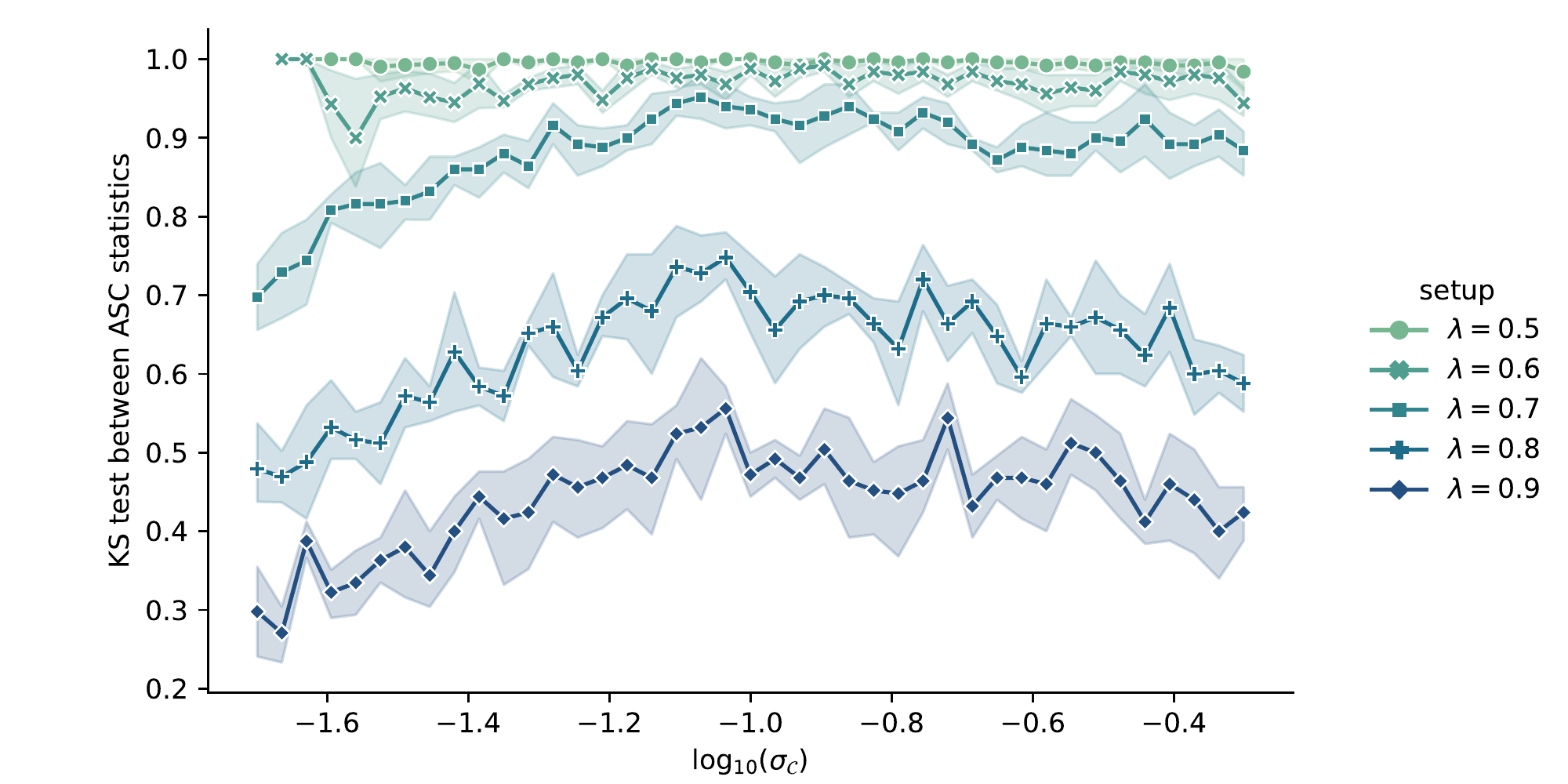}
	\caption{$\mathrm{ASC}$ statistics with $\phi(t)=t\log(t)$}
	\end{subfigure}
	\begin{subfigure}[t!]{0.5\textwidth}
	\centering 
	\includegraphics[trim=30 0 0 5, clip, width=0.95\textwidth]{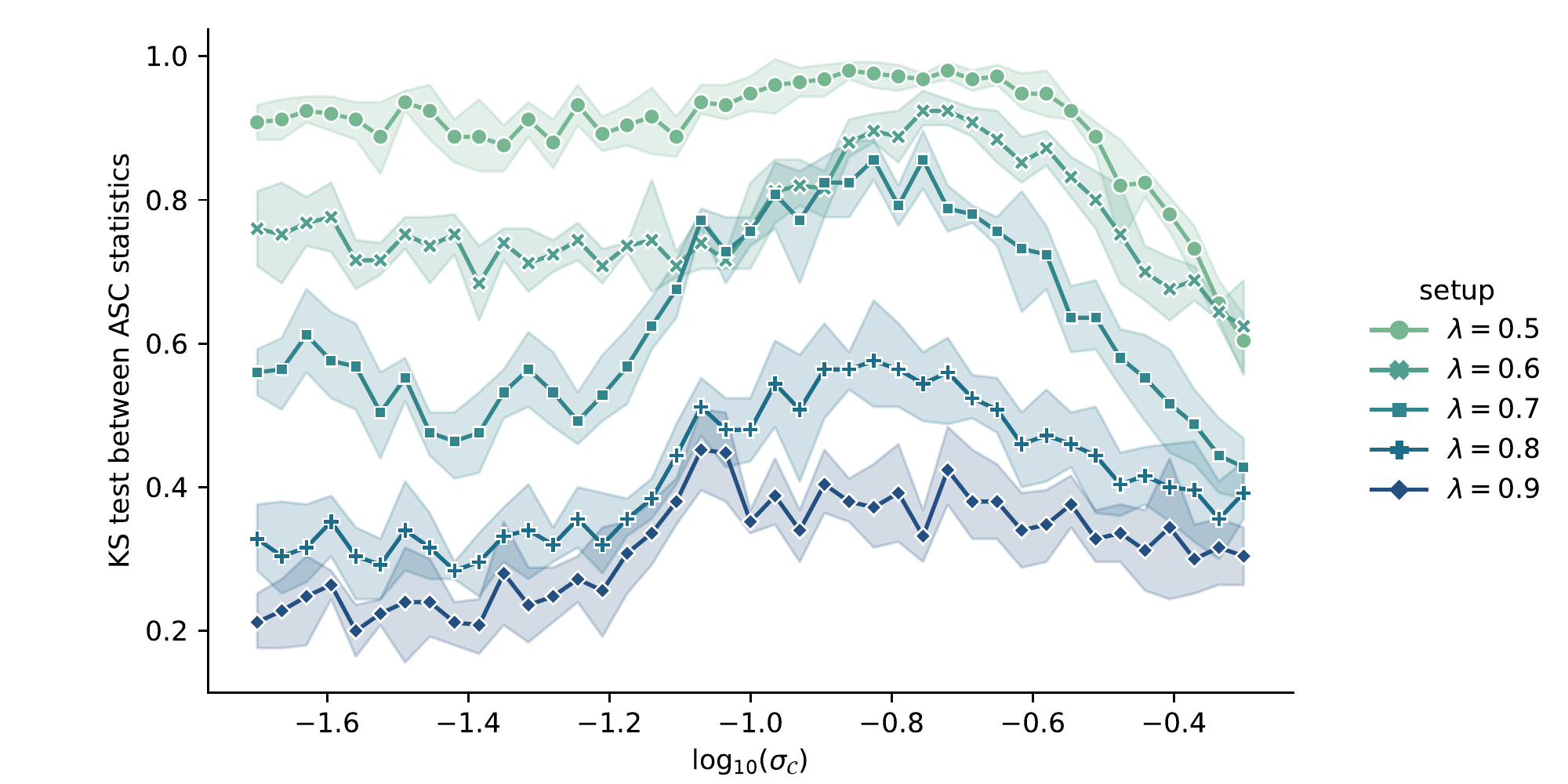}
	\caption{$\mathrm{ASC}$ statistics with $\phi(t)=(\sqrt{t}-1)^2$}
	\end{subfigure}
	
	\vspace{-0.3em}
	\caption{KS tests between distributions of statistics for KBC with different $\sigma_{\mC}$. (a) $\mathrm{LR}(Y_{H_0},\hat{\rho})$ vs $\mathrm{LR}(Y_{H_1},\hat{\rho})$ with $\lambda=0.8$ and different $m,N,R$, complementary to Fig. \ref{fig: 2d Q3 KS}. (b)-(d) $\hat{\mathrm{ASC}}_{\phi}(\hat{Y},Y_{H_0},\hat{\rho})$ vs $\hat{\mathrm{ASC}}_{\phi}(\hat{Y},Y_{H_1},\hat{\rho})$ for different $\phi$. Smaller values indicate the two compared distributions are closer. }
	\vspace{-0.3em}
	\label{fig: 2d Q3 KS MoG-8 appendix}
\end{figure}

\newpage
We visualize KS test results for KBC with different bandwidth $\sigma_{\mC}$ in Fig. \ref{fig: 2d Q3 KS MoG-8 appendix} (extension of Fig. \ref{fig: 2d Q3 KS}). The KS values are large for a wide range of $\sigma_{\mC}$, indicating KBC with these $\sigma_{\mC}$ can nicely distinguish pre-trained and re-trained model.  LR statistics are better than ASC statistics. In terms of $\lambda$, the models can be more easily distinguished when $\lambda$ is small, as expected.

\newpage
\subsection{CKB-8}\label{appendix: exp 2d CKB}

\paragraph{Setup.}

\begin{figure}[!h]
\vspace{-0.2em} 
    	\begin{subfigure}[b]{0.15\textwidth}
		\centering 
		\includegraphics[trim=25 20 105 20, clip, width=0.99\textwidth]{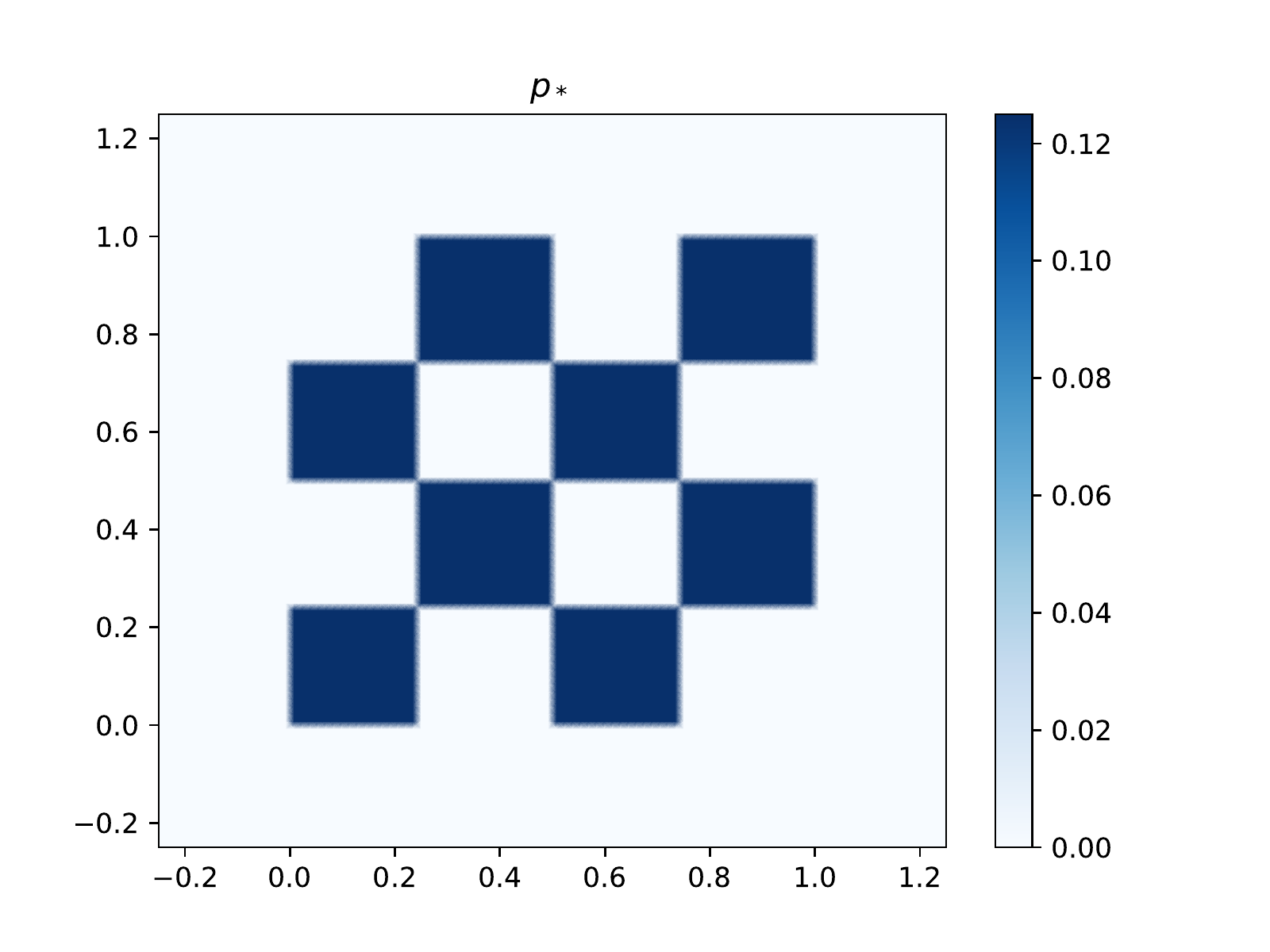}
		\caption{$p_*$}
	\end{subfigure}
	\begin{subfigure}[b]{0.15\textwidth}
		\centering 
		\includegraphics[trim=25 20 105 20, clip, width=0.99\textwidth]{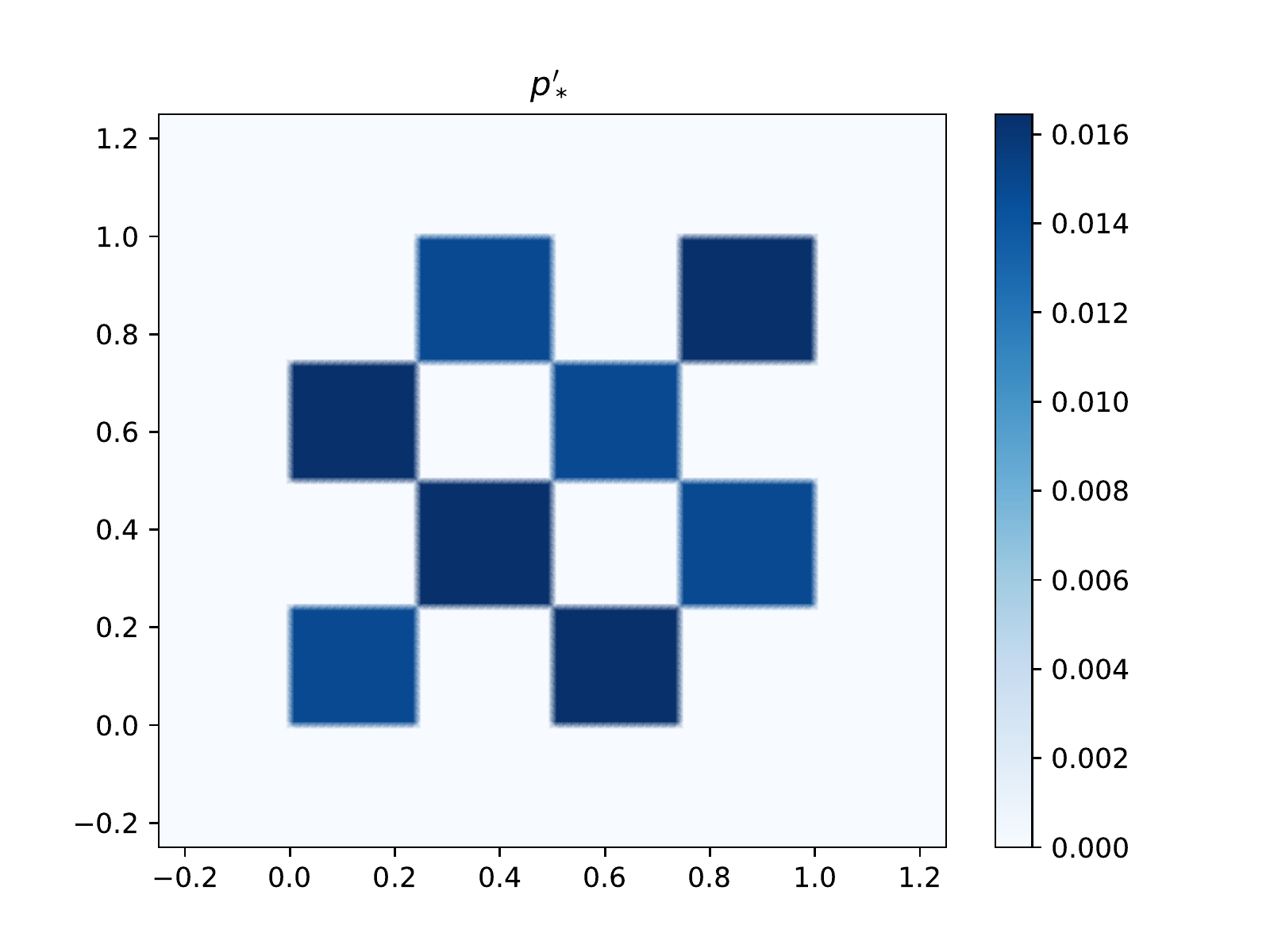}
		\caption{$p_*'(\lambda=0.9)$}
	\end{subfigure}
	\begin{subfigure}[b]{0.15\textwidth}
		\centering 
		\includegraphics[trim=25 20 105 20, clip, width=0.99\textwidth]{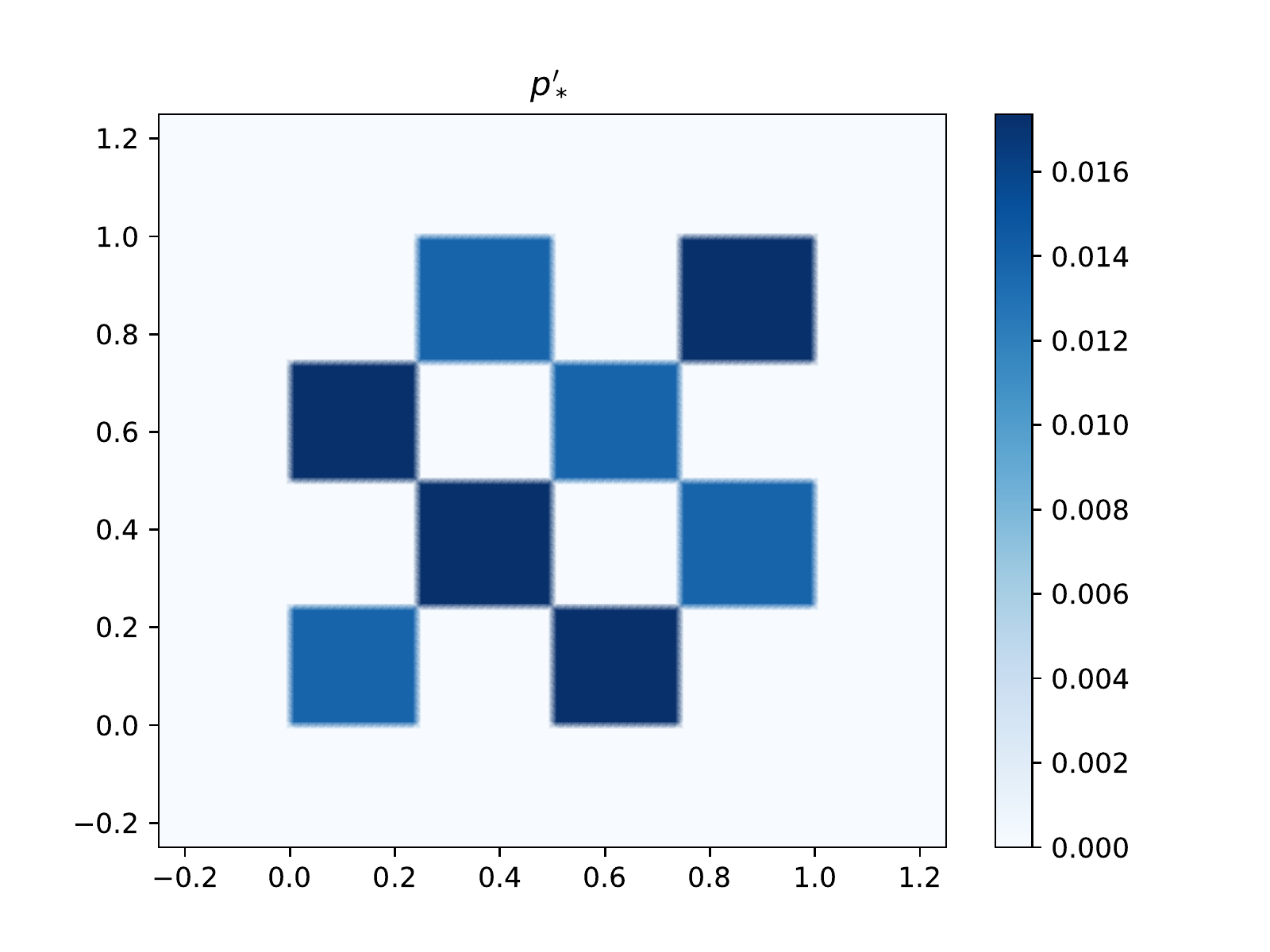}
		\caption{$p_*'(\lambda=0.8)$}
	\end{subfigure}
	\begin{subfigure}[b]{0.15\textwidth}
		\centering 
		\includegraphics[trim=25 20 105 20, clip, width=0.99\textwidth]{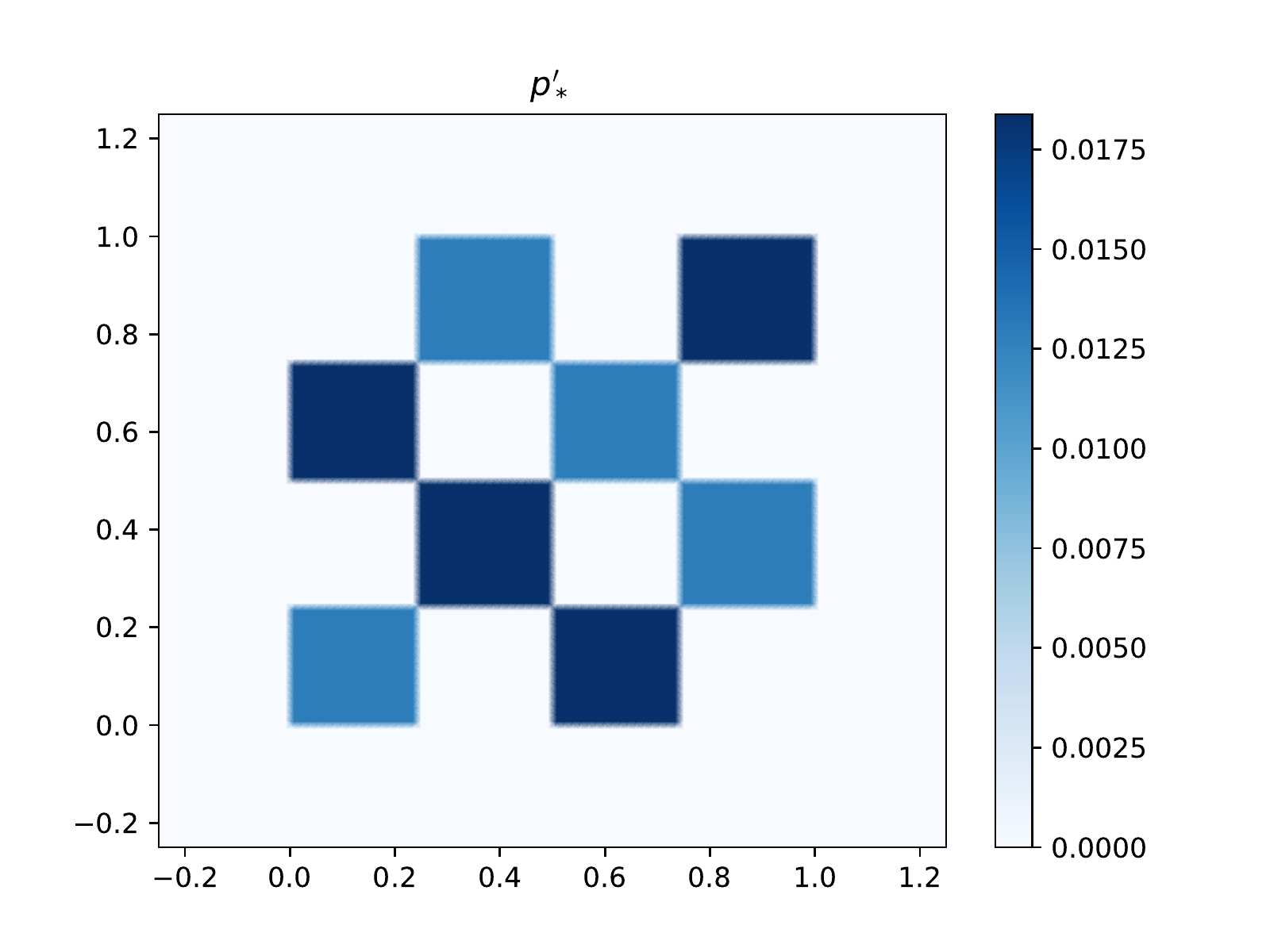}
		\caption{$p_*'(\lambda=0.7)$}
	\end{subfigure}
	\begin{subfigure}[b]{0.15\textwidth}
		\centering 
		\includegraphics[trim=25 20 105 20, clip, width=0.99\textwidth]{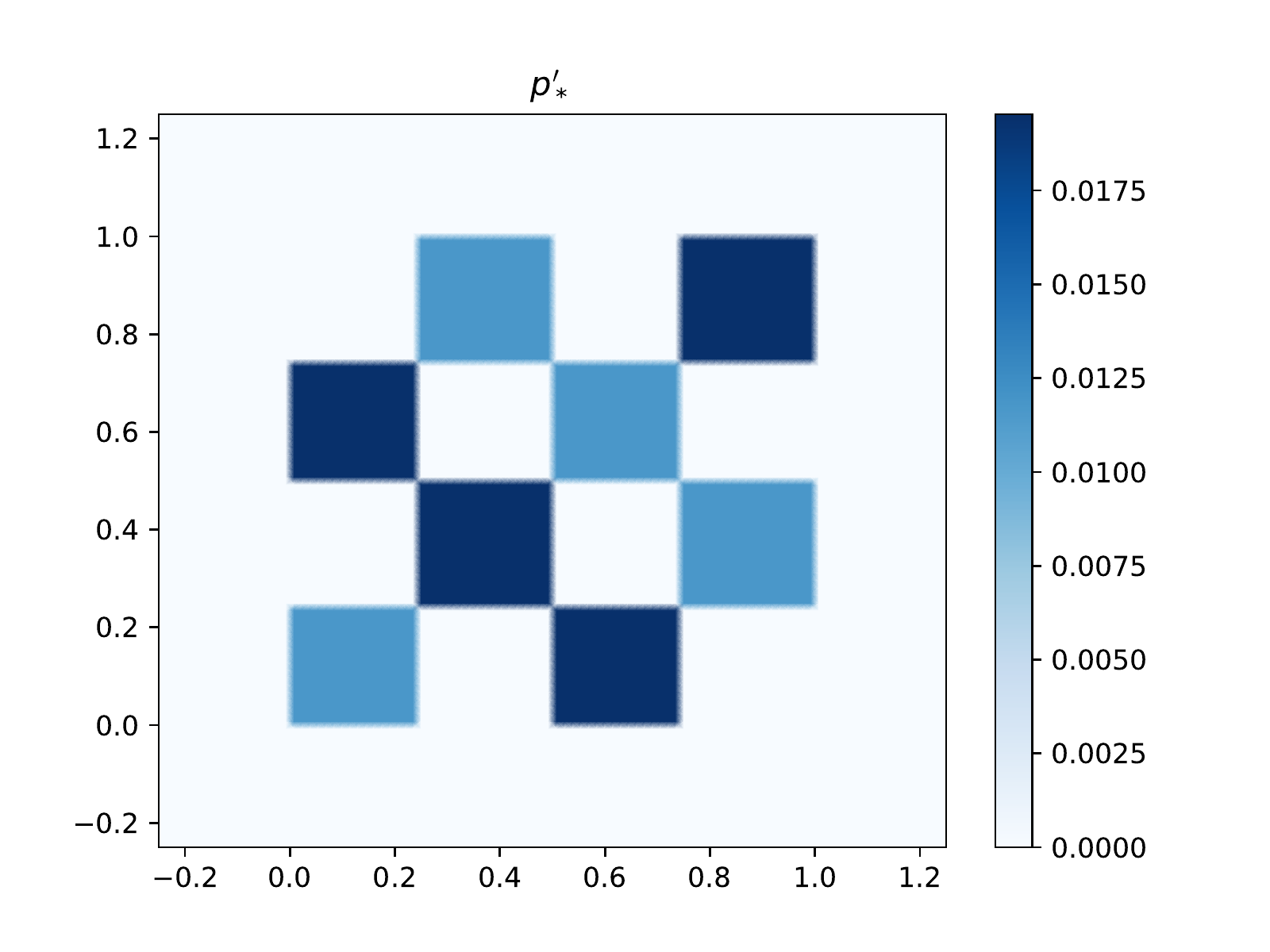}
		\caption{$p_*'(\lambda=0.6)$}
	\end{subfigure}
	\begin{subfigure}[b]{0.15\textwidth}
		\centering 
		\includegraphics[trim=25 20 105 20, clip, width=0.99\textwidth]{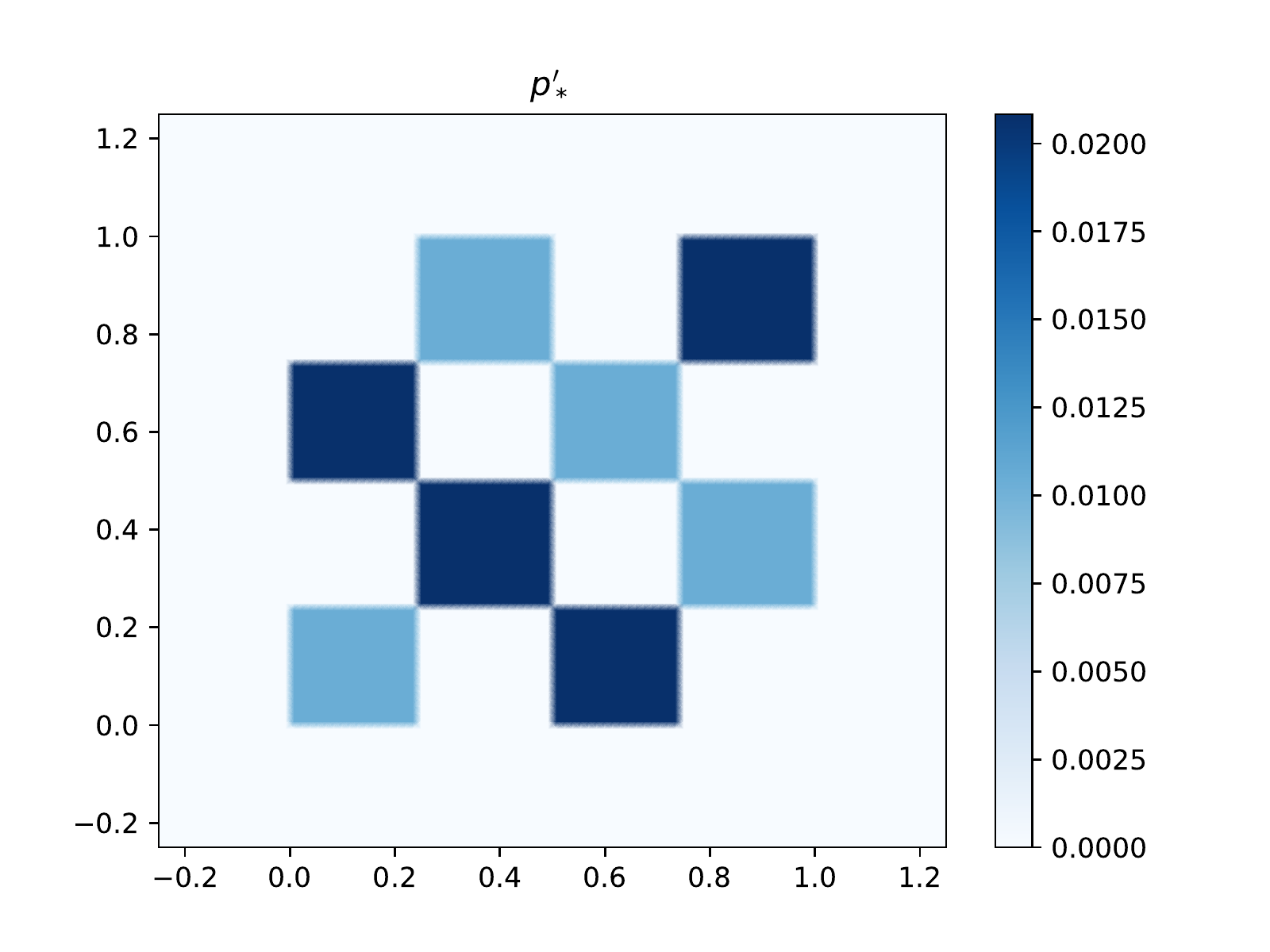}
		\caption{$p_*'(\lambda=0.5)$}
	\end{subfigure}

	\vspace{-0.3em}
	\caption{Visualization of the experimental setup of CKB-8. (a) Data distribution $p_*$. (b) - (f) $p_*'$ with different $\lambda$ values. A larger $\lambda$ means less data is deleted.}
	\label{fig: 2d setup CKB-8 appendix 1}
	\vspace{-0.2em}
\end{figure}

\begin{figure}[!h]
\vspace{-0.3em} 
    	\begin{subfigure}[b]{0.15\textwidth}
		\centering 
		\includegraphics[trim=25 20 105 20, clip, width=0.99\textwidth]{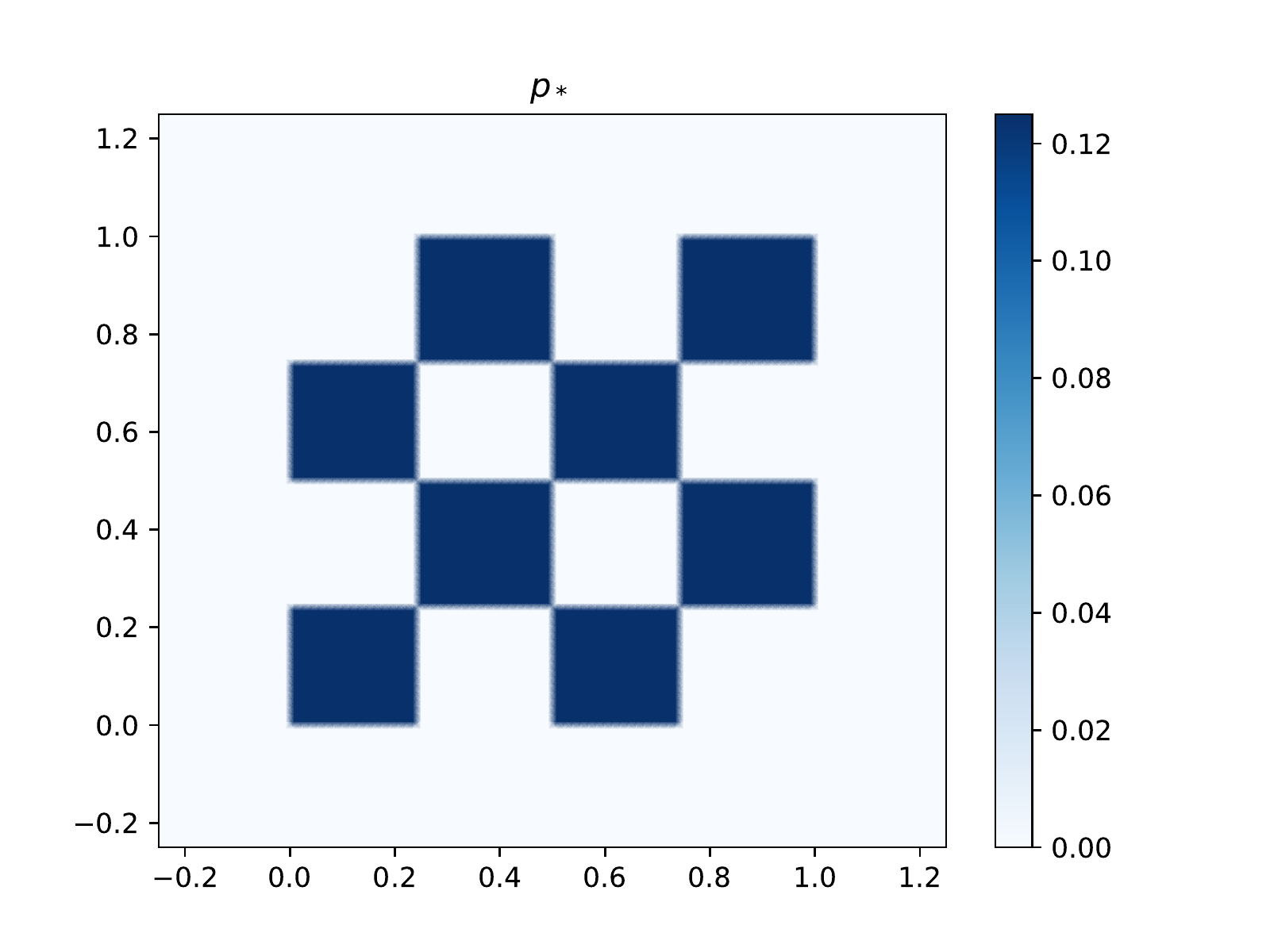}
		\caption{$p_*$}
	\end{subfigure}
	\begin{subfigure}[b]{0.15\textwidth}
		\centering 
		\includegraphics[trim=25 20 105 20, clip, width=0.99\textwidth]{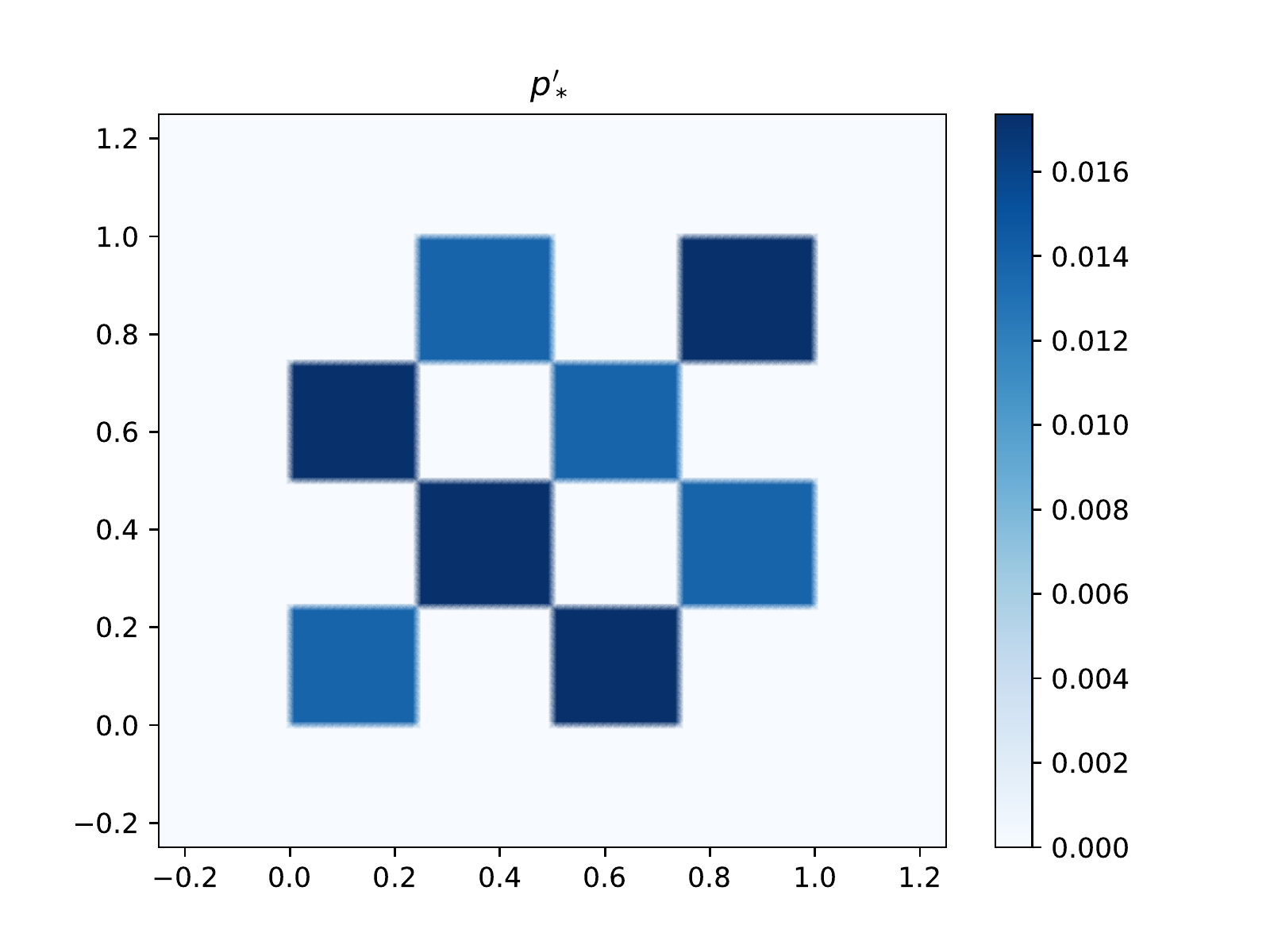}
		\caption{$p_*'$}
	\end{subfigure}
	\begin{subfigure}[b]{0.15\textwidth}
		\centering 
		\includegraphics[trim=25 20 105 20, clip, width=0.99\textwidth]{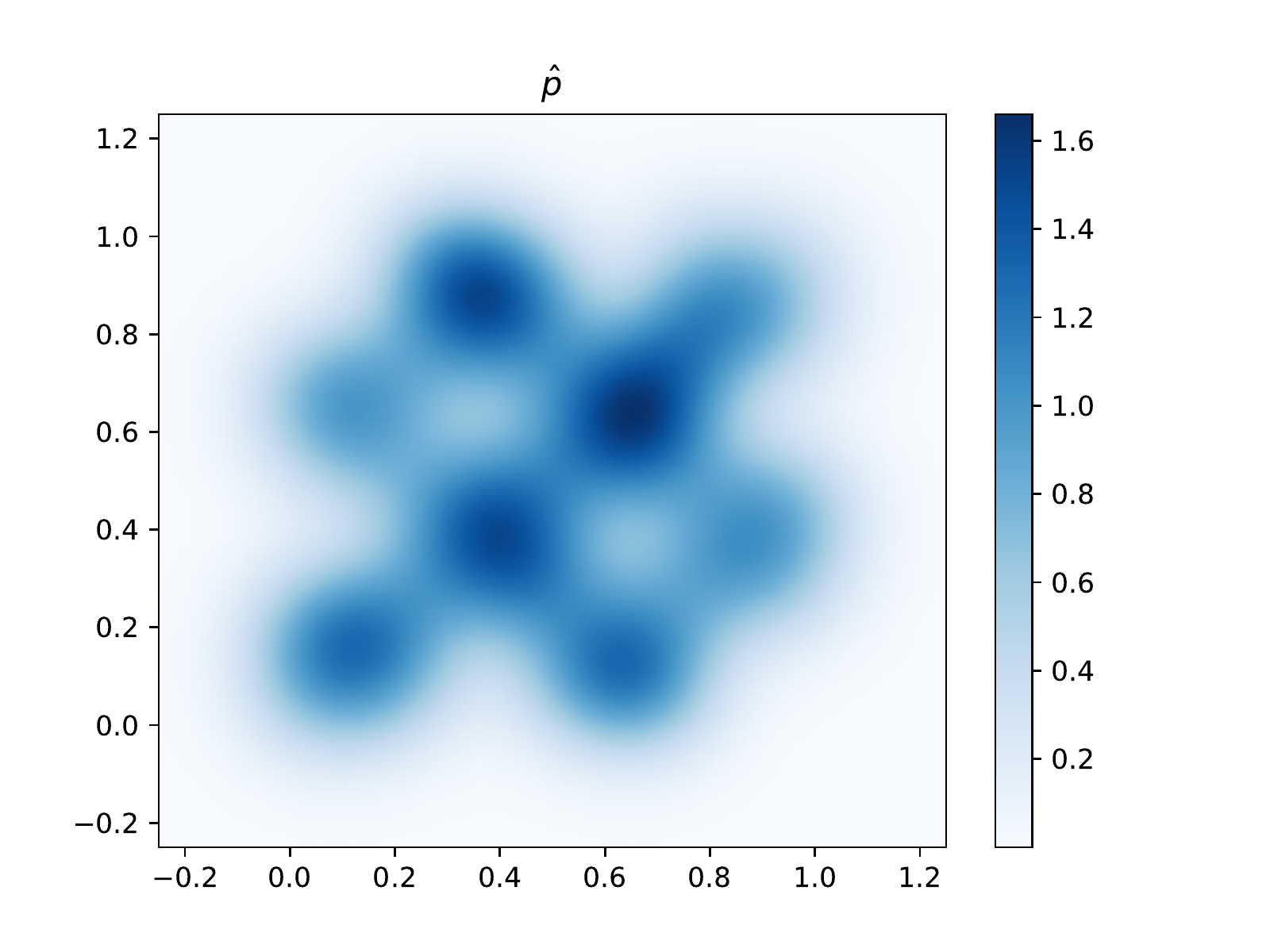}
		\caption{$\hat{p}$}
	\end{subfigure}
	\begin{subfigure}[b]{0.15\textwidth}
		\centering 
		\includegraphics[trim=25 20 105 20, clip, width=0.99\textwidth]{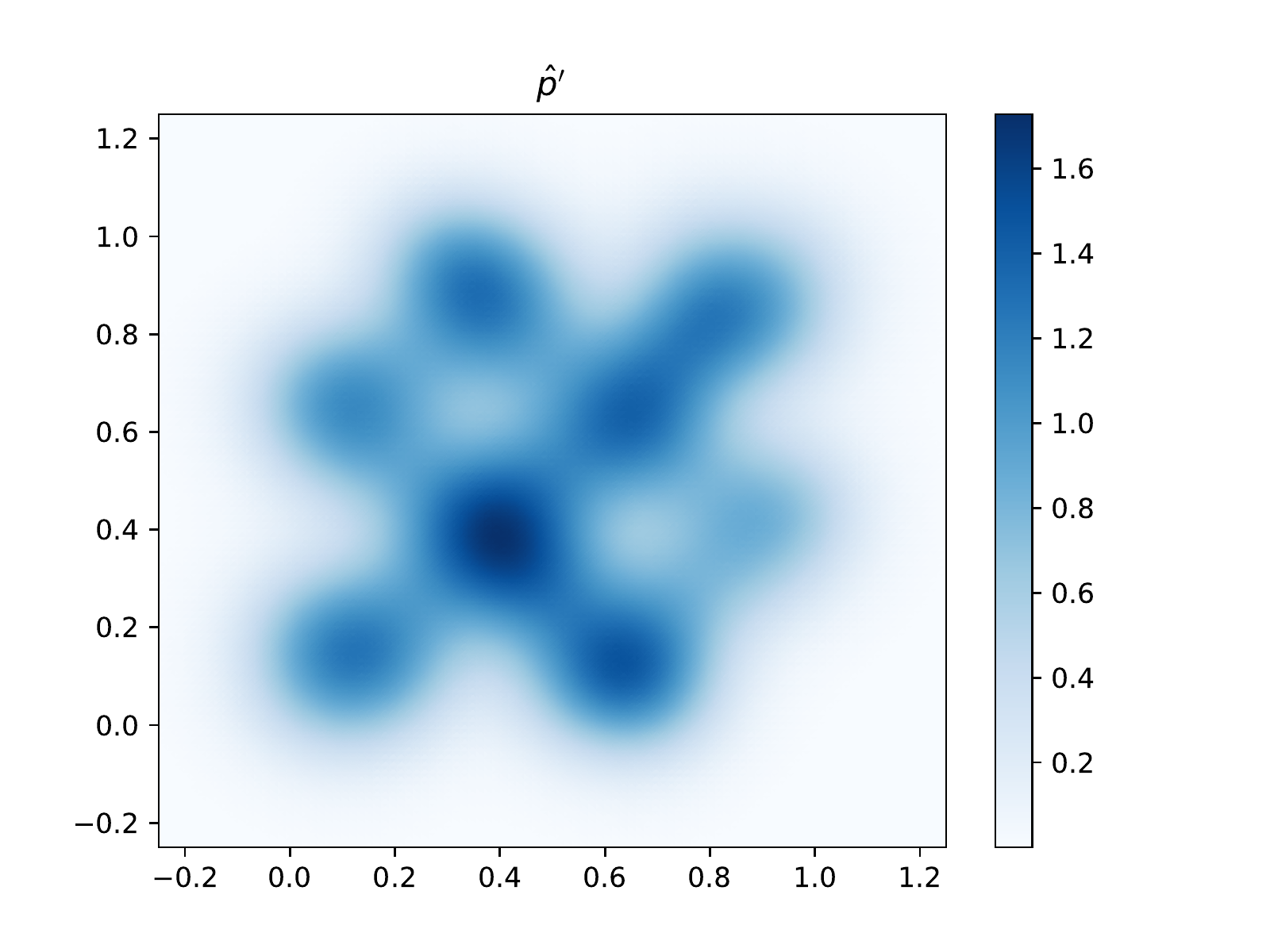}
		\caption{$\hat{p}'$}
	\end{subfigure}
	\begin{subfigure}[b]{0.175\textwidth}
		\centering 
		\includegraphics[trim=25 20 50 20, clip, width=0.99\textwidth]{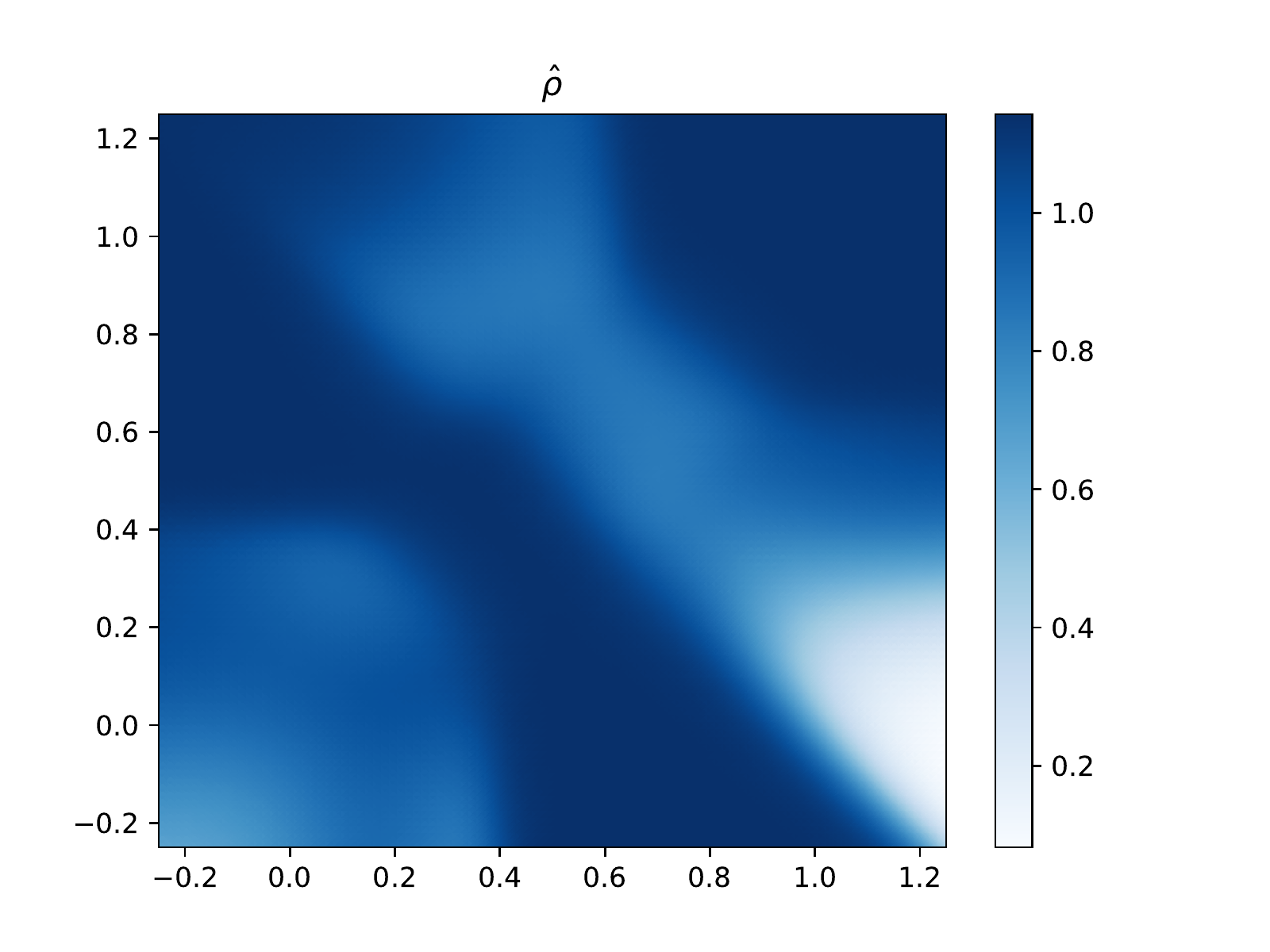}
		\caption{$\hat{\rho}$}
	\end{subfigure}
	\begin{subfigure}[b]{0.18\textwidth}
		\centering 
		\includegraphics[width=0.99\textwidth]{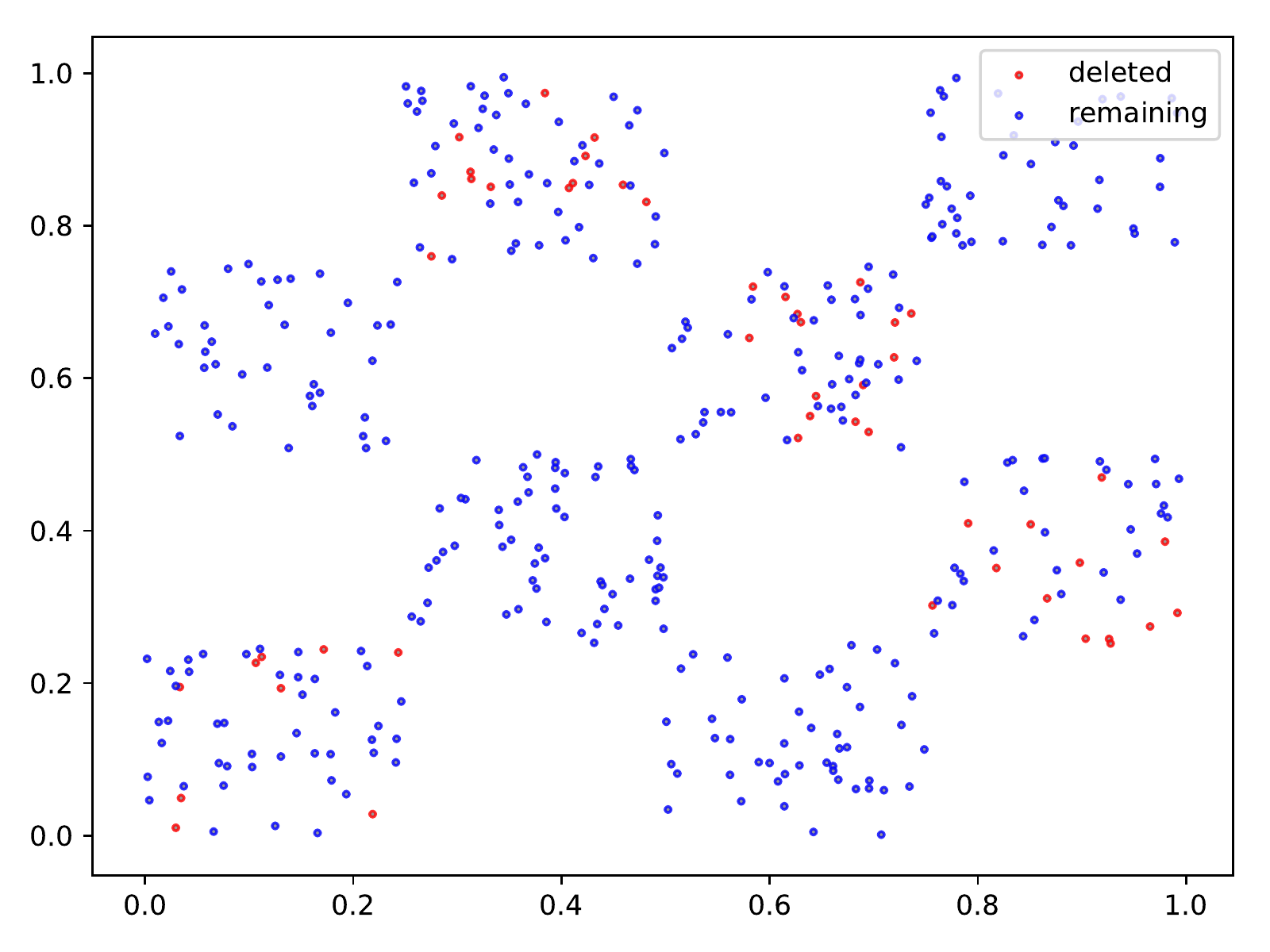}
		\caption{\textcolor{red}{$X'$} and \textcolor{blue}{$X\setminus X'$}}
	\end{subfigure}
	\vspace{-0.3em}
	\caption{Visualization of the experimental setup of CKB-8. (a) Data distribution $p_*$. (b) Distribution $p_*'$ with $\lambda=0.8$. (c) Pre-trained KDE $\hat{p}$ on $X$ with $\sigma_{\mA}=0.1$. (d) Re-trained KDE $\hat{p}'$ on $X\setminus X'$ with $\sigma_{\mA}=0.1$. (e) Density ratio $\hat{\rho}=\hat{p}'/\hat{p}$. (f) Deletion set $X'$ and the remaining set $X\setminus X'$.}
	\label{fig: 2d setup CKB-8 appendix 2}
\end{figure}

The data distribution is defined as
\[p_* = \texttt{Uniform}(\cup_{i=1}^8\Omega_i),\]
where $
\Omega_1=[0,0.25]\times[0,0.25],
\Omega_2=[0,0.25]\times[0.5,0.75],
\Omega_3=[0.25,0.5]\times[0.25,0.5],
\Omega_4=[0.25,0.5]\times[0.75,1],
\Omega_5=[0.5,0.75]\times[0,0.25],
\Omega_6=[0.5,0.75]\times[0.5,0.75],
\Omega_7=[0.75,1]\times[0.25,0.5],
\Omega_8=[0.75,1]\times[0.75,1]$.
The modified distribution $p_*'$ with weight $\lambda$ is defined as
\[p_*'(x) = \frac{1}{4(1+\lambda)}\sum_{i=1}^8 w_i\cdot\texttt{Uniform}(\Omega_i),\]
where $w_i=1$ for $i\in\{2,3,5,8\}$ and $\lambda$ for $i\in\{1,4,6,7\}$. The construction algorithm for $X$ is randomly sampling a square id between 1 and 8 and randomly drawing a sample from the corresponding uniform distribution. The construction algorithm for $X'$ is to include a sample $x\in X$ with probability $1-\lambda$ if $x$ is from $i$-th square for $i\in\{1,4,6,7\}$. The distributions and data with different $\lambda$ are shown in Fig. \ref{fig: 2d setup CKB-8 appendix 1} and \ref{fig: 2d setup CKB-8 appendix 2}.

Other hyperparameters are set as follows. The number of training samples $N=400$ unless specified. The number of samples for the deletion test $m=400$ unless specified. The number of repeats for each setup is $R=250$ unless specified. The learning algorithm KDE has bandwidth $\sigma_{\mA}=0.1$ unless specified.

\newpage
\paragraph{Question 1 (DRE Approximations).}

\begin{figure}[!h]
\vspace{-0.3em}
  	\begin{subfigure}[t!]{0.19\textwidth}
		\centering 
		\begin{overpic}[trim=25 20 50 37, clip, width=\textwidth]{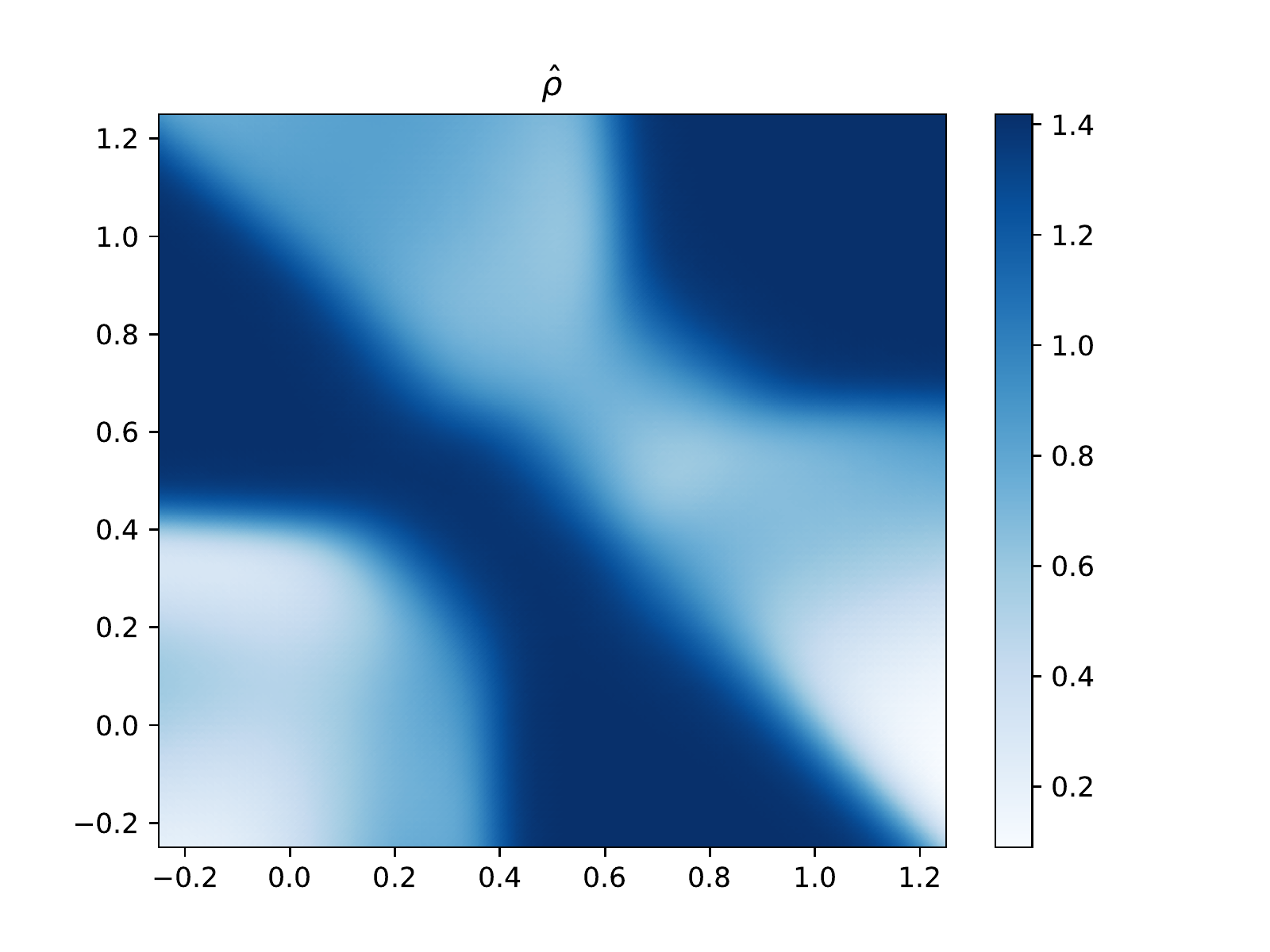} \put(-40,34){$\lambda=0.5$}\end{overpic}\\
		\begin{overpic}[trim=25 20 50 37, clip, width=\textwidth]{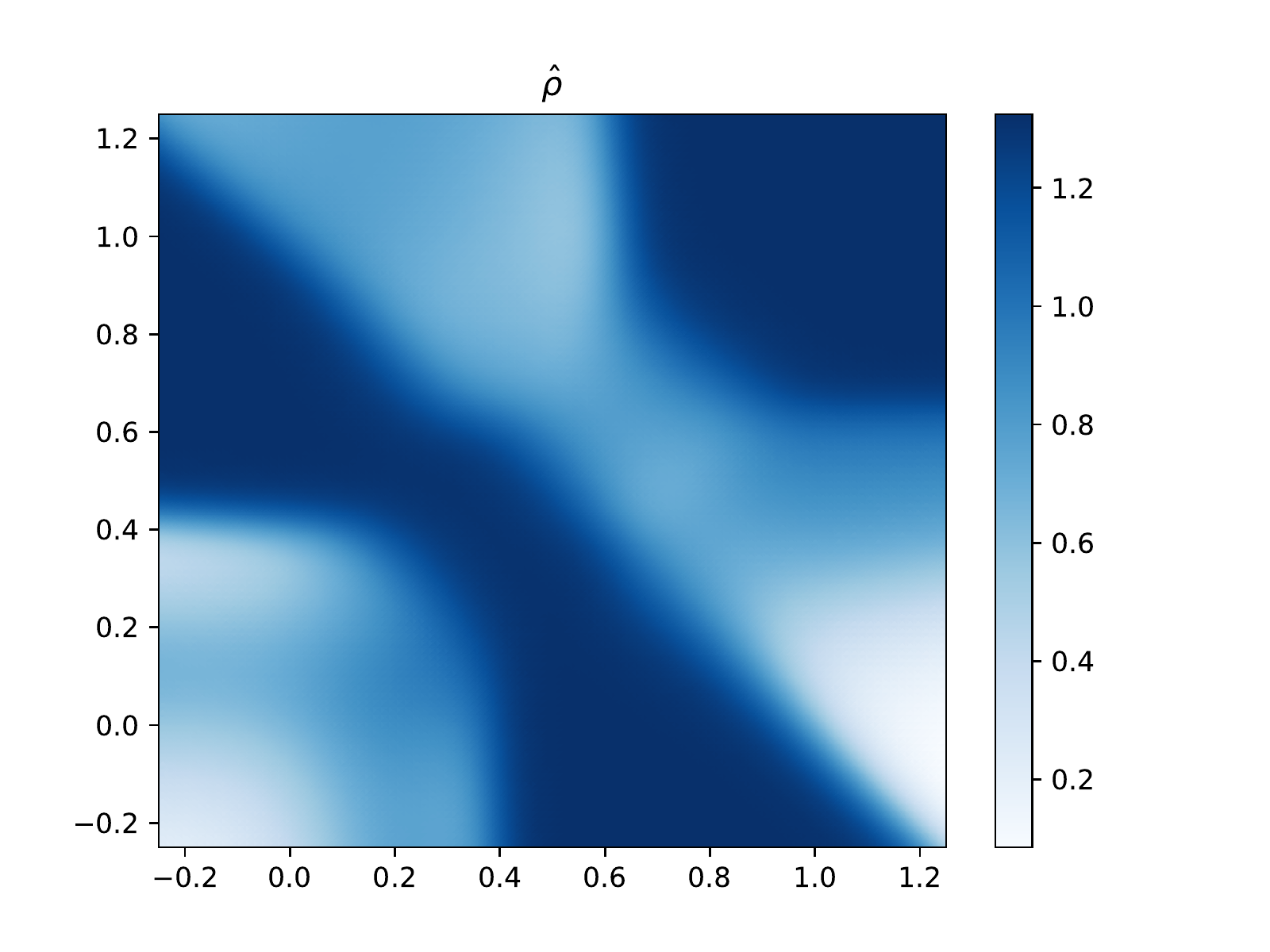} \put(-40,34){$\lambda=0.6$}\end{overpic}\\
		\begin{overpic}[trim=25 20 50 37, clip, width=\textwidth]{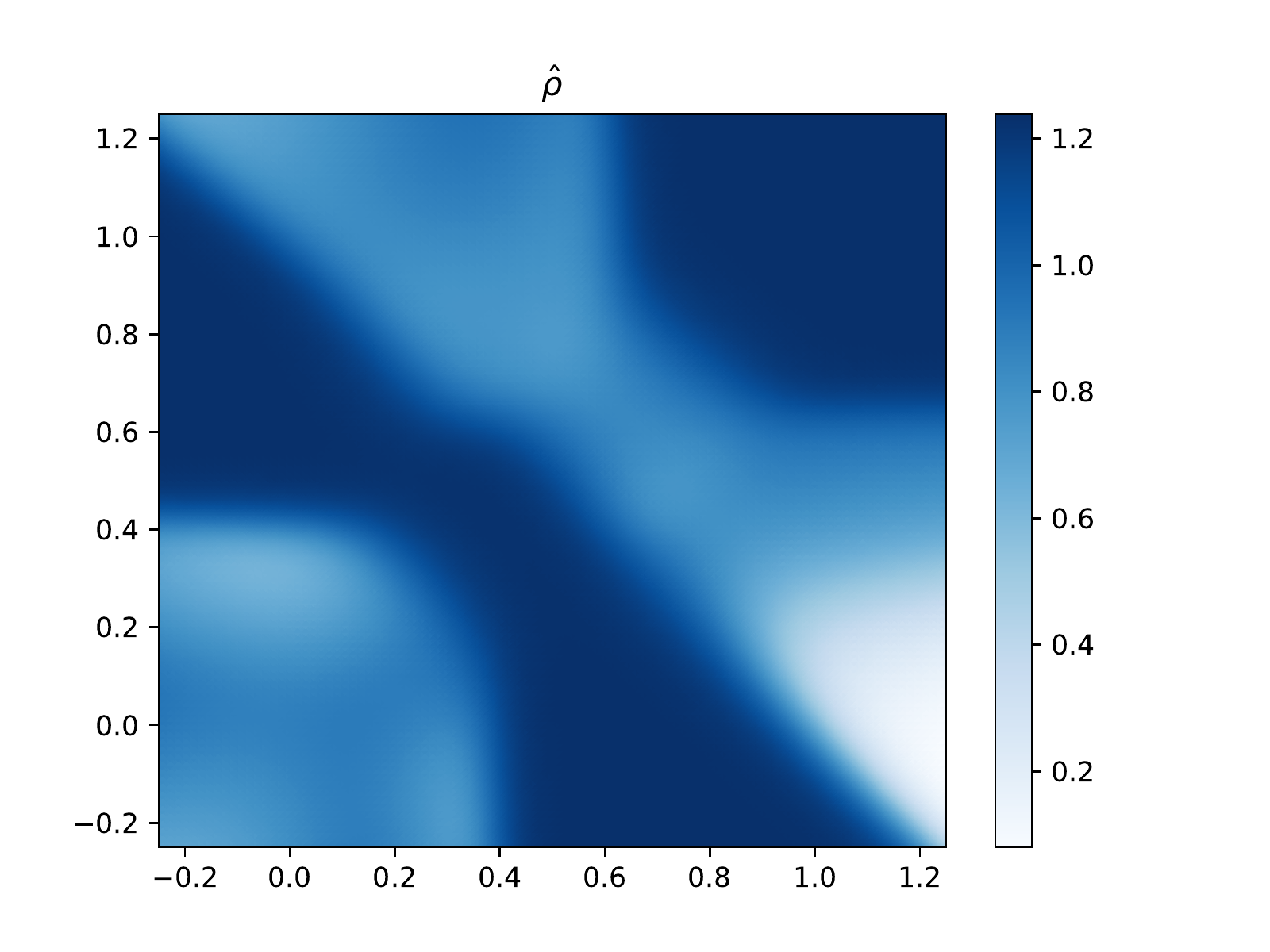} \put(-40,34){$\lambda=0.7$}\end{overpic}\\
		\begin{overpic}[trim=25 20 50 37, clip, width=\textwidth]{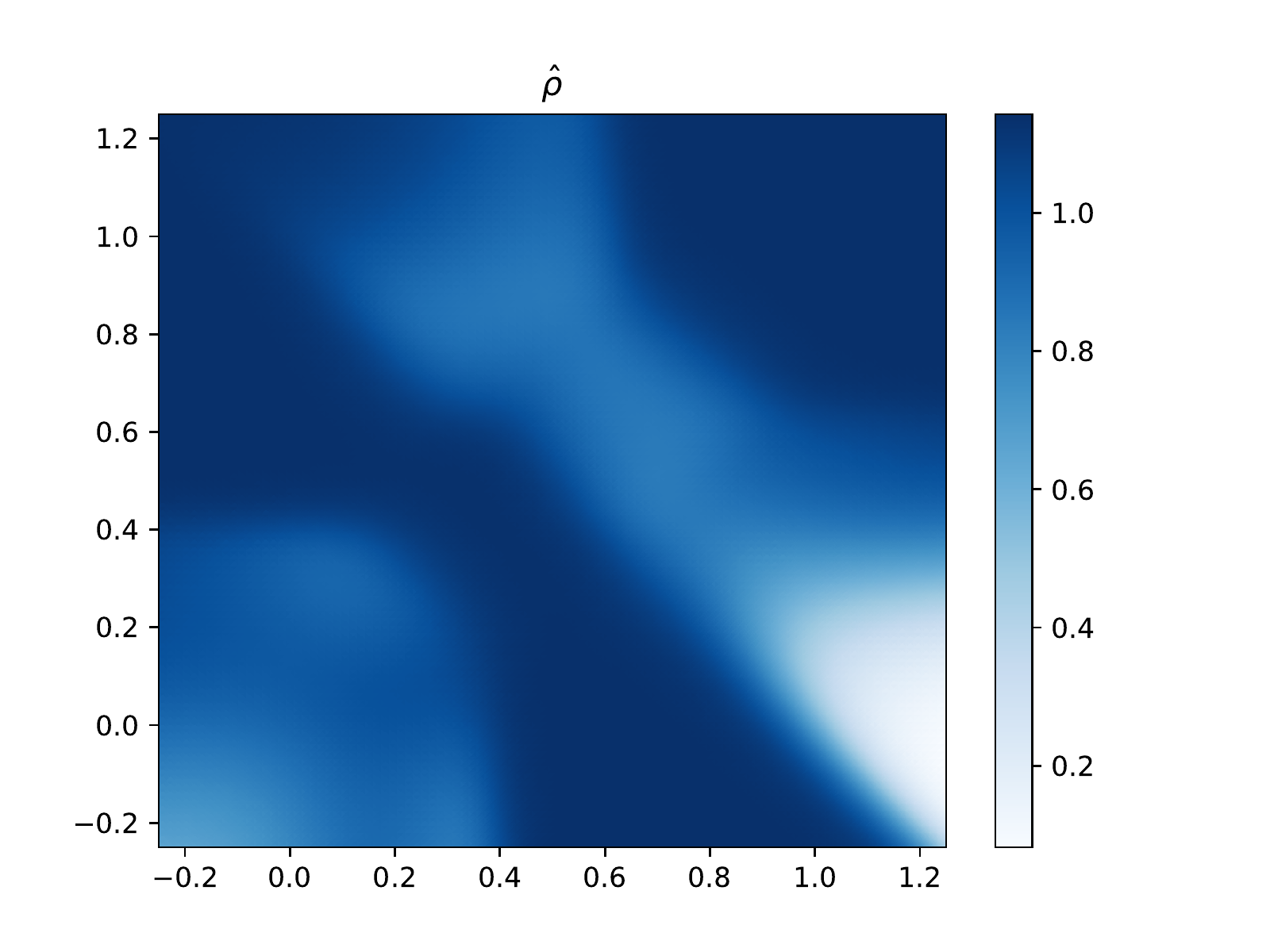} \put(-40,34){$\lambda=0.8$}\end{overpic}\\
		\begin{overpic}[trim=25 20 50 37, clip, width=\textwidth]{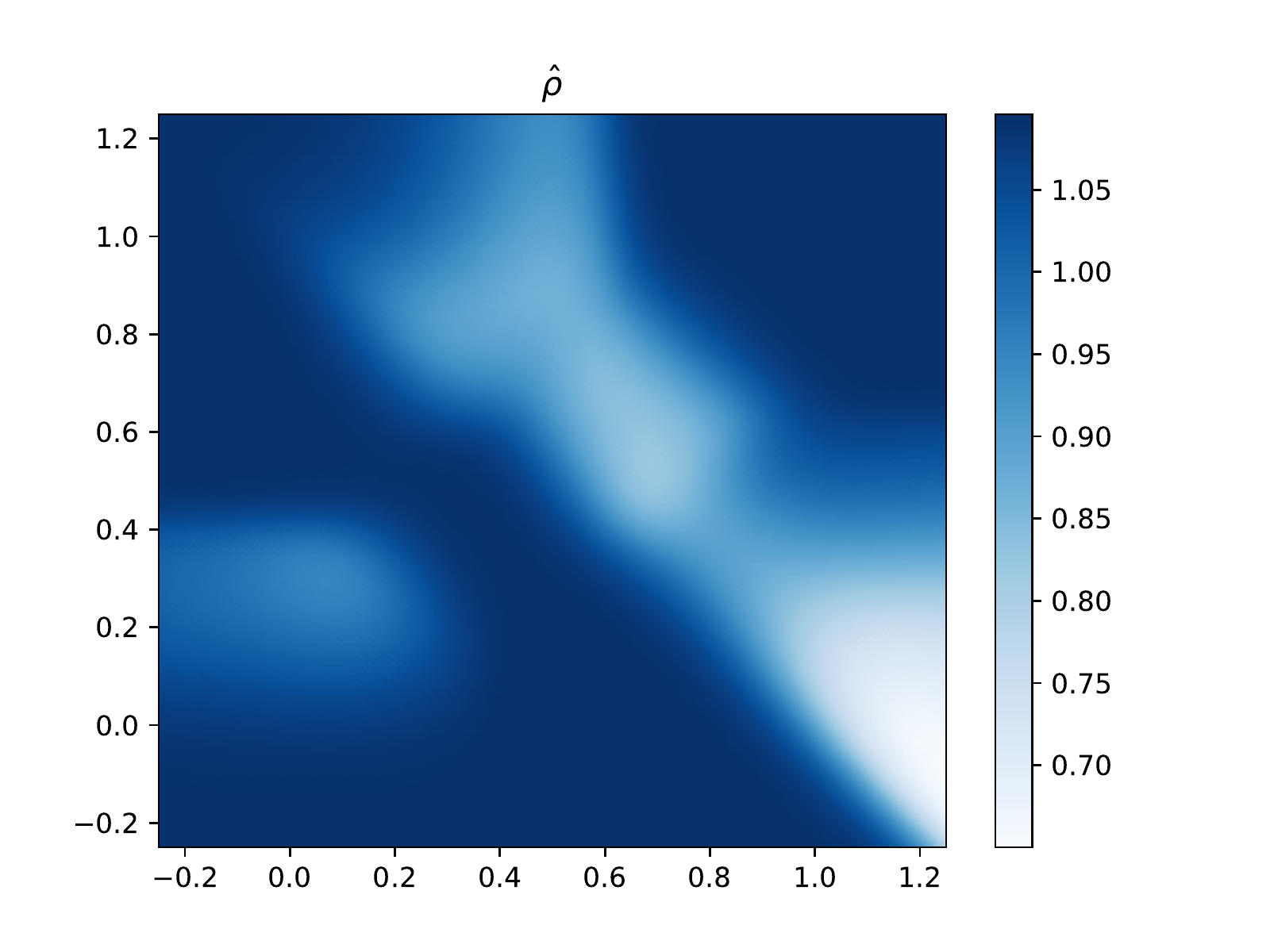} \put(-40,34){$\lambda=0.9$}\end{overpic}\\
		\caption{$\hat{\rho}$}
	\end{subfigure}
	\begin{subfigure}[t!]{0.19\textwidth}
		\centering 
		\includegraphics[trim=25 20 50 37, clip, width=\textwidth]{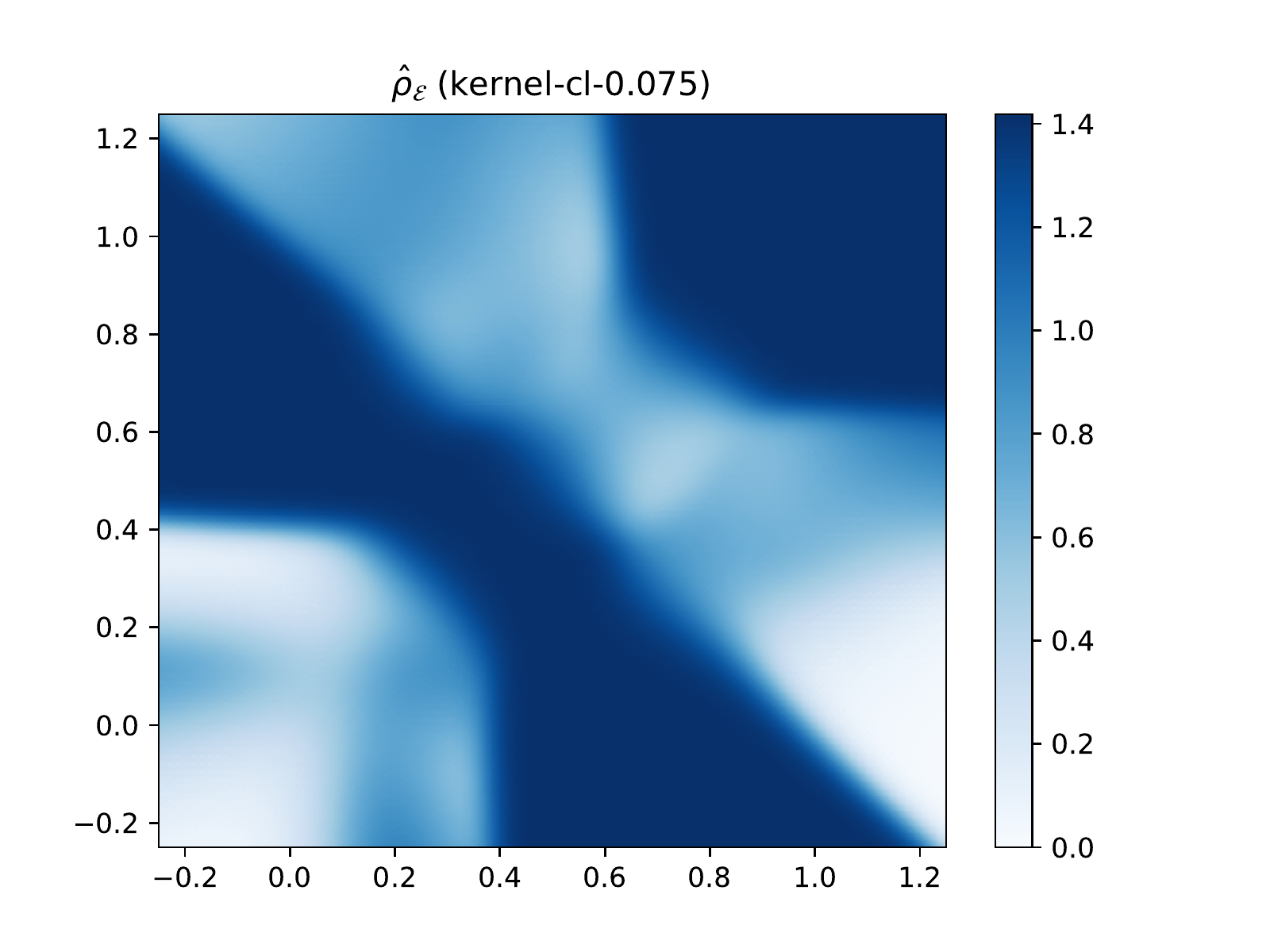}\\
		\includegraphics[trim=25 20 50 37, clip, width=\textwidth]{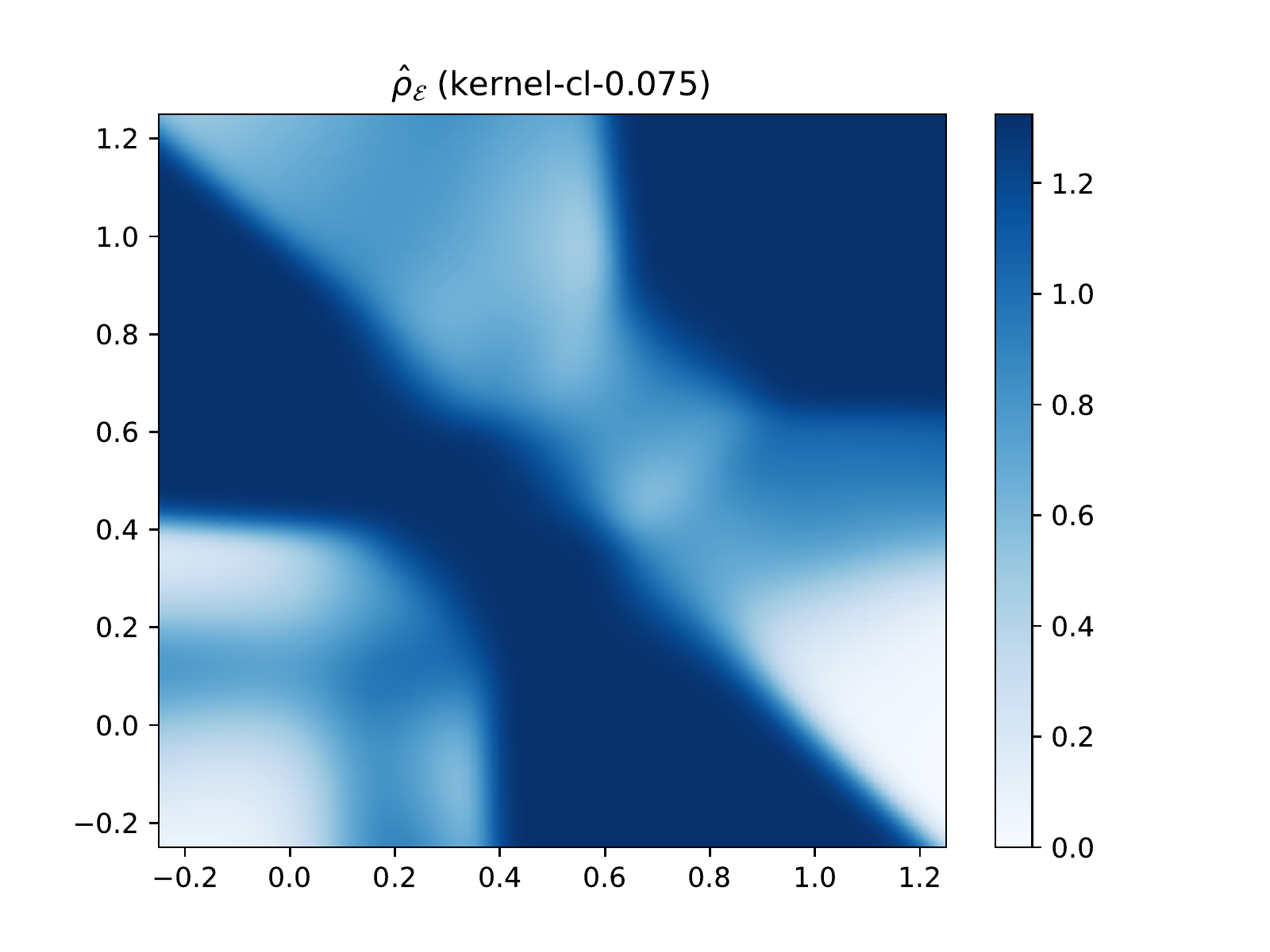}\\
		\includegraphics[trim=25 20 50 37, clip, width=\textwidth]{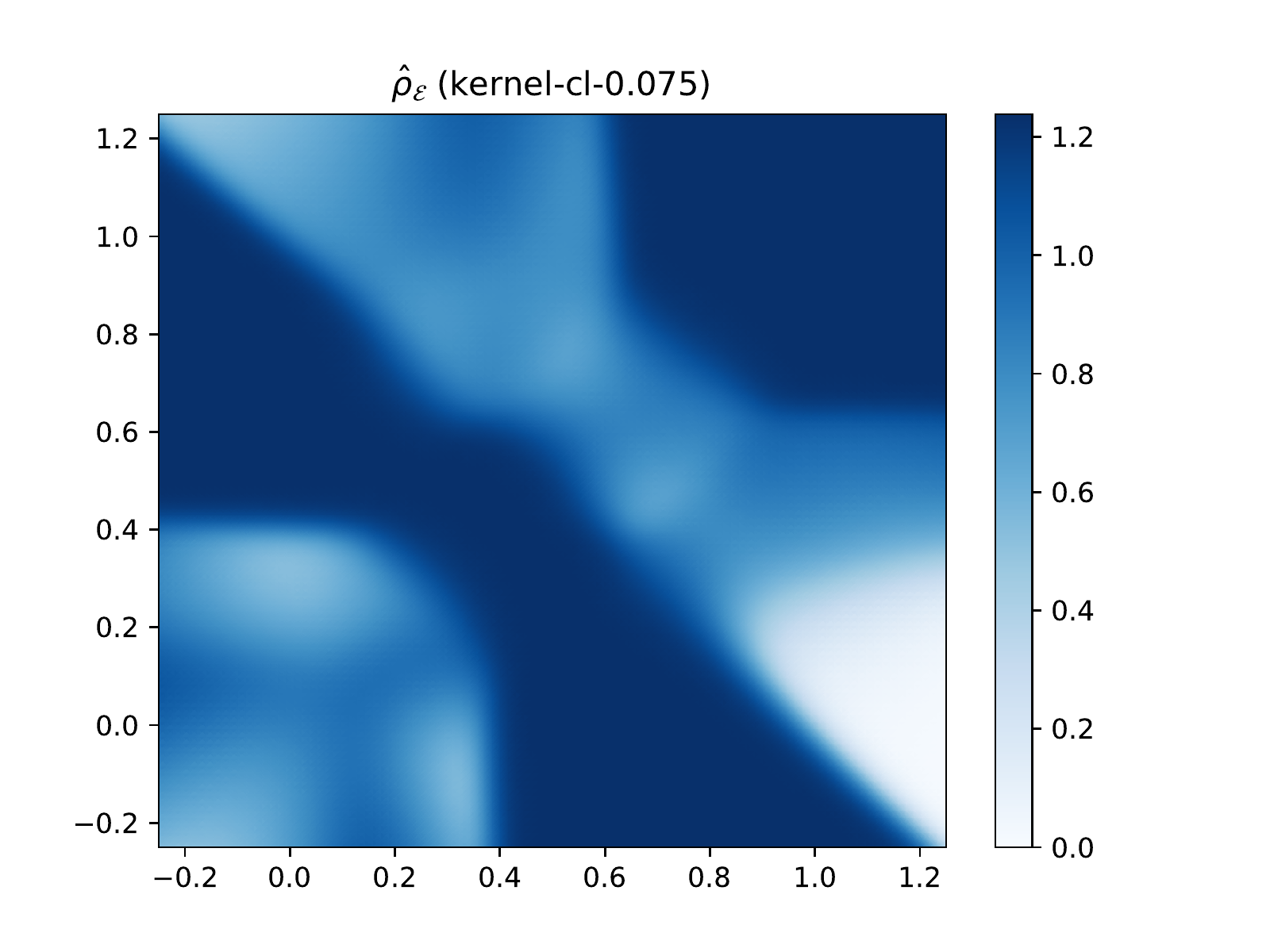}\\
		\includegraphics[trim=25 20 50 37, clip, width=\textwidth]{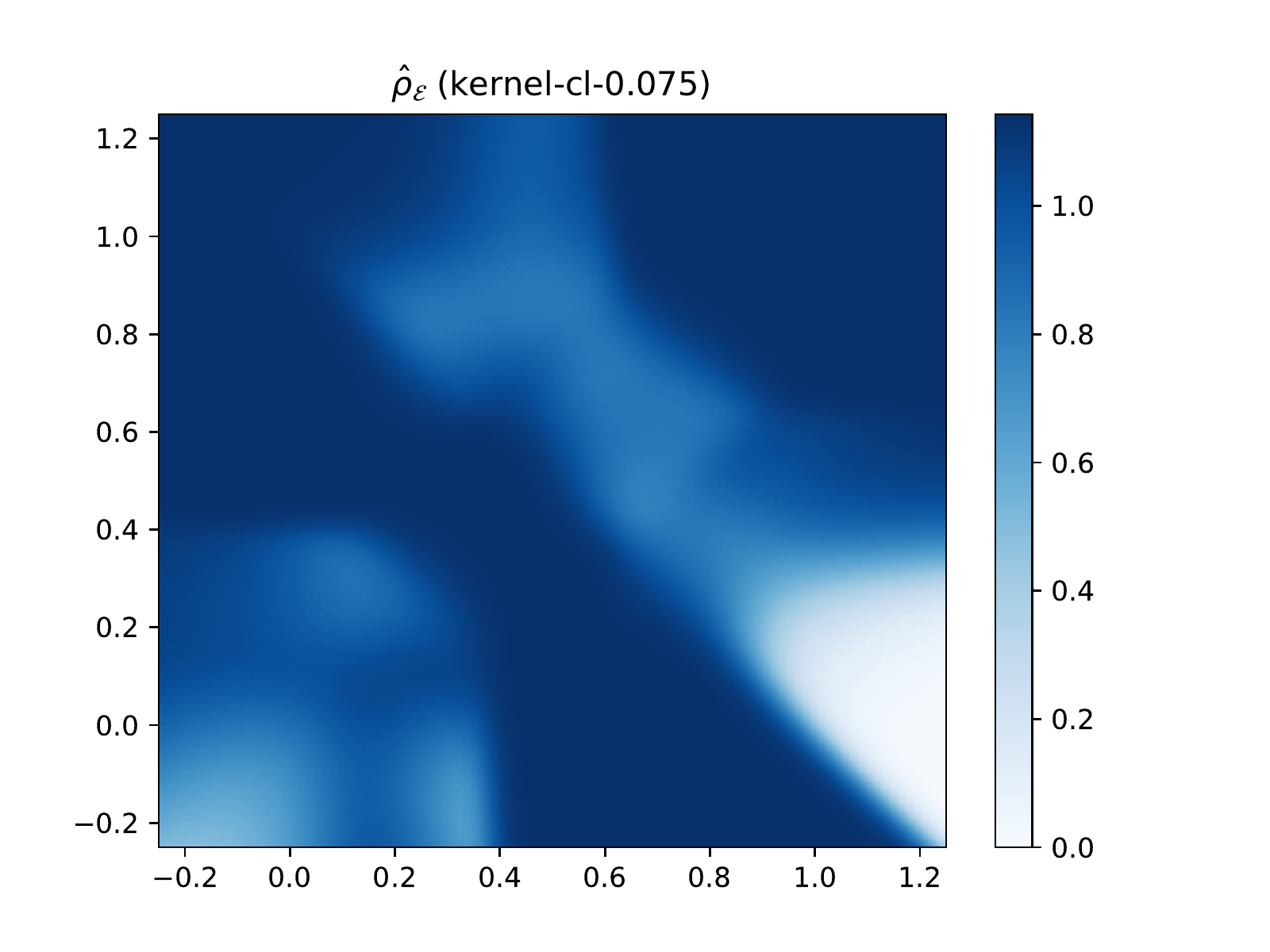}\\
		\includegraphics[trim=25 20 50 37, clip, width=\textwidth]{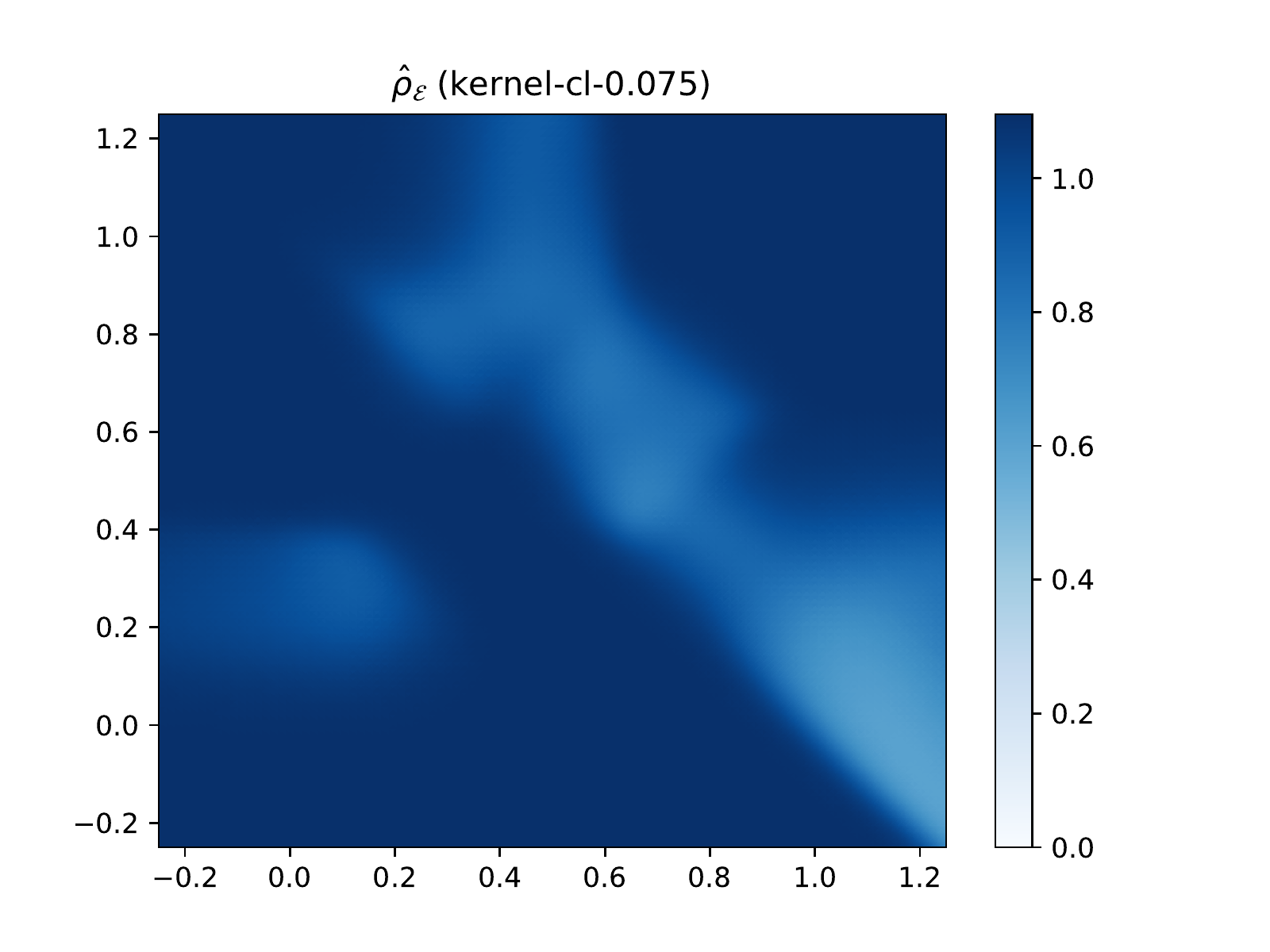}\\
		\caption{KBC ($\sigma_{\mC}$=$0.075$)}
	\end{subfigure}
	\begin{subfigure}[t!]{0.19\textwidth}
		\centering 
		\includegraphics[trim=25 20 50 37, clip, width=\textwidth]{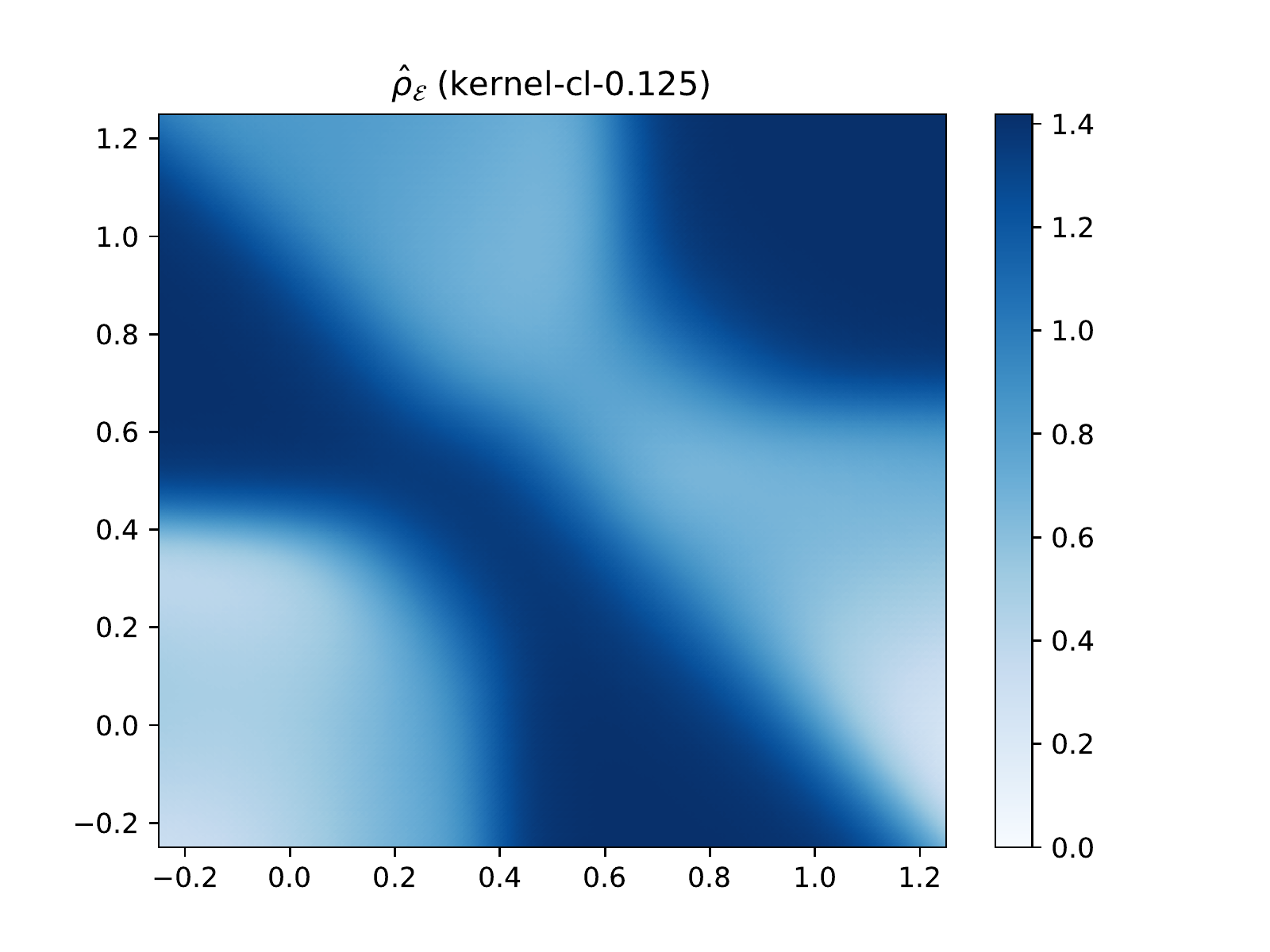}\\
		\includegraphics[trim=25 20 50 37, clip, width=\textwidth]{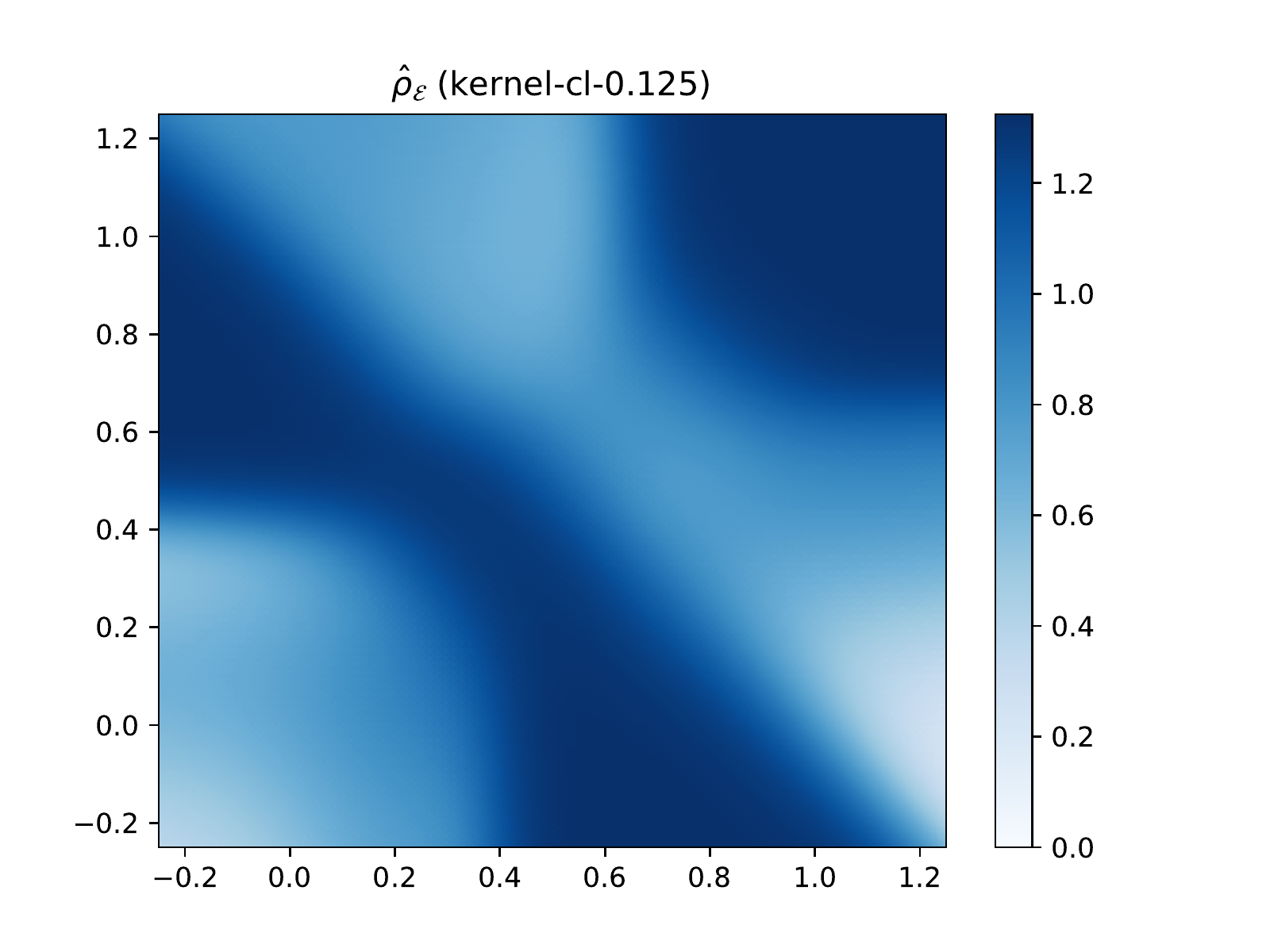}\\
		\includegraphics[trim=25 20 50 37, clip, width=\textwidth]{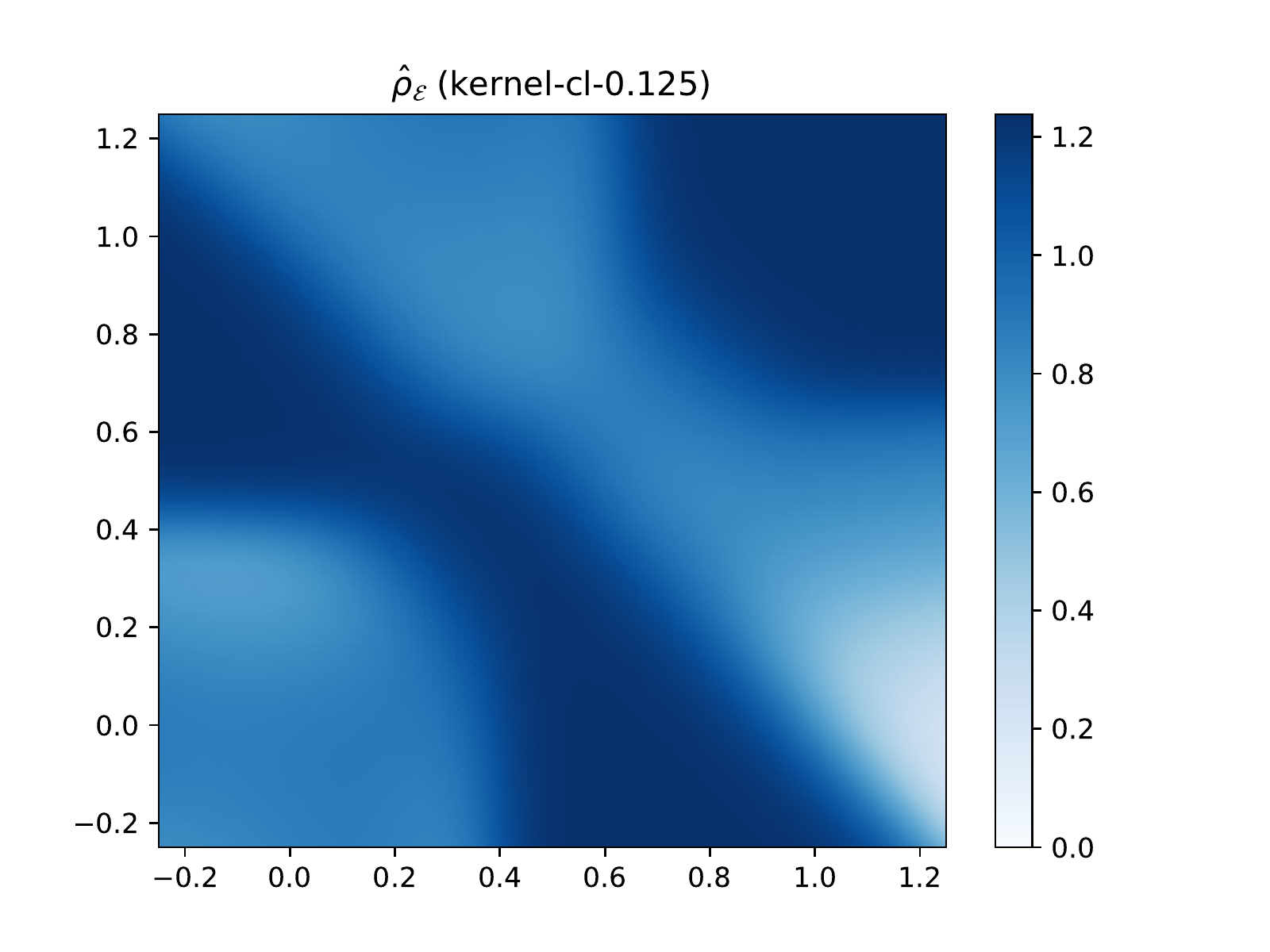}\\
		\includegraphics[trim=25 20 50 37, clip, width=\textwidth]{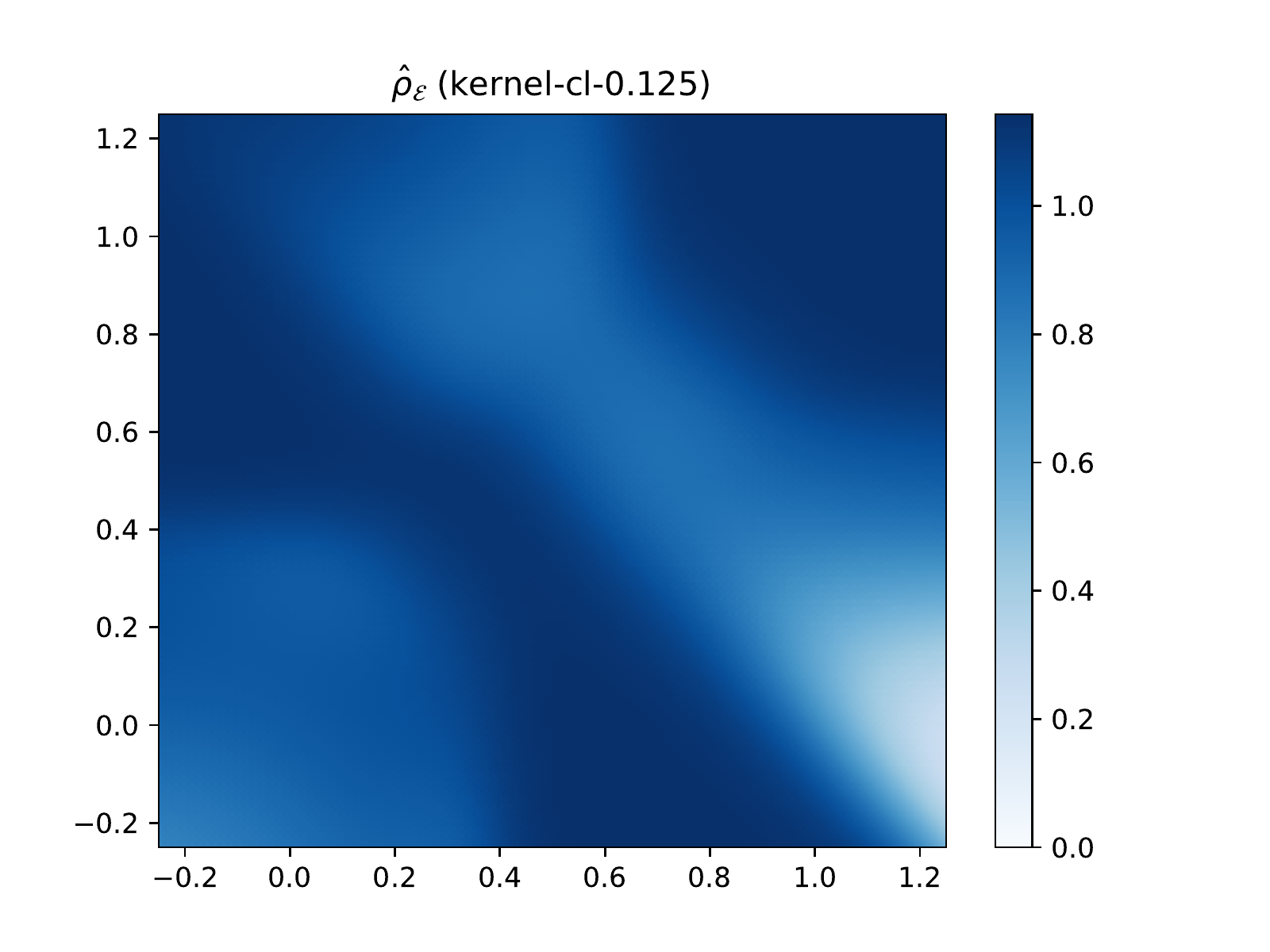}\\
		\includegraphics[trim=25 20 50 37, clip, width=\textwidth]{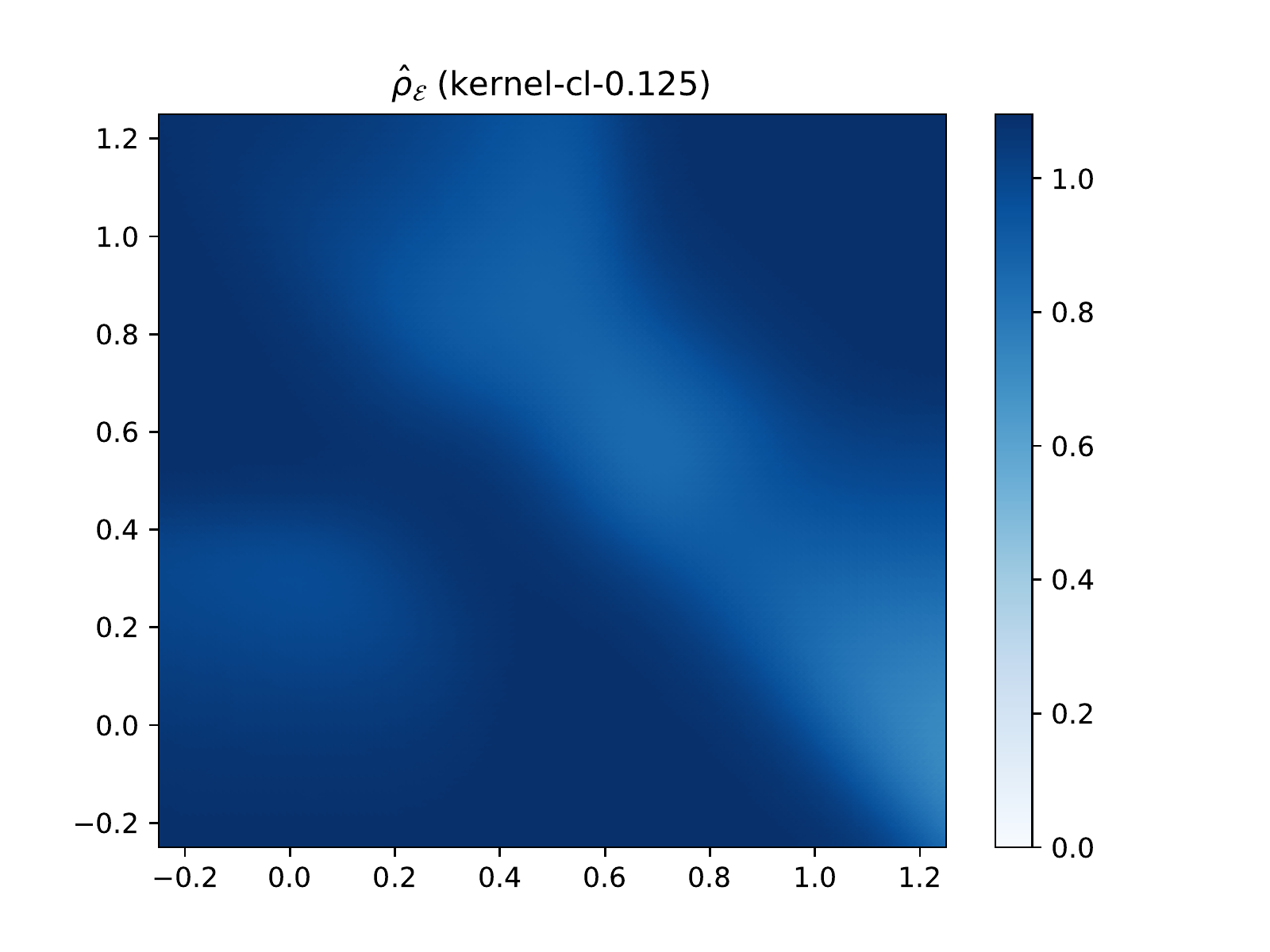}\\
		\caption{KBC ($\sigma_{\mC}$=$0.125$)}
	\end{subfigure}
	\begin{subfigure}[t!]{0.19\textwidth}
		\centering 
		\includegraphics[trim=25 20 50 37, clip, width=\textwidth]{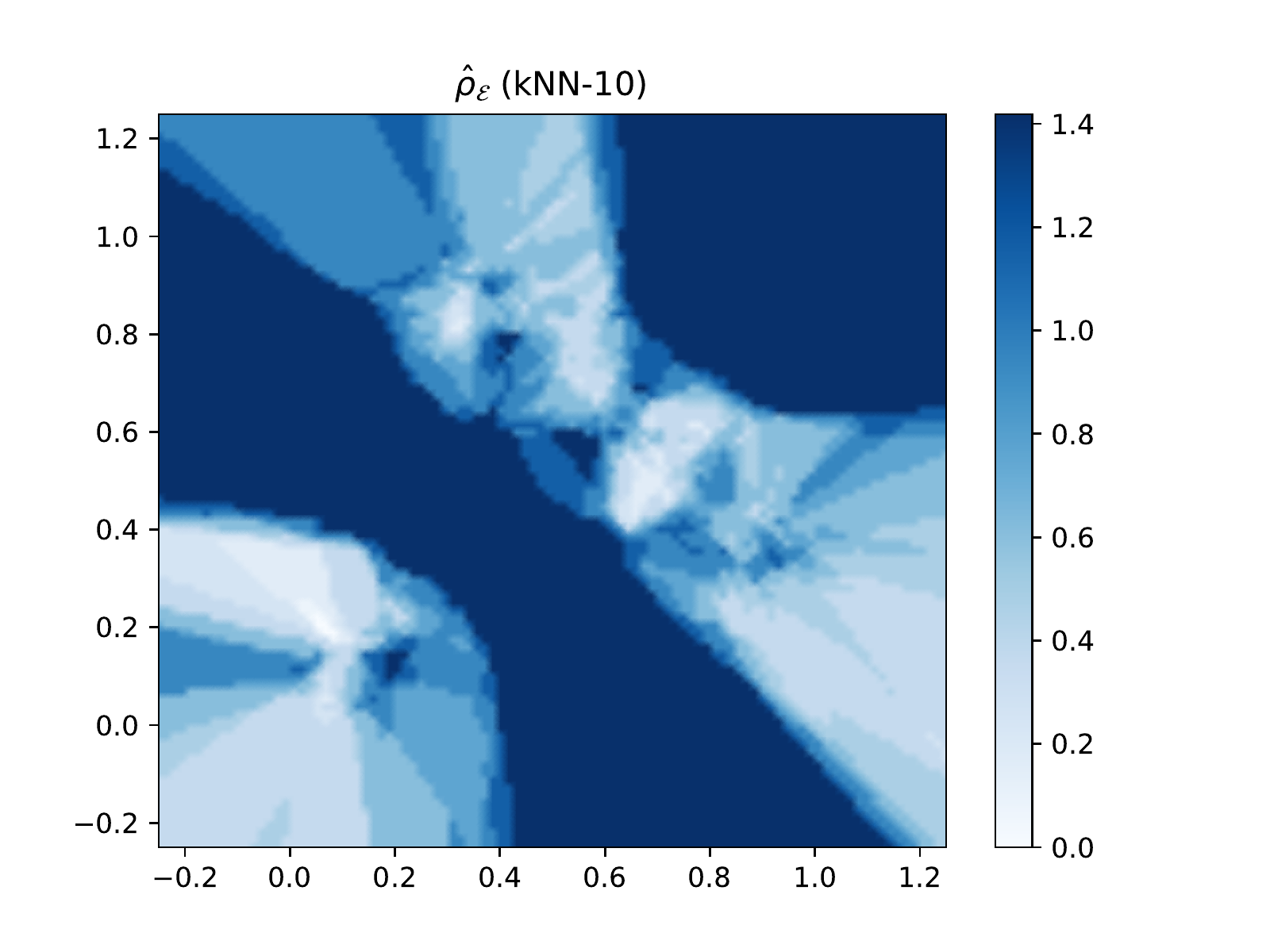}\\
		\includegraphics[trim=25 20 50 37, clip, width=\textwidth]{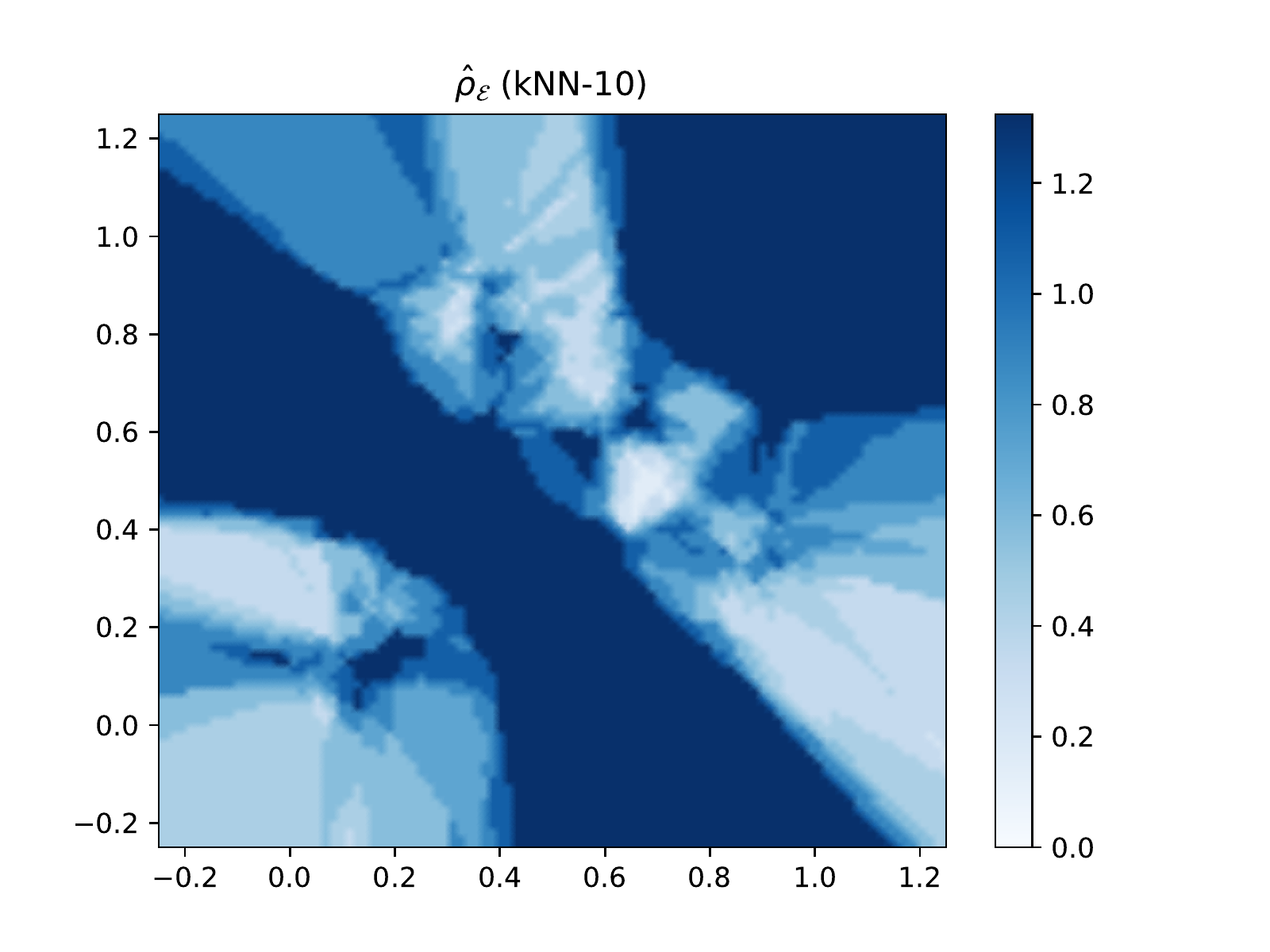}\\
		\includegraphics[trim=25 20 50 37, clip, width=\textwidth]{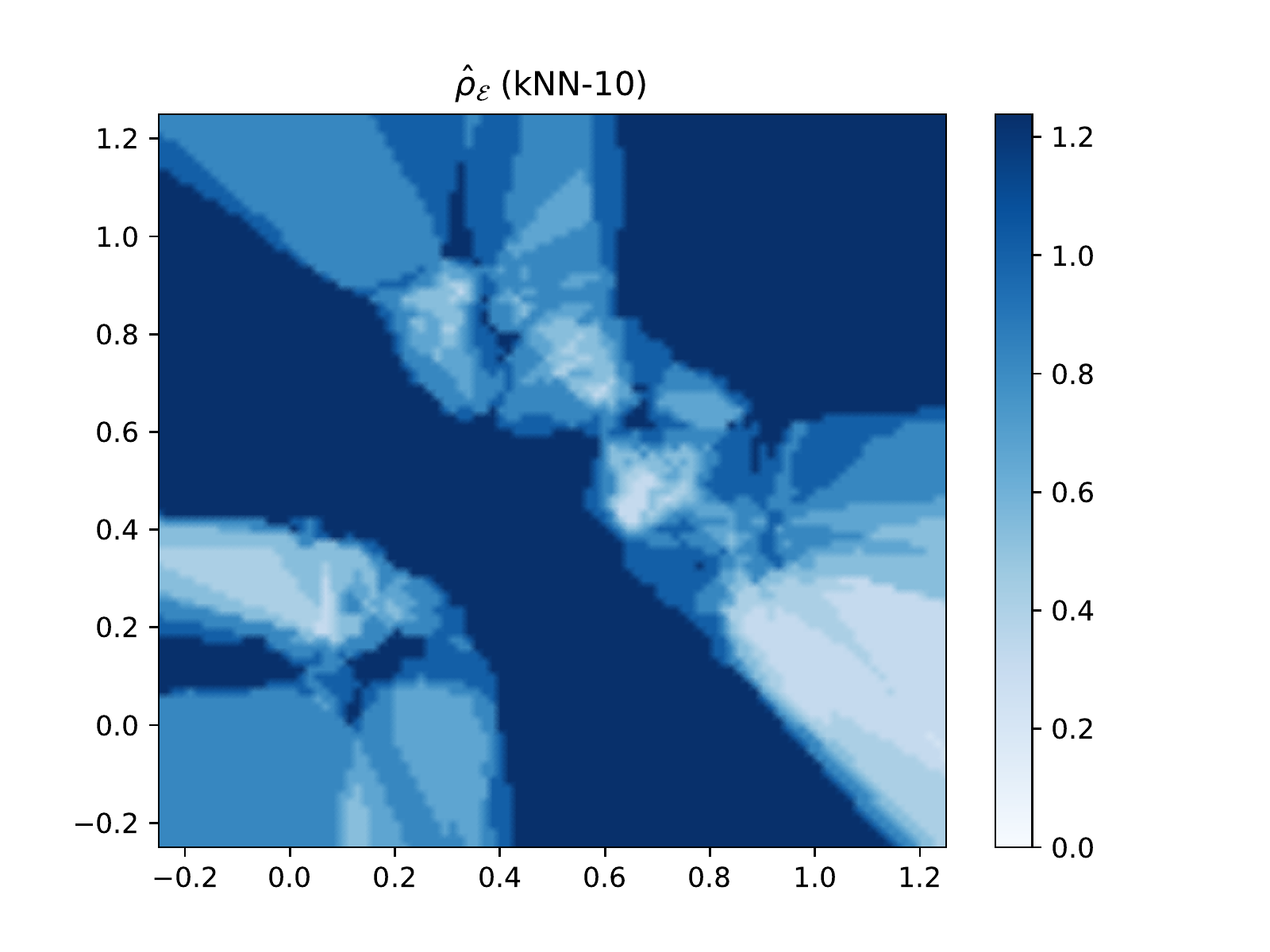}\\
		\includegraphics[trim=25 20 50 37, clip, width=\textwidth]{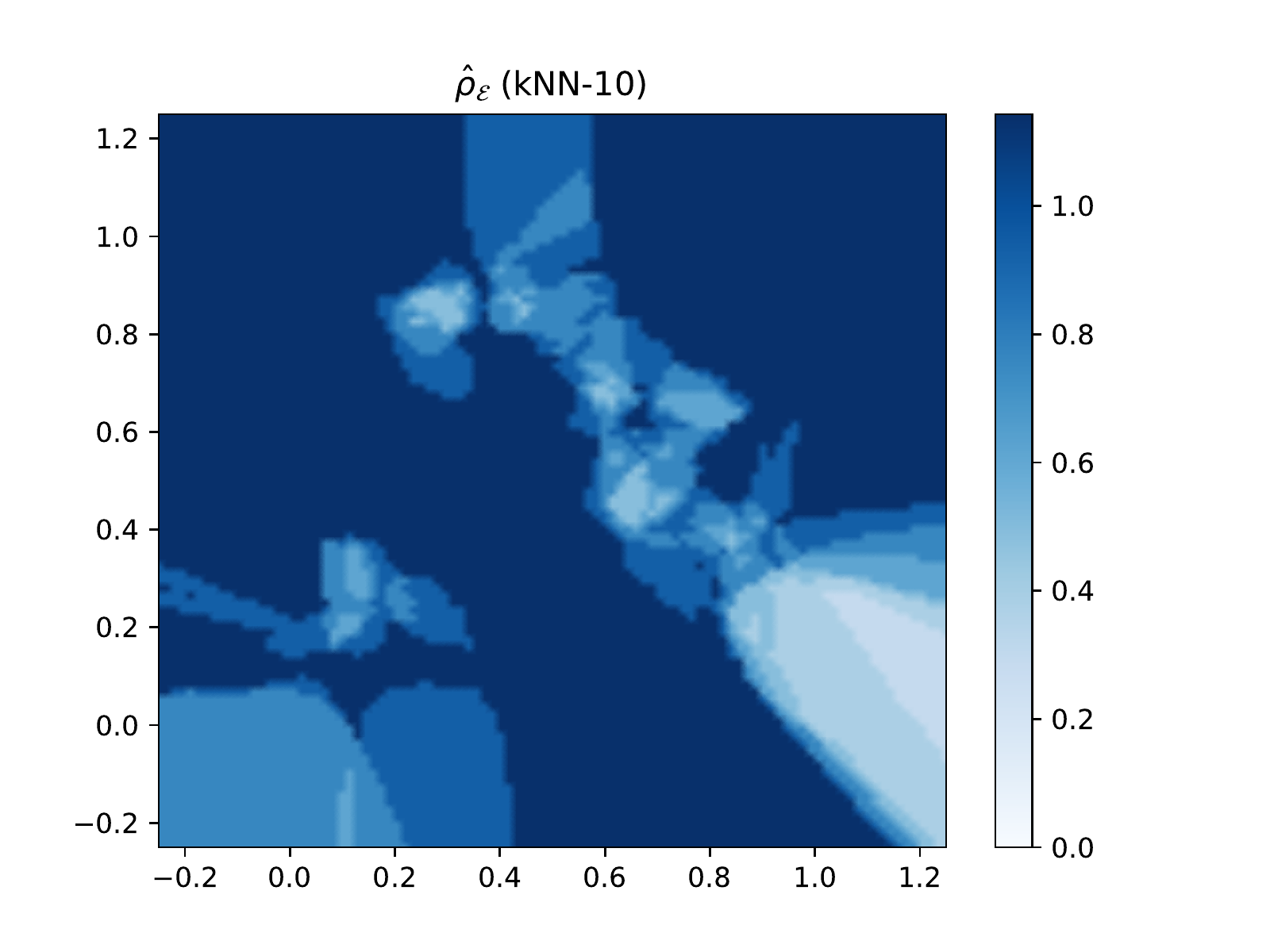}\\
		\includegraphics[trim=25 20 50 37, clip, width=\textwidth]{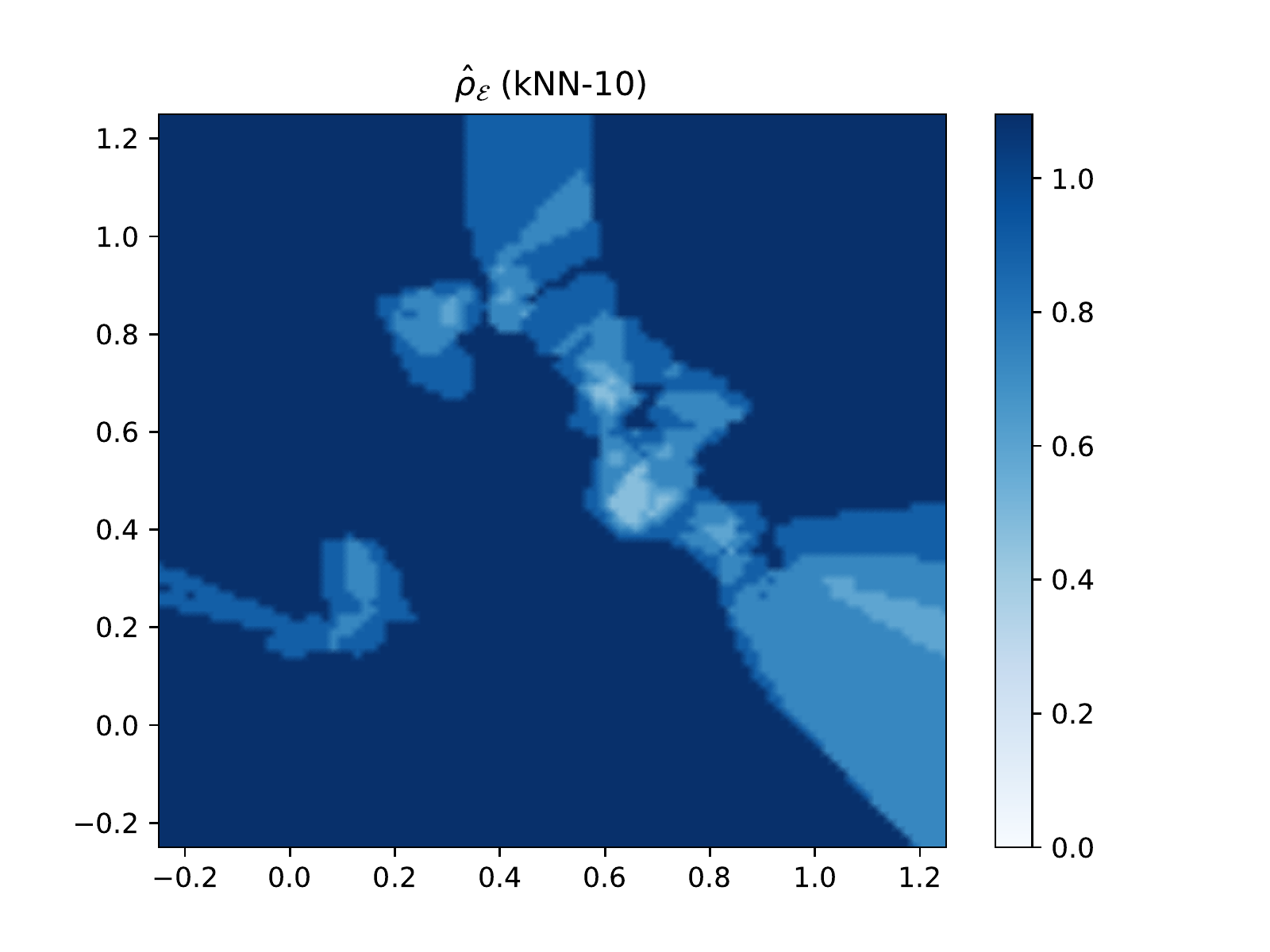}\\
		\caption{$k$-NN ($k$=$10$)}
	\end{subfigure}
	\begin{subfigure}[t!]{0.19\textwidth}
		\centering 
		\includegraphics[trim=25 20 50 37, clip, width=\textwidth]{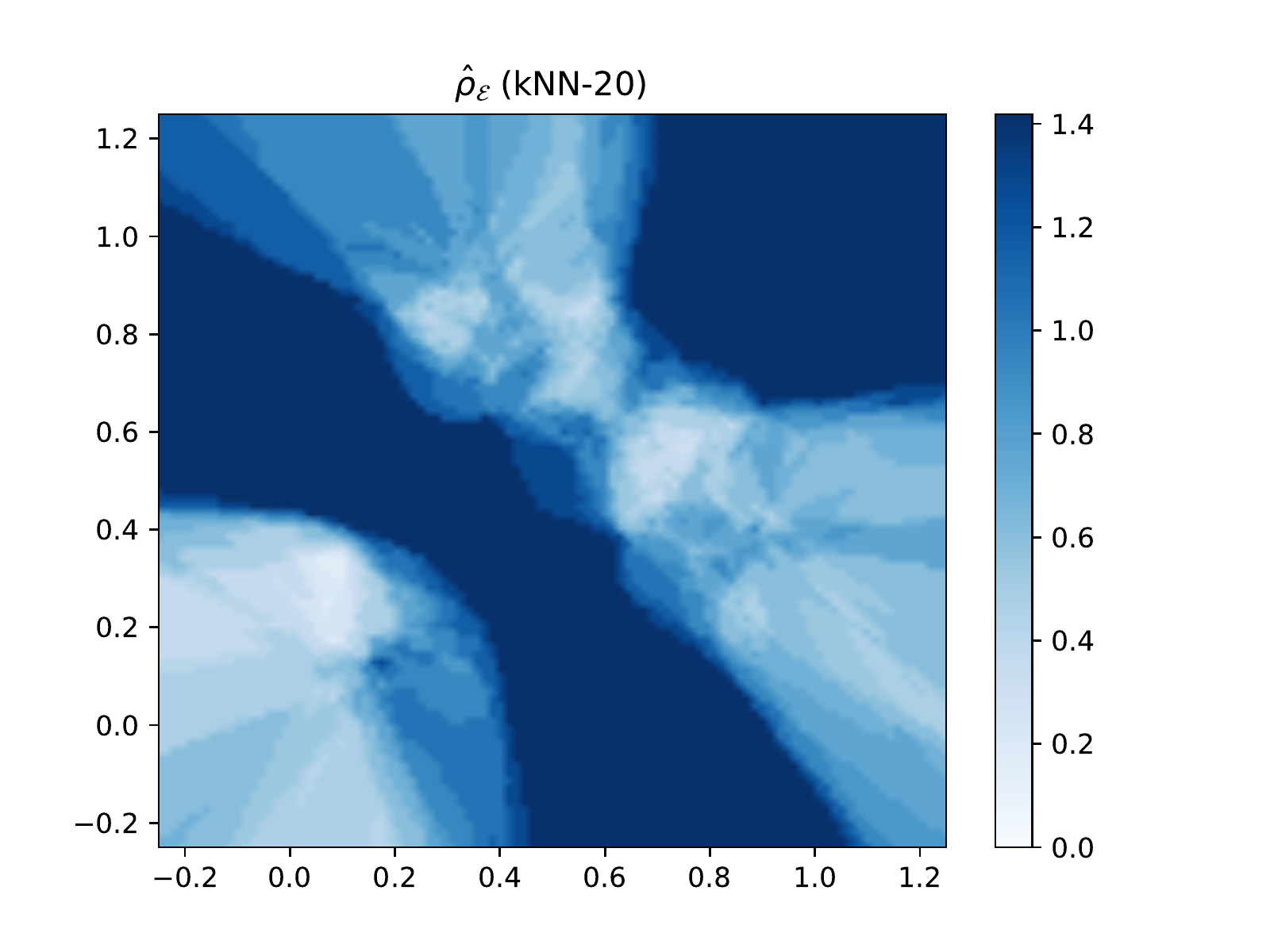}\\
		\includegraphics[trim=25 20 50 37, clip, width=\textwidth]{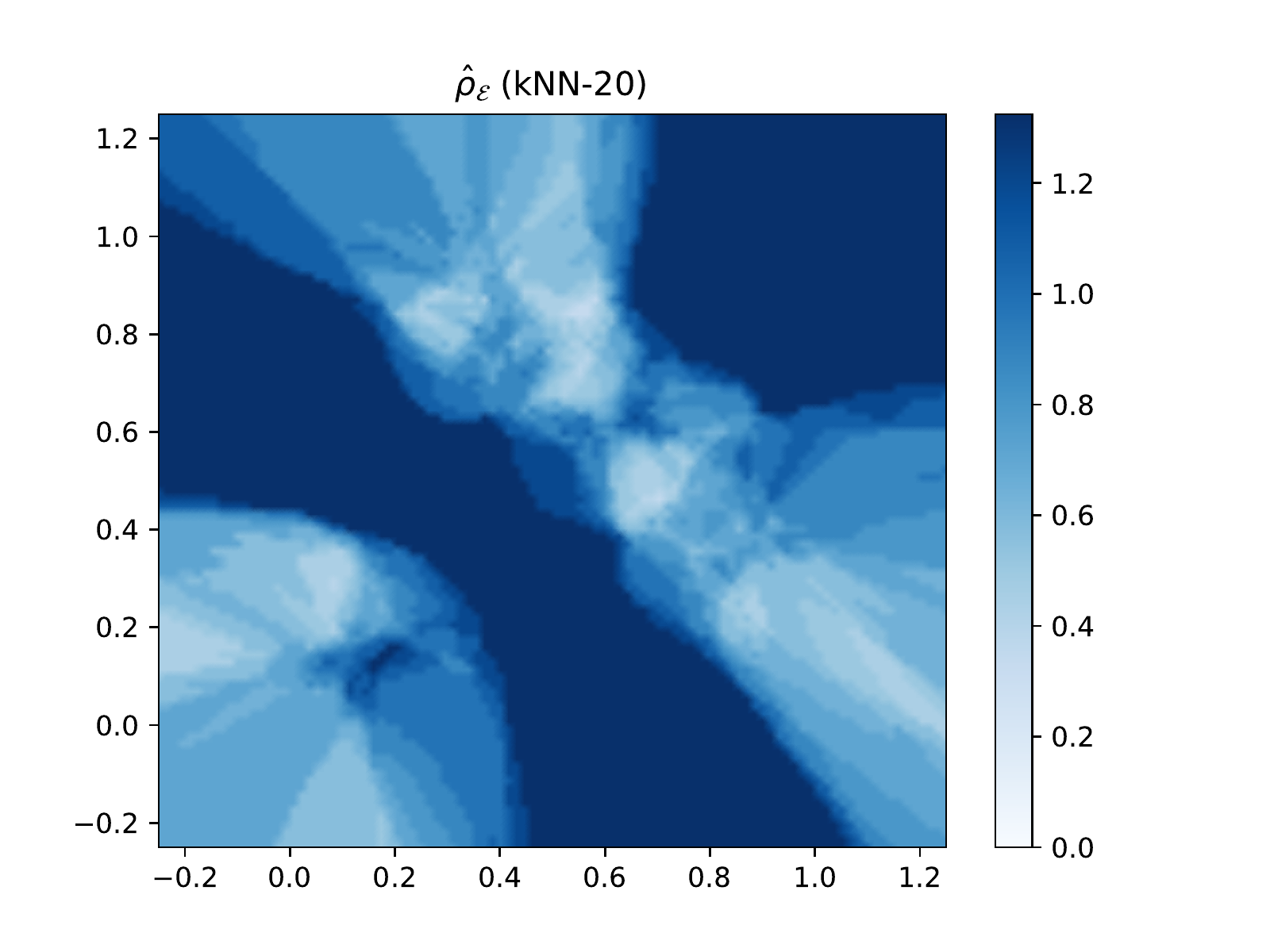}\\
		\includegraphics[trim=25 20 50 37, clip, width=\textwidth]{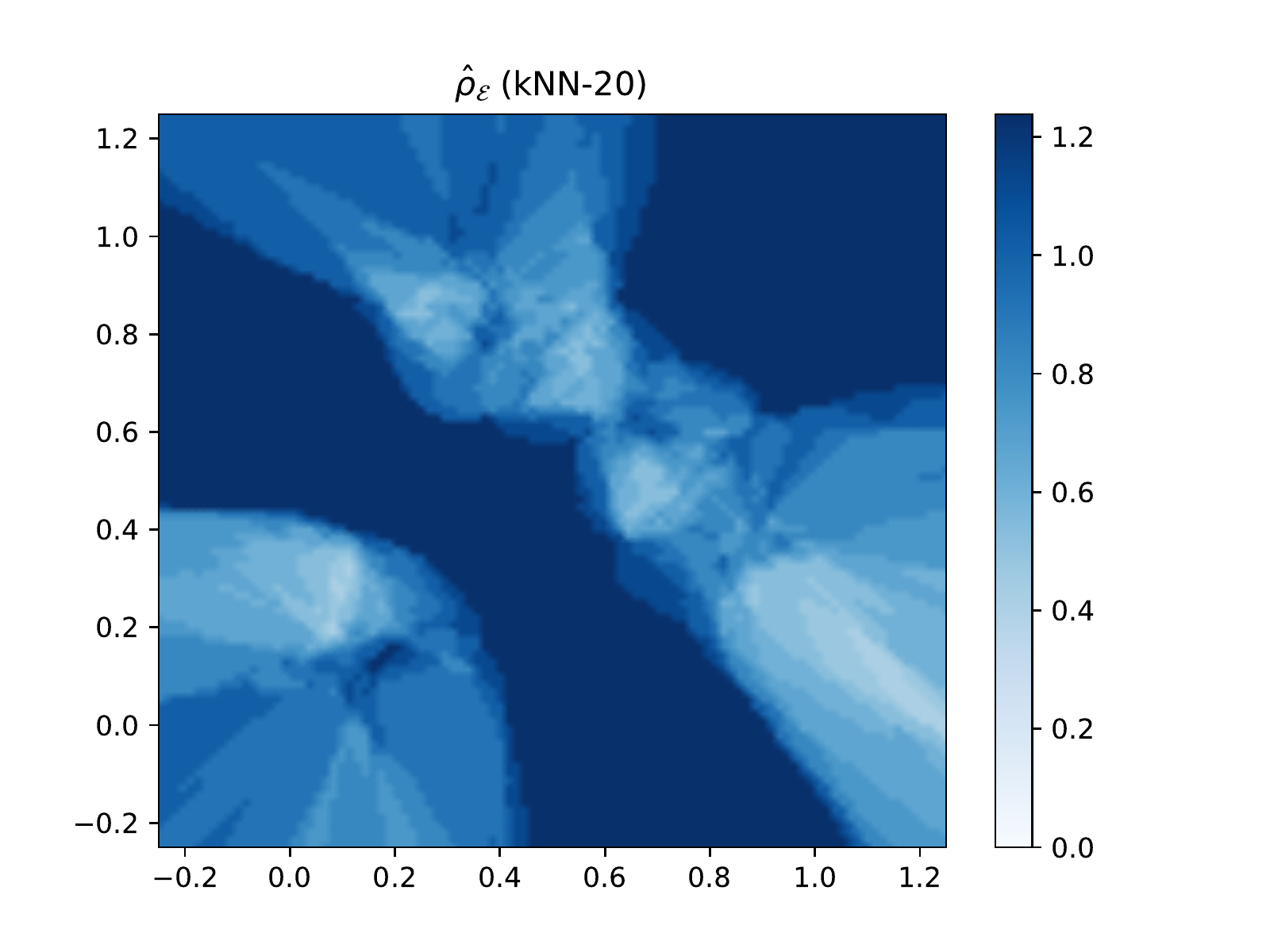}\\
		\includegraphics[trim=25 20 50 37, clip, width=\textwidth]{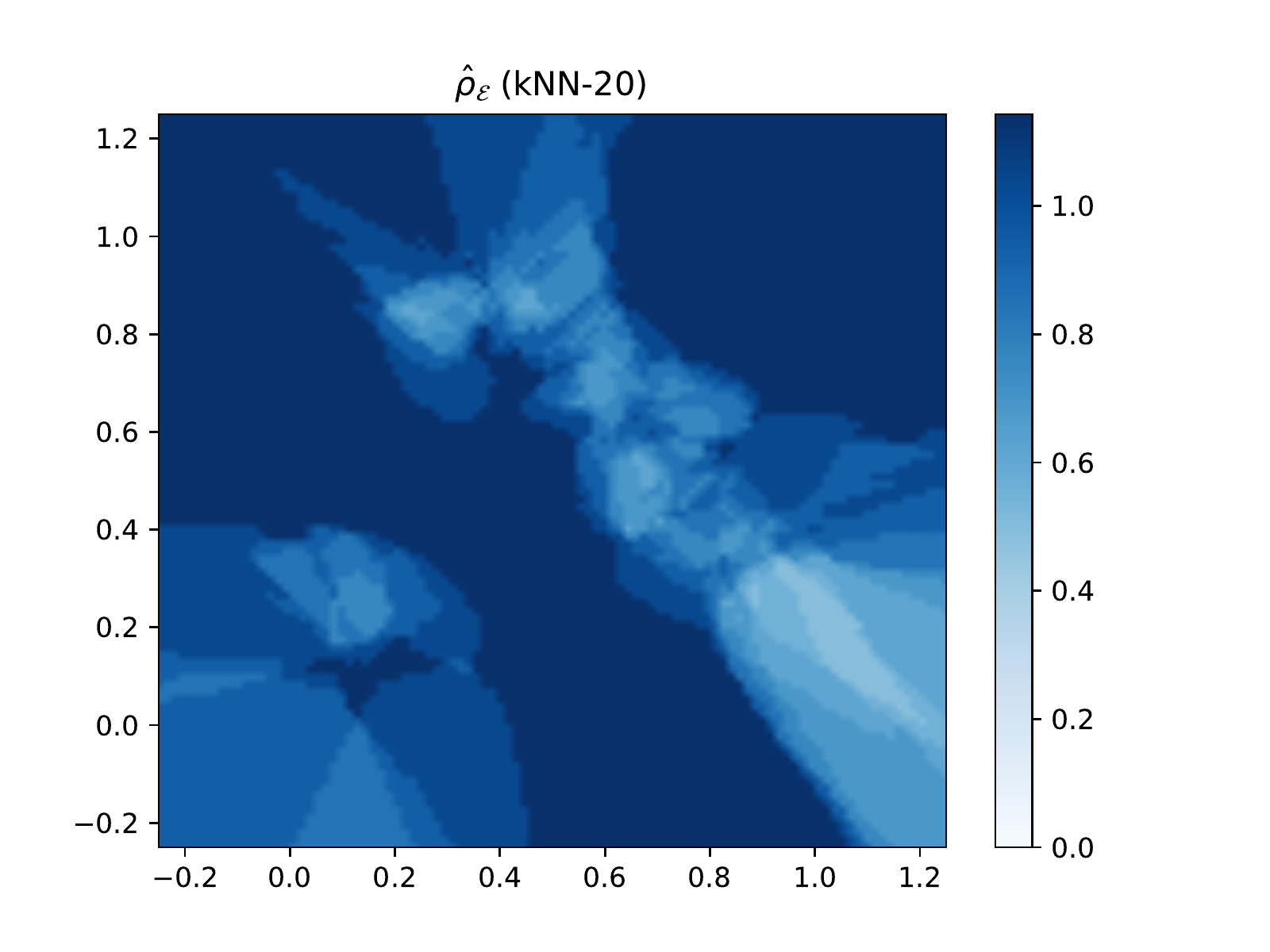}\\
		\includegraphics[trim=25 20 50 37, clip, width=\textwidth]{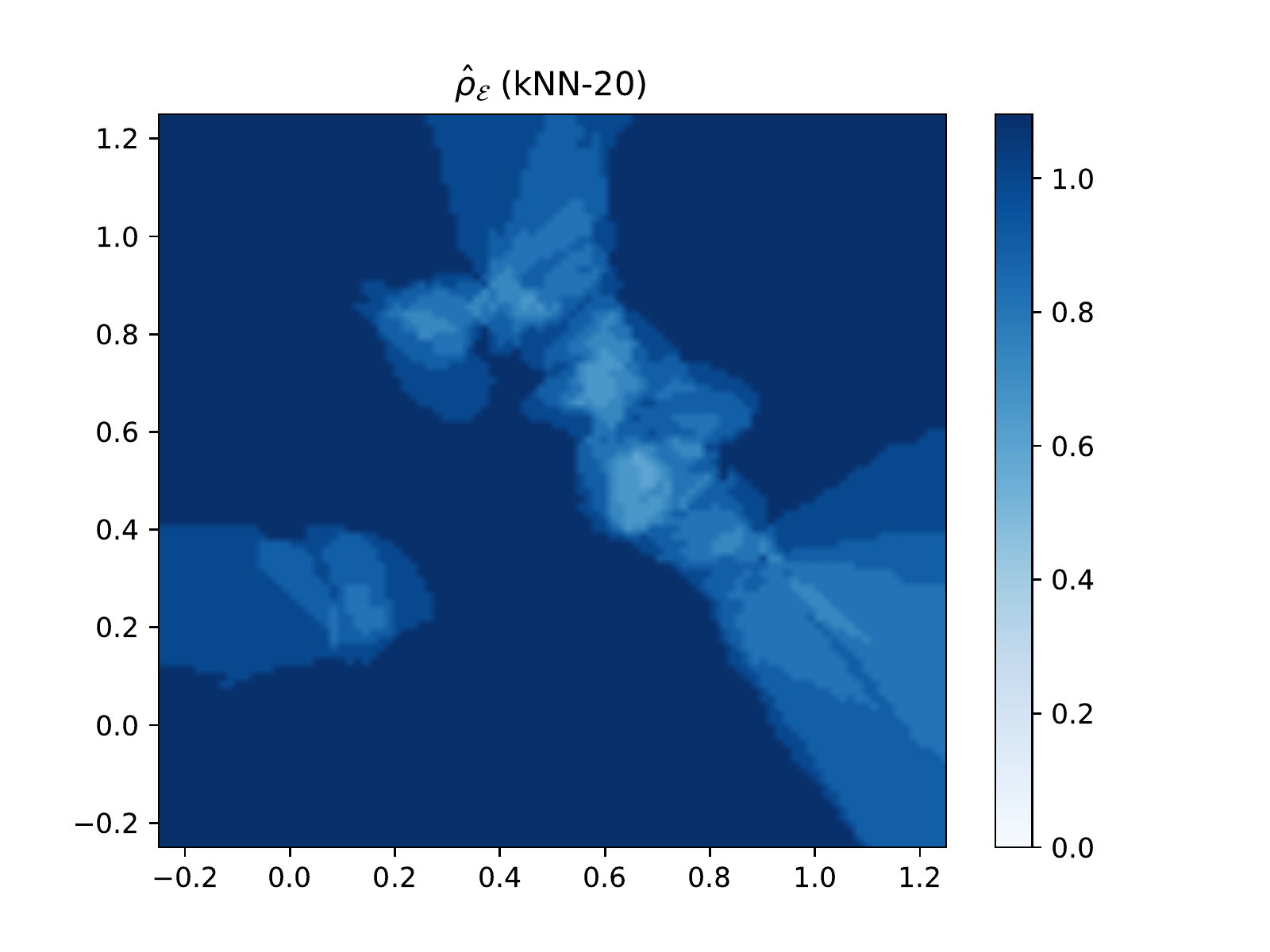}\\
		\caption{$k$-NN ($k$=$20$)}
	\end{subfigure}
	
	\vspace{-0.3em}
	\caption{Visualization of ratio $\hat{\rho}$ in (a) and $\hat{\rho}_{\mE}$ in (b)-(e) for different classifier-based DREs.}
	\label{fig: 2d Q1 DRE CKB-8 appendix}
	\vspace{-0.3em}
\end{figure}

We visualize $\hat{\rho}$ and $\hat{\rho}_{\mE}$ in Fig. \ref{fig: 2d Q1 DRE CKB-8 appendix} (extension of Fig. \ref{fig: 2d Q1 DRE} for CKB-8). These figures give qualitative answers to question 1 (DRE Approximations). 

\begin{figure}[!h]
\vspace{-0.3em}
  	\begin{subfigure}[t!]{0.5\textwidth}
	\centering 
	\includegraphics[trim=30 0 0 5, clip, width=0.95\textwidth]{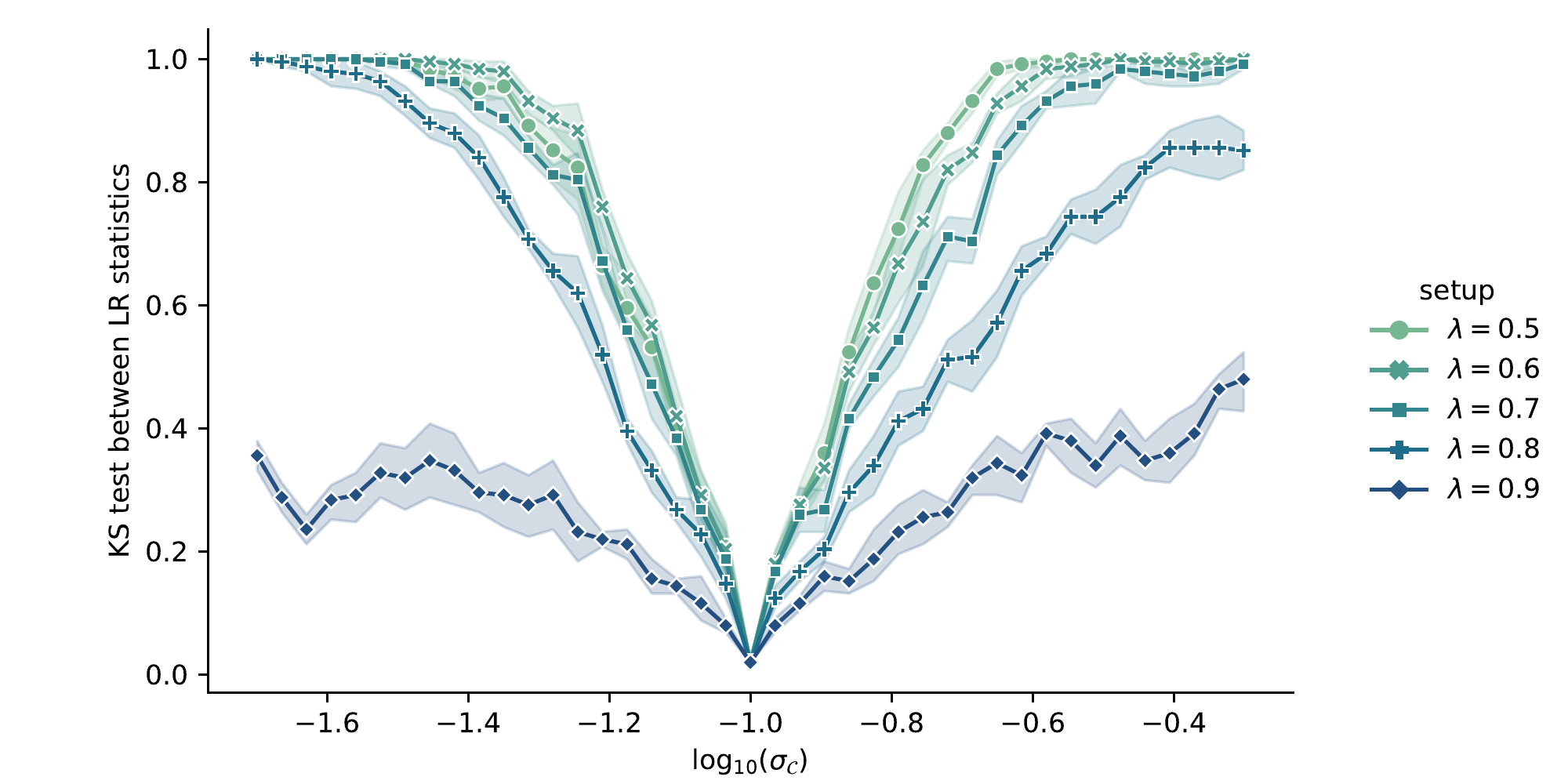}
	\caption{$\mathrm{LR}$ statistics}
	\end{subfigure}
	\begin{subfigure}[t!]{0.5\textwidth}
	\centering 
	\includegraphics[trim=30 0 0 5, clip, width=0.95\textwidth]{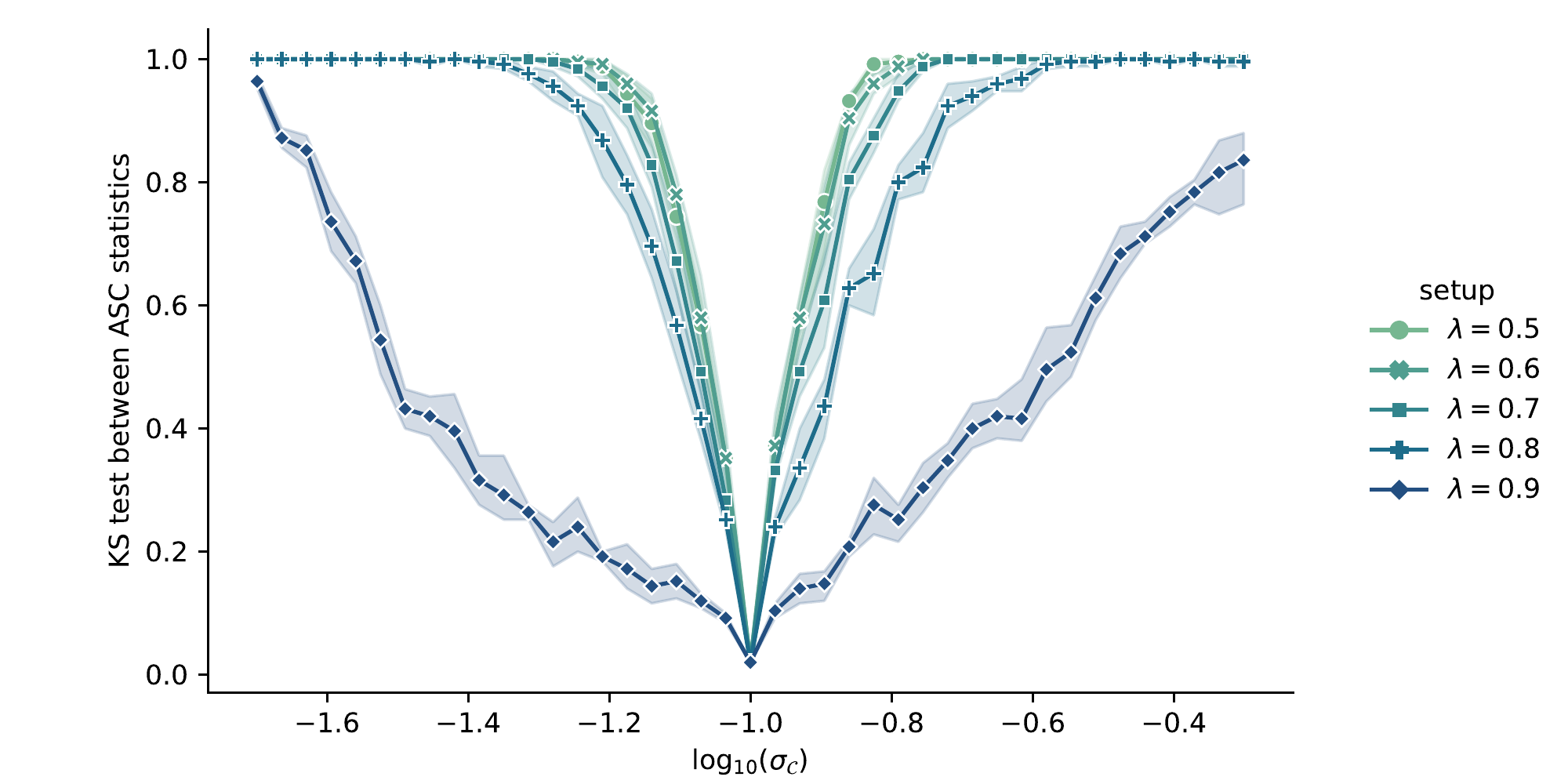}
	\caption{$\mathrm{ASC}$ statistics with $\phi(t)=\log(t)$}
	\end{subfigure}\\
	\begin{subfigure}[t!]{0.5\textwidth}
	\centering 
	\includegraphics[trim=30 0 0 5, clip, width=0.95\textwidth]{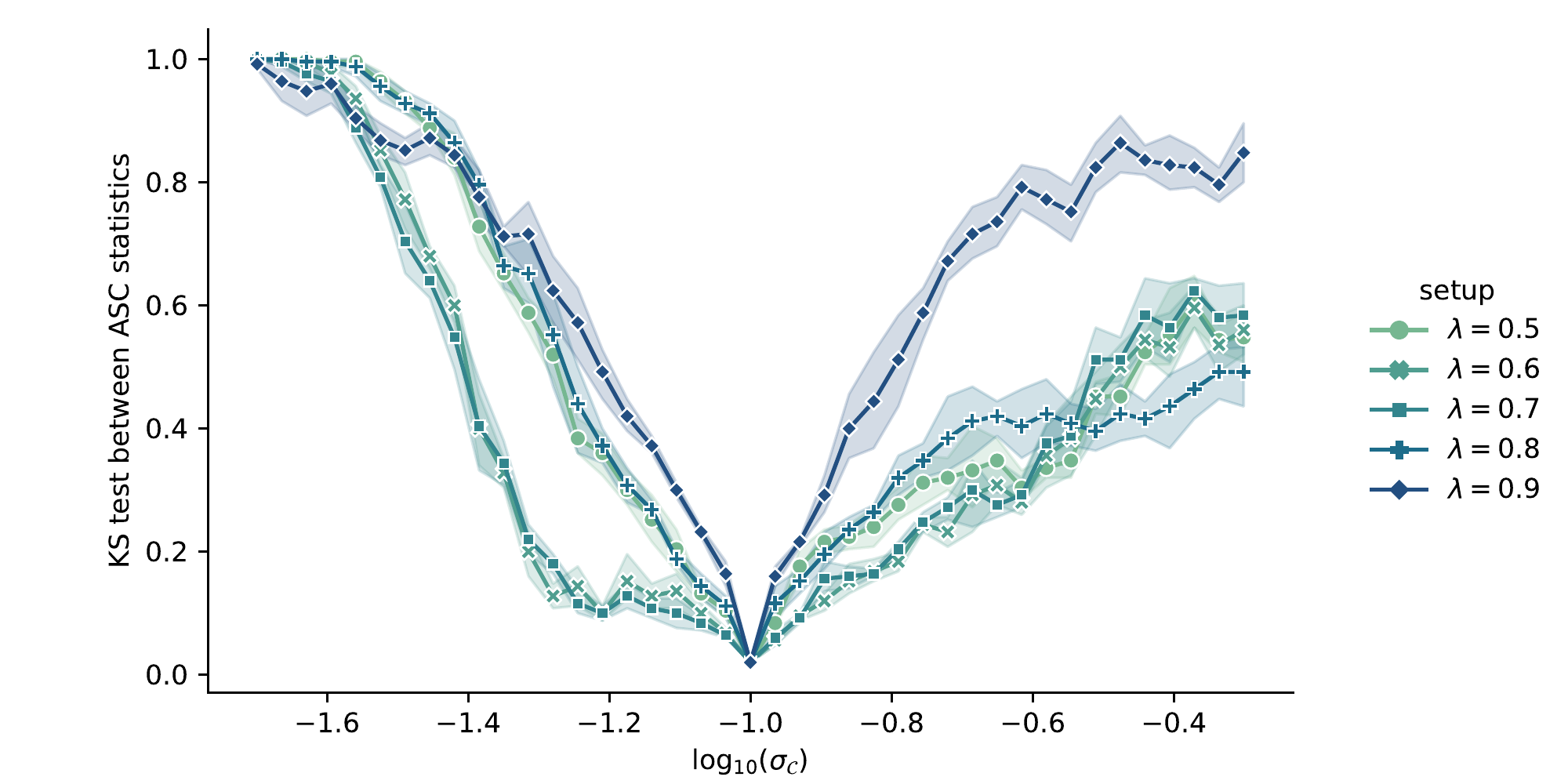}
	\caption{$\mathrm{ASC}$ statistics with $\phi(t)=t\log(t)$ (KL)}
	\end{subfigure}
	\begin{subfigure}[t!]{0.5\textwidth}
	\centering 
	\includegraphics[trim=30 0 0 5, clip, width=0.95\textwidth]{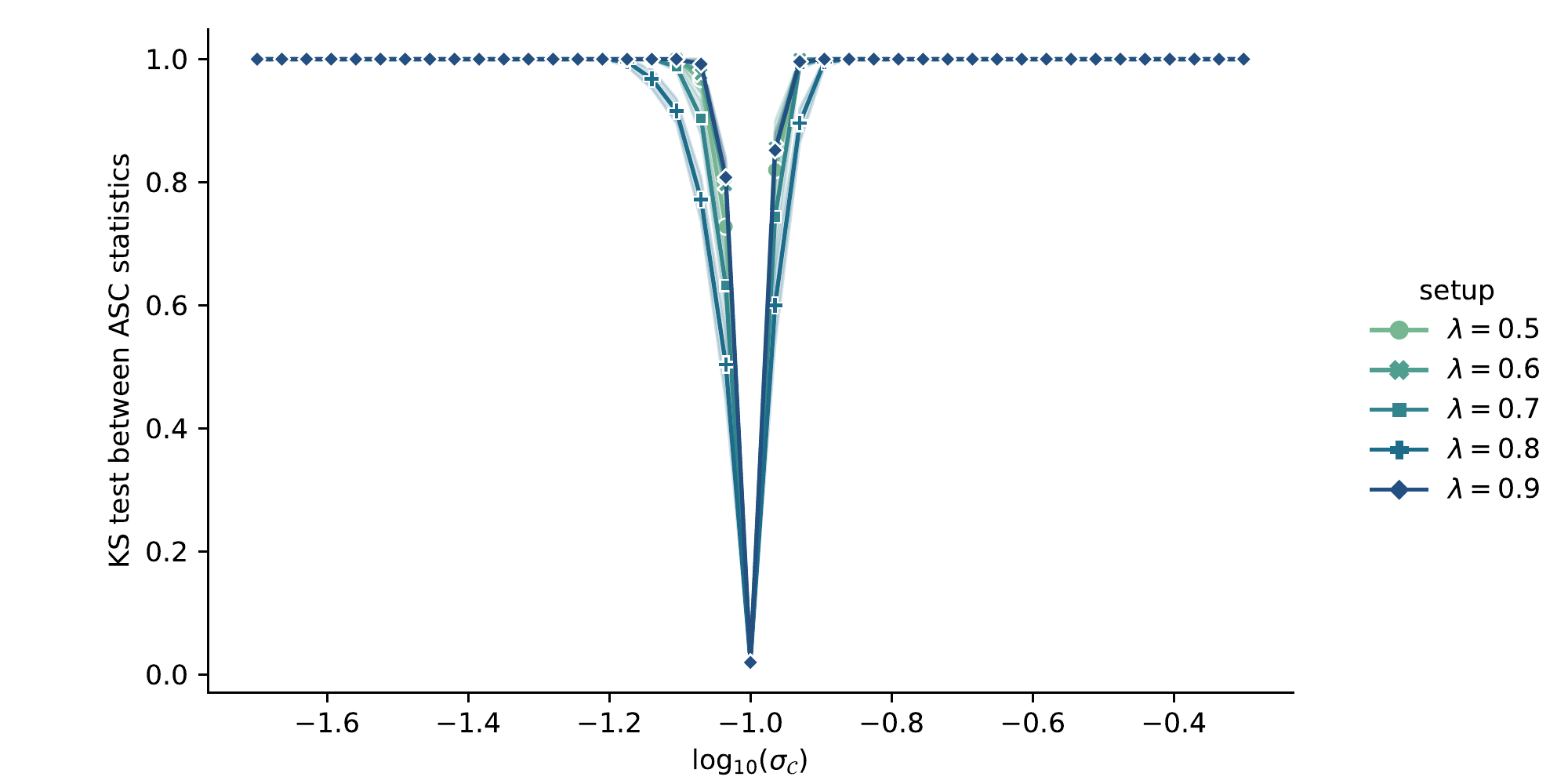}
	\caption{$\mathrm{ASC}$ statistics with $\phi(t)=(\sqrt{t}-1)^2$ (Hellinger)}
	\end{subfigure}
	
	\vspace{-0.3em}
	\caption{KS tests between distributions of statistics for KBC with different $\sigma_{\mC}$. (a) $\mathrm{LR}(Y_{H_0},\hat{\rho})$ vs $\mathrm{LR}(Y_{H_0},\hat{\rho}_{\mE})$ with different $\lambda$. (b)-(d) $\hat{\mathrm{ASC}}_{\phi}(\hat{Y},Y_{H_0},\hat{\rho})$ vs $\hat{\mathrm{ASC}}_{\phi}(\hat{Y},Y_{H_0},\hat{\rho}_{\mE})$ for different $\phi$. Smaller values indicate the two compared distributions are closer. }
	\vspace{-0.3em}
	\label{fig: 2d Q1 KS CKB-8 appendix}
\end{figure}

\newpage
We visualize KS test results for KBC with different bandwidth $\sigma_{\mC}$ in Fig. \ref{fig: 2d Q1 KS CKB-8 appendix} (extension of Fig. \ref{fig: 2d Q1 KS} for CKB-8). When $\sigma_{\mC}\approx\sigma_{\mA}=0.1$, the KS values are small, indicating KBC with these $\sigma_{\mC}$ can lead to classifier-based DRE $\hat{\rho}_{\mE}$ that is close to $\hat{\rho}$.  Comparing to MoG-8, the conclusion for CKB-8 is similar, but estimation is harder. One exception is when $\lambda=0.9$, the estimation is very accurate under LR or ASC with $\phi(t)=\log(t)$, but not as accurate as other $\lambda$ under ASC with $\phi(t)=t\log(t)$.

\newpage
\paragraph{Question 2 (Fast Deletion).}

\begin{figure}[!h]
\vspace{-0.3em}
	\begin{subfigure}[t!]{0.5\textwidth}
	\centering 
	\includegraphics[width=0.8\textwidth]{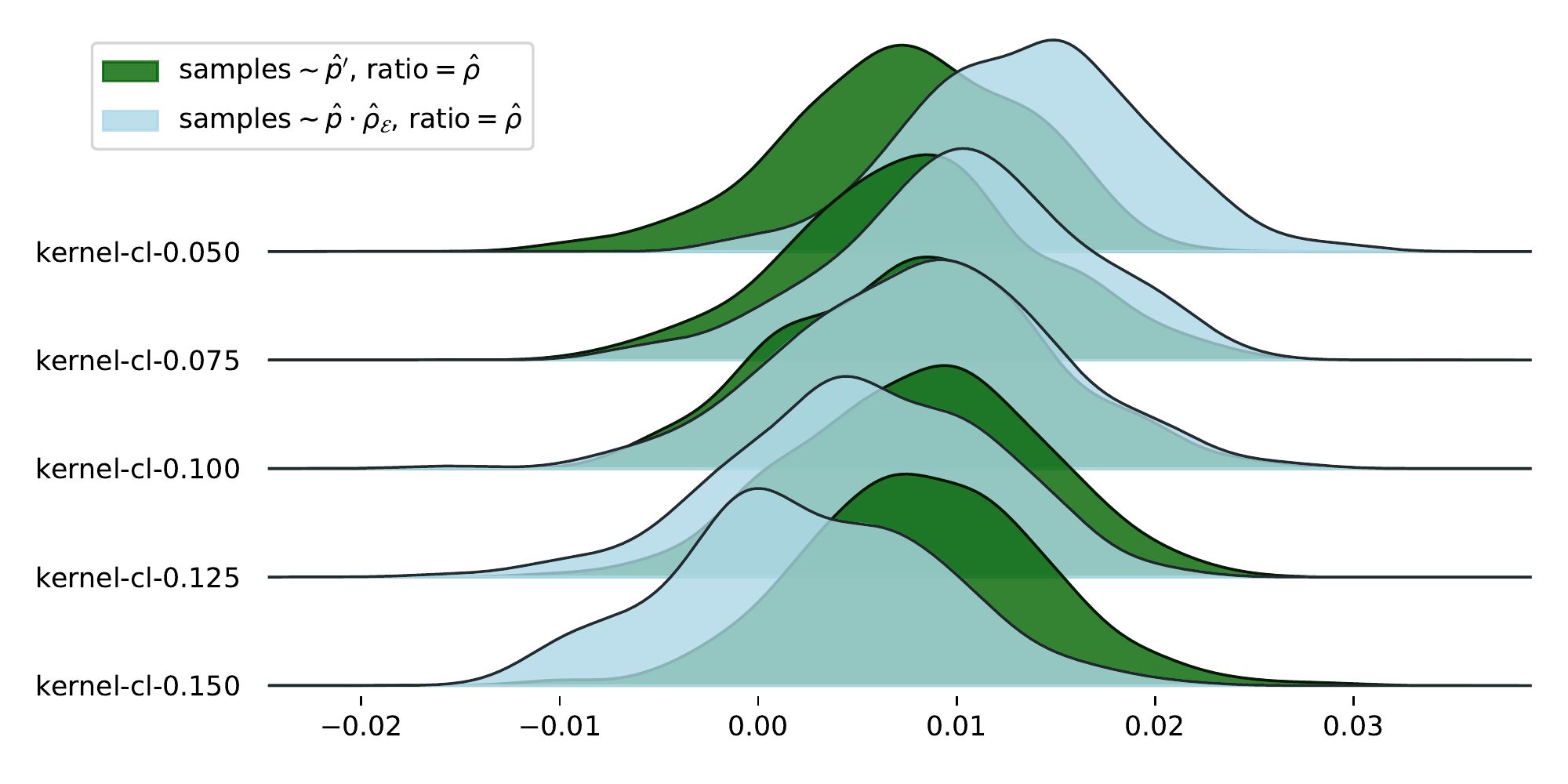}
	\caption{$\mathrm{LR}$ for KBC-based DRE ($\lambda=0.8$)}
	\end{subfigure}
	\begin{subfigure}[t!]{0.5\textwidth}
	\centering 
	\includegraphics[width=0.8\textwidth]{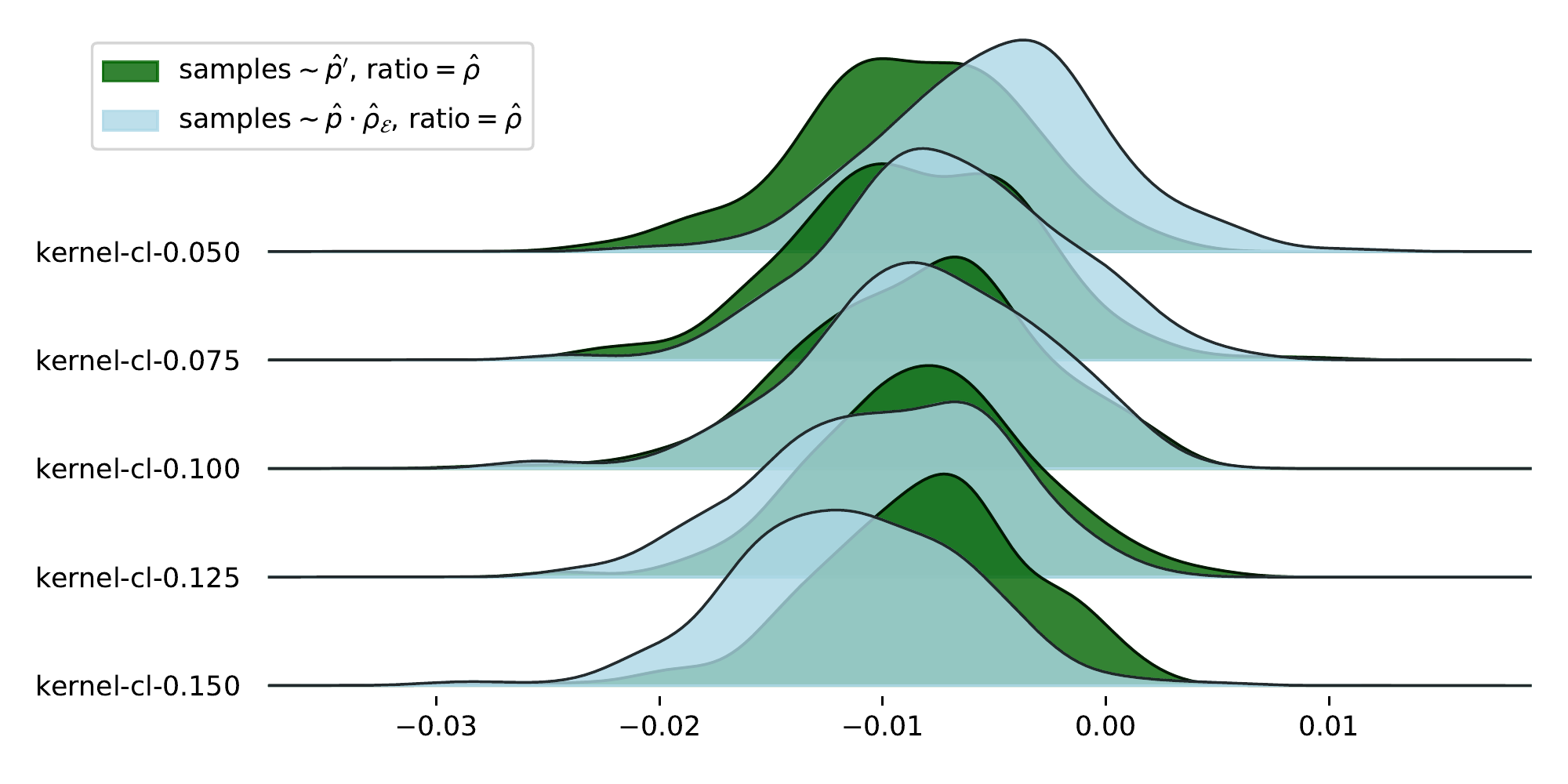}
	\caption{$\mathrm{ASC}$ for KBC-based DRE ($\phi(t)=\log(t)$)}
	\end{subfigure}\\
	\begin{subfigure}[t!]{0.5\textwidth}
	\centering 
	\includegraphics[width=0.8\textwidth]{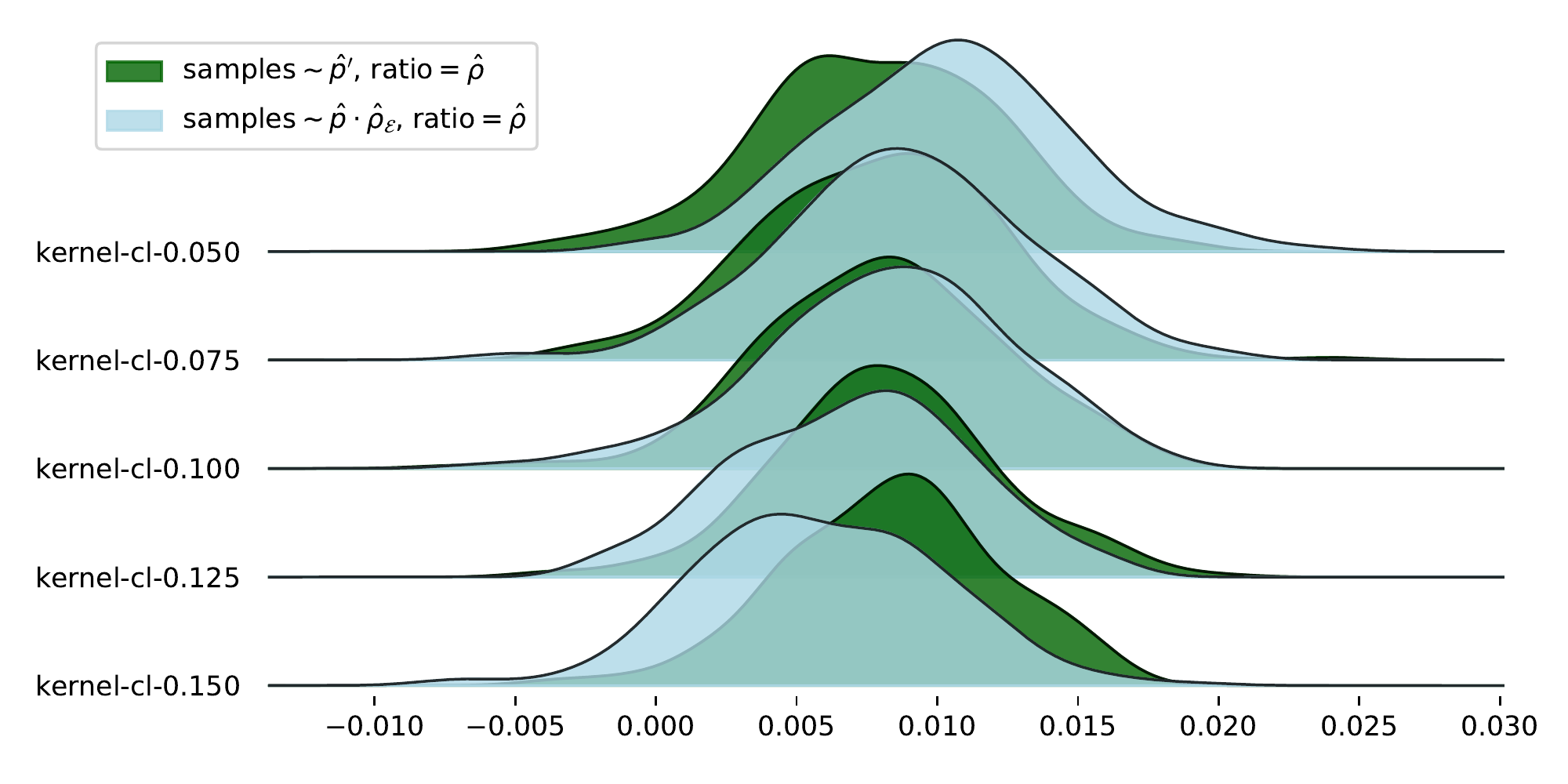}
	\caption{$\mathrm{ASC}$ for KBC-based DRE ($\phi(t)=t\log(t)$)}
	\end{subfigure}
	\begin{subfigure}[t!]{0.5\textwidth}
	\centering 
	\includegraphics[width=0.8\textwidth]{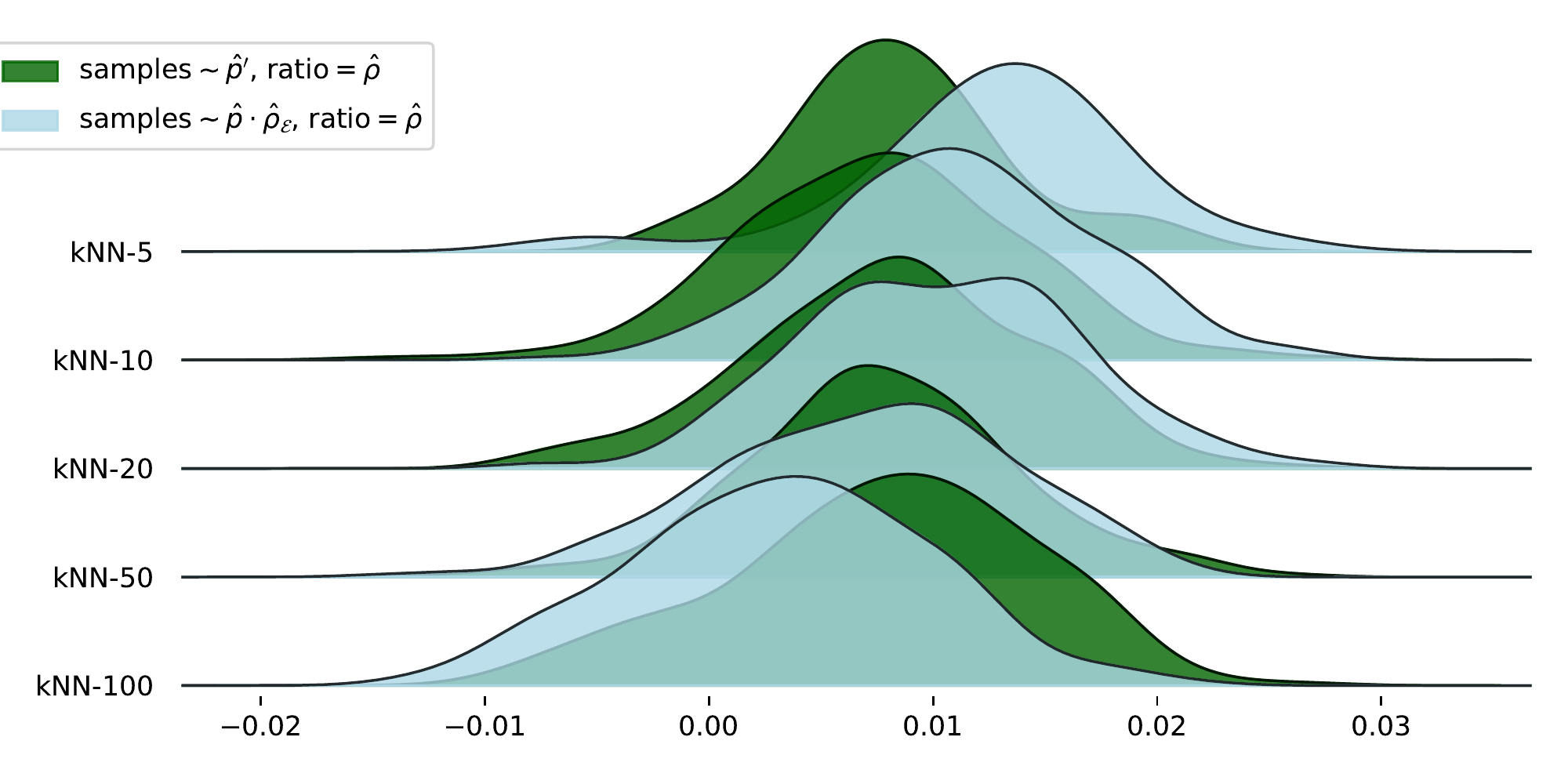}
	\caption{$\mathrm{LR}$ for $k$NN-based DRE ($\lambda=0.8$)}
	\end{subfigure}\\
	\begin{subfigure}[t!]{0.5\textwidth}
	\centering 
	\includegraphics[width=0.8\textwidth]{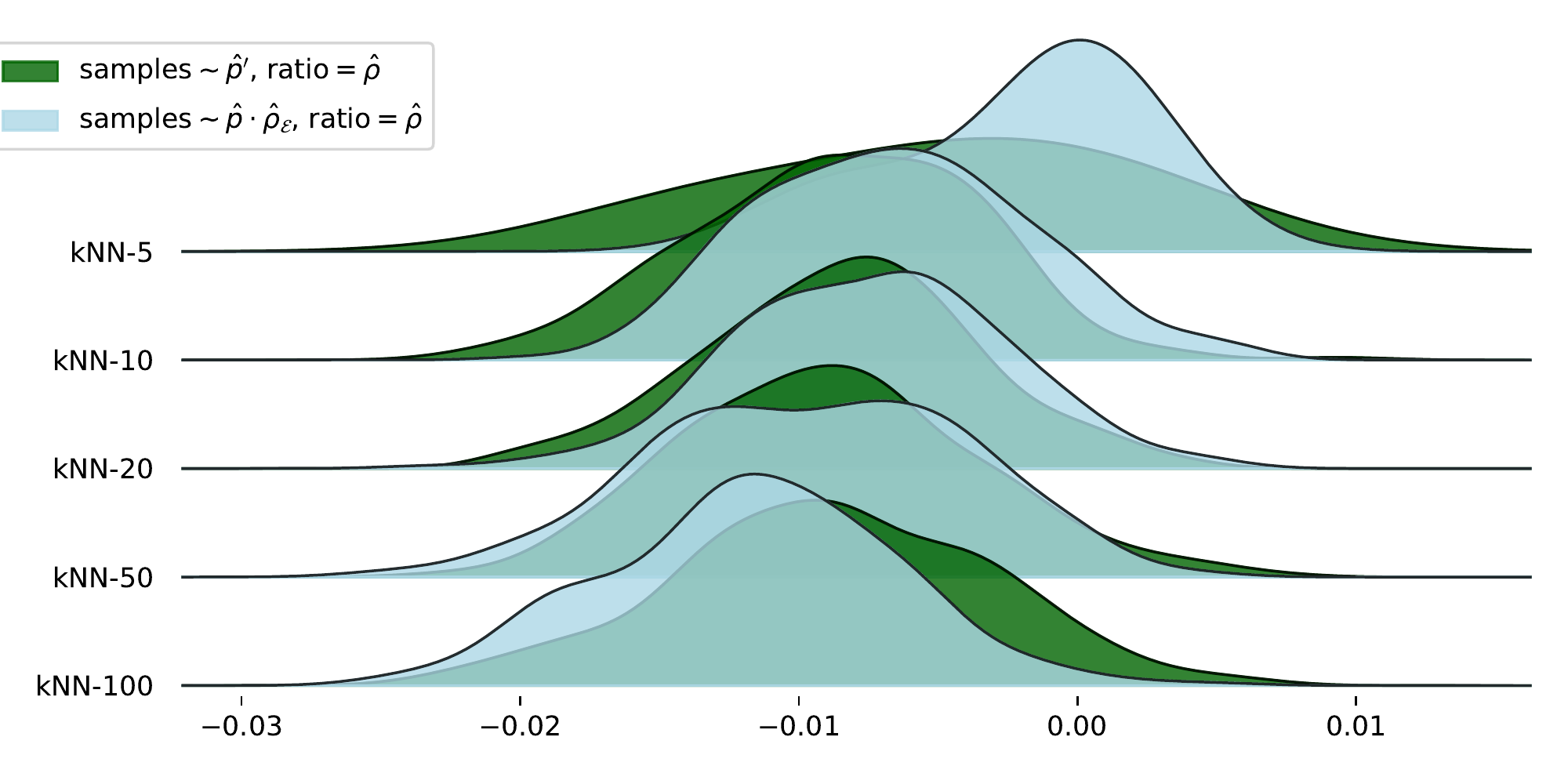}
	\caption{$\mathrm{ASC}$ for $k$NN-based DRE ($\phi(t)=\log(t)$)}
	\end{subfigure}
	\begin{subfigure}[t!]{0.5\textwidth}
	\centering 
	\includegraphics[width=0.8\textwidth]{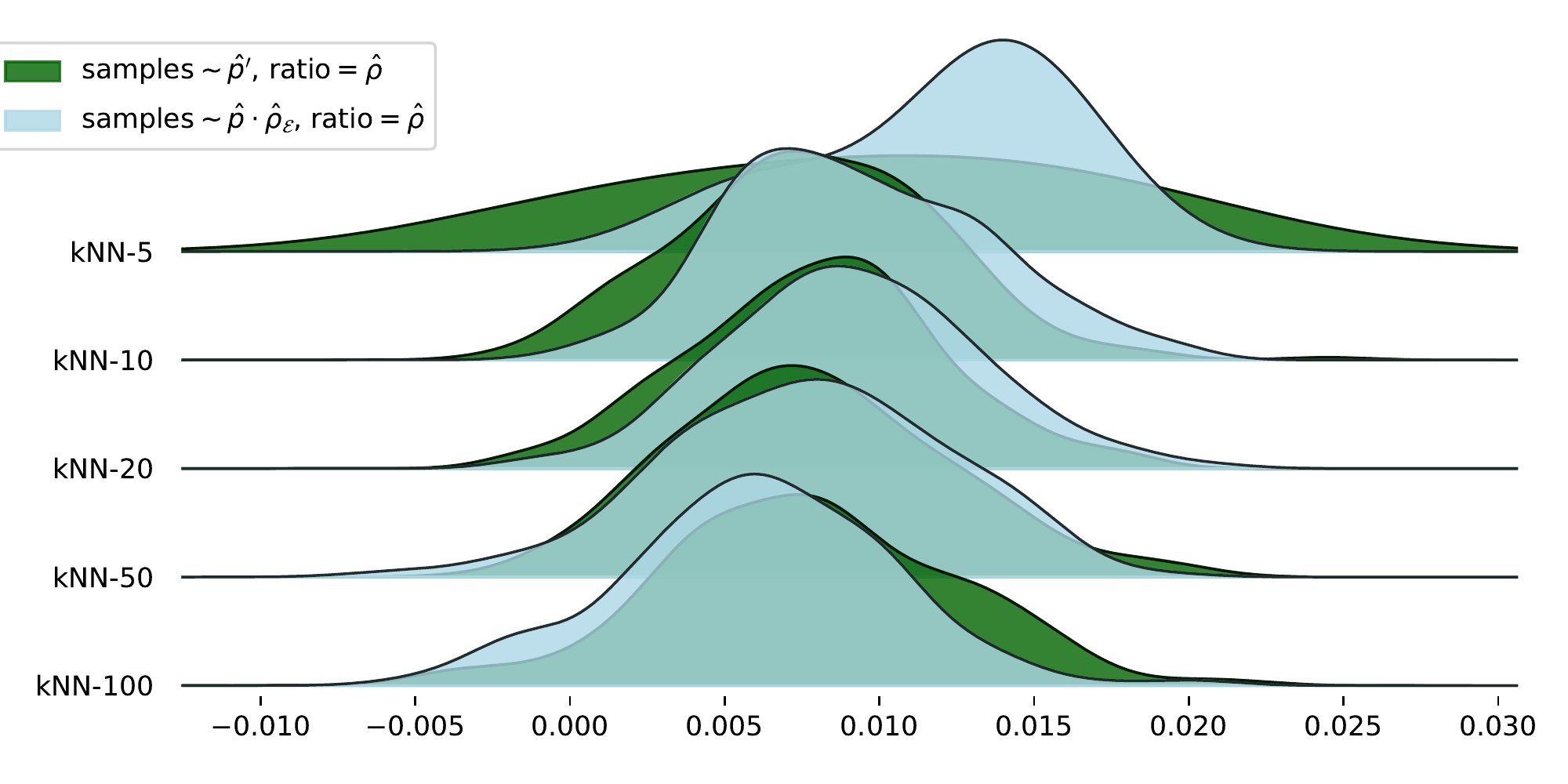}
	\caption{$\mathrm{ASC}$ for  $k$NN-based DRE ($\phi(t)=t\log(t)$)}
	\end{subfigure}
		
	\vspace{-0.3em}
	\caption{(a)-(c) KBC-based DRE. (d)-(f) $k$NN-based DRE. (a)\&(d) $\mathrm{LR}(Y_{H_1},\hat{\rho})$ vs $\mathrm{LR}(Y_{\mD},\hat{\rho})$. (b)-(c)\&(e)-(f) $\hat{\mathrm{ASC}}_{\phi}(\hat{Y},Y_{H_1},\hat{\rho})$ vs $\hat{\mathrm{ASC}}_{\phi}(\hat{Y},Y_{\mD},\hat{\rho})$.}
	\vspace{-0.3em}
	\label{fig: 2d Q2 joy CKB-8 appendix}
\end{figure}

We visualize distributions of LR and ASC statistics between $Y_{H_1}$ and $Y_{\mD}$ in Fig. \ref{fig: 2d Q2 joy CKB-8 appendix} (extension of Fig. \ref{fig: 2d Q2 joy} for CKB-8). The more overlapping between the distributions, the less distinguishable between the approximated and re-trained models. For both KBC and $k$NN, the distribution pairs are slightly more separated than MoG-8, indicating fast deletion is harder for CKB-8. For $k$NN a moderate $k$ (e.g. between 10 and 50) has better overlapping. 

\begin{figure}[!h]
\vspace{-0.3em}
  	\begin{subfigure}[t!]{0.5\textwidth}
	\centering 
	\includegraphics[trim=30 0 0 5, clip, width=0.95\textwidth]{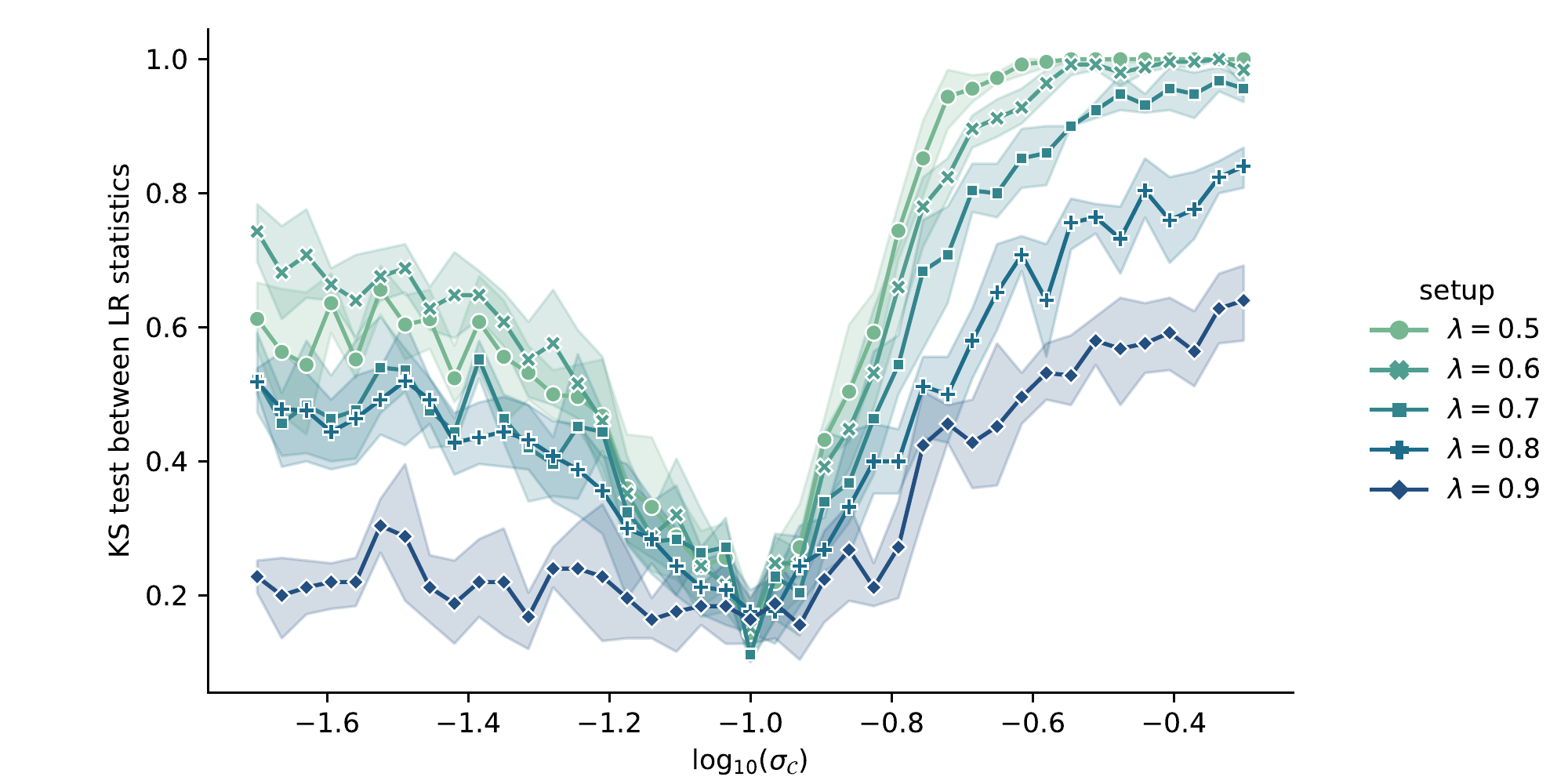}
	\caption{$\mathrm{LR}$ statistics}
	\end{subfigure}
	\begin{subfigure}[t!]{0.5\textwidth}
	\centering 
	\includegraphics[trim=30 0 0 5, clip, width=0.95\textwidth]{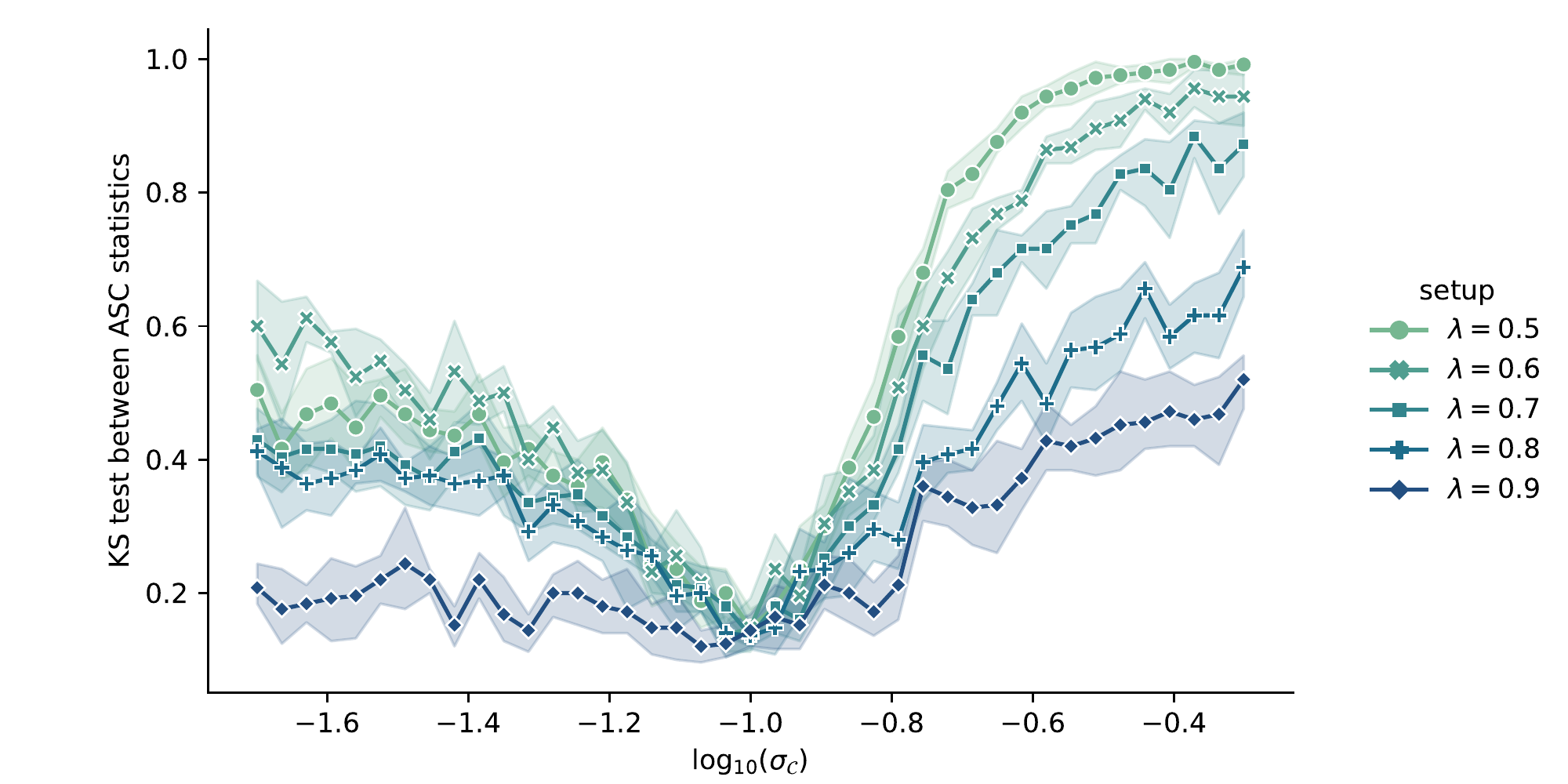}
	\caption{$\mathrm{ASC}$ statistics with $\phi(t)=\log(t)$}
	\end{subfigure}\\
	\begin{subfigure}[t!]{0.5\textwidth}
	\centering 
	\includegraphics[trim=30 0 0 5, clip, width=0.95\textwidth]{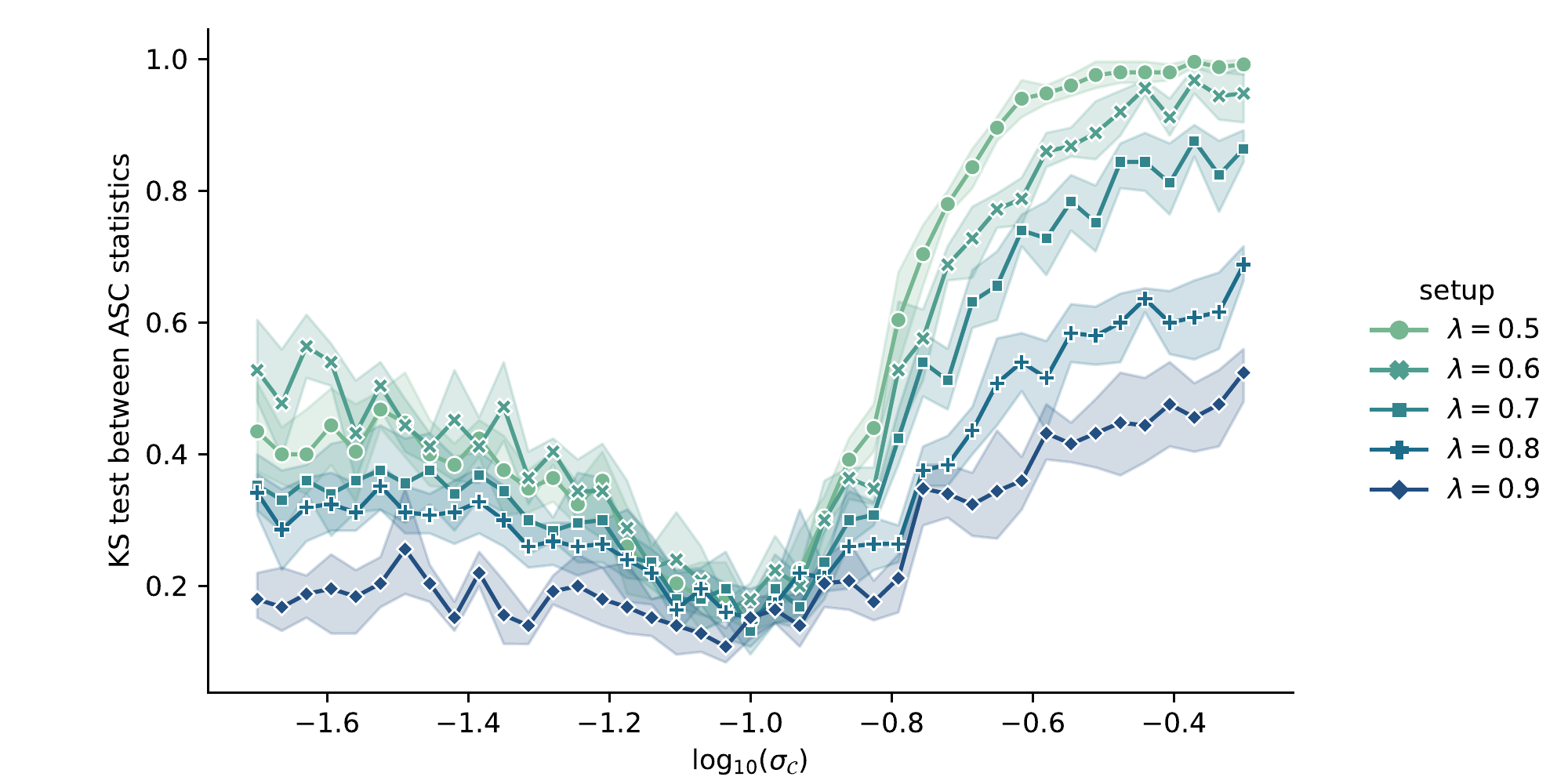}
	\caption{$\mathrm{ASC}$ statistics with $\phi(t)=t\log(t)$}
	\end{subfigure}
	\begin{subfigure}[t!]{0.5\textwidth}
	\centering 
	\includegraphics[trim=30 0 0 5, clip, width=0.95\textwidth]{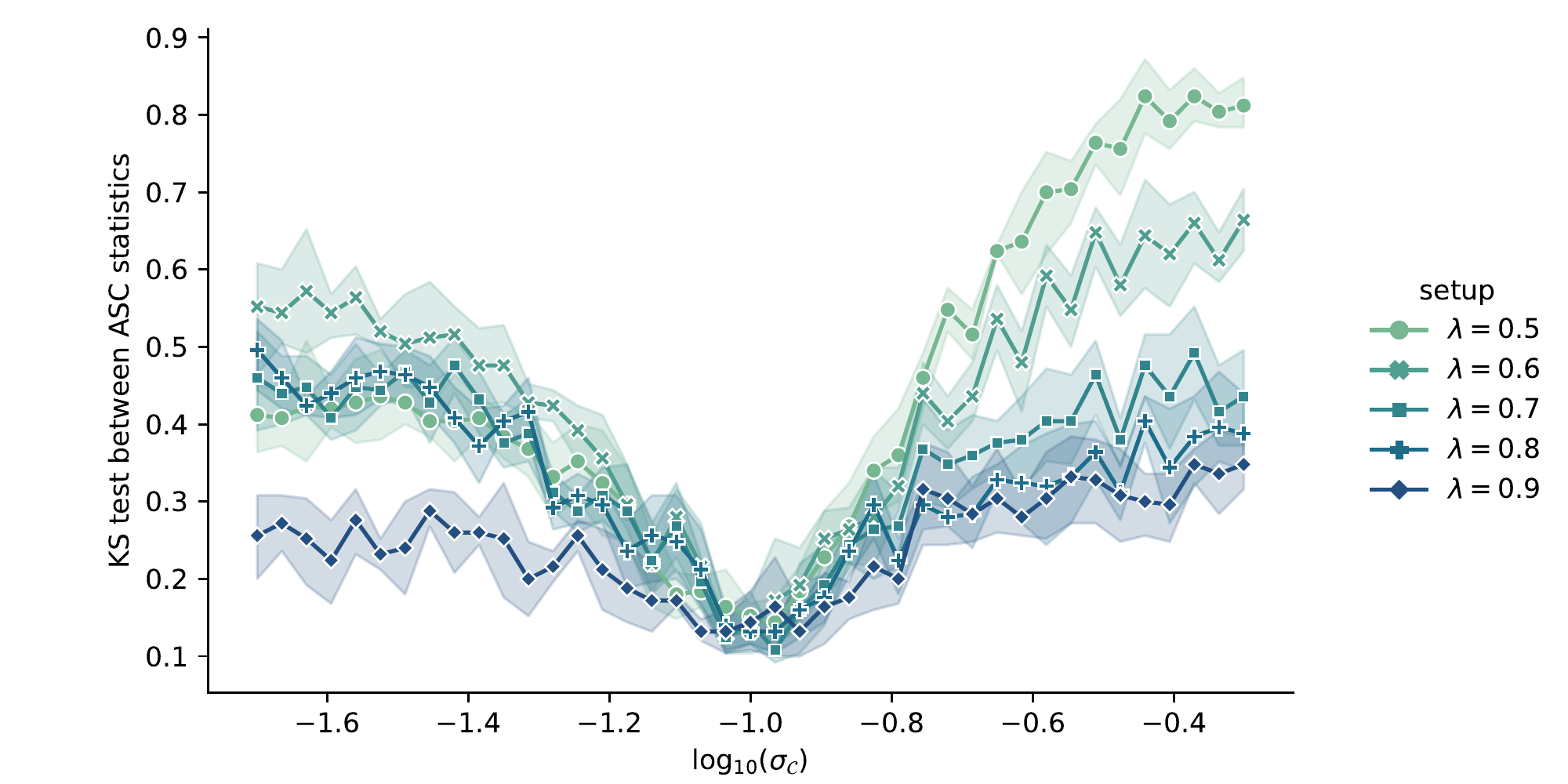}
	\caption{$\mathrm{ASC}$ statistics with $\phi(t)=(\sqrt{t}-1)^2$}
	\end{subfigure}
	
	\vspace{-0.3em}
	\caption{KS tests between distributions of statistics for KBC with different $\sigma_{\mC}$. (a) $\mathrm{LR}(Y_{H_1},\hat{\rho})$ vs $\mathrm{LR}(Y_{\mD},\hat{\rho})$ with $\lambda=0.8$. (b)-(d) $\hat{\mathrm{ASC}}_{\phi}(\hat{Y},Y_{H_1},\hat{\rho})$ vs $\hat{\mathrm{ASC}}_{\phi}(\hat{Y},Y_{\mD},\hat{\rho})$ for different $\phi$. Smaller values indicate the two compared distributions are closer. }
	\vspace{-0.3em}
	\label{fig: 2d Q2 KS CKB-8 appendix}
\end{figure}

\newpage
We visualize KS test results for KBC with different bandwidth $\sigma_{\mC}$ in Fig. \ref{fig: 2d Q2 KS CKB-8 appendix} (extension of Fig. \ref{fig: 2d Q2 KS} for CKB-8). The conclusions for CKB-8 are similar to MoG-8, except that the KS values are slightly higher, indicating the fast deletion is slightly harder for this dataset.

\newpage
\paragraph{Question 3 (Hypothesis Test).}

\begin{figure}[!h]
\vspace{-0.3em}
	\begin{subfigure}[t!]{0.5\textwidth}
	\centering 
	\includegraphics[width=0.8\textwidth]{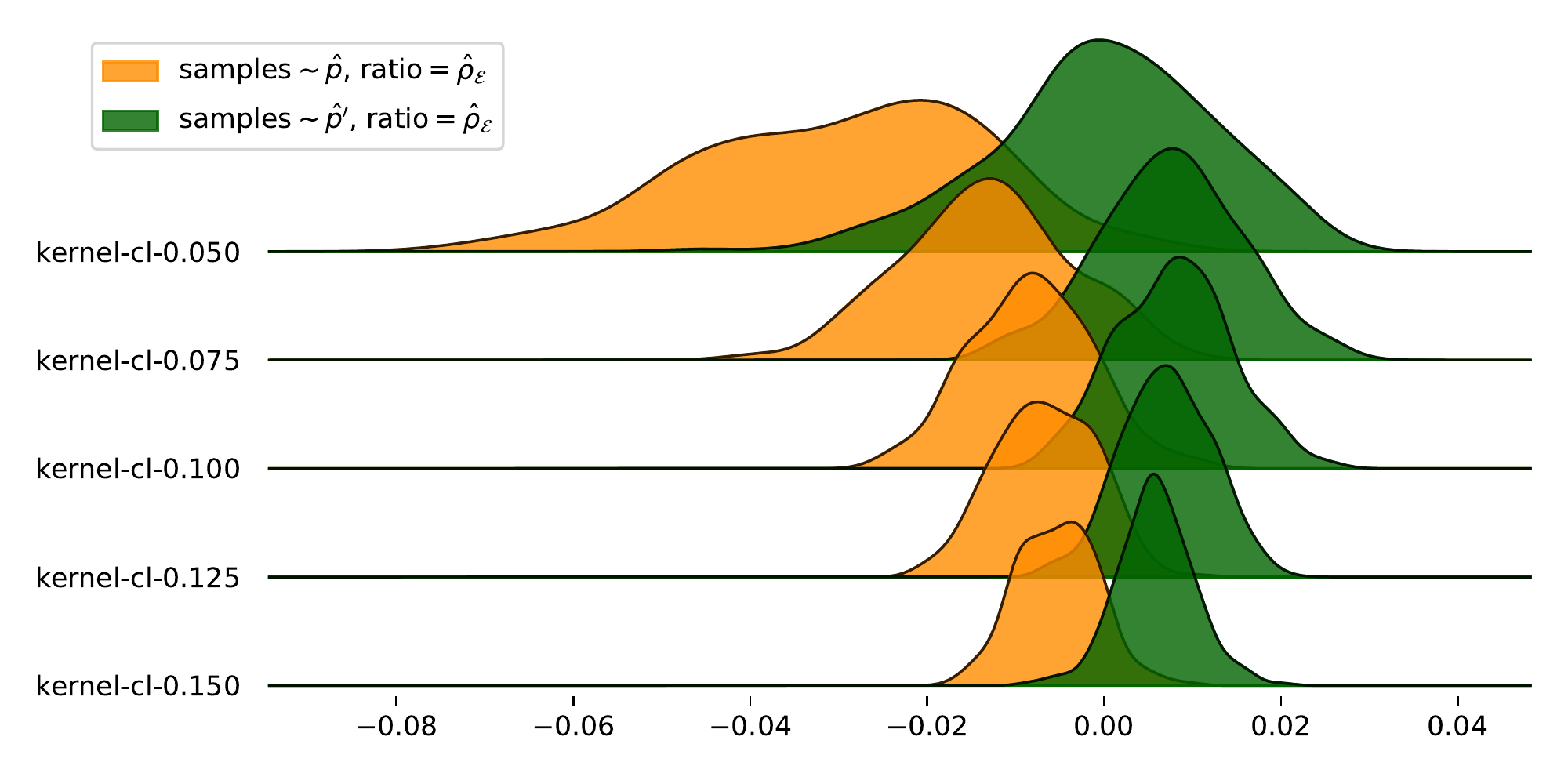}
	\caption{$\mathrm{LR}$ for KBC-based DRE ($\lambda=0.8$)}
	\end{subfigure}
	\begin{subfigure}[t!]{0.5\textwidth}
	\centering 
	\includegraphics[width=0.8\textwidth]{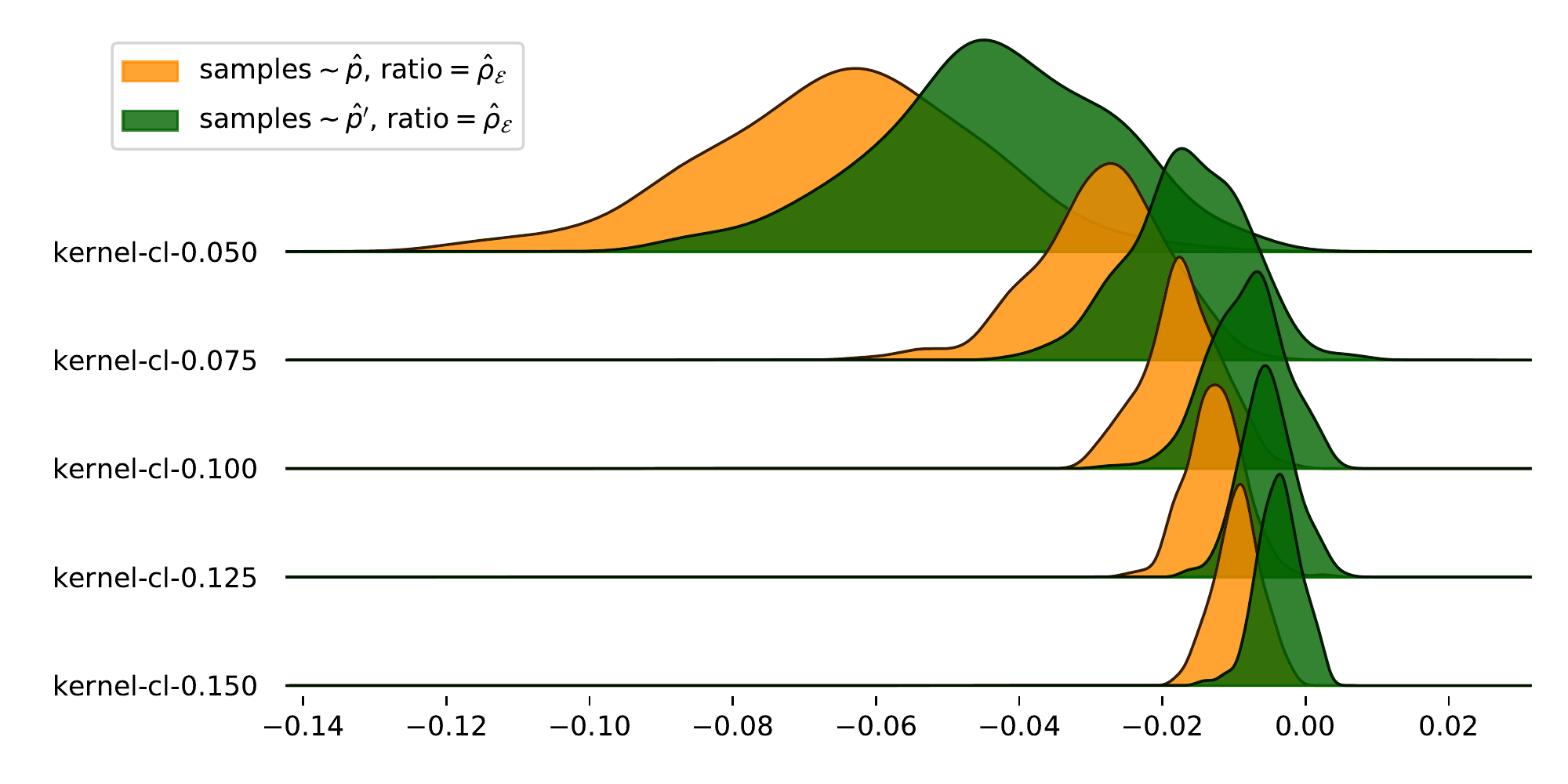}
	\caption{$\mathrm{ASC}$ for KBC-based DRE ($\phi(t)=\log(t)$)}
	\end{subfigure}\\
	\begin{subfigure}[t!]{0.5\textwidth}
	\centering 
	\includegraphics[width=0.8\textwidth]{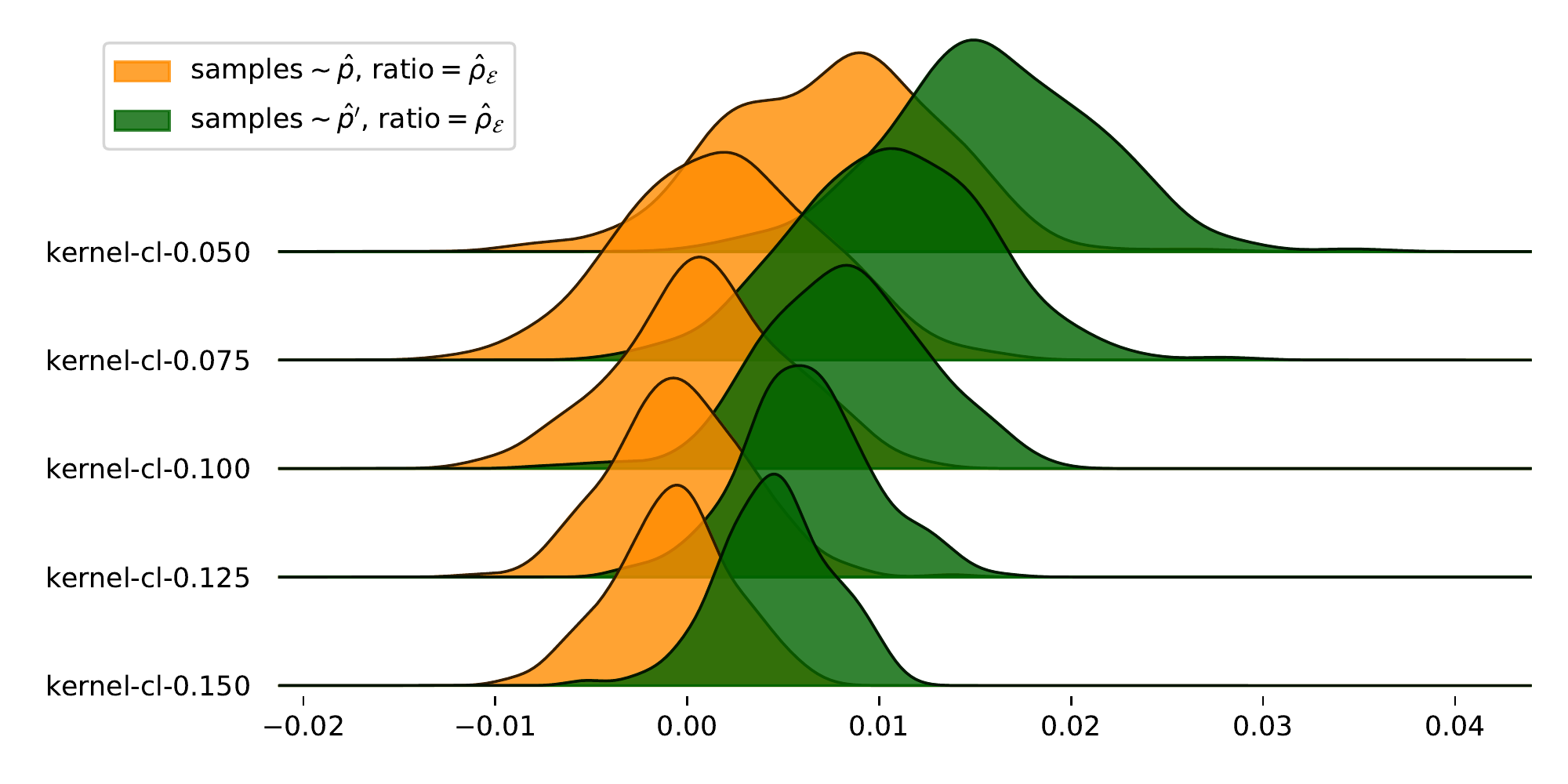}
	\caption{$\mathrm{ASC}$ for KBC-based DRE ($\phi(t)=t\log(t)$)}
	\end{subfigure}
	\begin{subfigure}[t!]{0.5\textwidth}
	\centering 
	\includegraphics[width=0.8\textwidth]{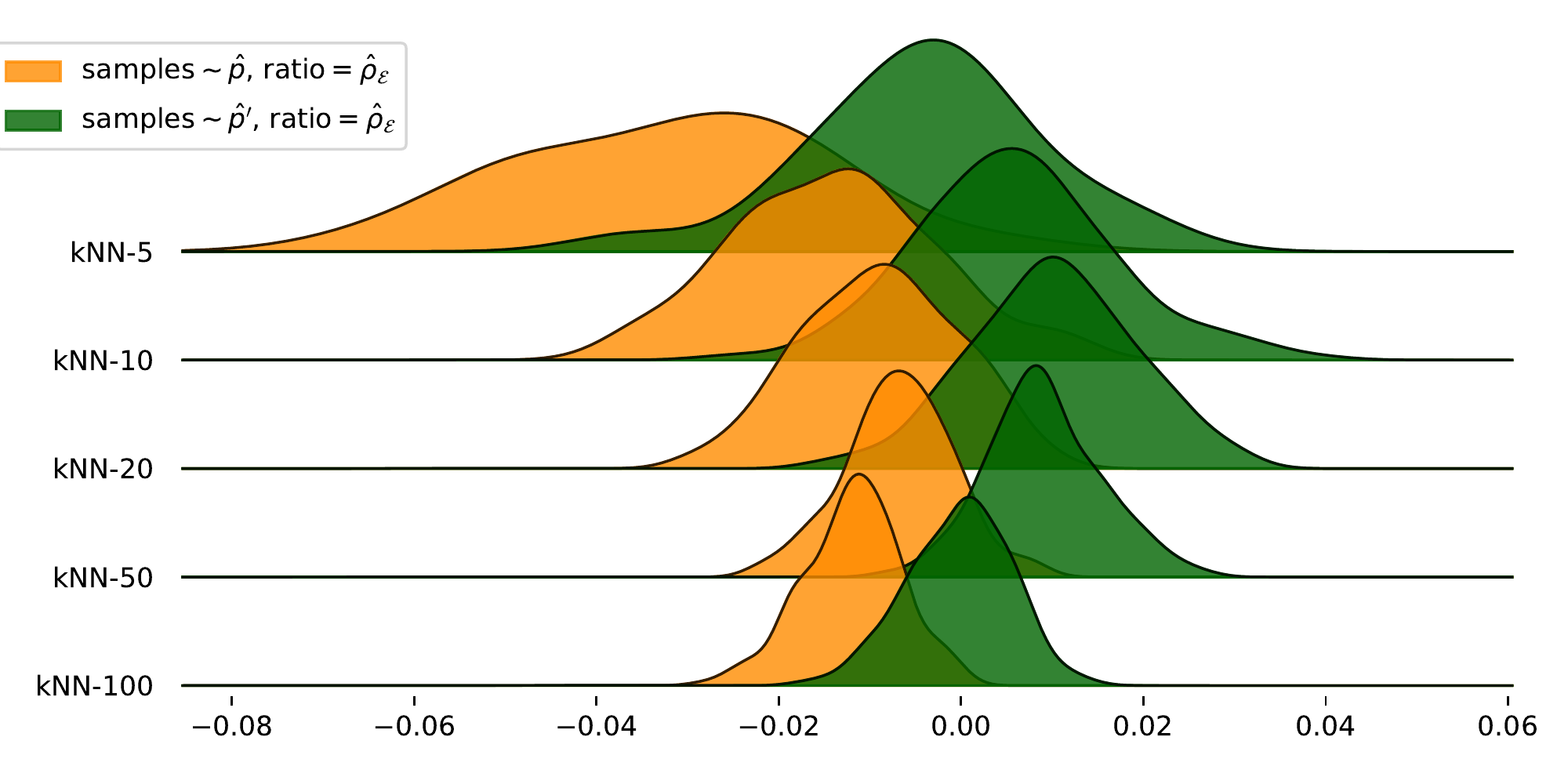}
	\caption{$\mathrm{LR}$ for $k$NN-based DRE ($\lambda=0.8$)}
	\end{subfigure}\\
	\begin{subfigure}[t!]{0.5\textwidth}
	\centering 
	\includegraphics[width=0.8\textwidth]{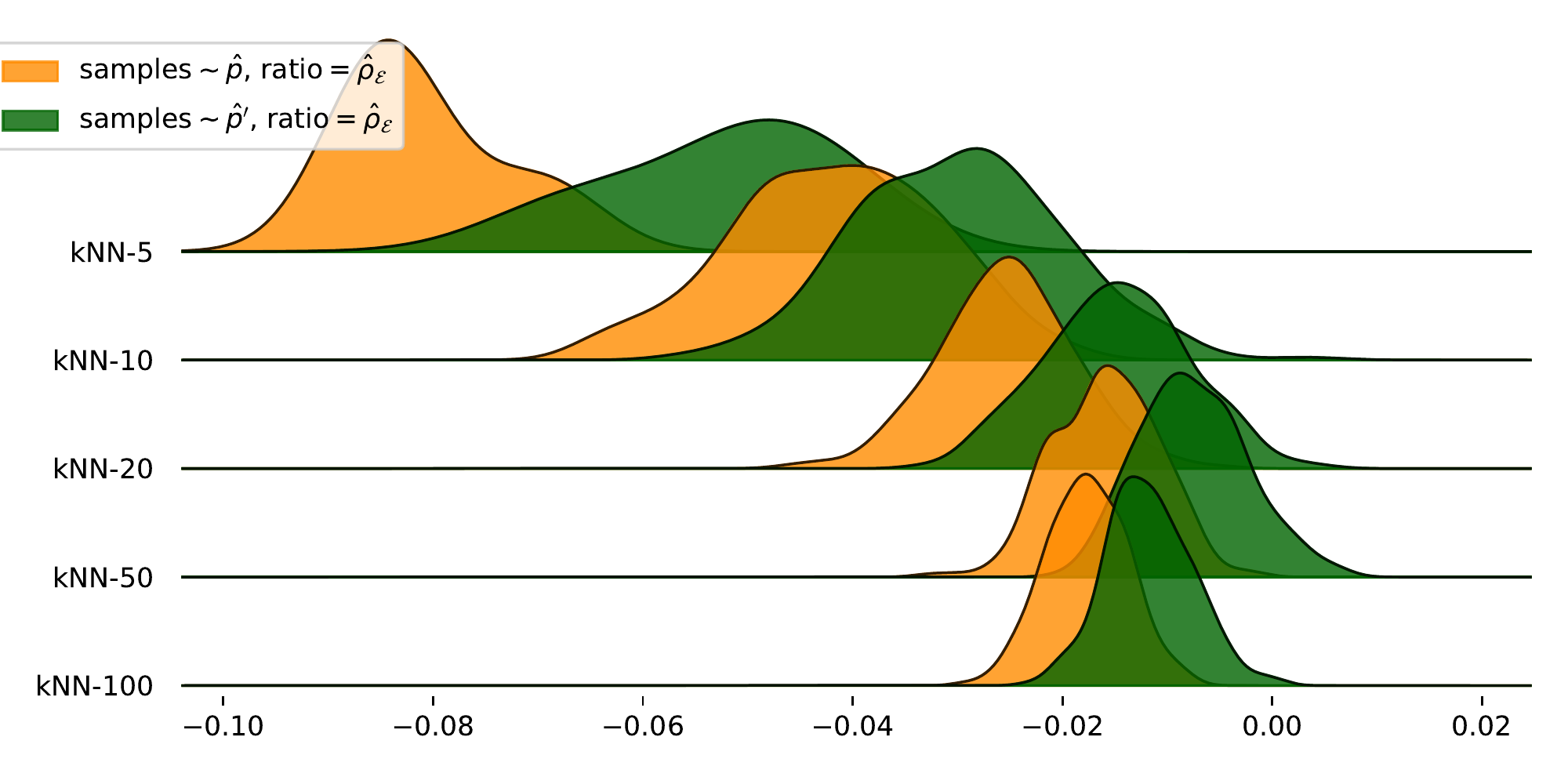}
	\caption{$\mathrm{ASC}$ for $k$NN-based DRE ($\phi(t)=\log(t)$)}
	\end{subfigure}
	\begin{subfigure}[t!]{0.5\textwidth}
	\centering 
	\includegraphics[width=0.8\textwidth]{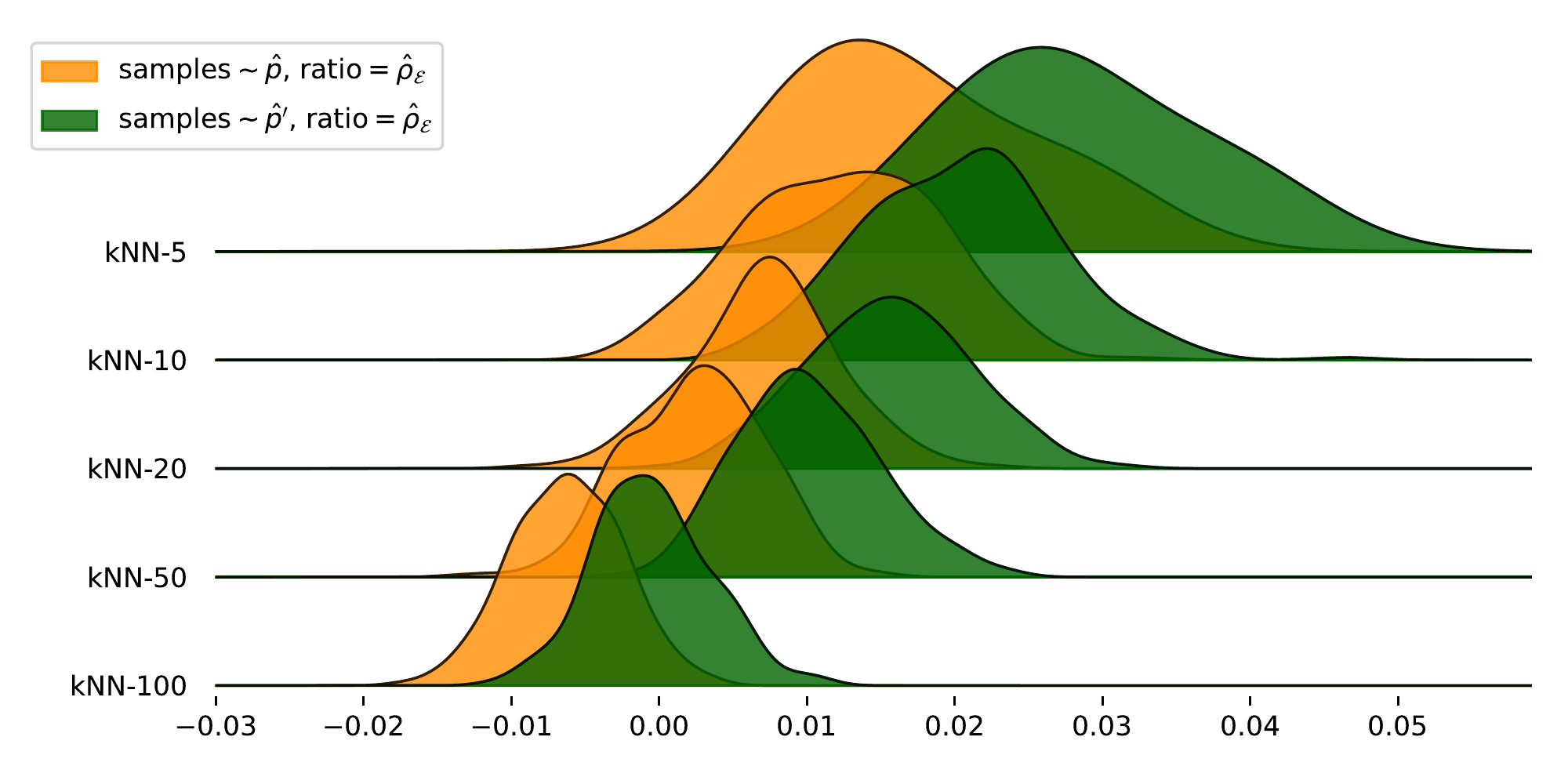}
	\caption{$\mathrm{ASC}$ for  $k$NN-based DRE ($\phi(t)=t\log(t)$)}
	\end{subfigure}
		
	\vspace{-0.3em}
	\caption{(a)-(c) KBC-based DRE. (d)-(f) $k$NN-based DRE. (a)\&(d) $\mathrm{LR}(Y_{H_0},\hat{\rho})$ vs $\mathrm{LR}(Y_{H_1},\hat{\rho})$. (b)-(c)\&(e)-(f) $\hat{\mathrm{ASC}}_{\phi}(\hat{Y},Y_{H_0},\hat{\rho})$ vs $\hat{\mathrm{ASC}}_{\phi}(\hat{Y},Y_{H_1},\hat{\rho})$.}
	\vspace{-0.3em}
	\label{fig: 2d Q3 joy CKB-8 appendix}
\end{figure}

We visualize distributions of LR and ASC statistics between $Y_{H_0}$ and $Y_{H_1}$ in Fig. \ref{fig: 2d Q3 joy CKB-8 appendix} (extension of Fig. \ref{fig: 2d Q3 joy} for CKB-8). The separation between the distributions indicates how the DRE can distinguish samples between pre-trained and re-trained models. Similar to MoG-8, LR statistics lead to better separation than ASC. 

\begin{figure}[!h]
\vspace{-0.3em}
  	\begin{subfigure}[t!]{0.5\textwidth}
	\centering 
	\includegraphics[trim=30 0 0 5, clip, width=0.95\textwidth]{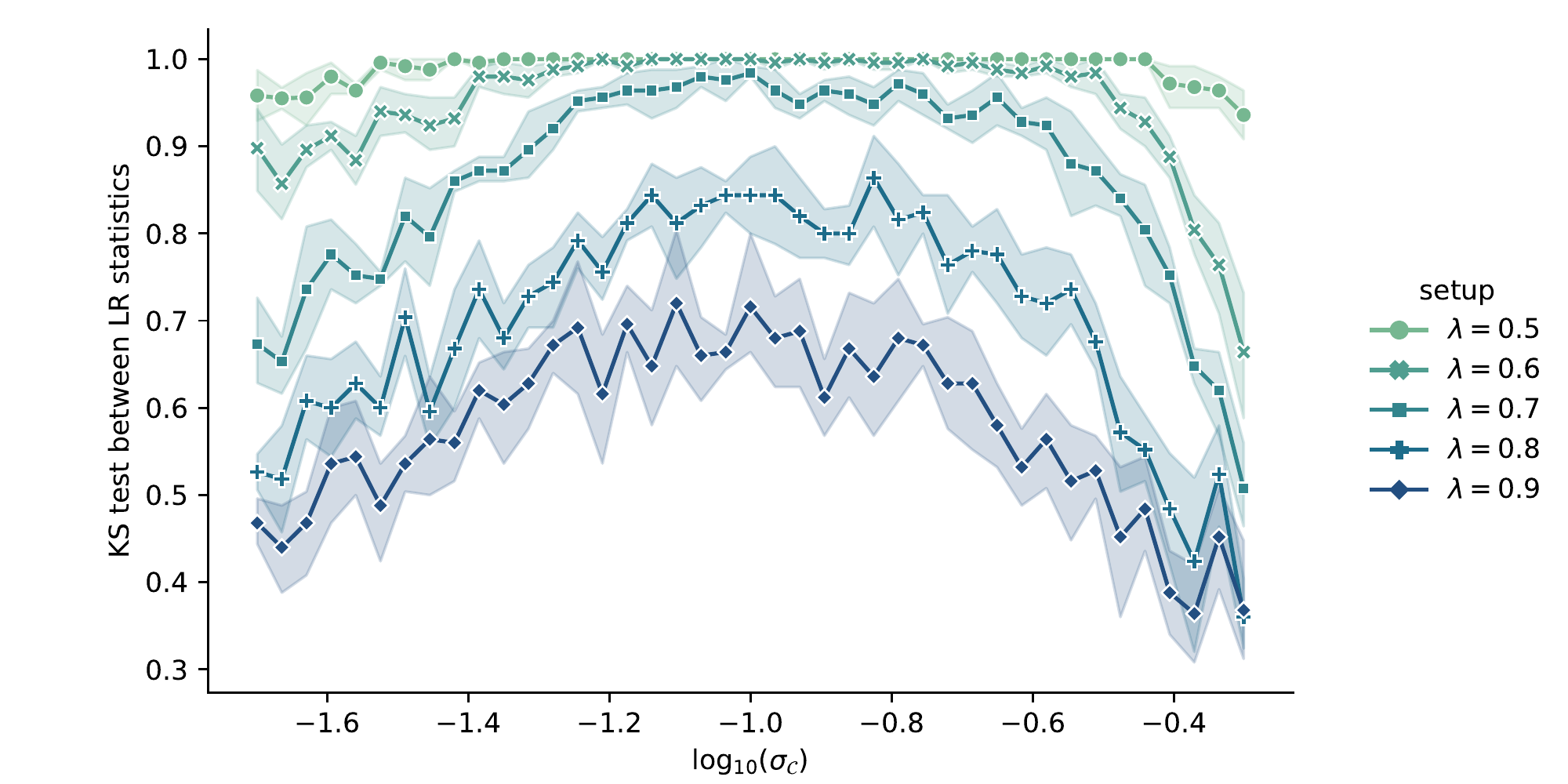}
	\caption{$\mathrm{LR}$ statistics}
	\end{subfigure}
	\begin{subfigure}[t!]{0.5\textwidth}
	\centering 
	\includegraphics[trim=30 0 0 5, clip, width=0.95\textwidth]{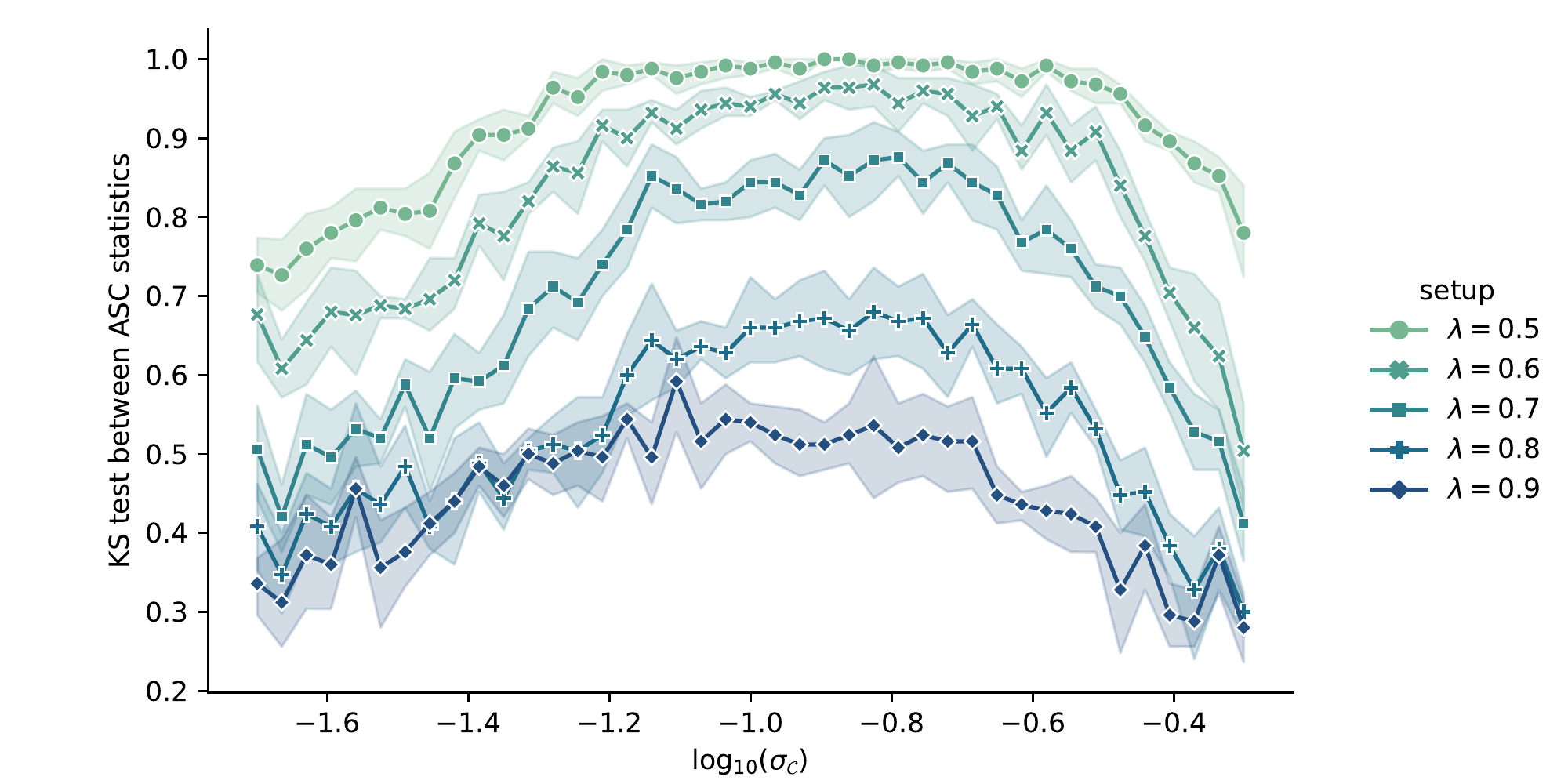}
	\caption{$\mathrm{ASC}$ statistics with $\phi(t)=\log(t)$}
	\end{subfigure}\\
	\begin{subfigure}[t!]{0.5\textwidth}
	\centering 
	\includegraphics[trim=30 0 0 5, clip, width=0.95\textwidth]{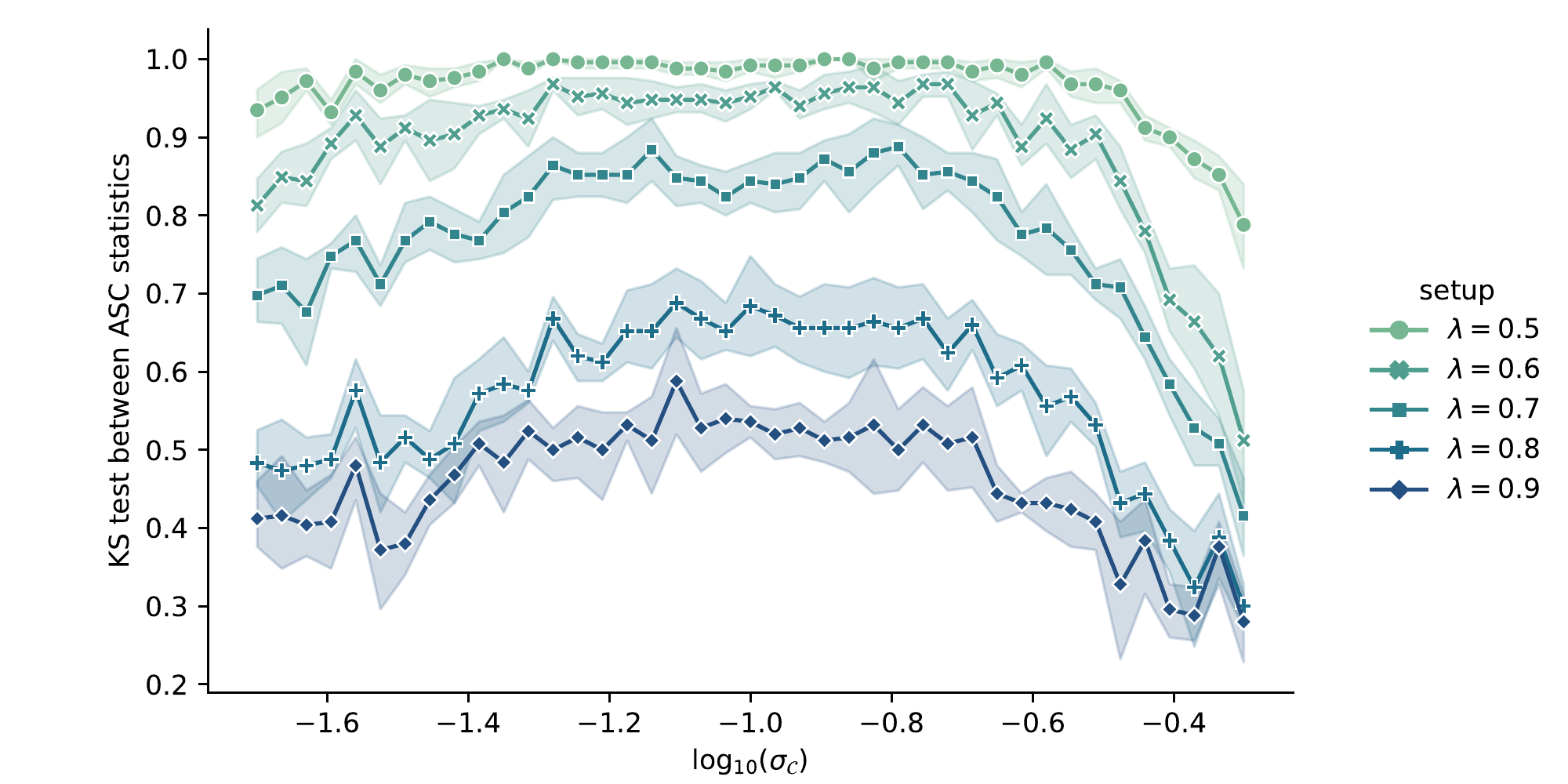}
	\caption{$\mathrm{ASC}$ statistics with $\phi(t)=t\log(t)$}
	\end{subfigure}
	\begin{subfigure}[t!]{0.5\textwidth}
	\centering 
	\includegraphics[trim=30 0 0 5, clip, width=0.95\textwidth]{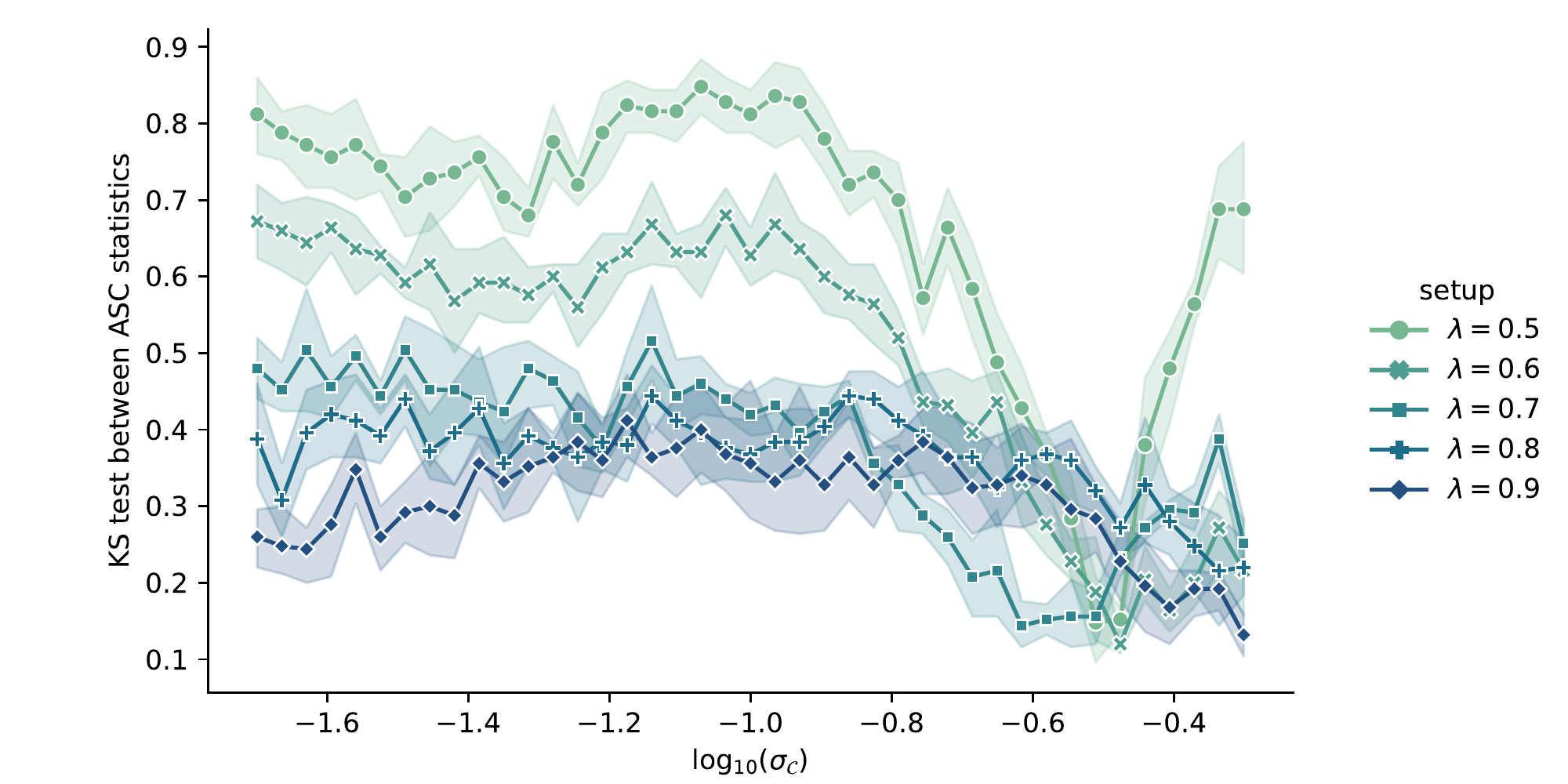}
	\caption{$\mathrm{ASC}$ statistics with $\phi(t)=(\sqrt{t}-1)^2$}
	\end{subfigure}
	
	\vspace{-0.3em}
	\caption{KS tests between distributions of statistics for KBC with different $\sigma_{\mC}$. (a) $\mathrm{LR}(Y_{H_0},\hat{\rho})$ vs $\mathrm{LR}(Y_{H_1},\hat{\rho})$ with $\lambda=0.8$. (b)-(d) $\hat{\mathrm{ASC}}_{\phi}(\hat{Y},Y_{H_0},\hat{\rho})$ vs $\hat{\mathrm{ASC}}_{\phi}(\hat{Y},Y_{H_1},\hat{\rho})$ for different $\phi$. Smaller values indicate the two compared distributions are closer. }
	\vspace{-0.3em}
	\label{fig: 2d Q3 KS CKB-8 appendix}
\end{figure}

\newpage
We visualize KS test results for KBC with different bandwidth $\sigma_{\mC}$ in Fig. \ref{fig: 2d Q3 KS CKB-8 appendix} (extension of Fig. \ref{fig: 2d Q3 KS} for CKB-8). Similar to MoG-8, LR statistics lead to better separation than ASC. 

\newpage
\section{Experiments on GAN}\label{appendix: exp GAN}

\subsection{Setup}

We run experiments on MNIST \citep{lecun2010mnist} and Fashion-MNIST \citep{xiao2017fashion}. Both datasets contain gray-scale $28\times28$ images with 10 labels $\{0,1,\cdots,9\}$. We define the $\texttt{even-}\lambda$ setting as the subset containing all samples with odd labels and a $\lambda$ fraction of samples with even labels randomly selected from the training set. The rest $1-\lambda$ fraction of samples with even labels form the deletion set $X'$. We have similar definition for $\texttt{odd-}\lambda$. In experiments, we let $\lambda\in\{0.6,0.7,0.8,0.9\}$.

The learner is a DCGAN \citep{radford2015unsupervised}. For pre-trained and re-trained models, we train each of them for 200 epochs. To obtain DRE, we optimize \eqref{eg: VDM KL}, where the network $T$ has the same architecture as the discriminator and is trained for 40 epochs. The learning rate is halved for stability. 

\newpage
\subsection{Results on MNIST}

\begin{figure}[!h]
\vspace{-0.3em}
  	\begin{subfigure}[t!]{0.5\textwidth}
	\centering 
	\includegraphics[width=0.95\textwidth]{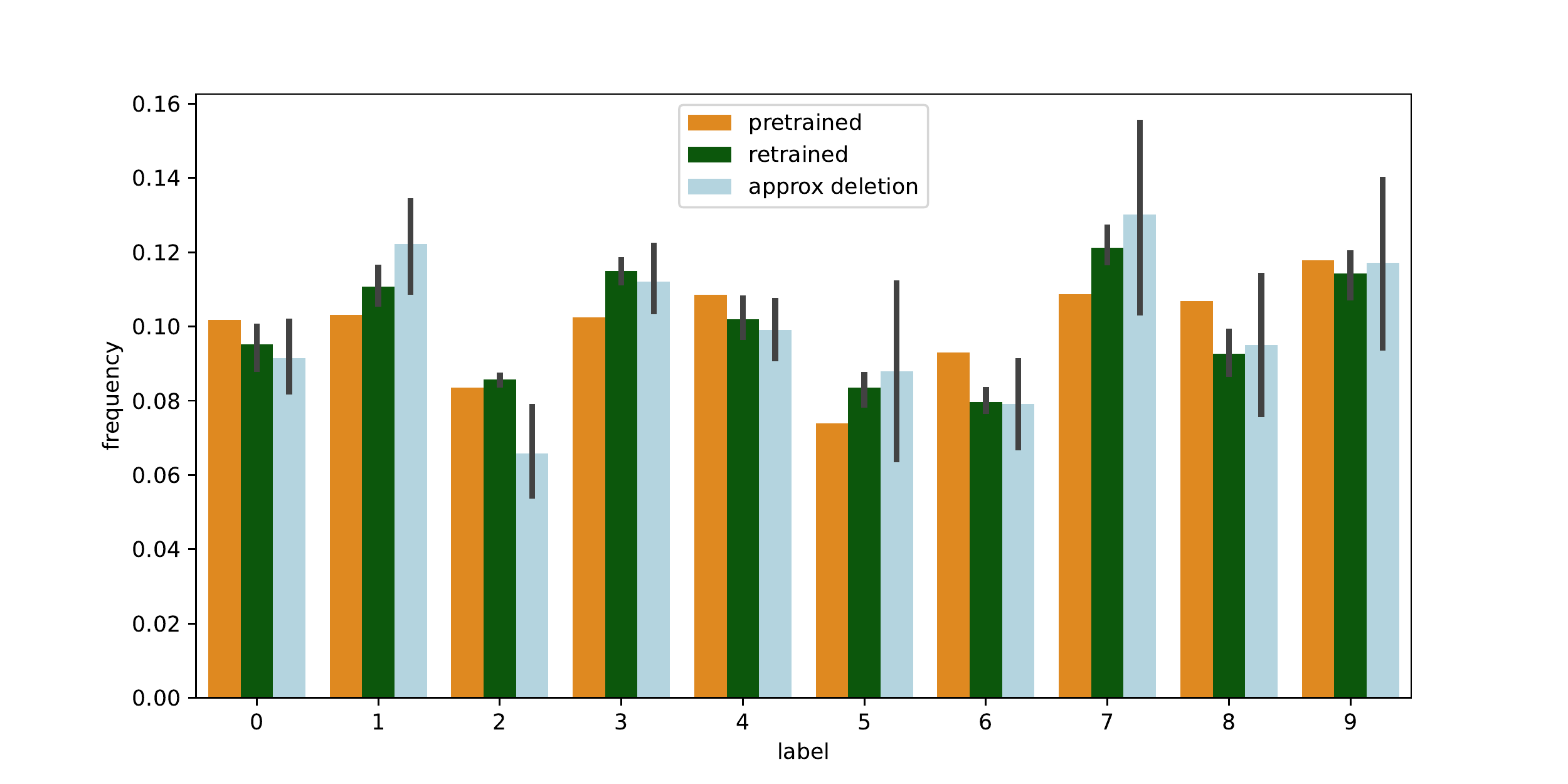}
	\caption{$\texttt{even-}0.9$}
	\end{subfigure}
	\begin{subfigure}[t!]{0.5\textwidth}
	\centering 
	\includegraphics[width=0.95\textwidth]{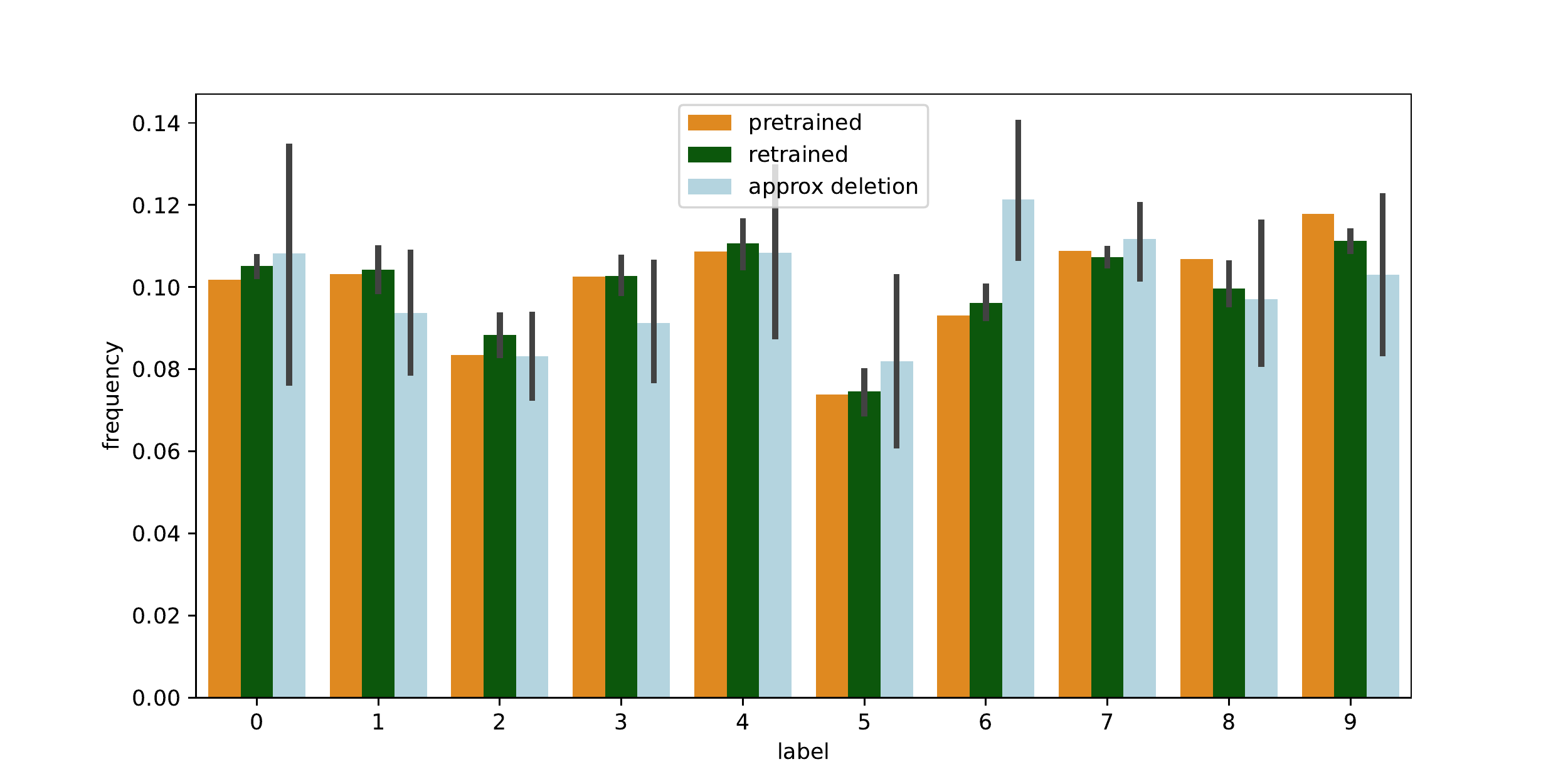}
	\caption{$\texttt{odd-}0.9$}
	\end{subfigure}\\
	\begin{subfigure}[t!]{0.5\textwidth}
	\centering 
	\includegraphics[width=0.95\textwidth]{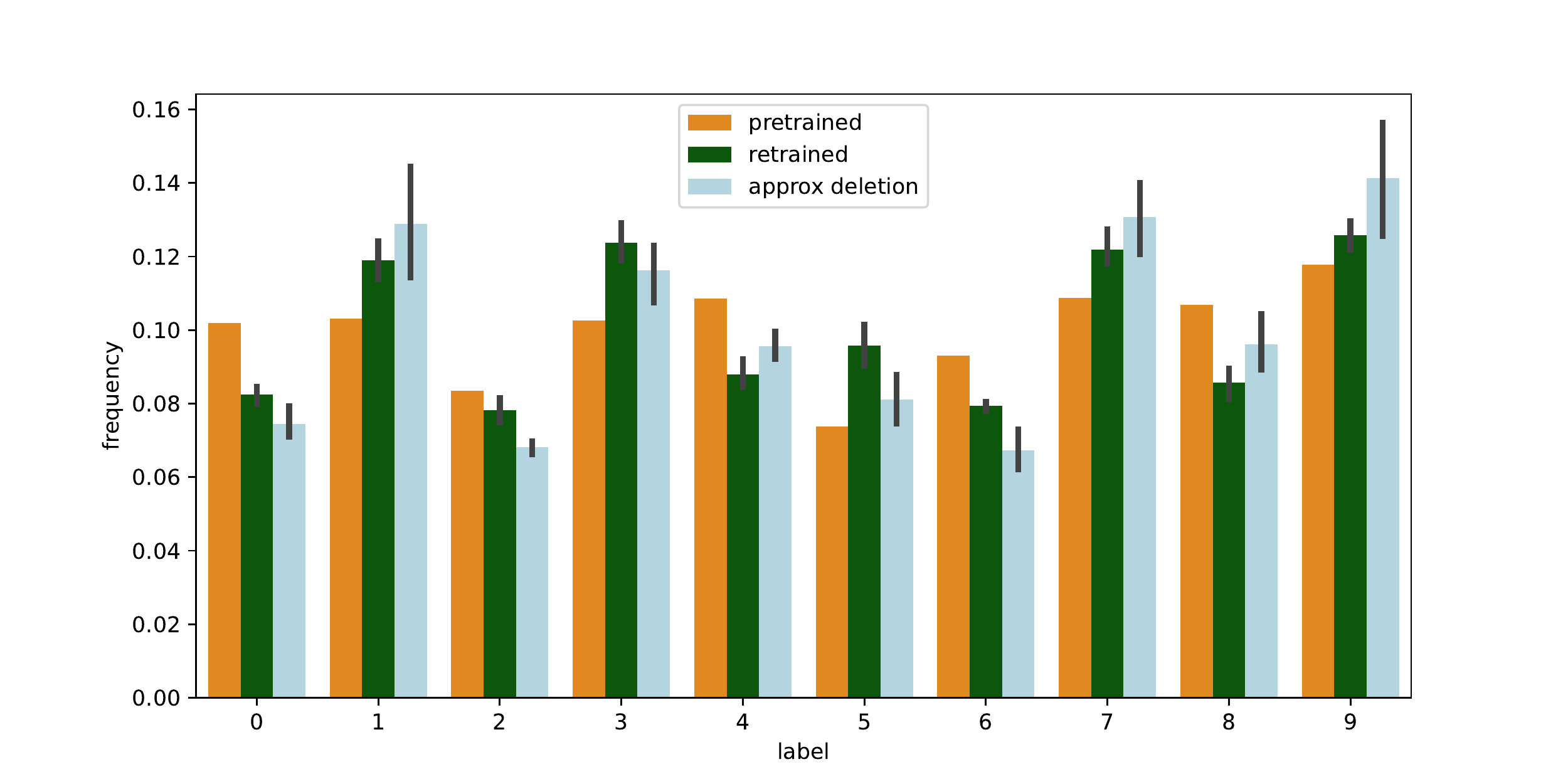}
	\caption{$\texttt{even-}0.8$}
	\end{subfigure}
	\begin{subfigure}[t!]{0.5\textwidth}
	\centering 
	\includegraphics[width=0.95\textwidth]{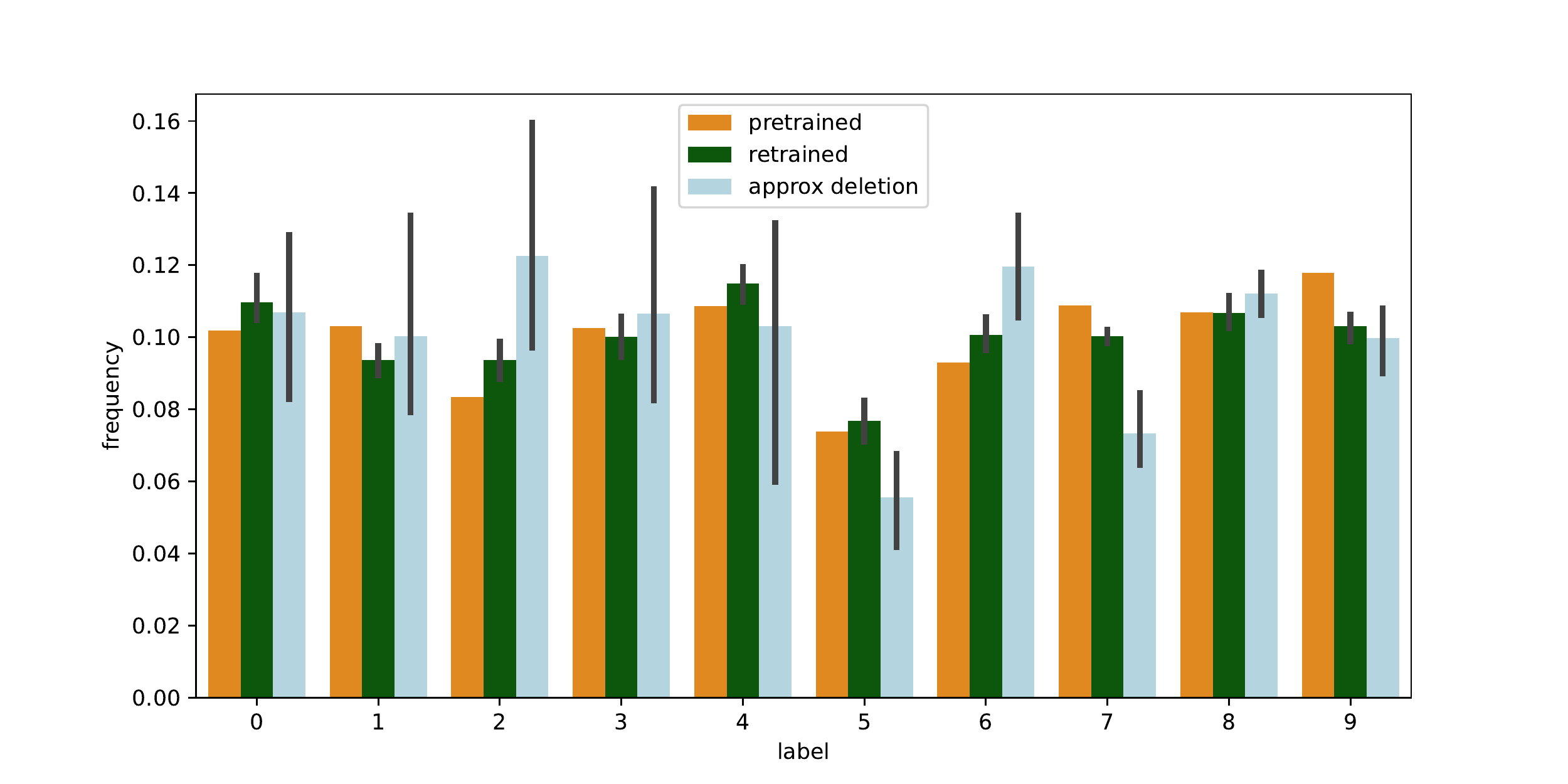}
	\caption{$\texttt{odd-}0.8$}
	\end{subfigure}\\
	\begin{subfigure}[t!]{0.5\textwidth}
	\centering 
	\includegraphics[width=0.95\textwidth]{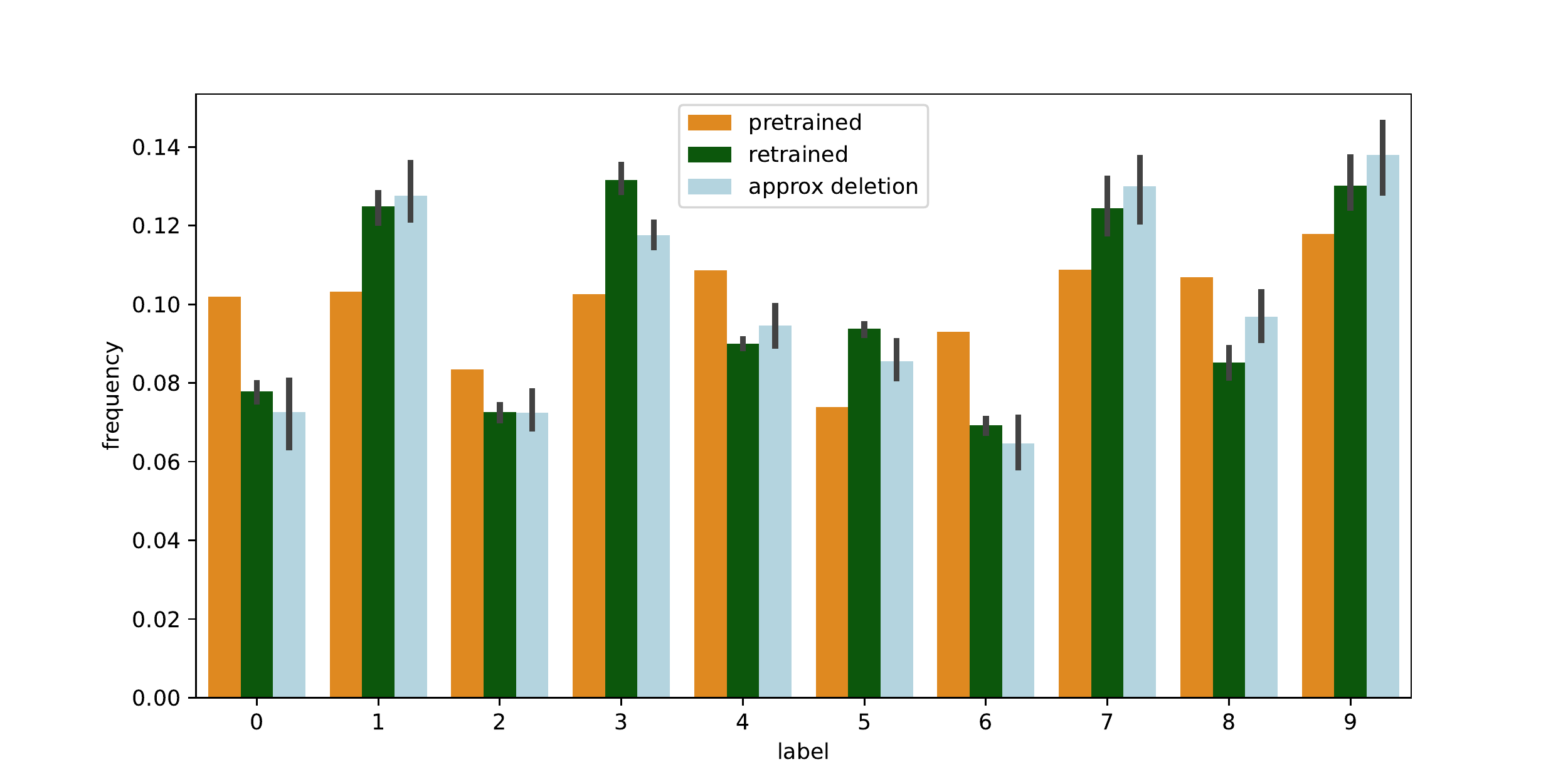}
	\caption{$\texttt{even-}0.7$}
	\end{subfigure}
	\begin{subfigure}[t!]{0.5\textwidth}
	\centering 
	\includegraphics[width=0.95\textwidth]{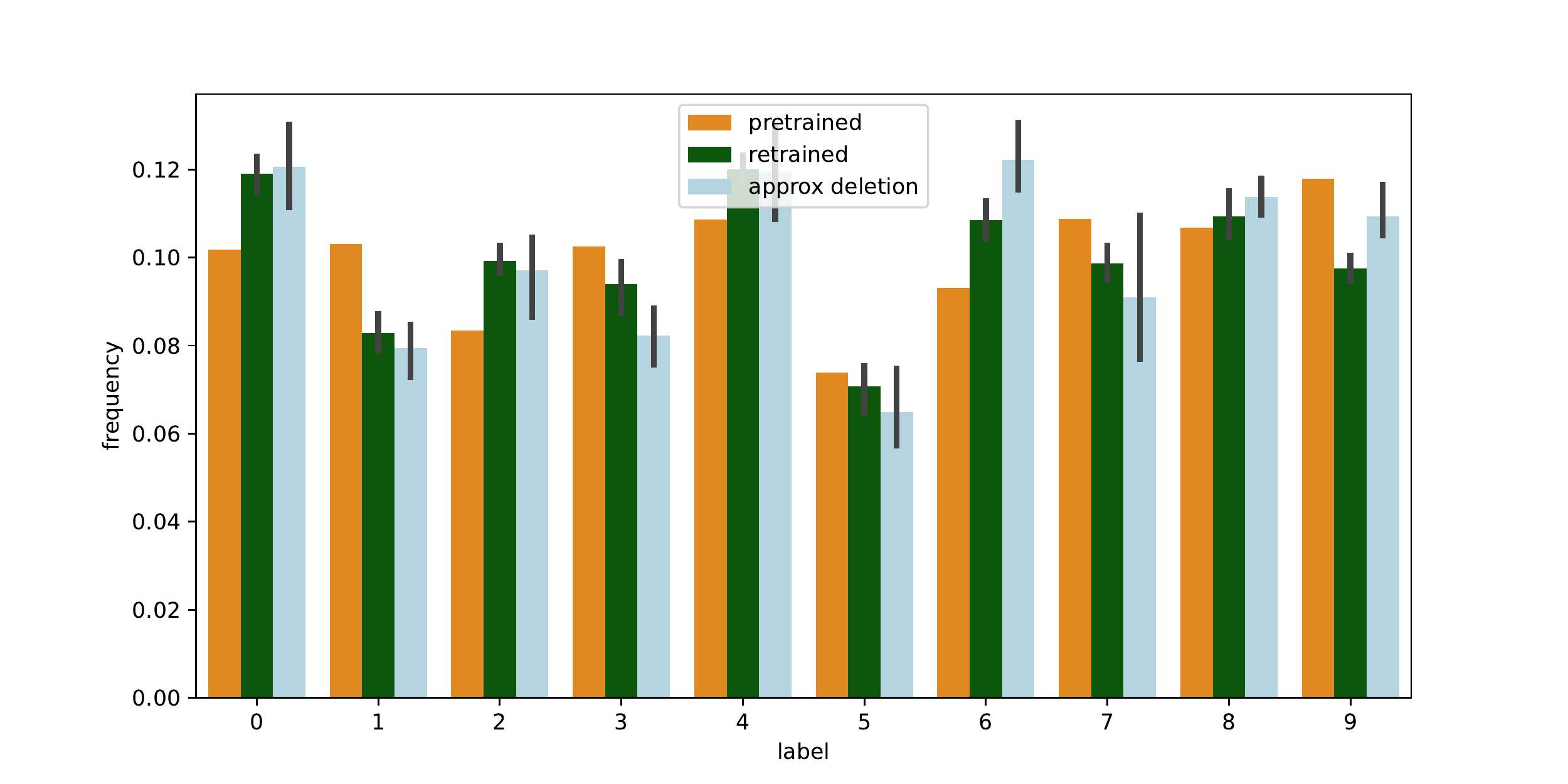}
	\caption{$\texttt{odd-}0.7$}
	\end{subfigure}\\
	\begin{subfigure}[t!]{0.5\textwidth}
	\centering 
	\includegraphics[width=0.95\textwidth]{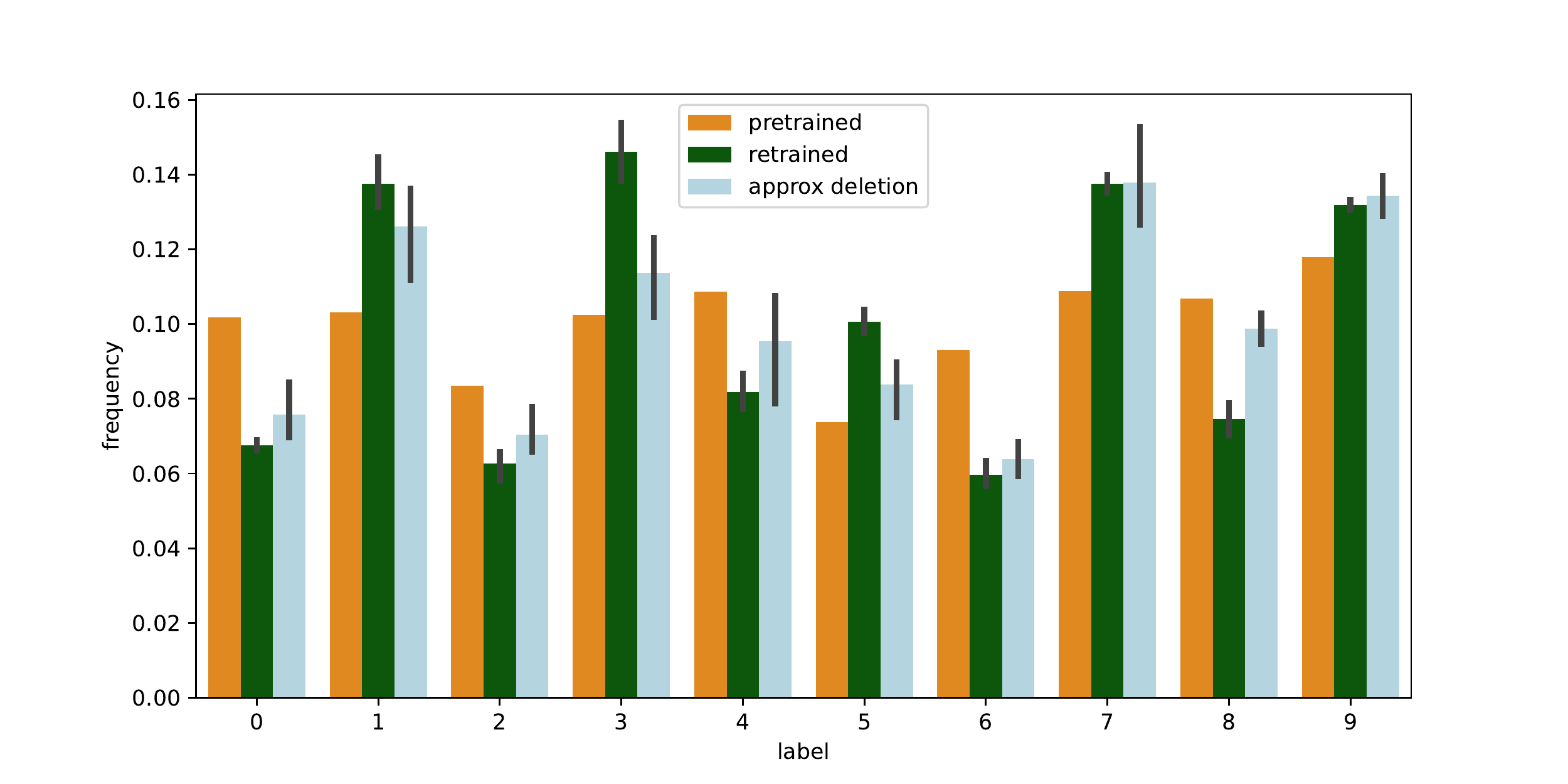}
	\caption{$\texttt{even-}0.6$}
	\end{subfigure}
	\begin{subfigure}[t!]{0.5\textwidth}
	\centering 
	\includegraphics[width=0.95\textwidth]{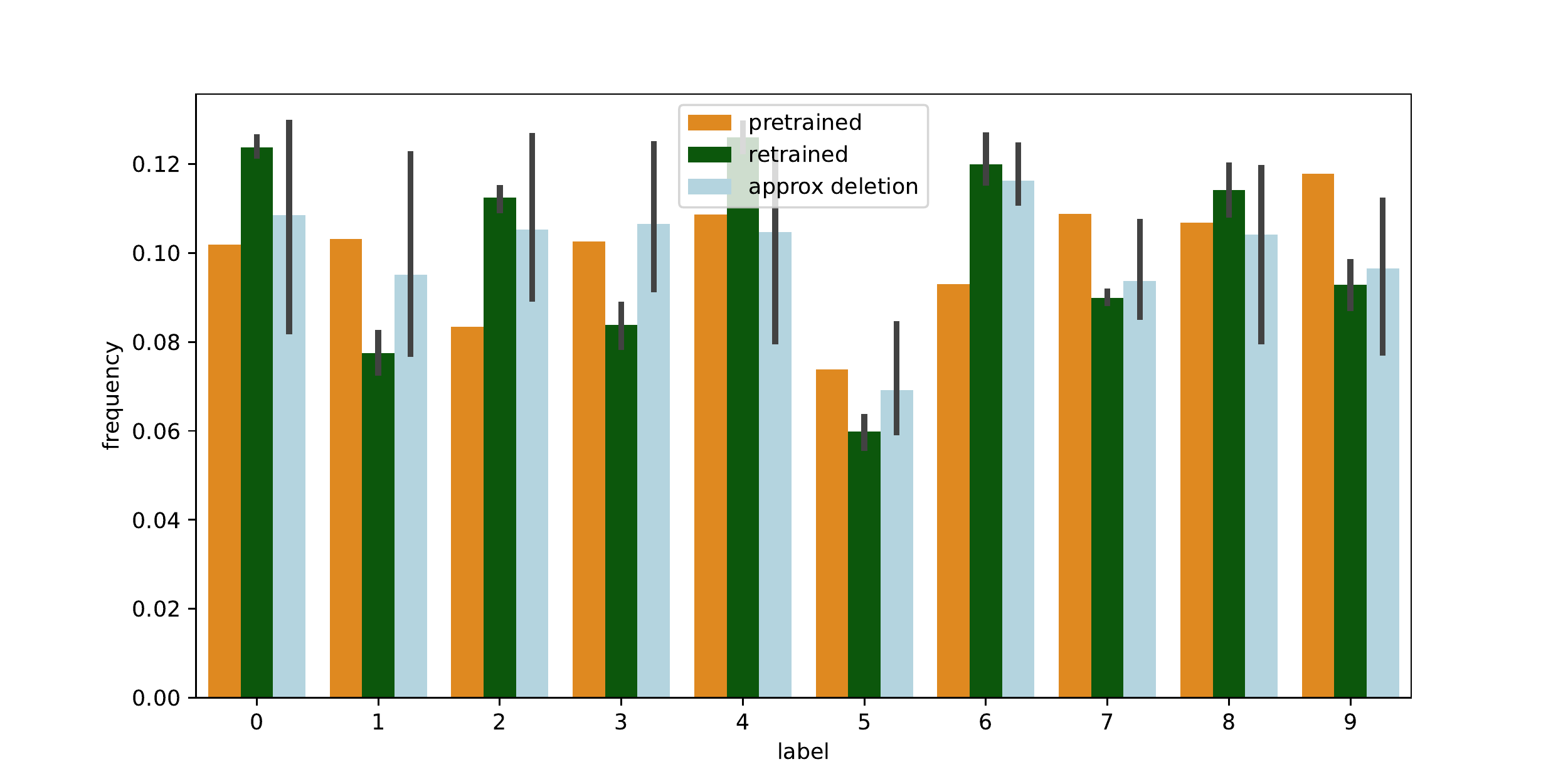}
	\caption{$\texttt{odd-}0.6$}
	\end{subfigure}\\
	
	\vspace{-0.3em}
	\caption{Label distributions of samples from pre-trained, re-trained, and approximated models. The closeness between green and light blue distributions indicate how well the fast deletion performs.}
	\vspace{-0.3em}
	\label{fig: GAN Q2 MNIST appendix}
\end{figure}

\paragraph{Question 2 (Fast Deletion).}
We generate $m=50$K samples from pre-retrained, re-trained, and approximated models (with rejection sampling bound $B=10$). We then compute the label distributions of these samples based on pre-trained classifiers. \footnote{\url{https://github.com/aaron-xichen/pytorch-playground} (MIT license)} Results for each deletion set (including means and standard errors for five random seeds) are shown in Fig. \ref{fig: GAN Q2 MNIST appendix} (extension of Fig. \ref{fig: GAN Q2 MNIST}). We find the approximated model generates less (even or odd) labels some data with these labels are deleted from the training set. The variances for deleting odd labels are higher than deleting even labels.

\begin{figure}[!h]
\vspace{-0.3em}
	\begin{subfigure}[t!]{0.5\textwidth}
	\centering 
	\includegraphics[width=0.95\textwidth]{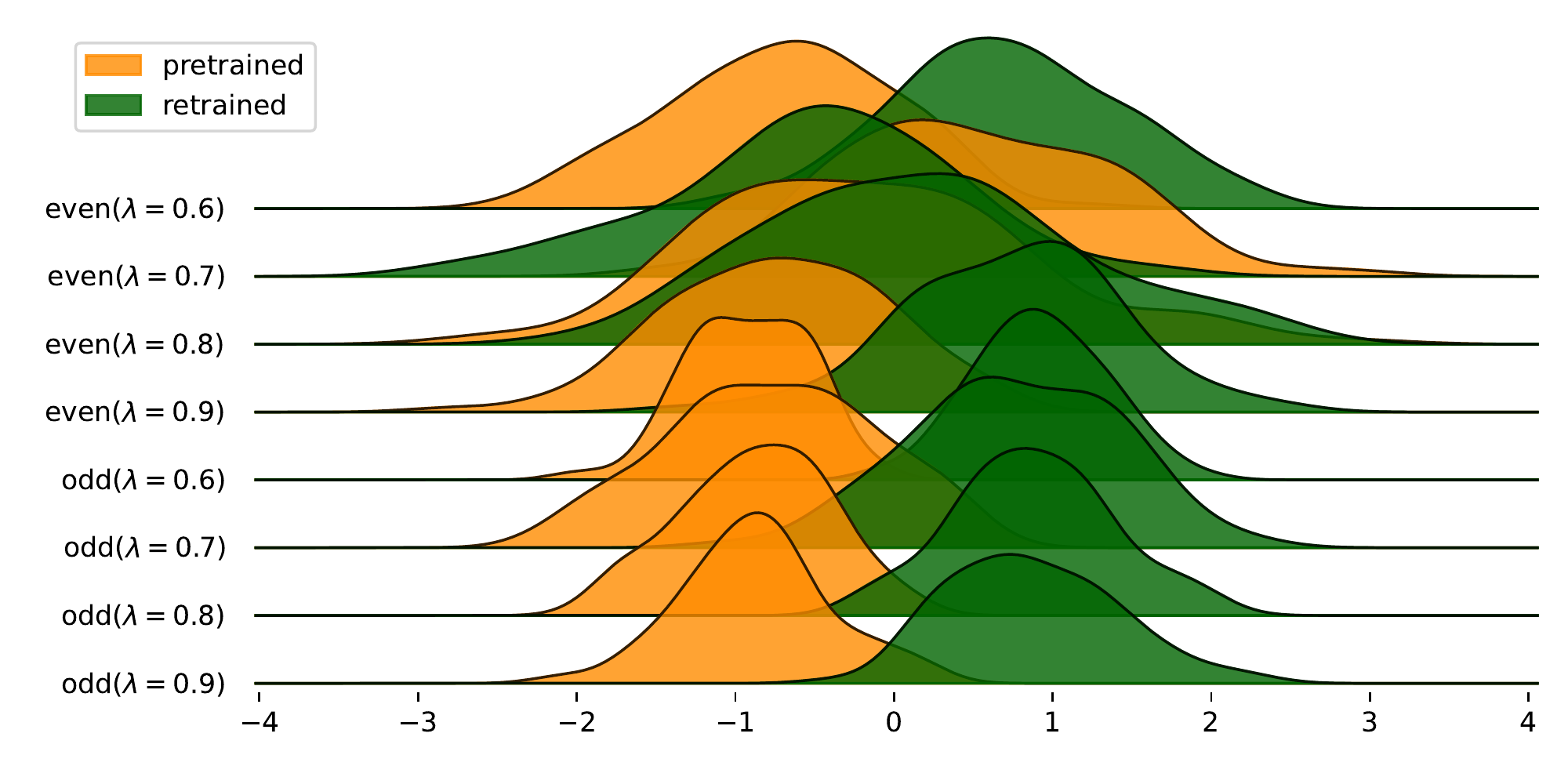}
	\caption{$\mathrm{ASC}$ for VDM-based DRE ($\phi(t)=\log(t)$)}
	\end{subfigure}
	\begin{subfigure}[t!]{0.5\textwidth}
	\centering 
	\includegraphics[width=0.95\textwidth]{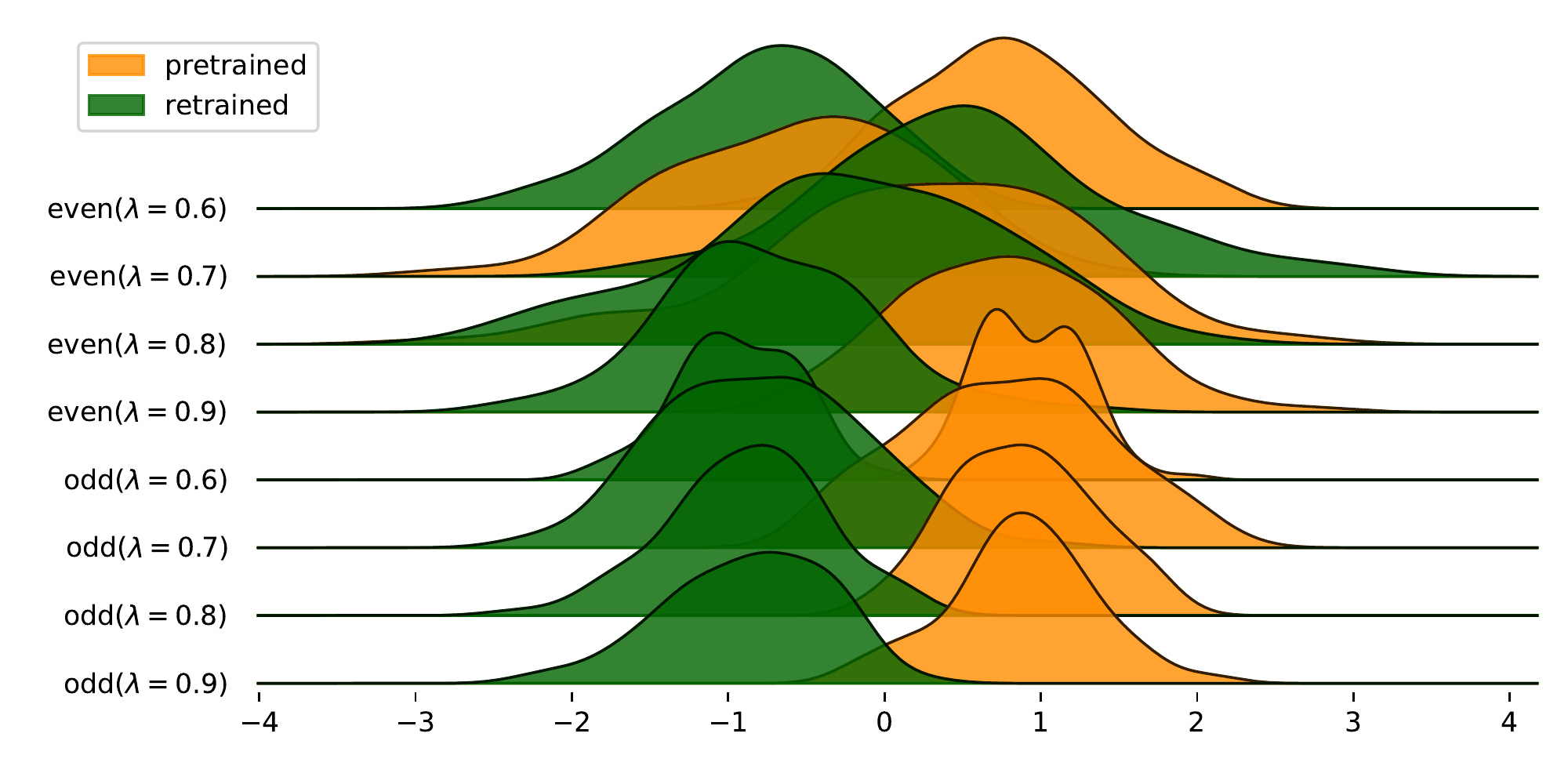}
	\caption{$\mathrm{ASC}$ for VDM-based DRE ($\phi(t)=t\log(t)$)}
	\end{subfigure}
		
	\vspace{-0.3em}
	\caption{(a)-(b) $\hat{\mathrm{ASC}}_{\phi}(\hat{Y},Y_{H_0},\hat{\rho})$ vs $\hat{\mathrm{ASC}}_{\phi}(\hat{Y},Y_{H_1},\hat{\rho})$.}
	\vspace{-0.3em}
	\label{fig: GAN Q3 MNIST appendix}
\end{figure}

\newpage
\paragraph{Question 3 (Hypothesis Test).}
We generate $m=1000$ samples for each $Y_{H_i}$, $i=1,2$, and $\hat{Y}$. We visualize distributions of LR and ASC statistics between $Y_{H_0}$ and $Y_{H_1}$ in Fig. \ref{fig: GAN Q3 MNIST appendix} (extension of Fig. \ref{fig: GAN Q3 LR MNIST}). The separation between the distributions indicates how the DRE can distinguish samples between pre-trained and re-trained models. The separation for $\texttt{odd-}\lambda$ is better than $\texttt{even-}\lambda$. In terms of statistics, the LR is slightly better than ASC. In terms of $\lambda$, a smaller $\lambda$ does not lead to more separation.

\newpage
\subsection{Results on Fashion-MNIST}

\begin{figure}[!h]
\vspace{-0.3em}
  	\begin{subfigure}[t!]{0.5\textwidth}
	\centering 
	\includegraphics[width=0.95\textwidth]{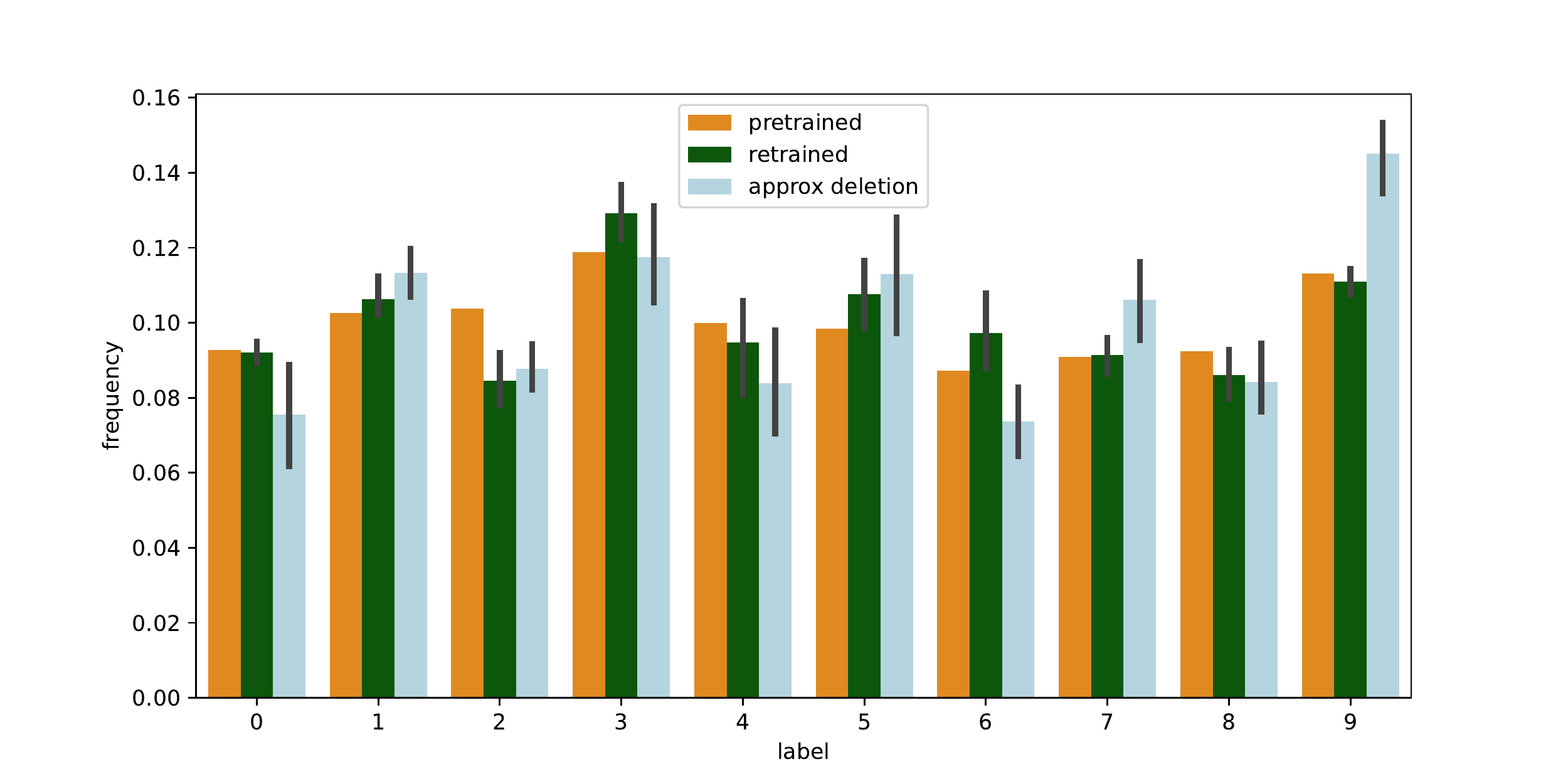}
	\caption{$\texttt{even-}0.9$}
	\end{subfigure}
	\begin{subfigure}[t!]{0.5\textwidth}
	\centering 
	\includegraphics[width=0.95\textwidth]{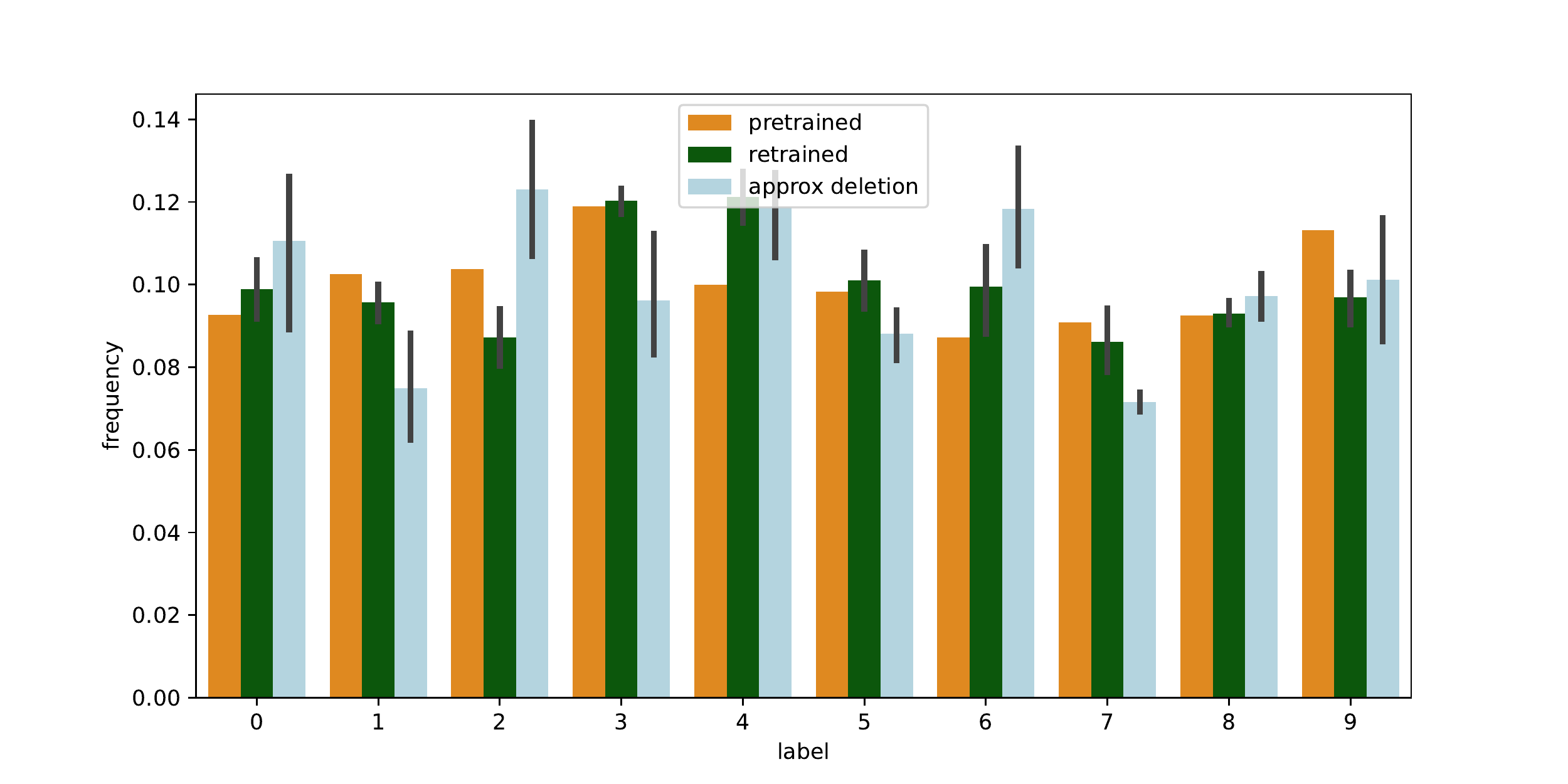}
	\caption{$\texttt{odd-}0.9$}
	\end{subfigure}\\
	\begin{subfigure}[t!]{0.5\textwidth}
	\centering 
	\includegraphics[width=0.95\textwidth]{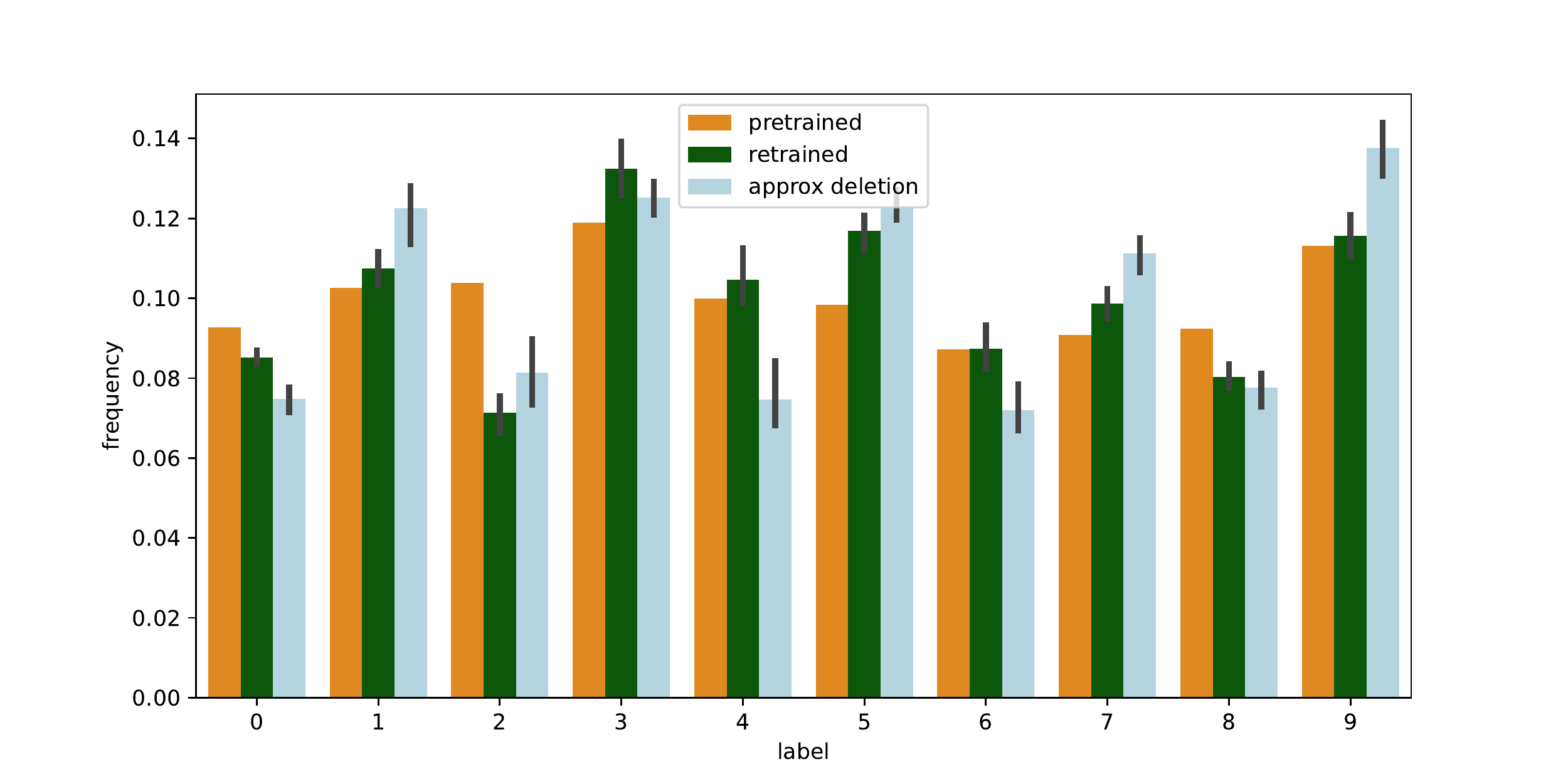}
	\caption{$\texttt{even-}0.8$}
	\end{subfigure}
	\begin{subfigure}[t!]{0.5\textwidth}
	\centering 
	\includegraphics[width=0.95\textwidth]{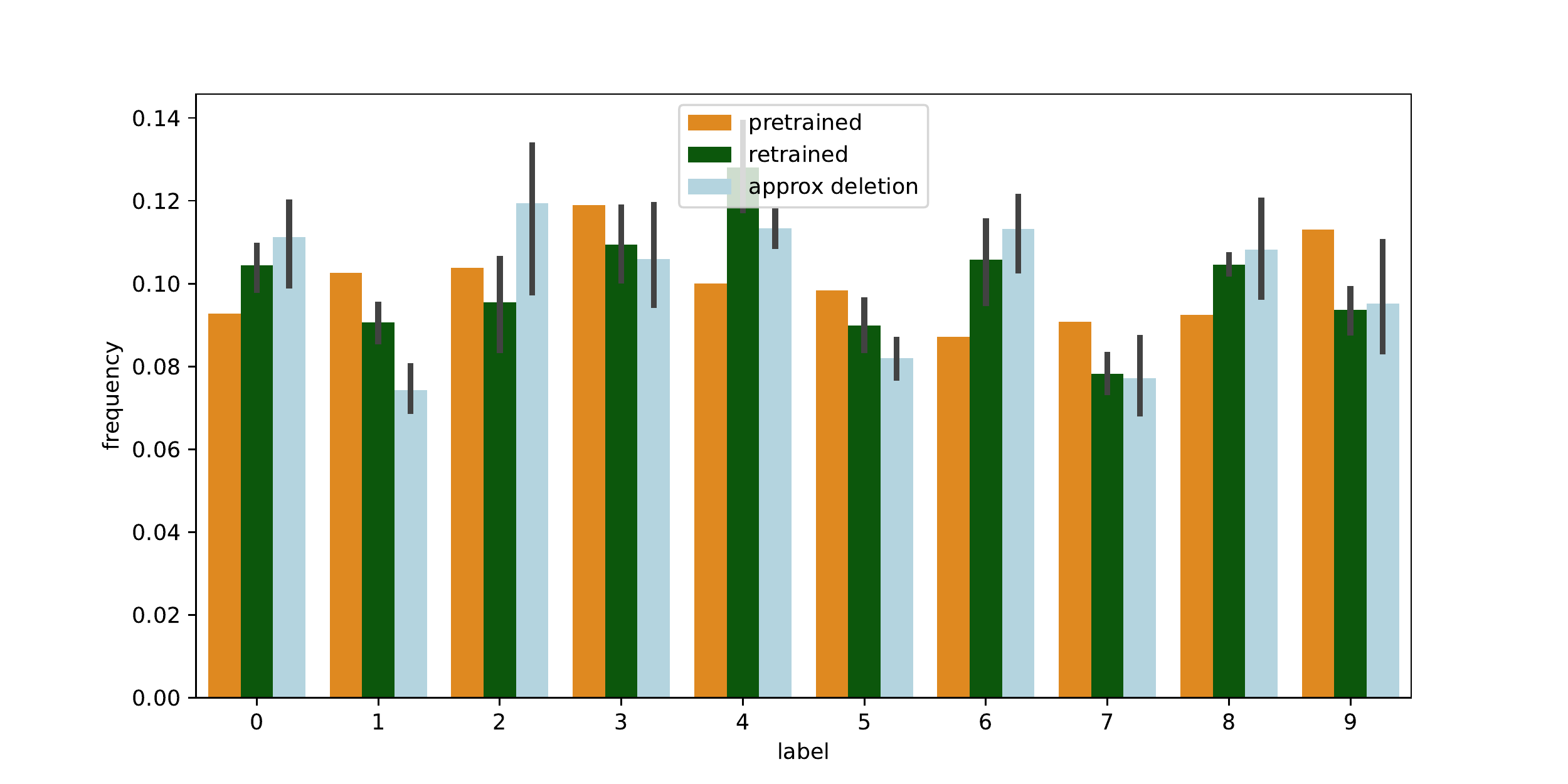}
	\caption{$\texttt{odd-}0.8$}
	\end{subfigure}\\
	\begin{subfigure}[t!]{0.5\textwidth}
	\centering 
	\includegraphics[width=0.95\textwidth]{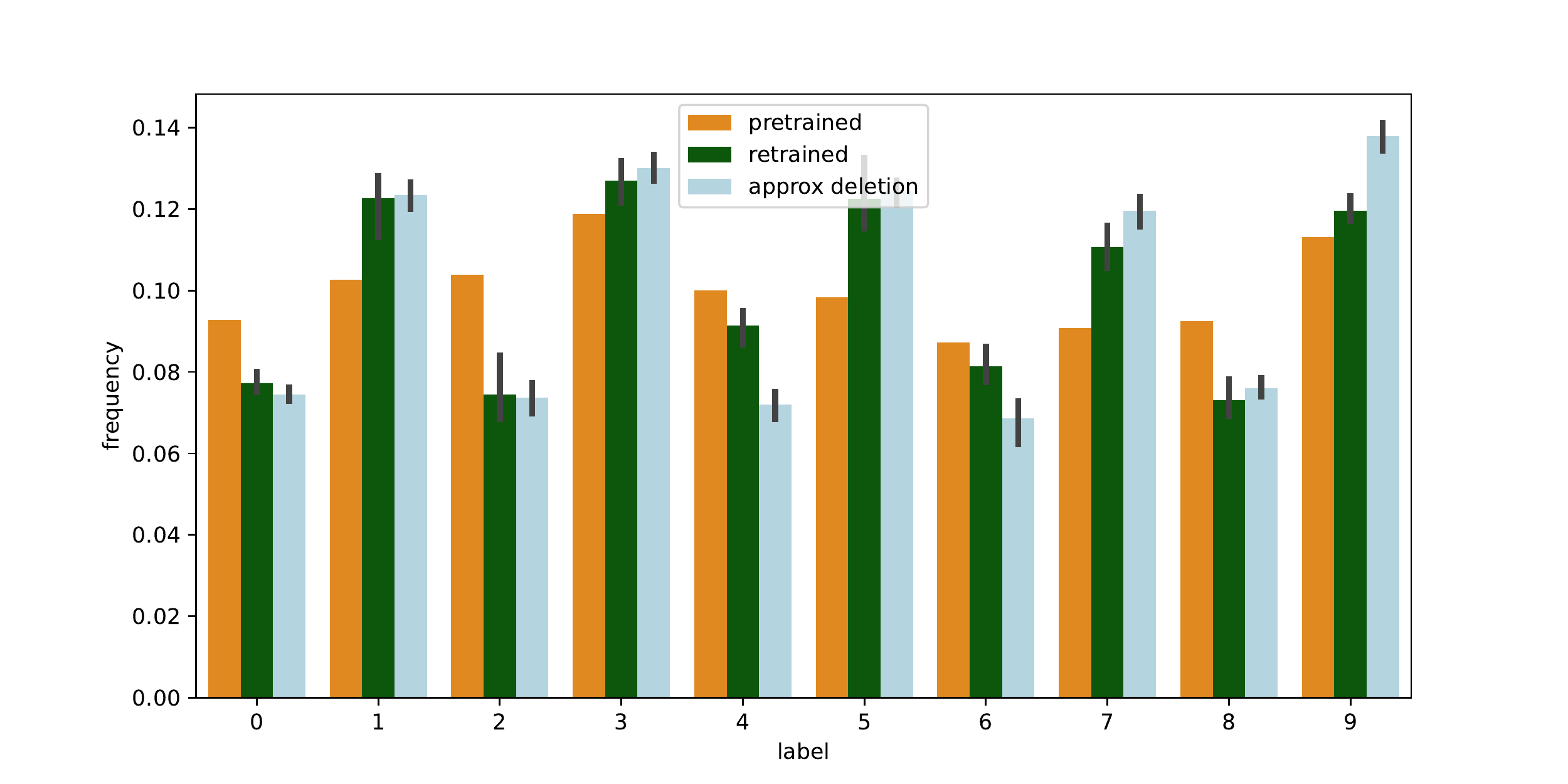}
	\caption{$\texttt{even-}0.7$}
	\end{subfigure}
	\begin{subfigure}[t!]{0.5\textwidth}
	\centering 
	\includegraphics[width=0.95\textwidth]{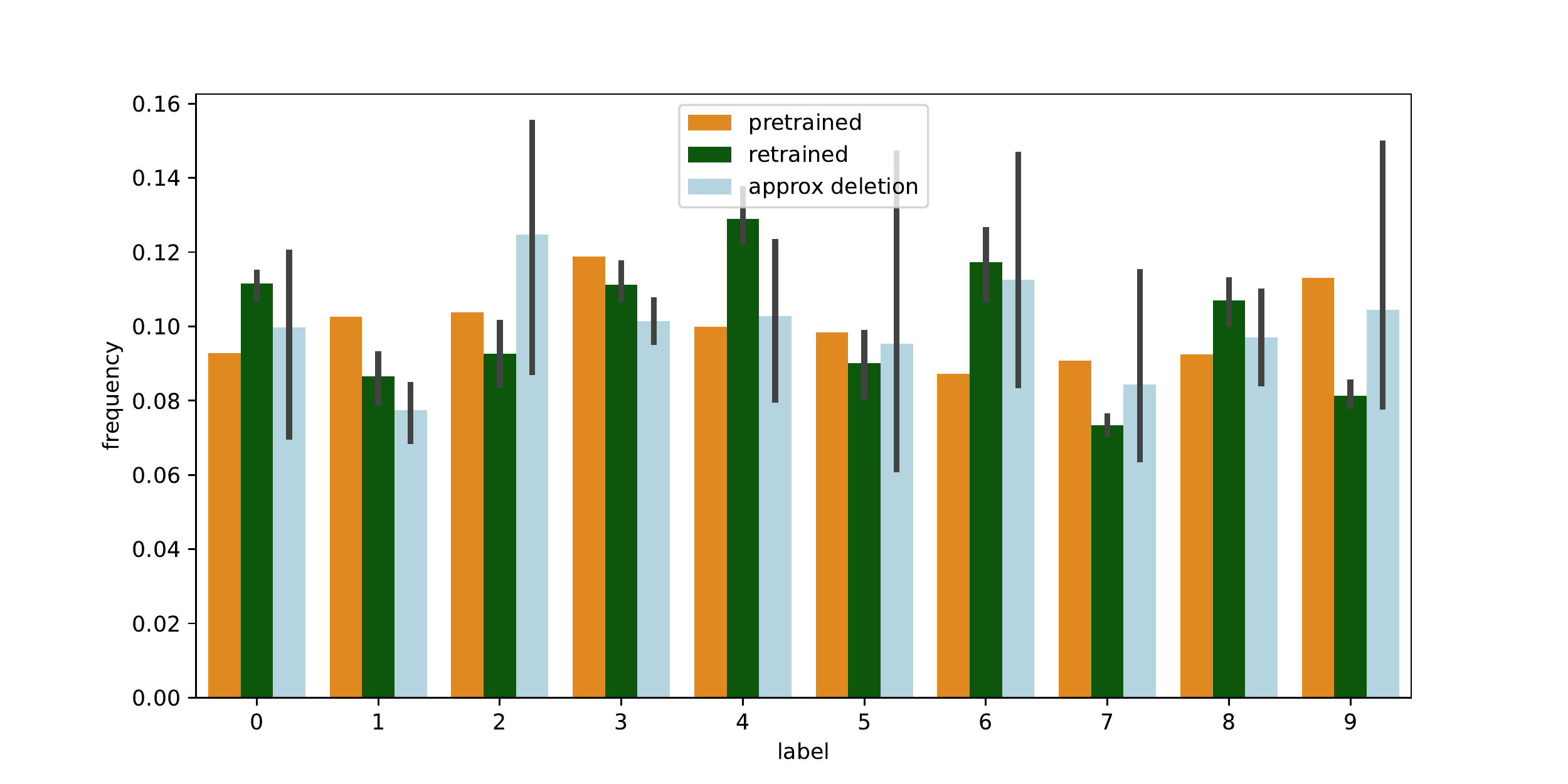}
	\caption{$\texttt{odd-}0.7$}
	\end{subfigure}\\
	\begin{subfigure}[t!]{0.5\textwidth}
	\centering 
	\includegraphics[width=0.95\textwidth]{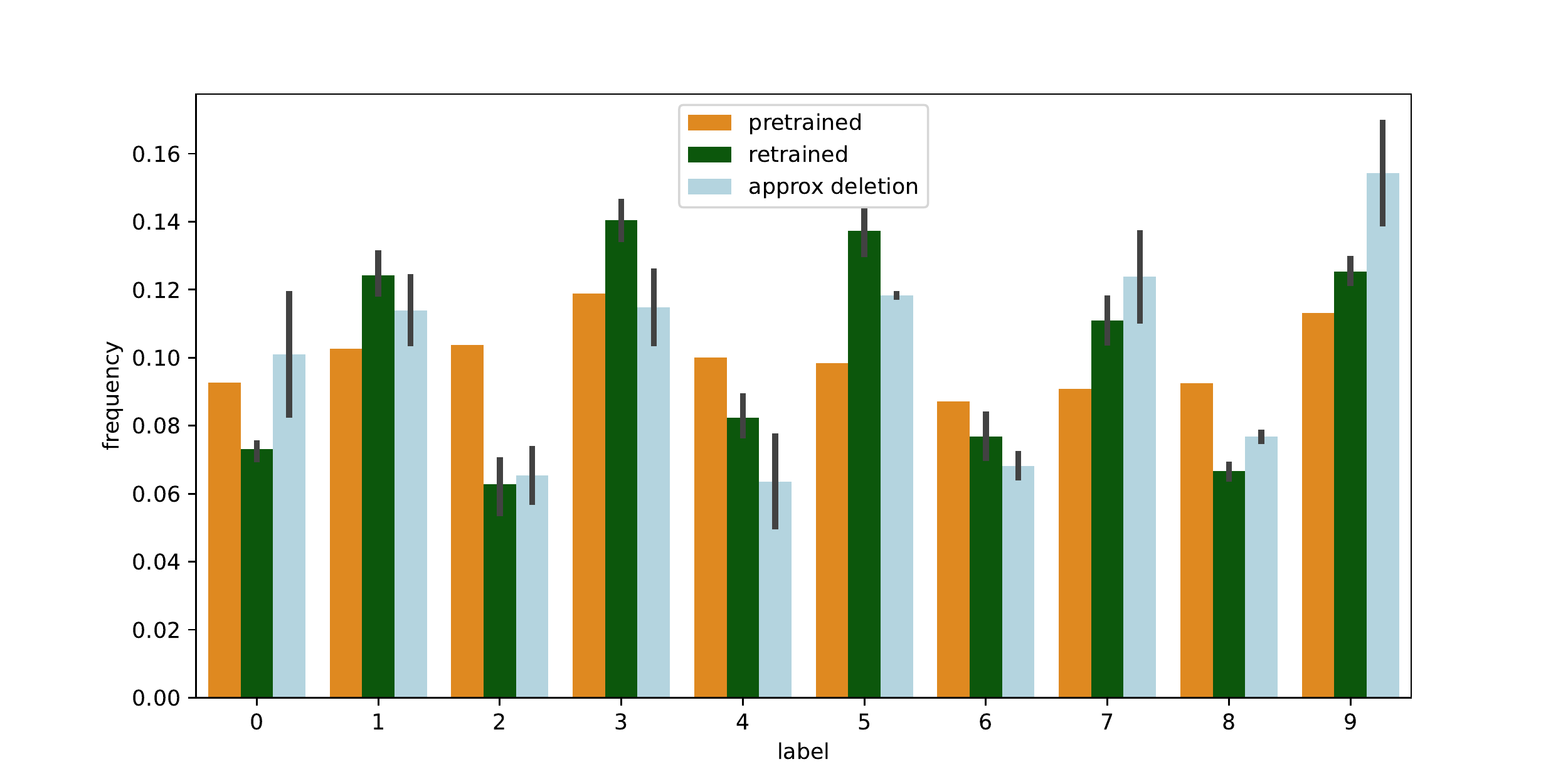}
	\caption{$\texttt{even-}0.6$}
	\end{subfigure}
	\begin{subfigure}[t!]{0.5\textwidth}
	\centering 
	\includegraphics[width=0.95\textwidth]{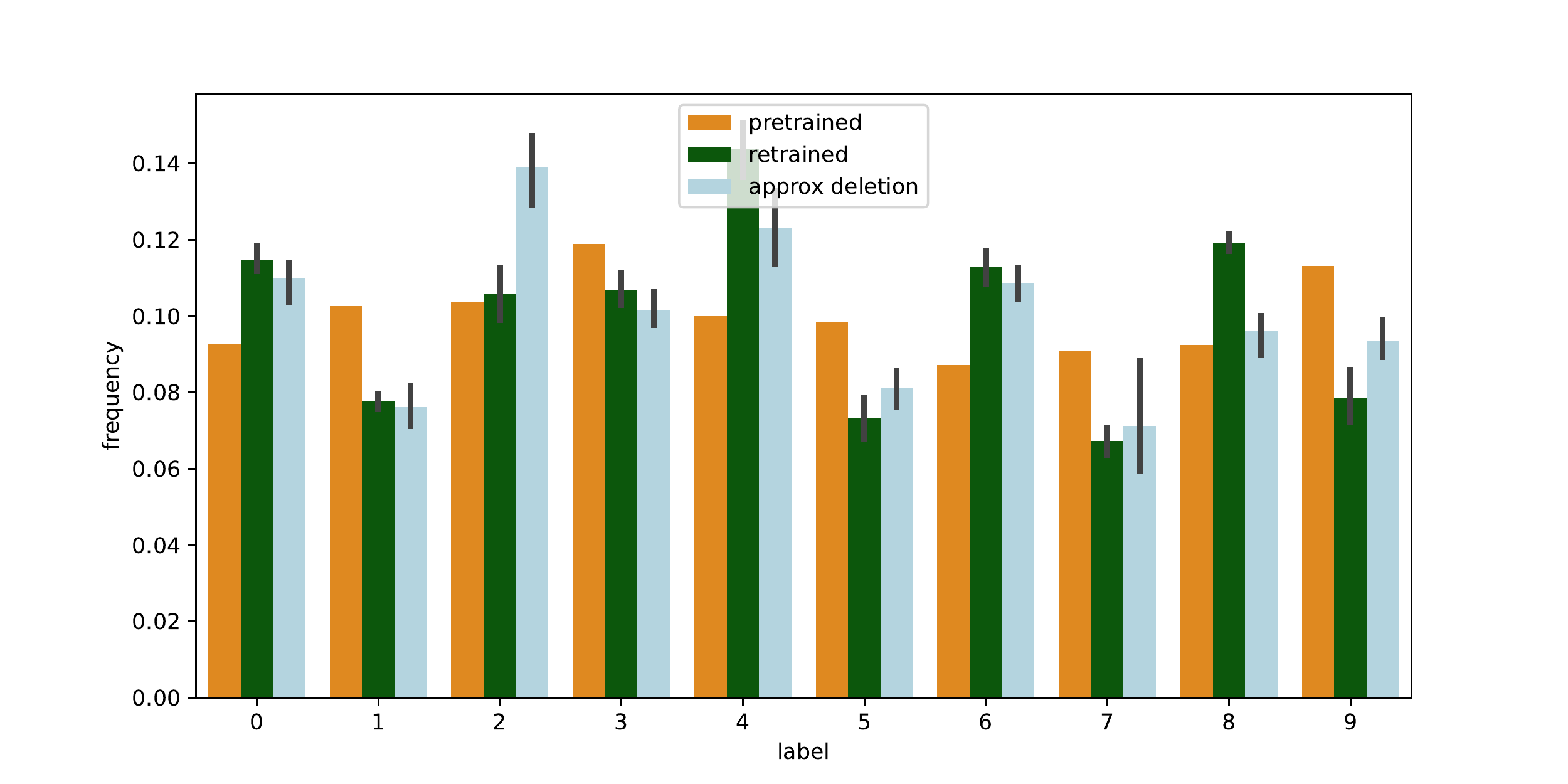}
	\caption{$\texttt{odd-}0.6$}
	\end{subfigure}\\
	
	\vspace{-0.3em}
	\caption{Label distributions of samples from pre-trained, re-trained, and approximated models. The closeness between green and light blue distributions indicate how well the fast deletion performs.}
	\vspace{-0.3em}
	\label{fig: GAN Q2 FMNIST appendix}
\end{figure}

\paragraph{Question 2 (Fast Deletion).}
Label distributions for each deletion set (including means and standard errors for five random seeds) are shown in Fig. \ref{fig: GAN Q2 FMNIST appendix} (extension of Fig. \ref{fig: GAN Q2 FMNIST}). Similar to MNIST, we find the approximated model generates less (even or odd) labels some data with these labels are deleted from the training set., and the variances for deleting odd labels are slightly higher than deleting even labels.

\begin{figure}[!h]
\vspace{-0.3em}
	\begin{subfigure}[t!]{0.5\textwidth}
	\centering 
	\includegraphics[width=0.95\textwidth]{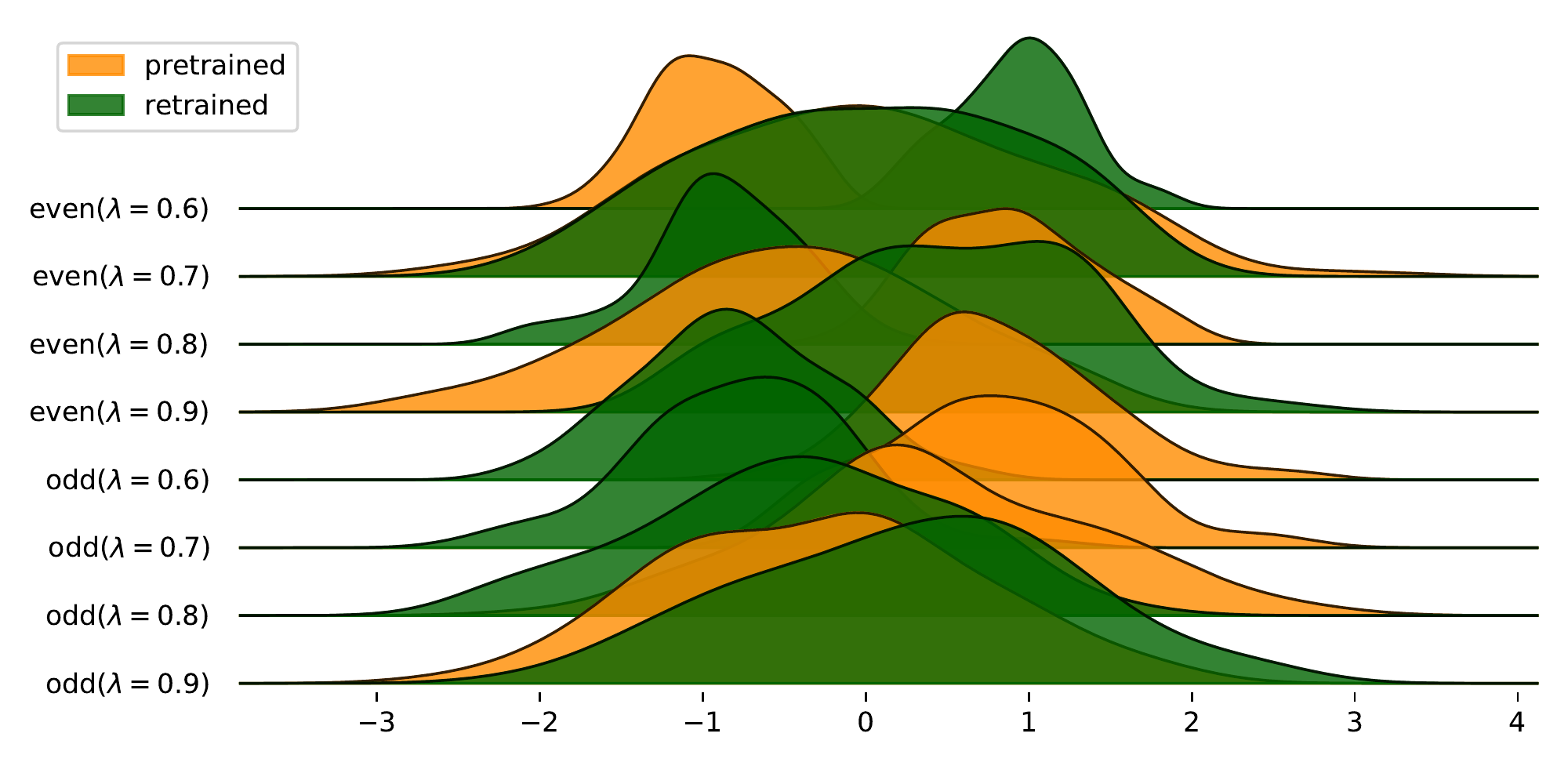}
	\caption{$\mathrm{ASC}$ for VDM-based DRE ($\phi(t)=\log(t)$)}
	\end{subfigure}
	\begin{subfigure}[t!]{0.5\textwidth}
	\centering 
	\includegraphics[width=0.95\textwidth]{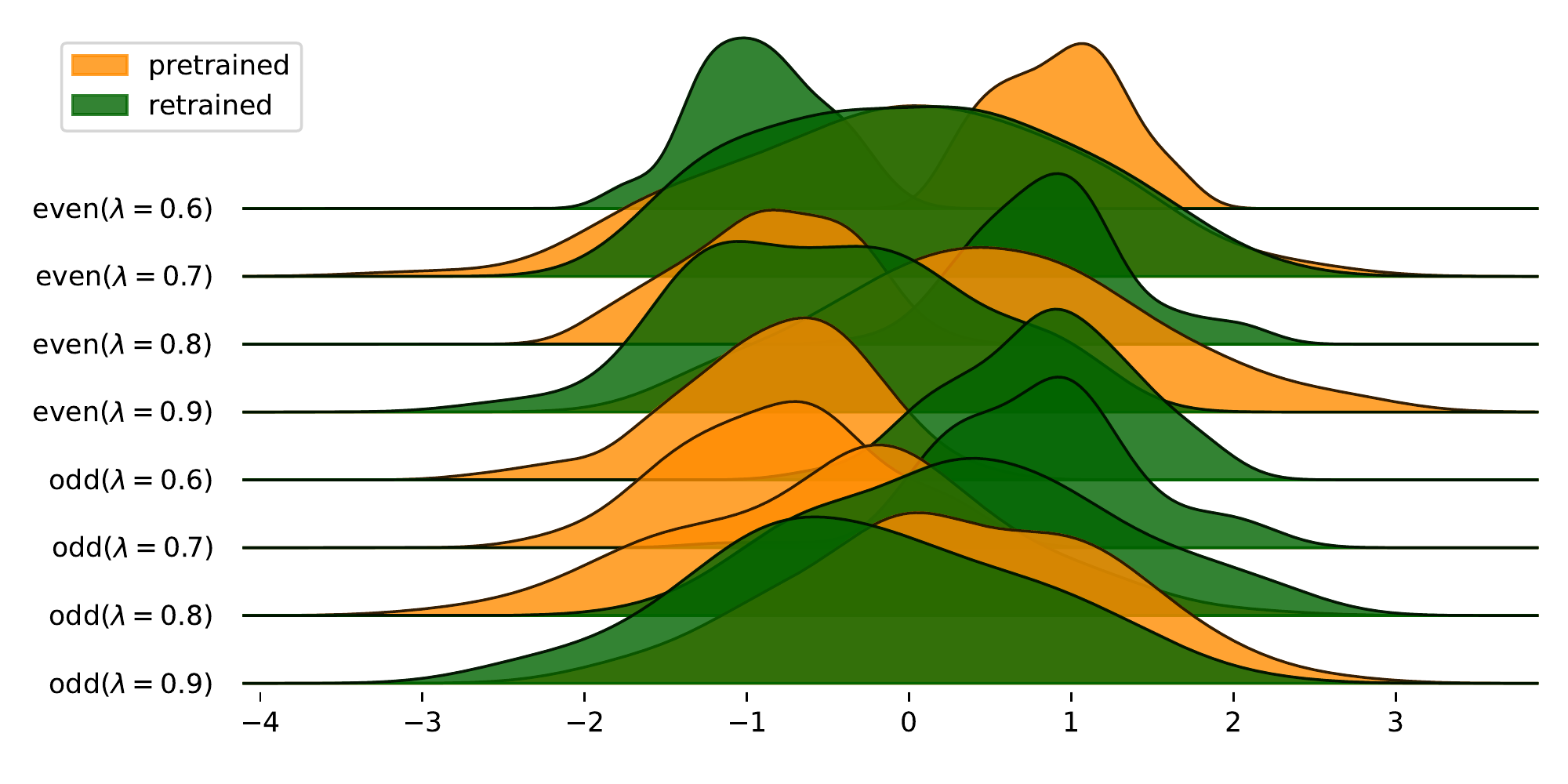}
	\caption{$\mathrm{ASC}$ for VDM-based DRE ($\phi(t)=t\log(t)$)}
	\end{subfigure} \\
	\begin{subfigure}[t!]{1.0\textwidth}
	\centering 
	\includegraphics[width=0.475\textwidth]{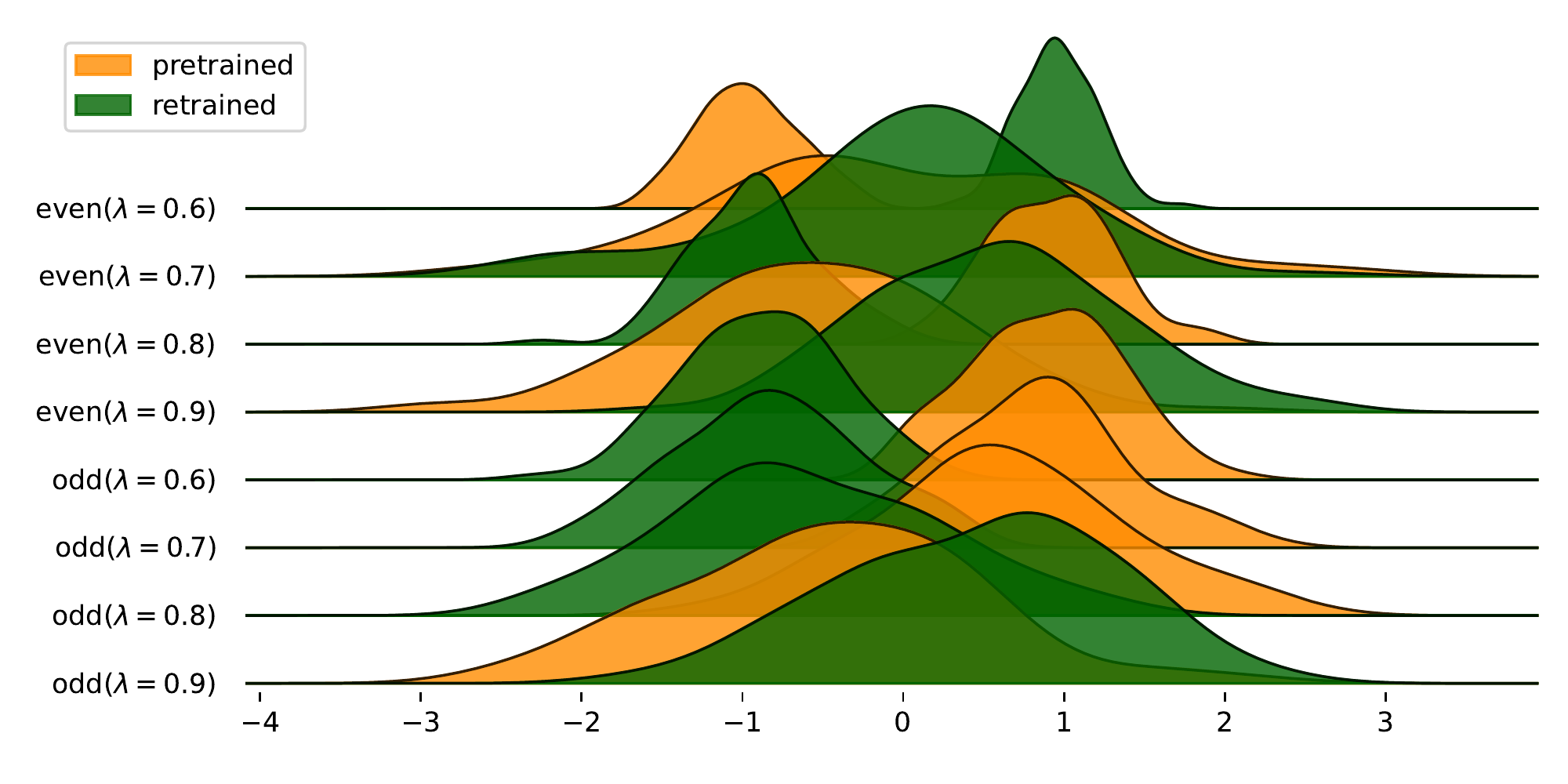}
	\caption{$\mathrm{LR}$ for VDM-based DRE}
	\end{subfigure} 
		
	\vspace{-0.3em}
	\caption{(a)-(b) $\hat{\mathrm{ASC}}_{\phi}(\hat{Y},Y_{H_0},\hat{\rho})$ vs $\hat{\mathrm{ASC}}_{\phi}(\hat{Y},Y_{H_1},\hat{\rho})$. (c) $\mathrm{LR}(Y_{H_0},\hat{\rho})$ vs $\mathrm{LR}(Y_{H_1},\hat{\rho})$}
	\vspace{-0.3em}
	\label{fig: GAN Q3 FMNIST appendix}
\end{figure}

\newpage
\paragraph{Question 3 (Hypothesis Test).}
We generate $m=1000$ samples for each $Y_{H_i}$, $i=1,2$, and $\hat{Y}$. We visualize distributions of LR and ASC statistics between $Y_{H_0}$ and $Y_{H_1}$ in Fig. \ref{fig: GAN Q3 FMNIST appendix} (extension of Fig. \ref{fig: GAN Q3 LR MNIST} for Fashion-MNIST). The separation between the distributions indicates how the DRE can distinguish samples between pre-trained and re-trained models. The separation is good for some deletion sets (e.g. $\lambda=0.6$) while not obvious for others (e.g. $\lambda=0.9$), indicating performing the deletion test for Fashion-MNIST is harder than MNIST. There is no significant differences between LR and ASC statistics. 

